%% file: main.tex
\documentclass{article}

\usepackage{preprint}
\usepackage[round,sort]{natbib}

\usepackage[T1]{fontenc}
\usepackage{amsmath}
\usepackage{mathtools}
\usepackage{amsthm}
\usepackage{newtxtext}
\usepackage{newtxmath}
\usepackage{microtype}
\usepackage{xcolor}
\usepackage{graphicx}
\usepackage{booktabs}
\usepackage{multirow}
\usepackage{array}
\usepackage[font=small,labelfont=bf,skip=4pt]{caption}
\usepackage{subcaption}
\usepackage{placeins}
\usepackage[hang,flushmargin]{footmisc}
\usepackage{enumitem}
\definecolor{linkblue}{HTML}{1F4E9A}
\usepackage[colorlinks=true,linkcolor=linkblue,citecolor=linkblue,urlcolor=linkblue]{hyperref}
\usepackage{bookmark}%
\usepackage{xurl}
\usepackage[capitalize,noabbrev]{cleveref}

\setlist{nosep,leftmargin=1.4em}
\makeatletter
\def\ci@split#1 [#2]#3\ci@end{%
  \if\relax\detokenize{#3}\relax\expandafter\@firstoftwo\else\expandafter\@secondoftwo\fi
  {\ci@unit{#1} &}{\ci@unit{#1} & {\scriptsize[#2]}}}
\def\ci@unit#1{\ci@pp#1\,pp\ci@end}
\def\ci@pp#1\,pp#2\ci@end{\ci@pct#1\%\ci@end}
\def\ci@pct#1\%#2\ci@end{#1}
\newcommand{\ci}[1]{\expandafter\ci@split#1 []\ci@end}
\makeatother

\input{generated/numbers}
\input{generated/neighbour_counts}

\input{generated/setup_facts}
\input{generated/round3b_numbers}

\newcommand{\memshift}{\textsc{MemShift}}
\newcommand{\repourl}{https://github.com/zichuanxu/memshift}
\newcommand{\holdmem}{\textsc{HoldMem}}
\newcommand{\Ssrc}{S_{\mathrm{src}}}
\newcommand{\Sdef}{S_{\mathrm{def}}}
\newcommand{\Sdel}{S_{\mathrm{del}}}
\newcommand{\Swr}{S_{\mathrm{wr}}}
\newcommand{\Simpl}{S_{\mathrm{impl}}}
\newcommand{\Scons}{S_{\mathrm{cons}}}
\definecolor{cB}{HTML}{0072B2}
\definecolor{cD}{HTML}{009E73}
\definecolor{cO}{HTML}{E69F00}
\definecolor{cgrey}{HTML}{6E6E6E}
\definecolor{cX}{HTML}{CC79A7}
\newcommand{\pass}{\checkmark}
\newcommand{\fail}{$\times$}
\input{memtab-style}

\hypersetup{
  pdftitle={MemShift: Separating Detection from Repair in LLM-Derived Personal Memory},
  pdfauthor={Zichuan Xu, Haochen Zhang, Yuyang Dong, Makoto Onizuka, Chuan Xiao},
  pdfkeywords={LLM memory, derived knowledge, invalidation, benchmark, counterfactual evaluation},
}

\title{MemShift: Separating Detection from Repair\\in LLM-Derived Personal Memory}
\author{%
  Zichuan Xu$^{1}$ \qquad Haochen Zhang$^{1,2}$ \qquad Yuyang Dong$^{2}$ \qquad Makoto Onizuka$^{1}$ \qquad Chuan Xiao$^{1,3}$ \\[2pt]
  \normalfont $^{1}$Osaka University \quad $^{2}$Initial S \quad $^{3}$Nagoya University \\[1pt]
  \normalfont\texttt{\{xu.zichuan,onizuka,chuanx\}@ist.osaka-u.ac.jp} \\
  \normalfont\texttt{\{hudson,dongyuyang\}@initial-s.com}
}

\begin{document}

\maketitle

\begin{abstract}
\input{sections/abstract}
\end{abstract}

\input{sections/intro}
\input{sections/related}
\input{sections/benchmark}
\input{sections/instrument}
\input{sections/setup}
\input{sections/results}
\input{sections/discussion}
\input{sections/conclusion}

\input{sections/statements}

{\small
\setlength{\bibsep}{2pt plus 1pt}
\bibliographystyle{plainnat}
\bibliography{refs/references}
}

\clearpage
\input{sections/appendix}

\end{document}

%% file: generated/numbers.tex
\newcommand{\CtlFGainRepDefDefinitionEst}{$+$1.0\,pp}
\newcommand{\CtlFGainRepSrcDefinitionEst}{0.0\,pp}
\newcommand{\CtlFRepDefDefinitionEst}{$+$93.6\,pp}
\newcommand{\CtlFRepSrcDefinitionEst}{$+$77.9\,pp}
\newcommand{\CtlFullLimit}{25,000}
\newcommand{\CtlFullTruncatedShareAll}{33.9\% [33.3, 34.8]}
\newcommand{\CtlGPoolAll}{$+$2.9\,pp [$+$1.7, $+$4.1]}
\newcommand{\CtlGStabDoneSrcDefinition}{$-$15.2\,pp [$-$27.5, $-$4.0]}
\newcommand{\CtlKCitedOrphanOther}{$+$3.6\,pp [0.0, $+$6.5]}
\newcommand{\CtlKCitedSrcDefinition}{$+$29.7\,pp [$+$24.5, $+$36.8]}
\newcommand{\CtlKUpdSrcDefinition}{0.0\,pp}
\newcommand{\CtlTUpdDefDefinition}{0.0\,pp}
\newcommand{\CtlTUpdOrphanAll}{$-$35.2\,pp [$-$39.3, $-$30.3]}
\newcommand{\CtlTUpdOrphanDefinition}{$-$0.6\,pp [$-$2.0, 0.0]}
\newcommand{\CtlTUpdSrcDefinition}{$+$0.6\,pp [0.0, $+$1.8]}
\newcommand{\CtlWorlds}{10}
\newcommand{\CtlZUpdSrcOtherEst}{$-$18.2\,pp}
\newcommand{\CtlZVsDonesDefDefinition}{$-$6.0\,pp [$-$8.3, $-$4.0]}
\newcommand{\CtlZVsDonesSrcDefinition}{0.0\,pp [$-$6.4, $+$5.5]}
\newcommand{\EFiveCostGraphitiCalls}{6.1}
\newcommand{\EFiveCostGraphitiUnits}{33}
\newcommand{\EFiveExMemBRateEst}{54.9\%}
\newcommand{\EFiveKappaBDev}{1.00}
\newcommand{\EFiveKappaBShort}{1.00}
\newcommand{\EFiveKappaDDev}{0.50}
\newcommand{\EFiveKappaDShort}{0.50}
\newcommand{\EFiveMemNoiseDifferOrphan}{21.5\%}
\newcommand{\EFiveMemNoiseDifferSrc}{23.6\%}
\newcommand{\EFiveMemNoiseUpdateOrphan}{$+$20.0\,pp}
\newcommand{\EFiveMemNoiseUpdateSrc}{$+$13.2\,pp}
\newcommand{\EFiveMemNoiseWorlds}{3}
\newcommand{\EFiveMemParseErrorRate}{3.6\%}
\newcommand{\EFiveMemParseErrors}{398}
\newcommand{\EFiveMemTFour}{1.14\%}
\newcommand{\EFiveMemTThreeEst}{$-$16.9\,pp}
\newcommand{\EFiveMemUpdatePredSrc}{57.6\% [51.5, 63.4]}
\newcommand{\EFiveMemUpdateSrc}{70.6\% [64.2, 76.2]}
\newcommand{\EFiveMemWorlds}{15}
\newcommand{\EFiveRagTThreeEst}{$-$20.7\,pp}
\newcommand{\EFiveZeroTThreeEst}{$-$16.8\,pp}
\newcommand{\EFourXCalendar}{$-$2.5\,pp [$-$11.2, $+$8.8]}
\newcommand{\EFourXCalendarDropBone}{$-$21.2\,pp [$-$27.4, $-$13.2]}
\newcommand{\EFourXCalendarDropBzero}{$-$18.8\,pp [$-$29.4, $-$12.5]}
\newcommand{\EFourXCalendarDropZero}{$-$16.2\,pp [$-$20.0, $-$11.8]}
\newcommand{\EFourXCalendarN}{80}
\newcommand{\EFourXPooled}{$-$3.4\,pp [$-$8.1, $+$2.4]}
\newcommand{\EFourXWorlds}{5}
\newcommand{\EOneBEffectDoneDefEst}{$+$26.9\,pp}
\newcommand{\EOneBEffectDzeroDef}{$+$0.2\,pp [$-$2.9, $+$3.4]}
\newcommand{\EOneBEffectDzeroDefEst}{$+$0.2\,pp}
\newcommand{\EOneBEffectPooledCitedDef}{$-$1.3\,pp [$-$4.8, $+$2.0]}
\newcommand{\EOneBEffectPooledCitedOrphan}{$+$42.5\,pp [$+$37.8, $+$47.0]}
\newcommand{\EOneBEffectPooledCitedOrphanEst}{$+$42.5\,pp}
\newcommand{\EOneBEffectPooledCitedSrc}{0.0\,pp [$-$3.0, $+$3.1]}
\newcommand{\EOneBEffectPooledDef}{$+$13.5\,pp [$+$10.6, $+$16.5]}
\newcommand{\EOneBEffectPooledOrphan}{$+$51.7\,pp [$+$47.3, $+$56.0]}
\newcommand{\EOneBEffectPooledOrphanEst}{$+$51.7\,pp}
\newcommand{\EOneBEffectPooledSrc}{$+$9.2\,pp [$+$6.7, $+$11.9]}
\newcommand{\EOneExCheckOnlyRepairDefinitionDefEst}{$-$4.1\,pp}
\newcommand{\EOneExCheckOnlyRepairDefinitionSrcEst}{$-$10.9\,pp}
\newcommand{\EOneExFallbackRepairDefEst}{$+$31.4\,pp}
\newcommand{\EOneExFallbackRepairSrcEst}{$+$15.2\,pp}
\newcommand{\EOneGranFalseSrc}{$+$63.7\,pp [$+$59.9, $+$67.5]}
\newcommand{\EOneWorlds}{20}
\newcommand{\ESevenAbstainDoneBone}{0.0\%}
\newcommand{\ESevenAbstainZero}{5.0\% [0.0, 15.0]}
\newcommand{\ESevenConflictItems}{20}
\newcommand{\ESevenImplicitCeiling}{6.7\%}
\newcommand{\ESevenImplicitItems}{15}
\newcommand{\ESevenLeakDoneBone}{63.0\% [42.0, 76.3]}
\newcommand{\ESevenLeakDzeroBzero}{94.8\% [90.6, 98.5]}
\newcommand{\ESevenLeakZero}{3.2\% [0.0, 12.6]}
\newcommand{\ESevenOverBlockDoneBone}{36.0\% [0.0, 79.1]}
\newcommand{\ESevenThetaERagMainDef}{43.3\% [40.9, 46.1]}
\newcommand{\ESevenThetaERagMainSrc}{69.7\% [66.1, 74.1]}
\newcommand{\ESevenThetaERagSwapDef}{44.6\% [40.1, 50.0]}
\newcommand{\ESevenThetaERagSwapSrc}{68.2\% [65.7, 70.2]}
\newcommand{\ESevenThetaRBUpdateMainDef}{$+$17.4\,pp [$+$10.9, $+$22.2]}
\newcommand{\ESevenThetaRBUpdateMainSrc}{$+$9.1\,pp [$+$5.4, $+$13.9]}
\newcommand{\ESevenThetaRBUpdateSwapDef}{$+$16.8\,pp [$+$13.4, $+$20.5]}
\newcommand{\ESevenThetaRBUpdateSwapSrc}{$+$14.3\,pp [$+$10.9, $+$18.6]}
\newcommand{\ESevenThetaRDRepairMainDef}{$+$33.4\,pp [$+$30.1, $+$37.4]}
\newcommand{\ESevenThetaRDRepairMainSrc}{$+$13.7\,pp [$+$11.9, $+$16.2]}
\newcommand{\ESevenThetaRDRepairSwapDef}{$+$35.1\,pp [$+$29.7, $+$40.0]}
\newcommand{\ESevenThetaRDRepairSwapSrc}{$+$16.2\,pp [$+$14.8, $+$18.3]}
\newcommand{\ESevenThetaRRagMainDef}{43.3\% [40.9, 46.1]}
\newcommand{\ESevenThetaRRagSwapDef}{50.7\% [45.0, 57.0]}
\newcommand{\ESevenThetaRWorlds}{5}
\newcommand{\ESevenThetaRZeroMainDef}{49.3\% [45.8, 53.6]}
\newcommand{\ESevenThetaRZeroSwapDef}{40.9\% [37.1, 45.1]}
\newcommand{\ESevenWorlds}{5}
\newcommand{\ESixNoExtMissed}{$+$100.0\,pp}
\newcommand{\ESixWithExtPost}{0.0\,pp}
\newcommand{\ESixWithExtPostEst}{0.0\,pp}
\newcommand{\EThreeExNoDefTThreePassHold}{1 of 3}
\newcommand{\EThreeExTThreeHoldDefinitionOneContribution}{$+$8.1\,pp}
\newcommand{\EThreeExTThreeHoldDefinitionThreeContribution}{$+$6.9\,pp}
\newcommand{\EThreeExTThreeHoldDefinitionTwelveContribution}{$+$7.7\,pp}
\newcommand{\EThreeTThreePassHold}{3 of 3}
\newcommand{\EThreeZeroMinusHoldTwelve}{$-$25.8\,pp [$-$28.5, $-$23.4]}
\newcommand{\EThreeZeroMinusHoldTwelveEst}{$-$25.8\,pp}
\newcommand{\ETwoCheckerPassDown}{77.0\%}
\newcommand{\ETwoCheckerPassSame}{79.3\%}
\newcommand{\ETwoCheckerPassUp}{82.0\%}
\newcommand{\ETwoDoneSurvivalDown}{99.6\%}
\newcommand{\ETwoDoneSurvivalSame}{98.6\%}
\newcommand{\ETwoDoneSurvivalUp}{99.6\%}
\newcommand{\ETwoExFailDefinition}{32}
\newcommand{\ETwoExFailTotal}{41}
\newcommand{\ETwoExGainsDoneBoneSame}{91}
\newcommand{\ETwoExLossesDoneBoneSame}{11}
\newcommand{\ETwoExNetDoneBoneSameEst}{$+$8.3\,pp}
\newcommand{\ETwoExRetentionDoneBoneFreetextSameEst}{85.1\%}
\newcommand{\ETwoExRetentionDzeroBoneFreetextSameEst}{95.5\%}
\newcommand{\ETwoExWriterAgreePooledAurocEst}{0.832}
\newcommand{\ETwoExWriterAgreePooledSlope}{1.01 [0.90, 1.18]}
\newcommand{\ETwoExWriterNewAccPooledAurocEst}{0.887}
\newcommand{\ETwoExWriterNoDefAgreePooledAurocEst}{0.275}
\newcommand{\ETwoExWriterNonDefPooledAurocEst}{0.869}
\newcommand{\ETwoExWriterOldAccPooledAurocEst}{0.800}
\newcommand{\ETwoPathThreeN}{2,446}
\newcommand{\ETwoTOneSame}{$-$1.4\,pp [$-$1.9, $-$0.9]}
\newcommand{\ExAcctFitTwoCells}{24}
\newcommand{\ExAcctFitTwoDefinitionSpearman}{$-$0.21 [$-$0.53, 0.31]}
\newcommand{\ExAcctFitTwoIntercept}{$+$3.2\,pp [$+$0.6, $+$5.0]}
\newcommand{\ExAcctFitTwoOtherSpearman}{0.75 [0.43, 0.97]}
\newcommand{\ExAcctFitTwoSlope}{0.30 [0.26, 0.34]}
\newcommand{\ExAcctFitTwoSpearman}{0.75 [0.68, 0.84]}
\newcommand{\ExAcctGemSrcDefinitionBxDEst}{$+$7.3\,pp}
\newcommand{\ExAcctGptSrcDefinitionBxDEst}{$+$44.8\,pp}
\newcommand{\ExAcctLlamaDefOtherBxD}{$+$19.3\,pp [$-$1.4, $+$41.0]}
\newcommand{\ExAcctLlamaOrphanOtherBxDEst}{$+$27.6\,pp}
\newcommand{\ExAcctMainDefDefinitionBxD}{$+$38.8\,pp [$+$32.0, $+$44.8]}
\newcommand{\ExAcctMainDefOtherBxD}{$-$0.6\,pp [$-$4.0, $+$2.7]}
\newcommand{\ExAcctMainOrphanDefinitionBxD}{$+$35.4\,pp [$+$30.4, $+$40.0]}
\newcommand{\ExAcctMainOrphanOtherBxD}{$-$1.4\,pp [$-$5.4, $+$2.3]}
\newcommand{\ExAcctMainSrcDefinitionBxD}{$+$27.3\,pp [$+$21.6, $+$32.6]}
\newcommand{\ExAcctMainSrcOtherBxD}{$-$0.6\,pp [$-$3.8, $+$2.5]}
\newcommand{\ExAttrDefChangedShareOfAll}{14.5\%}
\newcommand{\ExAttrDefDriftAllBDone}{$+$23.5\,pp [$+$17.4, $+$30.2]}
\newcommand{\ExAttrDefDriftAllBDzero}{$+$4.8\,pp [$+$0.9, $+$8.9]}
\newcommand{\ExAttrDefDriftDefinitionBDone}{$+$46.1\,pp [$+$35.9, $+$57.1]}
\newcommand{\ExAttrDefDriftDefinitionBDzero}{$+$7.3\,pp [0.0, $+$14.8]}
\newcommand{\ExAttrDefDriftDefinitionBxDEst}{$+$38.8\,pp}
\newcommand{\ExAttrDefDriftOtherBDone}{$+$1.8\,pp [$-$3.4, $+$7.1]}
\newcommand{\ExAttrDefDriftOtherBDzero}{$+$2.3\,pp [$-$2.4, $+$7.4]}
\newcommand{\ExAttrDefDriftShareOfAll}{8.8\%}
\newcommand{\ExAttrDefKappaKeptBDone}{$+$48.7\,pp [$+$39.1, $+$59.0]}
\newcommand{\ExAttrDefKappaKeptBDzero}{$+$9.2\,pp [$+$1.5, $+$17.1]}
\newcommand{\ExAttrDefKappaKeptBxD}{$+$39.5\,pp [$+$30.2, $+$48.7]}
\newcommand{\ExAttrDefKappaKeptItems}{152}
\newcommand{\ExAttrDefKappaRedefinedBDone}{$+$31.1\,pp [$+$24.8, $+$36.9]}
\newcommand{\ExAttrDefKappaRedefinedBDzero}{$-$7.2\,pp [$-$12.5, $-$1.7]}
\newcommand{\ExAttrDefKappaRedefinedBxD}{$+$38.3\,pp [$+$30.5, $+$46.0]}
\newcommand{\ExAttrDefKappaRedefinedItems}{235}
\newcommand{\ExAttrDefShiftDefinitionBDone}{$+$32.0\,pp [$+$25.4, $+$38.5]}
\newcommand{\ExAttrDefShiftDefinitionBDzero}{$-$6.8\,pp [$-$12.2, $-$1.3]}
\newcommand{\ExAttrDefShiftDefinitionBxDEst}{$+$38.7\,pp}
\newcommand{\ExAttrDefShiftShareOfAll}{5.8\%}
\newcommand{\ExAttrOrphanChangedShareOfAll}{46.0\%}
\newcommand{\ExAttrOrphanDriftAllBDone}{$+$24.4\,pp [$+$18.5, $+$30.6]}
\newcommand{\ExAttrOrphanDriftAllBDzero}{$+$2.0\,pp [$-$2.0, $+$5.8]}
\newcommand{\ExAttrOrphanDriftDefinitionBDone}{$+$40.0\,pp [$+$34.5, $+$45.7]}
\newcommand{\ExAttrOrphanDriftDefinitionBDzero}{$+$5.4\,pp [0.0, $+$10.5]}
\newcommand{\ExAttrOrphanDriftOtherBDone}{$-$5.1\,pp [$-$14.3, $+$3.9]}
\newcommand{\ExAttrOrphanDriftOtherBDzero}{$-$4.4\,pp [$-$12.6, $+$3.8]}
\newcommand{\ExAttrOrphanDriftShareOfAll}{10.3\%}
\newcommand{\ExAttrOrphanShiftAllBDone}{$+$62.8\,pp [$+$57.1, $+$67.8]}
\newcommand{\ExAttrOrphanShiftAllBDzero}{$+$62.9\,pp [$+$57.7, $+$67.8]}
\newcommand{\ExAttrOrphanShiftDefinitionBDone}{$+$58.3\,pp [$+$38.7, $+$71.1]}
\newcommand{\ExAttrOrphanShiftDefinitionBDzero}{$+$18.8\,pp [$+$8.3, $+$28.0]}
\newcommand{\ExAttrOrphanShiftOtherBDone}{$+$63.0\,pp [$+$57.0, $+$68.3]}
\newcommand{\ExAttrOrphanShiftOtherBDzero}{$+$64.5\,pp [$+$59.1, $+$69.5]}
\newcommand{\ExAttrOrphanShiftShareOfAll}{35.7\%}
\newcommand{\ExAttrSrcChangedShareOfAll}{30.4\%}
\newcommand{\ExAttrSrcDriftAllBDone}{$+$18.1\,pp [$+$12.1, $+$23.6]}
\newcommand{\ExAttrSrcDriftAllBDzero}{$+$0.8\,pp [$-$3.0, $+$4.4]}
\newcommand{\ExAttrSrcDriftDefinitionBDone}{$+$32.0\,pp [$+$25.5, $+$37.9]}
\newcommand{\ExAttrSrcDriftDefinitionBDzero}{$+$3.2\,pp [$-$1.5, $+$8.1]}
\newcommand{\ExAttrSrcDriftOtherBDone}{$-$1.3\,pp [$-$7.5, $+$5.0]}
\newcommand{\ExAttrSrcDriftOtherBDzero}{$-$2.5\,pp [$-$8.1, $+$3.3]}
\newcommand{\ExAttrSrcDriftShareOfAll}{9.9\%}
\newcommand{\ExAttrSrcItemTimes}{3,840}
\newcommand{\ExAttrSrcShiftAllBDone}{$+$10.4\,pp [$+$7.2, $+$13.8]}
\newcommand{\ExAttrSrcShiftAllBDzero}{$+$7.9\,pp [$+$4.6, $+$11.5]}
\newcommand{\ExAttrSrcShiftAllBxD}{$+$2.5\,pp [$+$0.5, $+$4.4]}
\newcommand{\ExAttrSrcShiftDefinitionBDone}{$+$36.0\,pp [$+$26.6, $+$44.0]}
\newcommand{\ExAttrSrcShiftDefinitionBDzero}{$+$11.7\,pp [$+$2.9, $+$20.3]}
\newcommand{\ExAttrSrcShiftOtherBDone}{$+$6.2\,pp [$+$2.7, $+$9.8]}
\newcommand{\ExAttrSrcShiftOtherBDzero}{$+$7.2\,pp [$+$3.9, $+$10.9]}
\newcommand{\ExAttrSrcShiftShareOfAll}{20.5\%}
\newcommand{\ExAttrSrcTargetSlotShare}{20.8\%}
\newcommand{\ExAttrSrcTargetSlots}{10}
\newcommand{\ExAttrWorldSlots}{48}
\newcommand{\ExClsDmDefAllBDoneEst}{$+$25.4\,pp}
\newcommand{\ExClsDmDefDefinitionBxDEst}{$+$34.9\,pp}
\newcommand{\ExClsDmOrphanOtherBxD}{$-$32.8\,pp [$-$37.5, $-$27.7]}
\newcommand{\ExClsDmOrphanOtherDBone}{$-$29.7\,pp [$-$33.9, $-$25.2]}
\newcommand{\ExClsDmSrcDefinitionBxDEst}{$+$27.3\,pp}
\newcommand{\ExClsDmSrcDefinitionDBone}{$+$57.4\,pp [$+$49.0, $+$65.7]}
\newcommand{\ExClsDmSrcDefinitionDBoneEst}{$+$57.4\,pp}
\newcommand{\ExClsGemDefAllBxD}{$+$14.4\,pp [$+$8.3, $+$20.1]}
\newcommand{\ExClsGemDefAllBxDEst}{$+$14.4\,pp}
\newcommand{\ExClsGemDefAllDRepair}{$+$31.9\,pp [$+$28.4, $+$35.0]}
\newcommand{\ExClsGemOrphanAllBxD}{$+$7.8\,pp [$+$4.9, $+$10.7]}
\newcommand{\ExClsGemSrcAllBPooled}{$+$9.5\,pp [$+$5.5, $+$13.6]}
\newcommand{\ExClsGemSrcAllBxD}{$+$0.8\,pp [$-$2.1, $+$3.8]}
\newcommand{\ExClsGemSrcAllDRepair}{$+$13.1\,pp [$+$11.6, $+$14.6]}
\newcommand{\ExClsGptDefAllBxD}{$+$28.5\,pp [$+$23.1, $+$32.8]}
\newcommand{\ExClsGptDefAllBxDEst}{$+$28.5\,pp}
\newcommand{\ExClsGptDefAllDRepair}{$+$26.5\,pp [$+$23.7, $+$29.5]}
\newcommand{\ExClsGptOrphanAllBxD}{$+$13.0\,pp [$+$9.8, $+$15.4]}
\newcommand{\ExClsGptSrcAllBPooled}{$+$12.2\,pp [$+$8.0, $+$16.6]}
\newcommand{\ExClsGptSrcAllBxD}{$+$12.6\,pp [$+$9.6, $+$16.5]}
\newcommand{\ExClsGptSrcAllDRepair}{$+$12.0\,pp [$+$10.5, $+$13.5]}
\newcommand{\ExClsLlamaDefAllBxD}{$+$20.8\,pp [$+$11.7, $+$29.4]}
\newcommand{\ExClsLlamaDefAllDRepair}{$+$40.0\,pp [$+$36.1, $+$43.3]}
\newcommand{\ExClsLlamaOrphanAllBxD}{$+$26.4\,pp [$+$20.5, $+$31.9]}
\newcommand{\ExClsLlamaOrphanAllBxDEst}{$+$26.4\,pp}
\newcommand{\ExClsLlamaSrcAllBPooled}{$+$29.6\,pp [$+$25.1, $+$34.1]}
\newcommand{\ExClsLlamaSrcAllBxD}{$+$4.2\,pp [$-$4.2, $+$13.1]}
\newcommand{\ExClsLlamaSrcAllDRepair}{$+$20.4\,pp [$+$18.0, $+$23.1]}
\newcommand{\ExClsMainDefDefinitionBDzeroEst}{$-$0.8\,pp}
\newcommand{\ExClsMainDefDefinitionBxDEst}{$+$38.8\,pp}
\newcommand{\ExClsMainDefDefinitionDBzeroEst}{$+$54.3\,pp}
\newcommand{\ExClsMainDefDefinitionDRepairChanged}{$+$93.7\,pp [$+$89.7, $+$97.4]}
\newcommand{\ExClsMainOrphanAllBxD}{$+$5.0\,pp [$+$1.7, $+$8.1]}
\newcommand{\ExClsMainOrphanAllBxDEst}{$+$5.0\,pp}
\newcommand{\ExClsMainOrphanDefinitionBDzeroEst}{$+$7.5\,pp}
\newcommand{\ExClsMainOrphanDefinitionDBzeroEst}{$+$52.3\,pp}
\newcommand{\ExClsMainSrcDefinitionBDone}{$+$33.3\,pp [$+$28.2, $+$38.0]}
\newcommand{\ExClsMainSrcDefinitionBDzero}{$+$6.0\,pp [$+$2.5, $+$9.6]}
\newcommand{\ExClsMainSrcDefinitionBDzeroEst}{$+$6.0\,pp}
\newcommand{\ExClsMainSrcDefinitionDBone}{$+$82.9\,pp [$+$77.0, $+$89.1]}
\newcommand{\ExClsMainSrcDefinitionDBoneEst}{$+$82.9\,pp}
\newcommand{\ExClsMainSrcDefinitionDBzeroEst}{$+$55.6\,pp}
\newcommand{\ExClsShortDefAllBDzero}{$+$8.7\,pp [$+$3.5, $+$14.8]}
\newcommand{\ExClsShortDefDefinitionDRepairChanged}{$+$78.5\,pp [$+$73.9, $+$82.7]}
\newcommand{\ExCostModelMainTwelveHoldRatioZero}{3.64}
\newcommand{\ExDecMainDefDefinitionDoneBoneRepairCitedEst}{82.3\%}
\newcommand{\ExDetCitedDefDriftAllItems}{336}
\newcommand{\ExDetCitedDefDriftAllMissShareLast}{37.8\% [30.5, 45.1]}
\newcommand{\ExDetCitedDefDriftAllMissedLast}{127}
\newcommand{\ExDetCitedDefShiftAllItems}{222}
\newcommand{\ExDetCitedDefShiftAllMissShareLast}{24.3\% [17.4, 32.0]}
\newcommand{\ExDetCitedDefShiftAllMissedLast}{54}
\newcommand{\ExDetKeyOracleDefAllBxD}{$+$29.9\,pp [$+$25.0, $+$34.6]}
\newcommand{\ExDetKeyOracleDefDefinitionBDone}{$+$37.3\,pp [$+$31.1, $+$43.4]}
\newcommand{\ExDetKeyOracleDefDefinitionBDzero}{$-$1.5\,pp [$-$7.4, $+$4.5]}
\newcommand{\ExDetKeyOracleDefDefinitionBxD}{$+$38.8\,pp [$+$34.1, $+$43.7]}
\newcommand{\ExDetKeyOracleDefinitionBDzeroMax}{$+$6.7\,pp}
\newcommand{\ExDetKeyOracleDefinitionBDzeroMin}{$-$1.5\,pp}
\newcommand{\ExDetKeyOracleOrphanAllBxD}{$+$10.9\,pp [$+$6.9, $+$15.1]}
\newcommand{\ExDetKeyOracleOrphanDefinitionBDone}{$+$44.2\,pp [$+$37.0, $+$50.9]}
\newcommand{\ExDetKeyOracleOrphanDefinitionBDzero}{$+$2.6\,pp [$-$1.4, $+$8.6]}
\newcommand{\ExDetKeyOracleOrphanDefinitionBxD}{$+$41.6\,pp [$+$36.4, $+$46.0]}
\newcommand{\ExDetKeyOracleSrcAllBxD}{$+$6.6\,pp [$+$2.9, $+$10.6]}
\newcommand{\ExDetKeyOracleSrcDefinitionBDone}{$+$30.3\,pp [$+$23.8, $+$37.2]}
\newcommand{\ExDetKeyOracleSrcDefinitionBDzero}{$+$6.7\,pp [$+$1.9, $+$11.6]}
\newcommand{\ExDetKeyOracleSrcDefinitionBxD}{$+$23.6\,pp [$+$15.0, $+$33.6]}
\newcommand{\ExDetKeySourceDefAllBxD}{$+$27.1\,pp [$+$22.9, $+$32.1]}
\newcommand{\ExDetKeySourceDefDefinitionBDone}{$+$37.3\,pp [$+$31.1, $+$43.4]}
\newcommand{\ExDetKeySourceDefDefinitionBDzero}{$-$1.5\,pp [$-$7.4, $+$4.5]}
\newcommand{\ExDetKeySourceDefDefinitionBxD}{$+$38.8\,pp [$+$34.1, $+$43.7]}
\newcommand{\ExDetKeySourceDefDefinitionBxDEst}{$+$38.8\,pp}
\newcommand{\ExDetKeySourceOrphanAllBxD}{$+$4.8\,pp [$+$0.1, $+$9.9]}
\newcommand{\ExDetKeySourceOrphanDefinitionBDone}{$+$44.2\,pp [$+$37.0, $+$50.9]}
\newcommand{\ExDetKeySourceOrphanDefinitionBDzero}{$+$3.2\,pp [$-$0.7, $+$9.0]}
\newcommand{\ExDetKeySourceOrphanDefinitionBxD}{$+$40.9\,pp [$+$36.0, $+$45.2]}
\newcommand{\ExDetKeySourceOrphanDefinitionBxDEst}{$+$40.9\,pp}
\newcommand{\ExDetKeySourceSrcAllBxD}{$+$6.2\,pp [$+$2.5, $+$10.3]}
\newcommand{\ExDetKeySourceSrcDefinitionBDone}{$+$30.3\,pp [$+$23.8, $+$37.2]}
\newcommand{\ExDetKeySourceSrcDefinitionBDzero}{$+$4.8\,pp [0.0, $+$10.1]}
\newcommand{\ExDetKeySourceSrcDefinitionBxD}{$+$25.5\,pp [$+$16.9, $+$34.6]}
\newcommand{\ExDetKeySourceSrcDefinitionBxDEst}{$+$25.5\,pp}
\newcommand{\ExDetKeyWorlds}{10}
\newcommand{\ExDetMainDefAllDoneBoneMissedCitedEst}{32.4\%}
\newcommand{\ExDetMainDefDefinitionDoneBoneRepairCitedEst}{100.0\%}
\newcommand{\ExDetMainDefDefinitionDoneBoneRepairEst}{100.0\%}
\newcommand{\ExDetMainDefDefinitionDoneBzeroUpdateEst}{61.8\%}
\newcommand{\ExDetMainDefDefinitionDzeroBoneMissedCitedEst}{29.5\%}
\newcommand{\ExDetMainDefDefinitionDzeroBoneRepairEst}{6.7\%}
\newcommand{\ExDetMainDefDefinitionDzeroBoneUpdateEst}{6.7\%}
\newcommand{\ExDetMainDefDefinitionDzeroBzeroUpdateEst}{7.5\%}
\newcommand{\ExDetMainDefOtherDoneBoneRepairEst}{73.1\%}
\newcommand{\ExDetMainDefOtherDoneBzeroUpdateEst}{71.3\%}
\newcommand{\ExDetMainDefOtherDzeroBoneRepairEst}{73.1\%}
\newcommand{\ExDetMainDefOtherDzeroBzeroUpdateEst}{70.8\%}
\newcommand{\ExDetMainDefStabilityItems}{558}
\newcommand{\ExDetMainDefUpdateItems}{558}
\newcommand{\ExDetMainOrphanAllDoneBoneMissedCitedEst}{8.8\%}
\newcommand{\ExDetMainOrphanDefinitionDoneBoneRepairEst}{100.0\%}
\newcommand{\ExDetMainOrphanDefinitionDoneBzeroUpdateEst}{56.5\%}
\newcommand{\ExDetMainOrphanDefinitionDzeroBoneMissedCitedEst}{28.9\%}
\newcommand{\ExDetMainOrphanDefinitionDzeroBoneRepairEst}{11.8\%}
\newcommand{\ExDetMainOrphanDefinitionDzeroBzeroUpdateEst}{4.2\%}
\newcommand{\ExDetMainOrphanOtherDoneBoneRepairEst}{67.5\%}
\newcommand{\ExDetMainOrphanOtherDoneBzeroUpdateEst}{10.9\%}
\newcommand{\ExDetMainOrphanOtherDzeroBoneRepairEst}{68.1\%}
\newcommand{\ExDetMainOrphanOtherDzeroBzeroUpdateEst}{10.0\%}
\newcommand{\ExDetMainOrphanStabilityItems}{1,683}
\newcommand{\ExDetMainOrphanUpdateItems}{1,767}
\newcommand{\ExDetMainSourceMissedMax}{0.6\%}
\newcommand{\ExDetMainSrcAllDoneBoneMissedCitedEst}{16.9\%}
\newcommand{\ExDetMainSrcDefinitionDoneBoneRepairCitedEst}{99.6\%}
\newcommand{\ExDetMainSrcDefinitionDoneBoneRepairEst}{99.7\%}
\newcommand{\ExDetMainSrcDefinitionDoneBzeroUpdateEst}{65.8\%}
\newcommand{\ExDetMainSrcDefinitionDzeroBoneMissedCitedEst}{28.5\%}
\newcommand{\ExDetMainSrcDefinitionDzeroBoneRepairEst}{16.3\%}
\newcommand{\ExDetMainSrcDefinitionDzeroBzeroUpdateEst}{10.2\%}
\newcommand{\ExDetMainSrcOtherDoneBoneRepairEst}{90.7\%}
\newcommand{\ExDetMainSrcOtherDoneBzeroUpdateEst}{85.9\%}
\newcommand{\ExDetMainSrcOtherDzeroBoneRepairEst}{89.1\%}
\newcommand{\ExDetMainSrcOtherDzeroBzeroUpdateEst}{83.7\%}
\newcommand{\ExDetMainSrcStabilityItems}{1,168}
\newcommand{\ExDetMainSrcUpdateItems}{1,168}
\newcommand{\ExDetRepDefStabilityItems}{575}
\newcommand{\ExDetRepDefUpdateItems}{575}
\newcommand{\ExDetRepOrphanStabilityItems}{1,671}
\newcommand{\ExDetRepOrphanUpdateItems}{1,739}
\newcommand{\ExDetRepSourceMissedMax}{1.0\%}
\newcommand{\ExDetRepSrcStabilityItems}{1,170}
\newcommand{\ExDetRepSrcUpdateItems}{1,170}
\newcommand{\ExDetWriterDefinitionBDzeroMax}{$+$38.8\,pp}
\newcommand{\ExDetWriterDefinitionBDzeroMin}{$+$30.3\,pp}
\newcommand{\ExDetWriterOpusDefDefinitionBDzero}{$+$38.8\,pp [$+$30.1, $+$46.6]}
\newcommand{\ExDetWriterOpusOrphanDefinitionBDzero}{$+$32.5\,pp [$+$23.8, $+$39.8]}
\newcommand{\ExDetWriterOpusSrcDefinitionBDzero}{$+$30.3\,pp [$+$21.3, $+$37.6]}
\newcommand{\ExDetWriterSonnetDefDefinitionBDzero}{$+$38.8\,pp [$+$31.6, $+$46.1]}
\newcommand{\ExDetWriterSonnetOrphanDefinitionBDzero}{$+$32.5\,pp [$+$25.9, $+$37.8]}
\newcommand{\ExDetWriterSonnetSrcDefinitionBDzero}{$+$32.7\,pp [$+$24.5, $+$39.9]}
\newcommand{\ExHolmMainDraws}{10,000}
\newcommand{\ExHolmMainFamily}{30}
\newcommand{\ExHolmMainSurvive}{23}
\newcommand{\ExHolmRepSurvive}{24}
\newcommand{\ExIntMainDefAllMOneCitedEst}{20.8\%}
\newcommand{\ExIntMainDefDefinitionMOneCitedEst}{14.0\%}
\newcommand{\ExIntMainDefOtherMOneCitedEst}{36.3\%}
\newcommand{\ExIntMainOrphanAllMOneCitedEst}{7.7\%}
\newcommand{\ExIntMainSrcAllMOneCitedEst}{14.9\%}
\newcommand{\ExIntMainSrcDefinitionMOneCitedEst}{23.4\%}
\newcommand{\ExIntMainSrcOtherMOneCitedEst}{11.5\%}
\newcommand{\ExMissClassMainDefDefinitionAAndBEst}{9.3\%}
\newcommand{\ExMissClassMainOrphanDefinitionAAndBEst}{10.8\%}
\newcommand{\ExMissClassMainSrcDefinitionAAndBEst}{6.4\%}
\newcommand{\ExMissedMainSrcDoneDefinitionBEffect}{$+$6.6\,pp [$+$0.6, $+$12.2]}
\newcommand{\ExMissedMainSrcDoneOtherACitedEst}{1.0\%}
\newcommand{\ExMissedMainSrcDoneOtherAZeroOnMissedEst}{59.4\%}
\newcommand{\ExMissedMainSrcDoneOtherBEffect}{$-$3.4\,pp [$-$7.4, $+$0.7]}
\newcommand{\ExMixMainDefDefinitionShare}{69.4\%}
\newcommand{\ExMixMainDefOtherShare}{30.6\%}
\newcommand{\ExMixMainOrphanDefinitionShare}{17.4\%}
\newcommand{\ExMixMainOrphanOtherShare}{82.6\%}
\newcommand{\ExMixMainSrcDefinitionShare}{28.5\%}
\newcommand{\ExMixMainSrcOtherShare}{71.5\%}
\newcommand{\ExPerDefLooMainDefBxDMax}{$+$47.3\,pp}
\newcommand{\ExPerDefLooMainDefBxDMin}{$+$34.6\,pp}
\newcommand{\ExPerDefLooMainSrcBxDMax}{$+$33.8\,pp}
\newcommand{\ExPerDefLooMainSrcBxDMin}{$+$16.5\,pp}
\newcommand{\ExPerDefMainSrcActiveProjectBxDEst}{$-$20.0\,pp}
\newcommand{\ExPerDefMainSrcImportantContactBxDEst}{$+$75.4\,pp}
\newcommand{\ExRThreeDelCOneDefinition}{$+$42.5\,pp [$+$39.0, $+$45.5]}
\newcommand{\ExRThreeDelCOneOther}{$-$15.1\,pp [$-$17.3, $-$12.5]}
\newcommand{\ExRThreeDelCOneOtherEst}{$-$15.1\,pp}
\newcommand{\ExRThreeDelDmMinusDoneD}{$+$1.8\,pp [$-$2.3, $+$5.3]}
\newcommand{\ExRThreeDelDmOkShare}{34.9\% [31.2, 38.6]}
\newcommand{\ExRThreeDelDoneDCount}{192}
\newcommand{\ExRThreeDelDoneDNoValueCount}{162}
\newcommand{\ExRThreeDelDzeroOkShare}{66.7\% [61.8, 71.5]}
\newcommand{\ExRThreeDelFailedSlots}{78}
\newcommand{\ExRThreeDelItems}{1,365}
\newcommand{\ExRThreeDelLossShareVsDzeroA}{68.9\% [61.1, 76.6]}
\newcommand{\ExRThreeDelLossShareVsDzeroAReplacement}{67.7\% [60.0, 75.3]}
\newcommand{\ExRThreeDelLossShareVsDzeroAReplacementEst}{67.7\%}
\newcommand{\ExRThreeDelLossShareVsDzeroD}{30.5\% [22.9, 38.2]}
\newcommand{\ExRThreeDelLossVsDzero}{35.8\,pp [31.2, 39.9]}
\newcommand{\ExRThreeDelProgContactDeletedItems}{936}
\newcommand{\ExRThreeDelProgContactDeletedWrong}{28.5\% [24.2, 32.9]}
\newcommand{\ExRThreeDelProgContactLiveItems}{789}
\newcommand{\ExRThreeDelProgContactLiveWrong}{26.7\% [21.4, 32.6]}
\newcommand{\ExRThreeDelProgSeriesDeletedItems}{153}
\newcommand{\ExRThreeDelProgSeriesDeletedWrong}{0.0\%}
\newcommand{\ExRThreeDelProgSeriesLiveItems}{108}
\newcommand{\ExRThreeDelProgSeriesLiveWrong}{11.1\% [2.6, 21.7]}
\newcommand{\ExRThreeDelProgTaskDeletedItems}{225}
\newcommand{\ExRThreeDelProgTaskDeletedWrong}{89.8\% [81.9, 96.8]}
\newcommand{\ExRThreeDelProgTaskLiveItems}{180}
\newcommand{\ExRThreeDelProgTaskLiveWrong}{83.3\% [71.8, 93.2]}
\newcommand{\ExRThreeDelSlots}{960}
\newcommand{\ExRThreeDelStableAll}{$-$2.2\,pp [$-$4.0, $-$0.2]}
\newcommand{\ExRThreeDelStableLossShareC}{68.8\% [57.9, 81.2]}
\newcommand{\ExRThreeDelTzeroTaskAgree}{15.0\%}
\newcommand{\ExRThreeDelTzeroWorkingHoursItems}{20}
\newcommand{\ExRThreeDelValueContact}{197}
\newcommand{\ExRThreeDelValueContactFirstListed}{122}
\newcommand{\ExRThreeDelValueOtherRecord}{386}
\newcommand{\ExRThreeDelValueStored}{394}
\newcommand{\ExRThreeDelValueStoredShare}{28.9\% [24.5, 32.7]}
\newcommand{\ExRThreeDelValueTask}{189}
\newcommand{\ExRThreeDelValueTaskLastListed}{189}
\newcommand{\ExRThreeNatOrphanDriftAtPooledMixEst}{$+$5.4\,pp}
\newcommand{\ExRThreeNatOrphanShiftAtPooledMixEst}{$+$5.7\,pp}
\newcommand{\ExRThreeNatOrphanShiftContributionDefinition}{$+$1.4\,pp [$+$0.3, $+$2.8]}
\newcommand{\ExRThreeNatOrphanShiftContributionOther}{$-$1.5\,pp [$-$5.7, $+$2.4]}
\newcommand{\ExRThreeNatOrphanShiftDefinitionShare}{3.5\% [1.6, 5.5]}
\newcommand{\ExRThreeNatSrcDriftAtPooledMix}{$+$9.1\,pp [$+$3.8, $+$15.4]}
\newcommand{\ExRThreeNatSrcShiftAtPooledMix}{$+$6.2\,pp [$+$3.3, $+$9.2]}
\newcommand{\ExRThreeNatSrcShiftContributionDefinition}{$+$3.4\,pp [$+$1.5, $+$5.6]}
\newcommand{\ExRThreeNatSrcShiftContributionOther}{$-$0.9\,pp [$-$3.8, $+$1.8]}
\newcommand{\ExRThreeNatSrcShiftDefinitionShare}{14.1\% [9.5, 19.0]}
\newcommand{\ExRThreeSRDmOrphanOtherReaderPart}{$-$12.4\,pp [$-$16.2, $-$7.9]}
\newcommand{\ExRThreeSRDmOrphanOtherStorePart}{$-$20.4\,pp [$-$24.0, $-$16.7]}
\newcommand{\ExRThreeSRDmSrcDefinitionDoneBoneRStoreEst}{71.8\%}
\newcommand{\ExRThreeSRMainDefDefinitionJsonBoneUEst}{54.4\%}
\newcommand{\ExRThreeSRMainDefDefinitionReaderPart}{0.0\,pp}
\newcommand{\ExRThreeSRMainDefOtherDoneBoneRStoreEst}{96.5\%}
\newcommand{\ExRThreeSRMainDefOtherDoneBoneUAbstain}{7.0\% [0.0, 20.0]}
\newcommand{\ExRThreeSRMainDefOtherDoneBoneUAnswerEst}{100.0\%}
\newcommand{\ExRThreeSRMainDefOtherDzeroBoneRStoreEst}{73.7\%}
\newcommand{\ExRThreeSRMainDefOtherReaderPart}{$-$21.1\,pp [$-$31.7, $-$10.8]}
\newcommand{\ExRThreeSRMainOrphanDefinitionJsonBoneUEst}{67.0\%}
\newcommand{\ExRThreeSRMainOrphanDefinitionReaderPart}{$-$0.3\,pp [$-$1.0, 0.0]}
\newcommand{\ExRThreeSRMainOrphanOtherDoneBoneUAbstain}{81.7\% [77.2, 85.4]}
\newcommand{\ExRThreeSRMainShiftsOtherDoneBoneLostAbstain}{281}
\newcommand{\ExRThreeSRMainSrcDefinitionDoneBoneUEst}{99.7\%}
\newcommand{\ExRThreeSRMainSrcDefinitionDzeroBoneUEst}{98.1\%}
\newcommand{\ExRThreeSRMainSrcDefinitionJsonBoneUEst}{70.1\%}
\newcommand{\ExRThreeSRMainSrcDefinitionReaderPart}{0.0\,pp [$-$0.9, $+$0.9]}
\newcommand{\ExRThreeViewDefDefinitionUpdateVsDm}{$-$6.0\,pp [$-$8.3, $-$4.0]}
\newcommand{\ExRThreeViewDefOtherStabilityVsDmEst}{$-$28.0\,pp}
\newcommand{\ExRThreeViewDefOtherUpdateVsDmEst}{$-$12.0\,pp}
\newcommand{\ExRThreeViewOrphanOtherStabilityVsDmEst}{$-$23.6\,pp}
\newcommand{\ExRThreeViewPooledAllStabilityVsDm}{$-$21.3\,pp [$-$28.0, $-$14.8]}
\newcommand{\ExRThreeViewPooledAllUpdateVsDm}{$-$6.6\,pp [$-$9.2, $-$4.2]}
\newcommand{\ExRThreeViewPooledDefinitionStabilityItems}{100}
\newcommand{\ExRThreeViewPooledDefinitionStabilityVsDm}{$+$43.0\,pp [$+$33.3, $+$55.6]}
\newcommand{\ExRThreeViewServePooledVsDm}{$+$246 [$+$213, $+$289]}
\newcommand{\ExRThreeViewSrcOtherUpdateVsDmEst}{$-$22.2\,pp}
\newcommand{\ExRThreeViewTypeWorkingHoursStabilityItems}{45}
\newcommand{\ExRThreeViewWithoutWorkingHoursDefinitionStabilityVsDm}{$-$3.6\,pp [$-$10.3, 0.0]}
\newcommand{\ExRThreeViewWriteDmSrcIngest}{319.7k}
\newcommand{\ExRThreeViewWriteDmSrcMaintain}{1.3M}
\newcommand{\ExRThreeViewWriteViewSrcIngest}{56.7k}
\newcommand{\ExRevCauseDmDefDriftAllBxD}{$+$20.5\,pp [$+$11.8, $+$29.3]}
\newcommand{\ExRevCauseDmDefDriftDefinitionBxD}{$+$38.2\,pp [$+$25.3, $+$49.7]}
\newcommand{\ExRevCauseDmDefDriftOtherBxD}{$+$3.5\,pp [$-$3.3, $+$10.8]}
\newcommand{\ExRevCauseDmDefShiftAllBxD}{$+$32.4\,pp [$+$23.4, $+$41.5]}
\newcommand{\ExRevCauseDmDefShiftDefinitionBxD}{$+$32.4\,pp [$+$23.4, $+$41.5]}
\newcommand{\ExRevCauseDmOrphanDriftAllBxD}{$+$23.4\,pp [$+$17.6, $+$29.1]}
\newcommand{\ExRevCauseDmOrphanDriftDefinitionBxD}{$+$35.0\,pp [$+$29.0, $+$40.3]}
\newcommand{\ExRevCauseDmOrphanDriftOtherBxD}{$+$1.5\,pp [$-$5.6, $+$10.2]}
\newcommand{\ExRevCauseDmOrphanShiftAllBxD}{$-$33.6\,pp [$-$38.8, $-$28.5]}
\newcommand{\ExRevCauseDmOrphanShiftDefinitionBxD}{$+$39.6\,pp [$+$15.0, $+$60.0]}
\newcommand{\ExRevCauseDmOrphanShiftOtherBxD}{$-$36.3\,pp [$-$41.6, $-$30.7]}
\newcommand{\ExRevCauseDmSrcDriftAllBxD}{$+$18.1\,pp [$+$11.8, $+$24.6]}
\newcommand{\ExRevCauseDmSrcDriftDefinitionBxD}{$+$29.3\,pp [$+$20.7, $+$37.2]}
\newcommand{\ExRevCauseDmSrcDriftOtherBxD}{$+$2.5\,pp [$-$2.8, $+$8.7]}
\newcommand{\ExRevCauseDmSrcShiftAllBxD}{$+$7.5\,pp [$+$4.3, $+$10.4]}
\newcommand{\ExRevCauseDmSrcShiftDefinitionBxD}{$+$23.4\,pp [$+$11.4, $+$34.4]}
\newcommand{\ExRevCauseDmSrcShiftOtherBxD}{$+$4.9\,pp [$+$0.6, $+$9.2]}
\newcommand{\ExRevCauseMainDefDriftAllBxD}{$+$18.8\,pp [$+$12.4, $+$25.1]}
\newcommand{\ExRevCauseMainDefDriftAllItems}{336}
\newcommand{\ExRevCauseMainDefDriftDefinitionBxD}{$+$38.8\,pp [$+$29.3, $+$48.4]}
\newcommand{\ExRevCauseMainDefDriftDefinitionItems}{165}
\newcommand{\ExRevCauseMainDefDriftOtherBxD}{$-$0.6\,pp [$-$4.0, $+$2.7]}
\newcommand{\ExRevCauseMainDefDriftOtherItems}{171}
\newcommand{\ExRevCauseMainDefPooledAllBxD}{$+$26.7\,pp [$+$21.2, $+$31.8]}
\newcommand{\ExRevCauseMainDefPooledDefinitionBxD}{$+$38.8\,pp [$+$32.0, $+$44.8]}
\newcommand{\ExRevCauseMainDefShiftAllBxD}{$+$38.7\,pp [$+$30.4, $+$46.9]}
\newcommand{\ExRevCauseMainDefShiftAllItems}{222}
\newcommand{\ExRevCauseMainDefShiftDefinitionBxD}{$+$38.7\,pp [$+$30.4, $+$46.9]}
\newcommand{\ExRevCauseMainDefShiftDefinitionItems}{222}
\newcommand{\ExRevCauseMainOrphanDriftAllBxD}{$+$22.4\,pp [$+$17.5, $+$27.2]}
\newcommand{\ExRevCauseMainOrphanDriftAllItems}{397}
\newcommand{\ExRevCauseMainOrphanDriftDefinitionBxD}{$+$34.6\,pp [$+$29.4, $+$39.3]}
\newcommand{\ExRevCauseMainOrphanDriftDefinitionItems}{260}
\newcommand{\ExRevCauseMainOrphanDriftOtherBxD}{$-$0.7\,pp [$-$4.4, $+$3.0]}
\newcommand{\ExRevCauseMainOrphanDriftOtherItems}{137}
\newcommand{\ExRevCauseMainOrphanPooledAllBxD}{$+$5.0\,pp [$+$1.7, $+$8.1]}
\newcommand{\ExRevCauseMainOrphanShiftAllBxD}{$-$0.1\,pp [$-$4.5, $+$4.3]}
\newcommand{\ExRevCauseMainOrphanShiftAllItems}{1,370}
\newcommand{\ExRevCauseMainOrphanShiftDefinitionBxD}{$+$39.6\,pp [$+$15.0, $+$60.0]}
\newcommand{\ExRevCauseMainOrphanShiftDefinitionItems}{48}
\newcommand{\ExRevCauseMainOrphanShiftOtherBxD}{$-$1.5\,pp [$-$5.9, $+$2.5]}
\newcommand{\ExRevCauseMainOrphanShiftOtherItems}{1,322}
\newcommand{\ExRevCauseMainSrcDriftAllBxD}{$+$17.3\,pp [$+$11.2, $+$24.0]}
\newcommand{\ExRevCauseMainSrcDriftAllItems}{381}
\newcommand{\ExRevCauseMainSrcDriftDefinitionBxD}{$+$28.8\,pp [$+$20.5, $+$36.5]}
\newcommand{\ExRevCauseMainSrcDriftDefinitionItems}{222}
\newcommand{\ExRevCauseMainSrcDriftOtherBxD}{$+$1.3\,pp [$-$5.4, $+$8.3]}
\newcommand{\ExRevCauseMainSrcDriftOtherItems}{159}
\newcommand{\ExRevCauseMainSrcPooledAllBxD}{$+$7.4\,pp [$+$5.0, $+$9.8]}
\newcommand{\ExRevCauseMainSrcShiftAllBxD}{$+$2.5\,pp [$+$0.5, $+$4.4]}
\newcommand{\ExRevCauseMainSrcShiftAllItems}{787}
\newcommand{\ExRevCauseMainSrcShiftDefinitionBxD}{$+$24.3\,pp [$+$12.5, $+$35.7]}
\newcommand{\ExRevCauseMainSrcShiftDefinitionItems}{111}
\newcommand{\ExRevCauseMainSrcShiftOtherBxD}{$-$1.0\,pp [$-$4.7, $+$2.0]}
\newcommand{\ExRevCauseMainSrcShiftOtherItems}{676}
\newcommand{\ExRevCitedDefDriftAllItems}{336}
\newcommand{\ExRevCitedDefDriftAllMissShare}{34.5\% [27.0, 41.9]}
\newcommand{\ExRevCitedDefDriftAllMissed}{116}
\newcommand{\ExRevCitedDefDriftDefinitionMissShare}{32.7\% [21.8, 44.3]}
\newcommand{\ExRevCitedDefDriftOtherMissShare}{36.3\% [24.3, 50.0]}
\newcommand{\ExRevCitedDefPooledAllItems}{558}
\newcommand{\ExRevCitedDefPooledAllMissShare}{20.8\% [16.3, 24.9]}
\newcommand{\ExRevCitedDefPooledAllMissed}{116}
\newcommand{\ExRevCitedDefPooledDefinitionItems}{387}
\newcommand{\ExRevCitedDefPooledDefinitionMissShare}{14.0\% [9.2, 19.1]}
\newcommand{\ExRevCitedDefPooledDefinitionMissed}{54}
\newcommand{\ExRevCitedDefPooledOtherItems}{171}
\newcommand{\ExRevCitedDefPooledOtherMissShare}{36.3\% [24.3, 50.0]}
\newcommand{\ExRevCitedDefPooledOtherMissed}{62}
\newcommand{\ExRevCitedDefShiftAllMissShareEst}{0.0\%}
\newcommand{\ExRevCitedDefShiftDefinitionItems}{222}
\newcommand{\ExRevCitedDefShiftDefinitionMissShareEst}{0.0\%}
\newcommand{\ExRevCitedOrphanDriftAllMissShare}{33.0\% [27.8, 37.7]}
\newcommand{\ExRevCitedOrphanDriftDefinitionMissShare}{26.5\% [21.7, 31.7]}
\newcommand{\ExRevCitedOrphanDriftOtherMissShare}{45.3\% [31.5, 59.6]}
\newcommand{\ExRevCitedOrphanPooledAllItems}{1,767}
\newcommand{\ExRevCitedOrphanPooledAllMissShare}{7.7\% [6.2, 9.2]}
\newcommand{\ExRevCitedOrphanPooledAllMissed}{136}
\newcommand{\ExRevCitedOrphanPooledDefinitionItems}{308}
\newcommand{\ExRevCitedOrphanPooledDefinitionMissShare}{24.0\% [19.4, 29.0]}
\newcommand{\ExRevCitedOrphanPooledDefinitionMissed}{74}
\newcommand{\ExRevCitedOrphanPooledOtherItems}{1,459}
\newcommand{\ExRevCitedOrphanPooledOtherMissShare}{4.2\% [2.9, 5.7]}
\newcommand{\ExRevCitedOrphanPooledOtherMissed}{62}
\newcommand{\ExRevCitedOrphanShiftAllMissShare}{0.4\% [0.0, 1.0]}
\newcommand{\ExRevCitedOrphanShiftDefinitionMissShare}{10.4\% [0.0, 27.3]}
\newcommand{\ExRevCitedOrphanShiftOtherMissShareEst}{0.0\%}
\newcommand{\ExRevCitedSrcDriftAllMissShare}{30.2\% [24.1, 36.6]}
\newcommand{\ExRevCitedSrcDriftDefinitionMissShare}{26.6\% [19.1, 34.8]}
\newcommand{\ExRevCitedSrcDriftOtherMissShare}{35.2\% [22.2, 49.6]}
\newcommand{\ExRevCitedSrcPooledAllItems}{1,168}
\newcommand{\ExRevCitedSrcPooledAllMissShare}{14.9\% [11.7, 17.9]}
\newcommand{\ExRevCitedSrcPooledAllMissed}{174}
\newcommand{\ExRevCitedSrcPooledDefinitionItems}{333}
\newcommand{\ExRevCitedSrcPooledDefinitionMissShare}{23.4\% [18.8, 28.4]}
\newcommand{\ExRevCitedSrcPooledDefinitionMissed}{78}
\newcommand{\ExRevCitedSrcPooledOtherItems}{835}
\newcommand{\ExRevCitedSrcPooledOtherMissShare}{11.5\% [7.5, 15.2]}
\newcommand{\ExRevCitedSrcPooledOtherMissed}{96}
\newcommand{\ExRevCitedSrcShiftAllMissShare}{7.5\% [4.7, 10.2]}
\newcommand{\ExRevCitedSrcShiftDefinitionMissShare}{17.1\% [7.8, 26.7]}
\newcommand{\ExRevCitedSrcShiftOtherMissShare}{5.9\% [3.3, 8.6]}
\newcommand{\ExRevClockMainDefAllClockOnly}{69}
\newcommand{\ExRevClockMainDefDefinitionDzeroBoneMissedLast}{0.3\% [0.0, 0.8]}
\newcommand{\ExRevClockMainOrphanAllClockOnly}{75}
\newcommand{\ExRevClockMainOrphanDefinitionDzeroBoneMissedLast}{0.6\% [0.0, 1.6]}
\newcommand{\ExRevClockMainShiftsDefinitionChanged}{1,028}
\newcommand{\ExRevClockMainShiftsDefinitionClockOnly}{99}
\newcommand{\ExRevClockMainShiftsDefinitionClockOnlyShare}{9.6\% [6.1, 13.2]}
\newcommand{\ExRevClockMainShiftsDefinitionClockSilent}{8}
\newcommand{\ExRevClockMainShiftsErrorsOneOne}{6}
\newcommand{\ExRevClockMainShiftsErrorsOneOneClock}{5}
\newcommand{\ExRevClockMainSrcAllClockOnly}{72}
\newcommand{\ExRevClockMainSrcDefinitionDoneBoneRepairLast}{99.7\% [99.1, 100.0]}
\newcommand{\ExRevClockMainSrcDefinitionDzeroBoneMissedLast}{0.6\% [0.0, 1.5]}
\newcommand{\ExRevClockRepShiftsErrorsOneOne}{7}
\newcommand{\ExRevClockRepShiftsErrorsOneOneClock}{7}
\newcommand{\ExRevCovActiveProjectCompleteShareEst}{100.0\%}
\newcommand{\ExRevCovActiveProjectItemTimes}{240}
\newcommand{\ExRevCovActiveProjectScopeShareEst}{100.0\%}
\newcommand{\ExRevCovCloseFriendCompleteShareEst}{0.0\%}
\newcommand{\ExRevCovCloseFriendItemTimes}{240}
\newcommand{\ExRevCovCloseFriendScopeShare}{19.8\% [18.6, 20.8]}
\newcommand{\ExRevCovImportantContactCompleteShare}{45.4\% [28.3, 62.5]}
\newcommand{\ExRevCovImportantContactItemTimes}{240}
\newcommand{\ExRevCovImportantContactScopeShare}{90.2\% [86.3, 93.9]}
\newcommand{\ExRevCovOpusDefCompleteDoneBxD}{$-$20.2\,pp [$-$25.5, $-$15.5]}
\newcommand{\ExRevCovOpusDefCompleteItemsZero}{104}
\newcommand{\ExRevCovOpusDefCompleteROne}{97.1\% [94.7, 100.0]}
\newcommand{\ExRevCovOpusDefCompleteRZero}{97.1\% [92.9, 100.0]}
\newcommand{\ExRevCovOpusDefIncompleteDoneBxD}{$+$55.7\,pp [$+$45.8, $+$66.0]}
\newcommand{\ExRevCovOpusDefIncompleteItemsZero}{97}
\newcommand{\ExRevCovOpusDefIncompleteROne}{94.8\% [90.4, 98.9]}
\newcommand{\ExRevCovOpusDefIncompleteRZero}{6.2\% [3.1, 9.1]}
\newcommand{\ExRevCovOpusOrphanCompleteDoneBxD}{$-$16.1\,pp [$-$27.3, $-$7.6]}
\newcommand{\ExRevCovOpusOrphanCompleteItemsZero}{62}
\newcommand{\ExRevCovOpusOrphanCompleteROne}{91.9\% [82.6, 98.4]}
\newcommand{\ExRevCovOpusOrphanCompleteRZero}{95.2\% [86.4, 100.0]}
\newcommand{\ExRevCovOpusOrphanIncompleteDoneBxD}{$+$62.0\,pp [$+$51.7, $+$71.6]}
\newcommand{\ExRevCovOpusOrphanIncompleteItemsZero}{92}
\newcommand{\ExRevCovOpusOrphanIncompleteROne}{94.6\% [90.1, 98.8]}
\newcommand{\ExRevCovOpusOrphanIncompleteRZero}{7.6\% [3.4, 11.5]}
\newcommand{\ExRevCovOpusSrcCompleteDoneBxD}{$-$23.6\,pp [$-$31.9, $-$15.5]}
\newcommand{\ExRevCovOpusSrcCompleteItemsZero}{72}
\newcommand{\ExRevCovOpusSrcCompleteROne}{91.7\% [83.6, 97.1]}
\newcommand{\ExRevCovOpusSrcCompleteRZero}{91.7\% [83.7, 97.1]}
\newcommand{\ExRevCovOpusSrcCompleteRZeroEst}{91.7\%}
\newcommand{\ExRevCovOpusSrcIncompleteDoneBxD}{$+$60.2\,pp [$+$48.9, $+$69.1]}
\newcommand{\ExRevCovOpusSrcIncompleteItemsZero}{93}
\newcommand{\ExRevCovOpusSrcIncompleteROne}{95.7\% [92.4, 98.9]}
\newcommand{\ExRevCovOpusSrcIncompleteRZero}{5.4\% [1.1, 10.0]}
\newcommand{\ExRevCovOverdueTaskCompleteShareEst}{100.0\%}
\newcommand{\ExRevCovOverdueTaskItemTimes}{240}
\newcommand{\ExRevCovOverdueTaskScopeShareEst}{100.0\%}
\newcommand{\ExRevCovQwenDefCompleteDoneBxD}{$+$50.3\,pp [$+$37.2, $+$60.6]}
\newcommand{\ExRevCovQwenDefCompleteItemsZero}{197}
\newcommand{\ExRevCovQwenDefCompleteROne}{99.5\% [98.3, 100.0]}
\newcommand{\ExRevCovQwenDefCompleteRZero}{12.7\% [6.0, 19.8]}
\newcommand{\ExRevCovQwenDefIncompleteDoneBxD}{$+$26.8\,pp [$+$22.0, $+$31.7]}
\newcommand{\ExRevCovQwenDefIncompleteItemsZero}{190}
\newcommand{\ExRevCovQwenDefIncompleteROneEst}{100.0\%}
\newcommand{\ExRevCovQwenDefIncompleteRZero}{0.5\% [0.0, 1.6]}
\newcommand{\ExRevCovQwenOrphanCompleteDoneBxD}{$+$39.7\,pp [$+$24.5, $+$52.2]}
\newcommand{\ExRevCovQwenOrphanCompleteItemsZero}{121}
\newcommand{\ExRevCovQwenOrphanCompleteROne}{98.3\% [95.8, 100.0]}
\newcommand{\ExRevCovQwenOrphanCompleteRZero}{27.3\% [18.6, 36.7]}
\newcommand{\ExRevCovQwenOrphanIncompleteDoneBxD}{$+$32.6\,pp [$+$27.3, $+$37.2]}
\newcommand{\ExRevCovQwenOrphanIncompleteItemsZero}{187}
\newcommand{\ExRevCovQwenOrphanIncompleteROneEst}{100.0\%}
\newcommand{\ExRevCovQwenOrphanIncompleteRZero}{1.6\% [0.0, 3.4]}
\newcommand{\ExRevCovQwenSrcCompleteDoneBxD}{$+$22.2\,pp [$+$8.8, $+$35.3]}
\newcommand{\ExRevCovQwenSrcCompleteItemsZero}{144}
\newcommand{\ExRevCovQwenSrcCompleteROne}{97.9\% [95.5, 100.0]}
\newcommand{\ExRevCovQwenSrcCompleteRZero}{36.8\% [24.8, 47.2]}
\newcommand{\ExRevCovQwenSrcCompleteRZeroEst}{36.8\%}
\newcommand{\ExRevCovQwenSrcIncompleteDoneBxD}{$+$31.2\,pp [$+$25.1, $+$36.6]}
\newcommand{\ExRevCovQwenSrcIncompleteItemsZero}{189}
\newcommand{\ExRevCovQwenSrcIncompleteROneEst}{100.0\%}
\newcommand{\ExRevCovQwenSrcIncompleteRZero}{0.5\% [0.0, 1.6]}
\newcommand{\ExRevCovSonnetDefCompleteDoneBxD}{$-$21.2\,pp [$-$31.2, $-$12.4]}
\newcommand{\ExRevCovSonnetDefCompleteItemsZero}{104}
\newcommand{\ExRevCovSonnetDefCompleteROne}{96.2\% [92.3, 100.0]}
\newcommand{\ExRevCovSonnetDefCompleteRZero}{96.2\% [91.5, 100.0]}
\newcommand{\ExRevCovSonnetDefIncompleteDoneBxD}{$+$56.7\,pp [$+$48.0, $+$65.6]}
\newcommand{\ExRevCovSonnetDefIncompleteItemsZero}{97}
\newcommand{\ExRevCovSonnetDefIncompleteROne}{99.0\% [96.8, 100.0]}
\newcommand{\ExRevCovSonnetDefIncompleteRZero}{8.2\% [4.5, 11.4]}
\newcommand{\ExRevCovSonnetOrphanCompleteDoneBxD}{$-$17.7\,pp [$-$37.9, 0.0]}
\newcommand{\ExRevCovSonnetOrphanCompleteItemsZero}{62}
\newcommand{\ExRevCovSonnetOrphanCompleteROne}{96.8\% [92.4, 100.0]}
\newcommand{\ExRevCovSonnetOrphanCompleteRZero}{93.5\% [82.8, 100.0]}
\newcommand{\ExRevCovSonnetOrphanIncompleteDoneBxD}{$+$66.3\,pp [$+$57.3, $+$74.5]}
\newcommand{\ExRevCovSonnetOrphanIncompleteItemsZero}{92}
\newcommand{\ExRevCovSonnetOrphanIncompleteROne}{98.9\% [96.6, 100.0]}
\newcommand{\ExRevCovSonnetOrphanIncompleteRZero}{8.7\% [4.7, 12.0]}
\newcommand{\ExRevCovSonnetSrcCompleteDoneBxD}{$-$41.7\,pp [$-$50.9, $-$30.2]}
\newcommand{\ExRevCovSonnetSrcCompleteItemsZero}{72}
\newcommand{\ExRevCovSonnetSrcCompleteROne}{94.4\% [89.7, 98.5]}
\newcommand{\ExRevCovSonnetSrcCompleteRZero}{94.4\% [87.2, 100.0]}
\newcommand{\ExRevCovSonnetSrcCompleteRZeroEst}{94.4\%}
\newcommand{\ExRevCovSonnetSrcIncompleteDoneBxD}{$+$61.3\,pp [$+$52.1, $+$69.2]}
\newcommand{\ExRevCovSonnetSrcIncompleteItemsZero}{93}
\newcommand{\ExRevCovSonnetSrcIncompleteROne}{98.9\% [96.7, 100.0]}
\newcommand{\ExRevCovSonnetSrcIncompleteRZero}{7.5\% [4.3, 10.2]}
\newcommand{\ExRevCovVipSenderCompleteShareEst}{0.0\%}
\newcommand{\ExRevCovVipSenderItemTimes}{240}
\newcommand{\ExRevCovVipSenderScopeShare}{23.4\% [21.9, 25.0]}
\newcommand{\ExRevCovWorkingHoursCompleteShareEst}{100.0\%}
\newcommand{\ExRevCovWorkingHoursItemTimes}{240}
\newcommand{\ExRevCovWorkingHoursScopeShareEst}{100.0\%}
\newcommand{\ExRevForcedDoneDefDefinitionBzeroEst}{37.0\%}
\newcommand{\ExRevForcedDoneDefDefinitionCitedEst}{5.6\%}
\newcommand{\ExRevForcedDoneDefDefinitionForcedEst}{100.0\%}
\newcommand{\ExRevForcedDoneDefDefinitionGain}{$+$94.4\,pp [$+$84.5, $+$100.0]}
\newcommand{\ExRevForcedDoneDefOtherBzeroEst}{41.9\%}
\newcommand{\ExRevForcedDoneDefOtherCitedEst}{0.0\%}
\newcommand{\ExRevForcedDoneDefOtherForcedEst}{40.3\%}
\newcommand{\ExRevForcedDoneDefOtherGain}{$+$40.3\,pp [$+$25.4, $+$61.4]}
\newcommand{\ExRevForcedDoneOrphanDefinitionBzeroEst}{33.8\%}
\newcommand{\ExRevForcedDoneOrphanDefinitionCitedEst}{4.1\%}
\newcommand{\ExRevForcedDoneOrphanDefinitionForcedEst}{98.6\%}
\newcommand{\ExRevForcedDoneOrphanDefinitionGain}{$+$94.6\,pp [$+$86.7, $+$100.0]}
\newcommand{\ExRevForcedDoneOrphanOtherBzeroEst}{43.5\%}
\newcommand{\ExRevForcedDoneOrphanOtherCitedEst}{1.6\%}
\newcommand{\ExRevForcedDoneOrphanOtherForcedEst}{41.9\%}
\newcommand{\ExRevForcedDoneOrphanOtherGain}{$+$40.3\,pp [$+$24.4, $+$61.8]}
\newcommand{\ExRevForcedDoneSrcDefinitionBzeroEst}{32.1\%}
\newcommand{\ExRevForcedDoneSrcDefinitionCitedEst}{3.8\%}
\newcommand{\ExRevForcedDoneSrcDefinitionForcedEst}{98.7\%}
\newcommand{\ExRevForcedDoneSrcDefinitionGain}{$+$94.9\,pp [$+$87.3, $+$100.0]}
\newcommand{\ExRevForcedDoneSrcDefinitionGainEst}{$+$94.9\,pp}
\newcommand{\ExRevForcedDoneSrcOtherBzeroEst}{59.4\%}
\newcommand{\ExRevForcedDoneSrcOtherCitedEst}{1.0\%}
\newcommand{\ExRevForcedDoneSrcOtherForcedEst}{62.5\%}
\newcommand{\ExRevForcedDoneSrcOtherGain}{$+$61.5\,pp [$+$46.1, $+$78.7]}
\newcommand{\ExRevForcedDzeroDefDefinitionGain}{$+$1.9\,pp [0.0, $+$6.2]}
\newcommand{\ExRevForcedDzeroDefOtherGain}{$+$43.5\,pp [$+$26.2, $+$65.4]}
\newcommand{\ExRevForcedDzeroOrphanDefinitionGain}{$+$2.7\,pp [0.0, $+$6.4]}
\newcommand{\ExRevForcedDzeroOrphanOtherGain}{$+$41.9\,pp [$+$25.9, $+$63.4]}
\newcommand{\ExRevForcedDzeroSrcDefinitionGain}{$+$2.6\,pp [0.0, $+$6.2]}
\newcommand{\ExRevForcedDzeroSrcDefinitionGainEst}{$+$2.6\,pp}
\newcommand{\ExRevForcedDzeroSrcOtherBzero}{60.4\% [48.5, 75.0]}
\newcommand{\ExRevForcedDzeroSrcOtherGain}{$+$63.5\,pp [$+$48.5, $+$80.7]}
\newcommand{\ExRevFullActiveProjectRederived}{76}
\newcommand{\ExRevFullActiveProjectSameRequestShareEst}{100.0\%}
\newcommand{\ExRevFullActiveProjectTruncatedShareEst}{0.0\%}
\newcommand{\ExRevFullAllRederived}{472}
\newcommand{\ExRevFullAllSameRequestShare}{33.9\% [31.0, 37.1]}
\newcommand{\ExRevFullAllTruncatedShare}{33.9\% [33.3, 34.8]}
\newcommand{\ExRevFullCloseFriendRederived}{80}
\newcommand{\ExRevFullCloseFriendSameRequestShareEst}{0.0\%}
\newcommand{\ExRevFullCloseFriendTruncatedShareEst}{100.0\%}
\newcommand{\ExRevFullImportantContactRederived}{80}
\newcommand{\ExRevFullImportantContactSameRequestShare}{10.0\% [0.0, 25.0]}
\newcommand{\ExRevFullImportantContactTruncatedShareEst}{0.0\%}
\newcommand{\ExRevFullOverdueTaskRederived}{76}
\newcommand{\ExRevFullOverdueTaskSameRequestShareEst}{100.0\%}
\newcommand{\ExRevFullOverdueTaskTruncatedShareEst}{0.0\%}
\newcommand{\ExRevFullVipSenderRederived}{80}
\newcommand{\ExRevFullVipSenderSameRequestShareEst}{0.0\%}
\newcommand{\ExRevFullVipSenderTruncatedShareEst}{100.0\%}
\newcommand{\ExRevFullWorkingHoursRederived}{80}
\newcommand{\ExRevFullWorkingHoursSameRequestShareEst}{0.0\%}
\newcommand{\ExRevFullWorkingHoursTruncatedShareEst}{0.0\%}
\newcommand{\ExRevRefreshDmDefDefinitionBoneMinusRefresh}{$+$2.8\,pp [$+$1.0, $+$5.0]}
\newcommand{\ExRevRefreshDmDefOtherBoneMinusRefresh}{$+$0.6\,pp [$-$1.4, $+$2.5]}
\newcommand{\ExRevRefreshDmOrphanDefinitionBoneMinusRefresh}{$+$3.6\,pp [$+$2.0, $+$5.2]}
\newcommand{\ExRevRefreshDmOrphanOtherBoneMinusRefresh}{$+$0.8\,pp [$-$3.0, $+$4.2]}
\newcommand{\ExRevRefreshDmSrcDefinitionBoneMinusRefresh}{$+$2.1\,pp [$+$0.8, $+$3.8]}
\newcommand{\ExRevRefreshDmSrcOtherBoneMinusRefresh}{$+$1.2\,pp [0.0, $+$2.5]}
\newcommand{\ExRevRefreshDoneDefDefinitionBoneMinusRefresh}{$+$0.3\,pp [0.0, $+$0.8]}
\newcommand{\ExRevRefreshDoneDefDefinitionBxDRefresh}{$+$38.5\,pp [$+$31.9, $+$44.3]}
\newcommand{\ExRevRefreshDoneDefOtherBoneMinusRefresh}{0.0\,pp}
\newcommand{\ExRevRefreshDoneOrphanDefinitionBoneMinusRefresh}{0.0\,pp}
\newcommand{\ExRevRefreshDoneOrphanOtherBoneMinusRefresh}{$+$9.9\,pp [$+$4.9, $+$14.6]}
\newcommand{\ExRevRefreshDoneOrphanOtherBoneMinusRefreshEst}{$+$9.9\,pp}
\newcommand{\ExRevRefreshDoneSrcDefinitionBoneMinusRefresh}{0.0\,pp}
\newcommand{\ExRevRefreshDoneSrcOtherBoneMinusRefresh}{$+$0.7\,pp [0.0, $+$1.6]}
\newcommand{\ExRevRefreshDzeroDefDefinitionBoneMinusRefresh}{0.0\,pp}
\newcommand{\ExRevRefreshDzeroDefOtherBoneMinusRefresh}{$+$0.6\,pp [0.0, $+$2.0]}
\newcommand{\ExRevRefreshDzeroOrphanDefinitionBoneMinusRefresh}{0.0\,pp [$-$0.9, $+$1.0]}
\newcommand{\ExRevRefreshDzeroOrphanOtherBoneMinusRefresh}{$+$4.0\,pp [$+$1.5, $+$6.7]}
\newcommand{\ExRevRefreshDzeroSrcDefinitionBoneMinusRefresh}{$-$0.9\,pp [$-$3.6, $+$0.9]}
\newcommand{\ExRevRefreshDzeroSrcOtherBoneMinusRefresh}{$+$1.4\,pp [0.0, $+$3.2]}
\newcommand{\ExRevSizeHeldoutAllColumnsNoBodyFit}{66.7\%}
\newcommand{\ExRevSizeMainActiveProjectColumnsNoBodyFit}{100.0\%}
\newcommand{\ExRevSizeMainActiveProjectColumnsNoBodyMax}{2,327}
\newcommand{\ExRevSizeMainActiveProjectColumnsNoBodyMedian}{1,870}
\newcommand{\ExRevSizeMainCloseFriendColumnsNoBodyFit}{0.0\%}
\newcommand{\ExRevSizeMainCloseFriendColumnsNoBodyMax}{44,109}
\newcommand{\ExRevSizeMainCloseFriendColumnsNoBodyMedian}{37,314}
\newcommand{\ExRevSizeMainImportantContactColumnsNoBodyFit}{100.0\%}
\newcommand{\ExRevSizeMainImportantContactColumnsNoBodyMax}{15,705}
\newcommand{\ExRevSizeMainImportantContactColumnsNoBodyMedian}{12,477}
\newcommand{\ExRevSizeMainOverdueTaskColumnsNoBodyFit}{100.0\%}
\newcommand{\ExRevSizeMainOverdueTaskColumnsNoBodyMax}{2,327}
\newcommand{\ExRevSizeMainOverdueTaskColumnsNoBodyMedian}{1,870}
\newcommand{\ExRevSizeMainVipSenderColumnsNoBodyFit}{0.0\%}
\newcommand{\ExRevSizeMainVipSenderColumnsNoBodyMax}{47,326}
\newcommand{\ExRevSizeMainVipSenderColumnsNoBodyMedian}{38,532}
\newcommand{\ExRevSizeMainWorkingHoursColumnsNoBodyFit}{100.0\%}
\newcommand{\ExRevSizeMainWorkingHoursColumnsNoBodyMax}{16,663}
\newcommand{\ExRevSizeMainWorkingHoursColumnsNoBodyMedian}{13,370}
\newcommand{\ExZeroMainDefAllUpdateDiffHoldOneOne}{$-$66.3\,pp [$-$71.5, $-$61.3]}
\newcommand{\ExZeroMainOrphanAllUpdateDiffHoldOneOne}{$-$1.8\,pp [$-$5.6, $+$2.2]}
\newcommand{\ExZeroMainOrphanAllUpdateHoldOneOneEst}{73.1\%}
\newcommand{\ExZeroMainOrphanAllUpdateZeroEst}{71.3\%}
\newcommand{\ExZeroMainOrphanOtherUpdateHoldOneOneEst}{67.5\%}
\newcommand{\ExZeroMainOrphanOtherUpdateHoldZeroZeroEst}{10.0\%}
\newcommand{\ExZeroMainOrphanOtherUpdateZeroEst}{85.9\%}
\newcommand{\ExZeroMainSrcAllUpdateDiffHoldOneOne}{$-$32.0\,pp [$-$34.7, $-$29.5]}
\newcommand{\GatePathTwoMaxAbs}{0.0\,pp}
\newcommand{\KeyAuditChangedDef}{558}
\newcommand{\KeyAuditChangedOrphan}{1,767}
\newcommand{\KeyAuditChangedSrc}{1,168}
\newcommand{\KeyAuditDefShareOracleMax}{86.5\%}
\newcommand{\KeyAuditDefShareOracleMin}{77.7\%}
\newcommand{\KeyAuditFalseOracleDefAll}{4.3\%}
\newcommand{\KeyAuditFalseOracleDefDefinition}{61.3\%}
\newcommand{\KeyAuditFalseOracleOrphanAll}{7.0\%}
\newcommand{\KeyAuditFalseOracleOrphanDefinition}{49.4\%}
\newcommand{\KeyAuditFalseOracleSrcAll}{16.8\%}
\newcommand{\KeyAuditFalseSourceDefAll}{99.3\%}
\newcommand{\KeyAuditFalseSourceOrphanAll}{100.0\%}
\newcommand{\KeyAuditFalseSourceSrcAll}{99.7\%}
\newcommand{\KeyAuditRederivOracleSrcRule}{344}
\newcommand{\KeyAuditRederivSourceSrcRule}{2,930}
\newcommand{\KeyAuditWorlds}{20}
\newcommand{\MOneCaveats}{the switch to gemma-4-12B and the switch up to Qwen3.8-27B}
\newcommand{\MOneExWriterAgreePooledAurocEst}{0.827}
\newcommand{\MOneExWriterAgreePooledSlope}{0.85 [0.79, 0.94]}
\newcommand{\MOneExWriterNewAccPooledAurocDiffType}{$+$0.022 [$-$0.001, $+$0.072]}
\newcommand{\MOneExWriterNewAccPooledAurocEst}{0.849}
\newcommand{\MOneExWriterNonDefPooledAurocEst}{0.592}
\newcommand{\MOneExWriterOldAccPooledAurocEst}{0.607}
\newcommand{\MOneExWriterWorldsPooled}{20}
\newcommand{\MOneMBGemmaEst}{$-$0.1\,pp}
\newcommand{\MOneMBGptOssEst}{$-$0.7\,pp}
\newcommand{\MOneMBHolding}{6 of 8}
\newcommand{\MOneSwitches}{8}
\newcommand{\MThreeLlamaWorlds}{5}
\newcommand{\MTwoGemmaZeroMinusHoldDef}{$-$36.3\,pp [$-$40.1, $-$32.3]}
\newcommand{\MTwoGptOssZeroMinusHoldDef}{$-$30.1\,pp [$-$33.2, $-$27.2]}
\newcommand{\MTwoLlamaWorlds}{10}
\newcommand{\MTwoLlamaZeroMinusHoldDef}{$-$50.5\,pp [$-$53.3, $-$48.4]}
\newcommand{\MultExOrigBh}{36}
\newcommand{\MultExOrigContrasts}{42}
\newcommand{\MultExOrigHolm}{35}
\newcommand{\MultExOrigInterval}{36}
\newcommand{\MultExRepBh}{36}
\newcommand{\MultExRepHolm}{36}
\newcommand{\MultExRepInterval}{36}
\newcommand{\NatCogneeLabel}{(0,1)}
\newcommand{\NatExNullBIdentified}{4 of 6}
\newcommand{\NatExNullInsideOosOf}{7 of 24}
\newcommand{\NatExNullRagInsideOosOf}{12 of 24}
\newcommand{\NatHippoRagCostRatio}{0.20}
\newcommand{\NatLightRagCostCalls}{2.67}
\newcommand{\NatLightRagCostRatio}{0.75}
\newcommand{\NatMemCostTokens}{10.7k}
\newcommand{\NatMemosCostCalls}{1.00}
\newcommand{\NatNFourStabilityBOne}{86.4\%}
\newcommand{\NatNFourStabilityBZero}{73.2\%}
\newcommand{\NatNOneAgree}{5 of 5}
\newcommand{\NatNThreeInside}{12 of 36}
\newcommand{\NatNThreeShare}{33.3\%}
\newcommand{\RThreeGADefDmBxD}{$+$27.4\,pp [$+$15.2, $+$40.7]}
\newcommand{\RThreeGADefDoneBxD}{$+$33.0\,pp [$+$22.1, $+$44.9]}
\newcommand{\RThreeGADefDoneDefinitions}{18}
\newcommand{\RThreeGADefDonePositive}{17}
\newcommand{\RThreeGAOrphanDmBxD}{$+$29.0\,pp [$+$16.0, $+$45.9]}
\newcommand{\RThreeGAOrphanDoneBxD}{$+$33.1\,pp [$+$19.7, $+$48.6]}
\newcommand{\RThreeGASrcDmBxD}{$+$23.3\,pp [$+$12.9, $+$36.4]}
\newcommand{\RThreeGASrcDoneBxD}{$+$26.7\,pp [$+$16.6, $+$38.3]}
\newcommand{\RThreeGAWorlds}{10}
\newcommand{\RThreeGBDefBxD}{$+$23.1\,pp [$+$14.6, $+$31.2]}
\newcommand{\RThreeGBDefSets}{18}
\newcommand{\RThreeGBDefWriterEtaSq}{0.99}
\newcommand{\RThreeGBGemmaDefBxDEst}{$+$30.7\,pp}
\newcommand{\RThreeGBGptOssDefBxD}{$+$4.7\,pp [$-$0.4, $+$9.8]}
\newcommand{\RThreeGBLlamaDefBxD}{$+$5.4\,pp [$+$0.2, $+$10.7]}
\newcommand{\RThreeGBOpusDefBxDEst}{$+$38.8\,pp}
\newcommand{\RThreeGBOrphanBxD}{$+$19.4\,pp [$+$10.9, $+$27.6]}
\newcommand{\RThreeGBQwenDefBxDEst}{$+$20.4\,pp}
\newcommand{\RThreeGBSonnetDefBxDEst}{$+$38.6\,pp}
\newcommand{\RThreeGBSrcBxD}{$+$15.1\,pp [$+$8.0, $+$22.4]}
\newcommand{\RThreeGBWorlds}{20}
\newcommand{\RThreeGBWriters}{6}
\newcommand{\RThreeHOpusDefDefinitionDmBxD}{$+$16.4\,pp [$+$9.8, $+$23.0]}
\newcommand{\RThreeHOpusDefDefinitionDoneBxD}{$+$16.4\,pp [$+$9.8, $+$23.0]}
\newcommand{\RThreeHOpusOrphanDefinitionDoneBxD}{$+$30.5\,pp [$+$23.4, $+$38.1]}
\newcommand{\RThreeHOpusRZero}{43.0\% [32.9, 51.1]}
\newcommand{\RThreeHOpusRZeroDiff}{$+$30.3\,pp [$+$22.7, $+$37.8]}
\newcommand{\RThreeHOpusSrcDefinitionDoneBxD}{$+$23.6\,pp [$+$14.2, $+$32.7]}
\newcommand{\RThreeHOpusTokensMax}{2,118}
\newcommand{\RThreeHSonnetDefDefinitionDmBxD}{$+$12.4\,pp [$+$5.7, $+$18.5]}
\newcommand{\RThreeHSonnetDefDefinitionDoneBxD}{$+$16.4\,pp [$+$9.7, $+$22.4]}
\newcommand{\RThreeHSonnetDefDefinitionDoneBxDEst}{$+$16.4\,pp}
\newcommand{\RThreeHSonnetOrphanDefinitionDoneBxD}{$+$32.5\,pp [$+$26.9, $+$38.8]}
\newcommand{\RThreeHSonnetRZero}{45.5\% [36.0, 53.1]}
\newcommand{\RThreeHSonnetRZeroDiff}{$+$32.7\,pp [$+$26.6, $+$38.5]}
\newcommand{\RThreeHSonnetRZeroQwen}{12.7\% [3.8, 21.7]}
\newcommand{\RThreeHSonnetSrcDefinitionDoneBxD}{$+$16.4\,pp [$+$4.8, $+$28.7]}
\newcommand{\RThreeHSonnetTokensMax}{1,953}
\newcommand{\RThreeHSonnetWorlds}{10}
\newcommand{\RThreeHSonnetWrites}{944}
\newcommand{\RThreeNDefDefinitionDiff}{$+$13.5\,pp [$+$1.8, $+$25.3]}
\newcommand{\RThreeNDefDefinitionNatural}{$+$25.2\,pp [$+$16.2, $+$33.9]}
\newcommand{\RThreeNDefDefinitionNaturalMoved}{54.5\% [45.5, 63.3]}
\newcommand{\RThreeNOrphanDefinitionDiff}{$+$12.5\,pp [0.0, $+$21.4]}
\newcommand{\RThreeNOrphanDefinitionNatural}{$+$27.1\,pp [$+$3.6, $+$52.3]}
\newcommand{\RThreeNOrphanDefinitionNaturalMoved}{58.3\% [40.4, 83.8]}
\newcommand{\RThreeNOrphanOtherDiff}{$-$1.6\,pp [$-$6.7, $+$2.9]}
\newcommand{\RThreeNOrphanOtherNatural}{$+$0.1\,pp [$-$1.8, $+$1.9]}
\newcommand{\RThreeNOrphanOtherNaturalMoved}{2.6\% [1.3, 4.1]}
\newcommand{\RThreeNSrcAllWorlds}{20}
\newcommand{\RThreeNSrcDefinitionDiff}{$-$4.5\,pp [$-$17.5, $+$9.6]}
\newcommand{\RThreeNSrcDefinitionNatural}{$+$28.8\,pp [$+$15.7, $+$42.7]}
\newcommand{\RThreeNSrcDefinitionNaturalMoved}{60.4\% [47.7, 72.5]}
\newcommand{\RThreeNSrcDefinitionNaturalMovedEst}{60.4\%}
\newcommand{\RThreeNSrcOtherDiff}{$+$0.3\,pp [$-$3.4, $+$3.5]}
\newcommand{\RThreeNSrcOtherNatural}{$-$1.3\,pp [$-$3.6, $+$0.9]}
\newcommand{\RThreeNSrcOtherNaturalMoved}{1.8\% [0.0, 3.9]}
\newcommand{\RThreeQBEffectNamed}{$+$38.0\,pp [$+$32.6, $+$43.4]}
\newcommand{\RThreeQBEffectUnnamed}{$+$38.0\,pp [$+$32.6, $+$43.2]}
\newcommand{\RThreeQDmBxDUnnamed}{$+$36.2\,pp [$+$30.0, $+$42.0]}
\newcommand{\RThreeQDoneBxDUnnamed}{$+$39.5\,pp [$+$33.3, $+$45.2]}
\newcommand{\RThreeQOneDiff}{0.0\,pp [$-$1.3, $+$1.2]}
\newcommand{\RThreeQWorlds}{20}
\newcommand{\RThreeUDeltaDoneDefDefinitionNoneEffectShare}{93\%}
\newcommand{\RThreeUDeltaDoneDefDefinitionNoneUpdateBoneEst}{99.3\%}
\newcommand{\RThreeUDeltaDoneDefDefinitionYieldedShare}{64\%}
\newcommand{\RThreeUDeltaDoneDefDefinitionYieldedUpdateBzeroEst}{95.6\%}
\newcommand{\RThreeUDeltaDoneOrphanOtherNoneShare}{84\%}
\newcommand{\RThreeUDeltaDoneShiftsDefinitionMoved}{4}
\newcommand{\RThreeUDeltaDoneShiftsOtherMoved}{120}
\newcommand{\RThreeUDeltaDoneSrcDefinitionNoneEffectShare}{79\%}
\newcommand{\RThreeUDeltaDoneSrcDefinitionNoneUpdateBoneEst}{97.8\%}
\newcommand{\RThreeUDeltaDoneSrcDefinitionNoneUpdateUpsertEst}{2.2\%}
\newcommand{\RThreeUDeltaDoneSrcDefinitionYieldedShare}{73\%}
\newcommand{\RThreeUDeltaDoneSrcDefinitionYieldedUpdateBzeroEst}{90.1\%}
\newcommand{\RThreeUDmDefDefinitionR}{0.92 [0.86, 0.97]}
\newcommand{\RThreeUDmDefDefinitionUpdateBoneEst}{69.3\%}
\newcommand{\RThreeUDmDefDefinitionUpdateBzeroEst}{35.1\%}
\newcommand{\RThreeUDmDefDefinitionUpdateRefreshEst}{66.4\%}
\newcommand{\RThreeUDmDefDefinitionUpdateUpsertEst}{35.7\%}
\newcommand{\RThreeUDmDefDefinitionUpsertR}{0.02 [$-$0.07, 0.07]}
\newcommand{\RThreeUDmOrphanDefinitionR}{0.92 [0.88, 0.95]}
\newcommand{\RThreeUDmOrphanDefinitionUpdateBoneEst}{77.9\%}
\newcommand{\RThreeUDmOrphanDefinitionUpdateBzeroEst}{34.7\%}
\newcommand{\RThreeUDmOrphanDefinitionUpdateRefreshEst}{74.4\%}
\newcommand{\RThreeUDmOrphanDefinitionUpdateUpsertEst}{33.1\%}
\newcommand{\RThreeUDmOrphanDefinitionUpsertR}{$-$0.04 [$-$0.09, 0.01]}
\newcommand{\RThreeUDmSrcDefinitionR}{0.94 [0.88, 0.98]}
\newcommand{\RThreeUDmSrcDefinitionUpdateBoneEst}{73.6\%}
\newcommand{\RThreeUDmSrcDefinitionUpdateBzeroEst}{40.2\%}
\newcommand{\RThreeUDmSrcDefinitionUpdateRefreshEst}{71.5\%}
\newcommand{\RThreeUDmSrcDefinitionUpdateUpsertEst}{40.5\%}
\newcommand{\RThreeUDmSrcDefinitionUpsertR}{0.01 [$-$0.06, 0.07]}
\newcommand{\RThreeUDoneDefDefinitionR}{0.99 [0.98, 1.00]}
\newcommand{\RThreeUDoneDefDefinitionUpdateBoneEst}{99.7\%}
\newcommand{\RThreeUDoneDefDefinitionUpdateBzeroEst}{61.8\%}
\newcommand{\RThreeUDoneDefDefinitionUpdateRefreshEst}{99.5\%}
\newcommand{\RThreeUDoneDefDefinitionUpdateUpsertEst}{61.8\%}
\newcommand{\RThreeUDoneDefDefinitionUpsertR}{0.00}
\newcommand{\RThreeUDoneOrphanDefinitionR}{1.00}
\newcommand{\RThreeUDoneOrphanDefinitionUpdateBoneEst}{99.4\%}
\newcommand{\RThreeUDoneOrphanDefinitionUpdateBzeroEst}{56.5\%}
\newcommand{\RThreeUDoneOrphanDefinitionUpdateRefreshEst}{99.4\%}
\newcommand{\RThreeUDoneOrphanDefinitionUpdateUpsertEst}{56.8\%}
\newcommand{\RThreeUDoneOrphanDefinitionUpsertR}{0.01 [0.00, 0.03]}
\newcommand{\RThreeUDoneSrcDefinitionR}{1.00}
\newcommand{\RThreeUDoneSrcDefinitionUpdateBoneEst}{99.1\%}
\newcommand{\RThreeUDoneSrcDefinitionUpdateBzeroEst}{65.8\%}
\newcommand{\RThreeUDoneSrcDefinitionUpdateRefreshEst}{99.1\%}
\newcommand{\RThreeUDoneSrcDefinitionUpdateUpsertEst}{66.1\%}
\newcommand{\RThreeUDoneSrcDefinitionUpsertR}{0.01 [0.00, 0.03]}
\newcommand{\RThreeUDzeroDefDefinitionBEffect}{$-$0.8\,pp [$-$5.7, $+$4.2]}
\newcommand{\RThreeUDzeroDefDefinitionR}{---}
\newcommand{\RThreeUDzeroDefDefinitionUpdateBoneEst}{6.7\%}
\newcommand{\RThreeUDzeroDefDefinitionUpdateBzeroEst}{7.5\%}
\newcommand{\RThreeUDzeroDefDefinitionUpdateRefreshEst}{6.7\%}
\newcommand{\RThreeUDzeroDefDefinitionUpdateUpsertEst}{7.5\%}
\newcommand{\RThreeUDzeroDefDefinitionUpsertR}{---}
\newcommand{\RThreeUDzeroOrphanDefinitionR}{1.00 [0.83, 1.13]}
\newcommand{\RThreeUDzeroOrphanDefinitionUpdateBoneEst}{11.7\%}
\newcommand{\RThreeUDzeroOrphanDefinitionUpdateBzeroEst}{4.2\%}
\newcommand{\RThreeUDzeroOrphanDefinitionUpdateRefreshEst}{11.7\%}
\newcommand{\RThreeUDzeroOrphanDefinitionUpdateUpsertEst}{4.5\%}
\newcommand{\RThreeUDzeroOrphanDefinitionUpsertR}{0.04 [0.00, 0.13]}
\newcommand{\RThreeUDzeroSrcDefinitionR}{1.15 [0.86, 1.88]}
\newcommand{\RThreeUDzeroSrcDefinitionUpdateBoneEst}{16.2\%}
\newcommand{\RThreeUDzeroSrcDefinitionUpdateBzeroEst}{10.2\%}
\newcommand{\RThreeUDzeroSrcDefinitionUpdateRefreshEst}{17.1\%}
\newcommand{\RThreeUDzeroSrcDefinitionUpdateUpsertEst}{10.2\%}
\newcommand{\RThreeUDzeroSrcDefinitionUpsertR}{0.00}
\newcommand{\RThreeUFiveGemmaR}{0.07 [$-$0.04, 0.18]}
\newcommand{\RThreeUFiveGptOssR}{$-$0.03 [$-$0.08, 0.02]}
\newcommand{\RThreeUFiveLlamaR}{$-$0.16 [$-$0.44, 0.01]}
\newcommand{\RThreeUFourDiff}{$-$2.8\,pp [$-$4.3, $-$1.4]}
\newcommand{\RThreeUFourUpdateBzeroEst}{18.8\%}
\newcommand{\RThreeUFourUpdateUpsertEst}{16.0\%}
\newcommand{\RThreeUThreeR}{0.01 [$-$0.02, 0.04]}
\newcommand{\RThreeUTwoSrc}{1.15 [0.86, 1.88]}
\newcommand{\RThreeUWorlds}{20}
\newcommand{\RThreeVDefBxD}{$+$32.8\,pp [$+$24.2, $+$41.3]}
\newcommand{\RThreeVOpusAgreeDiffEst}{$-$16.0\,pp}
\newcommand{\RThreeVOpusUpdatePooledDiff}{$-$18.1\,pp [$-$20.5, $-$15.7]}
\newcommand{\RThreeVOrphanBxD}{$+$29.0\,pp [$+$19.8, $+$37.0]}
\newcommand{\RThreeVQwenAgreeDiffEst}{$-$4.4\,pp}
\newcommand{\RThreeVQwenUpdatePooledDiff}{$+$1.4\,pp [$-$0.8, $+$3.7]}
\newcommand{\RThreeVSonnetAgreeDiffEst}{$-$5.4\,pp}
\newcommand{\RThreeVSonnetUpdatePooledDiff}{$-$7.3\,pp [$-$8.3, $-$6.2]}
\newcommand{\RThreeVSrcBxD}{$+$22.4\,pp [$+$14.7, $+$29.6]}
\newcommand{\RThreeXDefDefinitionAEst}{8.9\%}
\newcommand{\RThreeXDefDefinitionDetREst}{25.0\%}
\newcommand{\RThreeXDefDefinitionRepEffect}{$+$64.4\,pp [$+$57.4, $+$71.0]}
\newcommand{\RThreeXDefDefinitionRepEffectEst}{$+$64.4\,pp}
\newcommand{\RThreeXDefOtherDetEffect}{$-$8.5\,pp [$-$15.1, $-$1.1]}
\newcommand{\RThreeXNativeMutations}{0}
\newcommand{\RThreeXOrphanDefinitionAEst}{10.7\%}
\newcommand{\RThreeXOrphanDefinitionDetEffect}{$+$3.6\,pp [0.0, $+$10.7]}
\newcommand{\RThreeXOrphanDefinitionDetREst}{30.8\%}
\newcommand{\RThreeXOrphanDefinitionRepEffect}{$+$71.4\,pp [$+$52.0, $+$88.6]}
\newcommand{\RThreeXOrphanOtherDetEffect}{$+$35.9\,pp [$+$26.1, $+$44.8]}
\newcommand{\RThreeXOrphanOtherDetEffectEst}{$+$35.9\,pp}
\newcommand{\RThreeXSrcDefinitionAEst}{12.2\%}
\newcommand{\RThreeXSrcDefinitionDetEffect}{$+$0.6\,pp [$-$3.5, $+$4.2]}
\newcommand{\RThreeXSrcDefinitionRepEffect}{$+$72.6\,pp [$+$63.1, $+$81.2]}
\newcommand{\RThreeXWorlds}{10}
\newcommand{\RThreeYAnswers}{68,474}
\newcommand{\RThreeYDefDefinitionDoneDiff}{$-$15.2\,pp [$-$19.1, $-$11.0]}
\newcommand{\RThreeYDefOtherDmBxD}{$+$8.8\,pp [$-$0.6, $+$18.8]}
\newcommand{\RThreeYDefOtherDoneBxD}{$+$9.9\,pp [$+$3.4, $+$17.3]}
\newcommand{\RThreeYOrphanOtherDmBxD}{$-$22.6\,pp [$-$26.4, $-$18.7]}
\newcommand{\RThreeYOrphanOtherDoneBxD}{$+$5.3\,pp [$+$2.3, $+$8.4]}
\newcommand{\RThreeYSRMainDefOtherReaderPart}{$-$10.5\,pp [$-$16.6, $-$4.6]}
\newcommand{\RThreeYSRMainShiftsOtherLostAbstain}{1}
\newcommand{\RThreeYSRMainShiftsOtherRightAbstain}{1,210}
\newcommand{\RThreeYSRMainSrcDefinitionDzeroBoneREst}{25.5\%}
\newcommand{\RThreeYSRMainSrcDefinitionDzeroBoneRStoreEst}{16.2\%}
\newcommand{\RThreeYSrcDefinitionDoneDiff}{$-$8.7\,pp [$-$13.4, $-$4.6]}
\newcommand{\RThreeYSrcOtherDoneBxD}{$+$1.2\,pp [$-$1.1, $+$3.4]}
\newcommand{\RankExRhoSrcDef}{0.96 [0.85, 0.98]}
\newcommand{\RankExRhoSrcDefHoldMem}{0.94 [0.75, 1.00]}
\newcommand{\RankExRhoSrcDefNative}{0.52 [$-$0.09, 0.94]}
\newcommand{\RankExRhoSrcDefNativeBase}{0.76 [0.24, 0.90]}
\newcommand{\RankExRhoSrcOrphan}{0.55 [0.43, 0.66]}
\newcommand{\RankExSrcDefHoldMemSystems}{6}
\newcommand{\RankExSrcDefNativeSystems}{6}
\newcommand{\RankExSystems}{15}
\newcommand{\RendBUpdateSrc}{$+$9.1\,pp [$+$5.3, $+$13.2]}
\newcommand{\RendBxDDef}{$+$26.8\,pp [$+$22.4, $+$31.3]}
\newcommand{\RendDRepairDef}{$+$29.2\,pp [$+$26.7, $+$32.1]}
\newcommand{\RendDRepairSrc}{$+$6.4\,pp [$+$5.2, $+$7.5]}
\newcommand{\RendDiffBxDDef}{$-$0.4\,pp [$-$1.5, $+$0.7]}
\newcommand{\RendDiffBxDSrc}{$+$0.8\,pp [$-$2.7, $+$3.7]}
\newcommand{\RendDiffDRepairDef}{$-$4.4\,pp [$-$6.0, $-$2.9]}
\newcommand{\RendDiffDRepairSrc}{$-$7.4\,pp [$-$8.5, $-$6.4]}
\newcommand{\RendDigestTokenShareMax}{99.2\%}
\newcommand{\RendDigestTokenShareMin}{99.0\%}
\newcommand{\RendFidelityMin}{100.0\%}
\newcommand{\RendRewrittenTokenShare}{0.9\%}
\newcommand{\RendTemplateBUpdateSrc}{$+$7.8\,pp [$+$4.4, $+$11.7]}
\newcommand{\RendTemplateBxDDef}{$+$27.1\,pp [$+$22.9, $+$32.1]}
\newcommand{\RendTemplateBxDSrc}{$+$6.2\,pp [$+$2.5, $+$10.3]}
\newcommand{\RendTemplateDRepairDef}{$+$33.6\,pp [$+$31.2, $+$36.3]}
\newcommand{\RendTemplateDRepairSrc}{$+$13.7\,pp [$+$12.5, $+$15.2]}
\newcommand{\RendTemplateZeroMinusHoldDef}{$-$45.1\,pp [$-$48.6, $-$42.2]}
\newcommand{\RendWorlds}{10}
\newcommand{\RendZeroMinusHoldDef}{$-$40.5\,pp [$-$45.6, $-$35.8]}
\newcommand{\RepBEffectUpdatePooledCitedDef}{$-$1.7\,pp [$-$5.9, $+$2.7]}
\newcommand{\RepBEffectUpdatePooledCitedDefOrig}{$-$1.3\,pp [$-$4.8, $+$2.0]}
\newcommand{\RepBEffectUpdatePooledCitedSrc}{$-$0.9\,pp [$-$3.9, $+$2.0]}
\newcommand{\RepBEffectUpdatePooledCitedSrcOrig}{0.0\,pp [$-$3.0, $+$3.1]}
\newcommand{\RepBxDInteractionUpdateCitedDef}{$+$6.8\,pp [$+$2.4, $+$11.6]}
\newcommand{\RepBxDInteractionUpdateCitedDefOrig}{$+$7.5\,pp [$+$1.1, $+$14.0]}
\newcommand{\RepBxDInteractionUpdateCitedSrc}{$+$0.5\,pp [$-$1.4, $+$2.7]}
\newcommand{\RepBxDInteractionUpdateCitedSrcOrig}{$-$0.9\,pp [$-$3.6, $+$1.6]}
\newcommand{\RepPathOneEstimate}{$+$8.2\,pp [$+$5.3, $+$11.2]}
\newcommand{\RepPathThreeAuroc}{0.79}
\newcommand{\RepPathThreeSlope}{1.05 (90\% lower bound 0.93)}
\newcommand{\RepPathTwoMaxAbs}{0.0\,pp}
\newcommand{\RepWorlds}{20}
\newcommand{\SynthAgreeAllEst}{60.5\%}
\newcommand{\SynthAgreeKappaVipSenderEst}{0.0\%}
\newcommand{\SynthCFourSrcEst}{0.0\,pp}
\newcommand{\SynthCOneDef}{$+$21.3\,pp [$+$19.2, $+$23.4]}
\newcommand{\SynthCOneSrc}{$+$7.2\,pp [$+$6.0, $+$8.4]}
\newcommand{\SynthCThreeDef}{$+$25.3\,pp [$+$18.8, $+$31.7]}
\newcommand{\SynthCThreeDefEst}{$+$25.3\,pp}
\newcommand{\SynthCThreeSrc}{$+$11.0\,pp [$+$8.0, $+$14.0]}
\newcommand{\SynthCThreeSrcEst}{$+$11.0\,pp}
\newcommand{\SynthCTwoDef}{$+$11.8\,pp [$+$10.5, $+$13.2]}
\newcommand{\SynthCTwoShareDefEst}{35.7\%}
\newcommand{\SynthCTwoShareSrcEst}{47.3\%}
\newcommand{\SynthCTwoSrc}{$+$6.4\,pp [$+$5.4, $+$7.5]}
\newcommand{\SynthExCOneOrphan}{$-$7.8\,pp [$-$10.0, $-$5.4]}
\newcommand{\SynthExCThreeOrphan}{$-$20.8\,pp [$-$25.1, $-$16.4]}
\newcommand{\SynthExecAllEst}{90.6\%}
\newcommand{\SynthExecKappaWorkingHoursEst}{0.0\%}
\newcommand{\SynthItemsAll}{960}
\newcommand{\SynthWorlds}{20}
\newcommand{\TestFidelity}{274,232/274,232}
\newcommand{\TestWorlds}{20}
\newcommand{\WorldEventsMax}{1,654}
\newcommand{\WorldEventsMin}{1,505}
\newcommand{\WorldFidelity}{100.0\%}
\newcommand{\WorldTokensM}{2.2}

%% file: generated/neighbour_counts.tex
\newcommand{\NbrTotal}{15}
\newcommand{\NbrGoldFirstYes}{6}
\newcommand{\NbrGoldFirstPart}{8}

\newcommand{\NbrGoldDepYes}{4}
\newcommand{\NbrGoldDepPart}{10}

\newcommand{\NbrRedefYes}{0}
\newcommand{\NbrRedefPart}{0}

\newcommand{\NbrWriterYes}{1}
\newcommand{\NbrWriterPart}{6}

\newcommand{\NbrImplYes}{0}
\newcommand{\NbrImplPart}{1}

\newcommand{\NbrBranchYes}{1}
\newcommand{\NbrBranchPart}{4}

\newcommand{\NbrJudgeNone}{3}
\newcommand{\NbrJudgeMinor}{2}

\newcommand{\MsNativeObs}{6}

\newcommand{\MsQ}{19,562}

%% file: generated/setup_facts.tex
\newcommand{\SonnetModelId}{\texttt{claude-sonnet-5-5}}
\newcommand{\OpusModelId}{\texttt{claude-opus-5-5}}
\newcommand{\ClaudeCallsFrom}{2026-10-16}
\newcommand{\ClaudeCallsTo}{2026-10-19}
\newcommand{\ClaudeTokensTotal}{278 million}
\newcommand{\ClaudeSpendUSD}{336}
\newcommand{\GpuModel}{NVIDIA A100 40GB}
\newcommand{\GpuHoursTotal}{140}
\newcommand{\GpuHoursOtherRuns}{79}
\newcommand{\GpuHoursNativeSystems}{34}

%% file: generated/round3b_numbers.tex
\newcommand{\TypedInWorlds}{10}
\newcommand{\TypedInQwenRSelBothAll}{8.2\% [3.3, 14.0]}
\newcommand{\TypedInQwenRTypedBothAll}{7.1\% [4.2, 10.3]}
\newcommand{\TypedInQwenRTypedBothAllEst}{7.1\%}
\newcommand{\TypedInQwenRAuthoredBothAll}{99.7\% [99.2, 100.0]}
\newcommand{\TypedInQwenWrites}{472}
\newcommand{\TypedInQwenUnreadableShare}{1.9\%}
\newcommand{\TypedInQwenAtLimitShare}{1.9\%}
\newcommand{\TypedInQwenTzeroCorrectAll}{26.7\% [21.7, 31.7]}
\newcommand{\TypedInSonnetRTypedBothImportantContactEst}{65.7\%}
\newcommand{\TypedInSonnetRAuthoredBothImportantContactEst}{88.1\%}
\newcommand{\TypedInSonnetRTypedBothCloseFriendEst}{97.5\%}
\newcommand{\TypedInSonnetRTypedBothVipSenderEst}{98.6\%}
\newcommand{\TypedInSonnetRSelBothAll}{50.0\% [45.2, 54.5]}
\newcommand{\TypedInSonnetRSelBothAllEst}{50.0\%}
\newcommand{\TypedInSonnetRTypedBothAll}{91.8\% [89.8, 93.7]}
\newcommand{\TypedInSonnetRTypedBothAllEst}{91.8\%}
\newcommand{\TypedInSonnetRAuthoredBothAll}{97.3\% [95.4, 98.9]}
\newcommand{\TypedInSonnetRAuthoredBothAllEst}{97.3\%}
\newcommand{\TypedInSonnetRAuthoredMinusTypedBothAll}{$+$5.5\,pp [$+$2.7, $+$8.4]}
\newcommand{\TypedInSonnetRTypedMinusSelBothAllSelComplete}{$-$0.6\,pp [$-$3.4, $+$2.4]}
\newcommand{\TypedInSonnetRSelBothAllSelIncompleteEst}{7.9\%}
\newcommand{\TypedInSonnetRTypedBothAllSelIncompleteEst}{88.9\%}
\newcommand{\TypedInSonnetUnreadableShare}{0.0\%}
\newcommand{\TypedInSonnetAtLimitShare}{0.0\%}
\newcommand{\TypedInSonnetTzeroCorrectAll}{100.0\%}
\newcommand{\TypedInTokensMedianCloseFriend}{35,446}
\newcommand{\TypedInTokensMedianVipSender}{37,340}
\newcommand{\TypedInTokensMedianAll}{14,063}
\newcommand{\TypedInTokensPninetyfiveAll}{40,340}
\newcommand{\TypedInTokensMaxAll}{44,357}
\newcommand{\ExecFbWorlds}{20}
\newcommand{\ExecFbValidWithWorkingHoursEst}{32.1\%}
\newcommand{\ExecFbValidWithoutAll}{85.8\% [84.8, 86.7]}
\newcommand{\ExecFbValidWithAll}{90.4\% [89.5, 91.2]}
\newcommand{\ExecFbRunsAll}{2,085}
\newcommand{\ExecFbRequestsAll}{430}
\newcommand{\ExecFbValidWithoutOtherItemsEst}{92.6\%}
\newcommand{\ExecFbValidWithOtherItemsEst}{96.8\%}
\newcommand{\ExecFbRDiffSrc}{$+$0.9\,pp [$-$0.3, $+$2.5]}
\newcommand{\ExecFbRDiffDef}{$-$0.3\,pp [$-$1.7, $+$1.0]}
\newcommand{\ExecFbDelUpdateDiffBone}{0.0\,pp [$-$1.6, $+$1.6]}
\newcommand{\ExecFbBxDDiffSrcDefItemsEst}{$+$0.9\,pp}
\newcommand{\ExecFbBxDDiffSrcOtherItemsEst}{$-$0.5\,pp}
\newcommand{\ExecFbBxDDiffDefDefItemsEst}{0.0\,pp}
\newcommand{\ExecFbBxDDiffDefOtherItemsEst}{$-$2.3\,pp}
\newcommand{\ExecFbBxDDiffDelDefItemsEst}{0.0\,pp}
\newcommand{\ExecFbBxDDiffDelOtherItemsEst}{$-$2.1\,pp}
\newcommand{\ExecFbMessages}{1,860}
\newcommand{\ExecFbMessagesWorldData}{0}
\newcommand{\FieldExWorlds}{10}
\newcommand{\FieldExBxDExtractedSrcDefItems}{$+$18.2\,pp [$+$7.7, $+$29.7]}
\newcommand{\FieldExBxDDiffSrcDefItemsEst}{$-$6.1\,pp}
\newcommand{\FieldExBxDDiffSrcOtherItems}{0.0\,pp [$-$1.0, $+$0.9]}
\newcommand{\FieldExRTypedSrcEst}{77.0\%}
\newcommand{\FieldExRExtractedSrcEst}{55.8\%}
\newcommand{\FieldExBxDExtractedDefDefItems}{$+$25.4\,pp [$+$18.6, $+$32.4]}
\newcommand{\FieldExBxDDiffDefDefItemsEst}{$-$10.4\,pp}
\newcommand{\FieldExBxDDiffDefOtherItems}{0.0\,pp}
\newcommand{\FieldExRTypedDefEst}{69.2\%}
\newcommand{\FieldExRExtractedDefEst}{52.7\%}
\newcommand{\FieldExRecordCorrectContactsEst}{100.0\%}
\newcommand{\FieldExRecordCorrectCalendarEst}{91.5\%}
\newcommand{\FieldExRecordCorrectTasksEst}{39.8\%}
\newcommand{\FieldExRecordCorrectEmailEst}{57.3\%}
\newcommand{\FieldExRewriteMissedEmail}{57.8\%}
\newcommand{\FieldExRecordCorrectMessagesEst}{92.0\%}
\newcommand{\FieldExRecordCorrectNotesEst}{63.1\%}
\newcommand{\FieldExRecordsAllSources}{10,903}
\newcommand{\FieldExRecordCorrectAllSources}{82.7\% [80.2, 85.0]}
\newcommand{\FieldExRewriteMissedAllSources}{16.6\%}

%% file: memtab-style.tex
\usepackage{colortbl}%

\colorlet{mtmuted}{cgrey}%
\definecolor{mtnote}{HTML}{333333}
\definecolor{mtfocus}{HTML}{EEF2F7}

\newcommand{\mtsmall}{\fontsize{7.5}{8.5}\selectfont}
\newcommand{\mtnotefont}{\fontsize{8}{9.5}\selectfont}
\newcommand{\mtsetup}{\footnotesize
  \setlength{\heavyrulewidth}{.08em}\setlength{\lightrulewidth}{.05em}\setlength{\cmidrulewidth}{.03em}}

\newcommand{\mtmuted}[1]{{\color{mtmuted}#1}}
\newcommand{\mtint}[1]{{\mtsmall\color{mtmuted}#1}}
\makeatletter
\def\mt@cisplit#1 [#2]#3\mt@ciend{%
  \if\relax\detokenize{#3}\relax\expandafter\@firstoftwo\else\expandafter\@secondoftwo\fi
  {\ci@unit{#1} &}{\ci@unit{#1} & \mtint{[#2]}}}
\newcommand{\mtci}[1]{\expandafter\mt@cisplit#1 []\mt@ciend}
\newlength{\mt@estw}\newlength{\mt@intw}
\newcommand{\mtpairwidths}[2]{\settowidth{\mt@estw}{#1}\settowidth{\mt@intw}{\mtsmall #2}}
\newcommand{\mtpair}[2]{\makebox[\mt@estw][r]{#1}\,\makebox[\mt@intw][l]{\mtint{#2}}}
\def\mt@pairsplit#1 [#2]#3\mt@pairend{%
  \if\relax\detokenize{#3}\relax\expandafter\@firstoftwo\else\expandafter\@secondoftwo\fi
  {\mtpair{\ci@unit{#1}}{}}{\mtpair{\ci@unit{#1}}{[#2]}}}
\newcommand{\mtcipair}[1]{\expandafter\mt@pairsplit#1 []\mt@pairend}
\makeatother

\newcolumntype{N}[1]{>{\setbox0\hbox{#1}\hbox to\wd0\bgroup\hfil\ignorespaces}c<{\unskip\egroup}}
\newcolumntype{H}[2]{>{\raggedright\arraybackslash\hangindent=#2\hangafter=1}p{#1}}

\newcommand{\mtgroup}[2]{\multicolumn{#1}{@{}l}{\itshape #2}\\}
\newcommand{\mtin}{\hspace*{1em}}
\newcommand{\mtgap}{\addlinespace[3pt]}
\newcommand{\mtna}{\multicolumn{1}{c}{---}}

\newcommand{\mthead}[3][c]{\begin{tabular}[b]{@{}#1@{}}#2\\[-1pt]{\mtsmall\color{mtmuted}#3}\end{tabular}}

\makeatletter
\newsavebox{\mt@box}
\newlength{\mtwidth}
\newenvironment{mttabular}[1]
  {\begin{lrbox}{\mt@box}\begin{tabular}{#1}}
  {\end{tabular}\end{lrbox}\global\mtwidth=\wd\mt@box
   \par{\centering\usebox{\mt@box}\par}}
\newenvironment{mttabular*}[2]
  {\begin{lrbox}{\mt@box}\begin{tabular*}{#1}{@{\extracolsep{\fill}}#2}}
  {\end{tabular*}\end{lrbox}\global\mtwidth=\wd\mt@box
   \par{\centering\usebox{\mt@box}\par}}
\newcommand{\mtnotes}[2][\mtwidth]{%
  \par\vspace{3pt}%
  {\centering\parbox{#1}{\mtnotefont\color{mtnote}\raggedright #2}\par}}
\makeatother

%% file: sections/abstract.tex
AI assistants store values that a language model derived from a person's records under the person's
own definitions, such as their important contacts. Keeping such a value correct takes two steps:
\emph{detection}, noticing that an input changed, and \emph{repair}, producing the correct new value.
Benchmarks that score only the final answer cannot tell which step failed, and they state every change
as new text. Seeing this limitation, we present \memshift{}, a benchmark of five shifts: source
revision, deletion, redefinition, writer switch, and export. The first three fork a synthetic world
into branches that differ only in the shift. Gold values and dependencies are fixed before any text,
and a deterministic scorer replaces an LLM judge. For a memory that reports its invalidations,
\memshift{} shows for each changed item whether the memory \emph{noticed the change} and whether its
\emph{repair was right}. Moreover, we build a reference memory with known detection and repair to validate the measure. By evaluating four open-weight LLMs, we find that detection helps only as much
as repair allows: the LLMs re-derive few values computed by the person's definitions correctly, and
programs that the LLM writes over typed records close most of the gap to exact recomputation. Code and
data are available at \url{\repourl}.

%% file: sections/intro.tex
\section{Introduction}
\label{sec:intro}

AI assistants increasingly keep a memory of values derived from a person's records. Suppose an
assistant stores the item ``my important contacts are Ana, Ben, and Chlo\'e''. A language model, the
\emph{writer}, derived it from calendar records using the person's own definition: someone with at least
$n$ one-off meetings in the last $w$ days. Five \emph{shifts} can make such an item wrong or put it at
risk. A meeting is rescheduled (a source revision) or deleted (a deletion), or the person changes what
``important'' means (a redefinition). The assistant moves to a new writer model (a writer switch), or
the person exports the memory to another product (an export). Practitioners report stale memories
\citep{mem0_state_2026}, contradictions that pile up when a memory only adds new
facts,\footnote{Mem0 GitHub issue \#4956, \url{https://github.com/mem0ai/mem0/issues/4956}.} and memories
moved between products by pasting text exports \citep{anthropic_memory_import,google_gemini_memory_import}.

\begin{figure*}[t]
\centering
\includegraphics{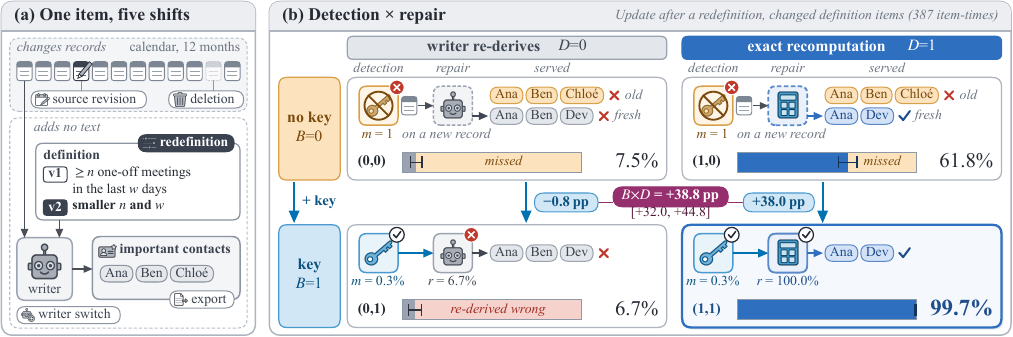}
\caption{Detection and repair are separate steps. (a)~One definition item and the five shifts. A source
revision and a deletion change records, and a redefinition, a writer switch, and an export add no text.
(b)~The four \holdmem{} cells $(D,B)$ after a redefinition. With a key ($B{=}1$), the memory invalidates
the item and re-derives it. Without a key ($B{=}0$), it keeps the old list and derives a new item only
when a new record arrives. The writer derives a wrong list ($D{=}0$), and exact recomputation the right
one ($D{=}1$). Each bar splits the changed definition items as in \cref{eq:decomp}: right (gray at
$D{=}0$, blue at $D{=}1$), wrong after a re-derivation (red), and wrong after a missed change (amber).
$m$: share of changes missed. $r$: share of re-derived items that are right.}
\label{fig:concept}
\end{figure*}

Keeping the item correct takes two steps: \emph{detection}, noticing that an input changed, and
\emph{repair}, producing the correct new value. A score on the final answer cannot tell which step
failed. Consider two memories after the person redefines ``important'' (\cref{fig:concept}). The first
never notices the change, and \ExDetMainDefDefinitionDzeroBzeroUpdateEst{} of the values it serves for
the changed definition items are right. The second notices nearly every change and asks its writer to
derive the value again, and \ExDetMainDefDefinitionDzeroBoneUpdateEst{} are right. The two get about the
same score, but they fail at different steps: the first at detection, the second at repair. Benchmarks
such as LoCoMo \citep{maharana2024locomo}, LongMemEval \citep{wu2025longmemeval}, and MemoryAgentBench
\citep{hu2025memoryagentbench} score the final answer after new information arrives, so they cannot
separate these memories. They also state every change as new text, but three of the five shifts add no
text. A redefinition leaves every record unchanged, a writer switch replaces only the model, and an
export replaces only the product. Stating a redefinition as a new message would change both the text
and the meaning, and the wording of a change alone already moves answers \citep{arxiv2607_16019}.

We thus ask: \emph{after a change, did a memory fail to notice it or fail to repair it, and what fixes
each failure?}

We propose a benchmark \memshift{} that answers this question item by item (\cref{sec:benchmark}). It
forks synthetic worlds into branches that are identical except for the shift. It fixes gold values and
their dependencies before any text is written, and it scores answers with a deterministic scorer instead
of an LLM judge. For each item whose gold value or status changed, it records whether the memory
invalidated the item, whether the re-derived value is right, and what the memory served otherwise. A
simple identity links these three terms, missed invalidation~$m$, repair correctness~$r$, and the served
accuracy~$a$ of items that were not re-derived, to accuracy after the shift. A memory that reports its
invalidations thus gets a diagnosis for every item. Any other memory still gets its carryover: its
accuracy on the items that the shift changed and on those it did not.

To validate the measure, we build a reference memory, \holdmem{}, in which detection and repair are two
switches with known effects (\cref{sec:instrument}). With $B{=}1$, each item has a key that invalidates
it when one of its inputs changes. With $B{=}0$, it has none. With $D{=}0$, the writer re-derives an
invalidated item. With $D{=}1$, the memory recomputes it exactly as gold does (possible for 43 of the
\ExAttrWorldSlots{} items), which gives an upper bound on repair. A third setting runs programs that the
writer writes itself.

We test four open-weight models from four families (Qwen3.5-9B, Llama-3.1-8B, gemma-4-12B, and
gpt-oss-20b), each as writer and reader, and two Claude models as writers of definition items. The
results show three patterns.
\begin{enumerate}[label=(\arabic*),leftmargin=1.6em]
\item Detection helps only as much as repair allows (\cref{sec:res-e1}). For every open-weight model, a
key raises Update more with exact recomputation than with the writer's own re-derivation. The difference
is \ExClsGemDefAllBxDEst{} to \ExClsGptDefAllBxDEst{} after a redefinition and
\ExClsMainOrphanAllBxDEst{} to \ExClsLlamaOrphanAllBxDEst{} after a deletion. A strong
writer makes the key useful on its own.
\item Repair fails on the items computed by the person's definitions (\cref{sec:res-construct}). Each
open-weight writer re-derives at most 25\% of these changed values correctly, while exact recomputation
gets at least 91\% right. Programs the writer writes over the memory's typed records close most of
this gap, and a Claude writer that sees every record of a definition comes close to exact recomputation.
\item Keys built from the writer’s citations notice every redefinition but miss records that join an
item later (\cref{sec:res-keys}).
\end{enumerate}
The interaction between detection and repair also holds on fresh worlds and on
held-out definitions, and most items that fail a writer switch are definition items
(\cref{sec:res-families}).

\paragraph{Contributions.} (1)~\memshift{}, a benchmark of five shifts with counterfactual branches,
gold dependencies, and a deterministic scorer, which separates detection from repair.
(2)~A reference memory with known detection and repair, which validates the benchmark and measures what
keys and programs are worth when a language model re-derives, together with design rules drawn from the
results.

%% file: sections/related.tex
\section{Related Work}
\label{sec:related}

\paragraph{Evaluating memory.} Agent memories page context in and out, reflect, link notes, extract
facts, keep temporal graphs, or act as an operating-system resource
\citep{packer2023memgpt,park2023generative,xu2025amem,arxiv2504_19413,arxiv2501_13956,li2025memos}. Many
build on RAG \citep{lewis2020rag,hipporag2024,graphrag2024}. Benchmarks test recall over long
conversations \citep{maharana2024locomo,wu2025longmemeval}, incremental interaction
\citep{hu2025memoryagentbench}, hallucination, long-horizon tasks, stale memories, governance, and
poisoning
\citep{chen2025halumem,he2026memoryarena,zhao2026amabench,xu2026evoarena,arxiv2605_06527,arxiv2607_16019,ren2026gatemem,chen2026memsecbench},
and personalization \citep{zhong2024memorybank,salemi2024lamp}. Most of their knowledge-update questions
state the change in the history and score only the final answer. A missed detection and a failed repair
then look the same.

\paragraph{Closest work.} Several recent systems pair detection with a repair step, and \memshift{} can
place each of them on $m$, $r$, and $a$. MemoRepair \citep{arxiv2605_07242} withdraws everything derived
from an invalidated source and rebuilds it by deterministic recomputation, regeneration, or removal. It
assumes complete provenance. In our terms, it is a key over complete dependencies with a repair that
mixes $D{=}1$ and $D{=}0$. The StateMem method of \citet{arxiv2608_19652} records which records an item
was derived from and recomputes the item with a language model, close to our citation key at $D{=}0$.
Its benchmark, StateMemBench, asks whether an answer is current or outdated, but every change arrives as
dialogue text, and an LLM judge scores the answer. PLANFENCE \citep{arxiv2609_03340} checks the stored
input versions of a derived plan and reports detection separately from completion, but its repair is
always a replan by a language model. User as Code \citep{arxiv2606_16707} keeps typed state with
executable rules, close to the writer's programs. MemTrace \citep{arxiv2606_17328} varies the age of
one fixed fact, the question, and the evidence, but never what the gold answer depends on. RECON
\citep{arya2026recon} asks which derived conclusions lose support after an invalidation but does not
compare a stale store with its repaired version. \citet{arxiv2609_05339} switch the writer over a fixed
history and find that the memory's representation decides what survives. We split survival into
detection and repair, much as studies of model updates separate accuracy from regressions
\citep{bansal2019updates,srivastava2020backward,yan2021pct,echterhoff2024muscle,xie2021regression}. What
\memshift{} adds is the measurement: detection crossed with repair, in branches that differ only in the
shift, with shifts that add no text, and with scoring against gold fixed before any text, as in script-first synthetic worlds
\citep{weston2015babi,arxiv2607_21962,arxiv2606_22877}, and without an LLM judge \citep{zheng2023judge}.

\paragraph{Dependency tracking and program-aided reasoning.} Build systems and incremental view
maintenance detect what a change affects \citep{mokhov2018build,gupta1995materialized}, and they
rebuild deterministically.
In a memory, a language model rebuilds, so the value of invalidation depends on how well it re-derives.
Language models are weak at aggregation over long contexts \citep{hsieh2024ruler,kuratov2024babilong},
which definition items require. Program-aided reasoning computes values with programs
\citep{gao2023pal,chen2023pot}, and \memshift{} measures what such programs are worth after each shift.

%% file: sections/benchmark.tex
\section{The \memshift{} Benchmark}
\label{sec:benchmark}

\memshift{} checks every item that a memory derived. After a shift, it asks whether the item is still
correct. If it is not, it asks which step failed: detection or repair. This section describes the data,
the shifts, the questions we ask, and the measures.

\subsection{Worlds and Derived Items}
\label{sec:worlds}

A \emph{world} is twelve months in the life of one synthetic person. It has six sources: e-mail, chat
messages, calendar, contacts, tasks, and notes. We first sample the true changes, as in
\citet{arxiv2607_21962}, and only then render them as text with deterministic templates. Each change inserts, revises, or
deletes a record, and each record has an identifier and a revision number. Background traffic, such as
newsletters, small talk, and routine mail, adds distracting text. A world has
\WorldEventsMin{}--\WorldEventsMax{} events and about \WorldTokensM{}M tokens.

From a world, a memory derives \ExAttrWorldSlots{} \emph{items} of two kinds. The 42 \emph{other items}
are facts that can be read from records: six attributes of the person (employer, manager, home city, time
zone, diet, and commute), the employer and city of each of twelve contacts, four recurring series, two
appointments, and six task due dates. Five of them are free text, and the rest are fields of structured
records. The six \emph{definition items} are computed with the person's own definitions: important
contacts, close friends, active projects, working hours, VIP senders, and overdue tasks. A definition is
a one-sentence rule over structured fields, such as calendar entries, contact cards, tasks, and message
headers. It names a source, a condition, and an aggregate, often with a time window and a threshold, as in the
opening example, and it has two versions. For every item and time, a deterministic evaluator computes the gold value,
the gold status, and the gold dependencies, that is, the records and revisions that the value depends
on. No human labels are needed.

\begin{table*}[t]
\caption{The five shifts. Each world forks at the end of month nine, and its branches share every event
outside the shift byte for byte.}
\label{tab:shifts}
\centering\mtsetup
\setlength{\tabcolsep}{4pt}
\newcolumntype{Y}[1]{>{\raggedright\arraybackslash}p{#1}}
\begin{mttabular*}{\linewidth}{@{}l Y{8.0cm} l Y{4.4cm}@{}}
\toprule
Shift & What changes & Records & Effect on gold \\
\midrule
\mtgroup{4}{Changes to the world}
\mtin source revision & ordinary events revise the records behind about a fifth of the items (a job
  change, a moved meeting, new messages) & revised & re-evaluated on the revised records \\
\addlinespace[2pt]
\mtin deletion & the record behind half of the items read from a series, task, or contact card is
  deleted & some deleted & \texttt{abstain} for those items \\
\addlinespace[2pt]
\mtin redefinition & half of the definitions move to their second version, and the memory is told &
  unchanged & re-evaluated under the new version \\
\midrule
\mtgroup{4}{Changes to the memory}
\mtin writer switch & the writer model, at the same tier, one larger, and one smaller & unchanged &
  none \\
\addlinespace[2pt]
\mtin export & the product: the memory moves to the Portable Memory format, and a second
  implementation imports it & unchanged & none \\
\bottomrule
\end{mttabular*}
\end{table*}

\subsection{Shifts}
\label{sec:shifts}

\Cref{tab:shifts} lists the five shifts. A source revision, a deletion, and a redefinition each fork the
world into a branch, and each draws from its own random stream. A branch thus differs from the others only in the events of its
shift. Three shifts add no text. A redefinition changes what a value means while every record stays the
same. The memory is told of the new version, and every question about a definition item names the
version in effect, so even a system that stores no definitions learns of it. A writer switch and an
export to the Portable Memory format \citep{portablememory_spec} change the memory, not the world, so no
gold value changes. They test whether what a memory derived survives a new model or a new product.

\subsection{Questions and Carryover}
\label{sec:probes}

A \emph{probe} is a question about one item at a given time, with a gold value and a gold status. The
status is \texttt{answer}, \texttt{abstain} (the record is gone), or \texttt{blocked} (the asking agent's
consent was revoked). We ask probes at the fork and at four checkpoints one, two, four, and eight weeks
later. Every system is queried through the same \emph{reader}, a model that answers each probe from the
memory's evidence within a 4,096-token budget. A deterministic scorer compares dates, times, entities,
and sets as typed values. An answer is correct only if both its status and its value match gold. For
example, the gold answer for the important contacts is a set, and an answer that names the same people
in any order is correct. No LLM judge is used.

\emph{Carryover} is how much of what a memory derived stays correct after a shift. We count each item
once at each checkpoint. An item is \emph{changed} at a checkpoint if its gold value or status differs
from that at the fork. \emph{Update} is the share of correct answers on the changed items.
\emph{Stability} is the same share on a random sample of the unchanged items, at most as many as the
changed items at that checkpoint. Both are percentages, and their differences are in percentage points
(pp). A writer switch changes no gold value, so for it we report \emph{retention}: the share of
answers that were correct before the switch and are still correct after it.

\subsection{Detection, Repair, and What Is Served}
\label{sec:identity}\label{sec:theory}

Consider the changed items of a memory that reports which items it invalidated. Three numbers describe
what happens to them. The \emph{missed invalidation}~$m$ is the share of changed items that the memory
did not invalidate after their gold value last changed. The \emph{repair correctness}~$r$ is the share
of invalidated and re-derived items that are answered correctly. The \emph{served accuracy}~$a$ is the
share of the other changed items, those not invalidated, that are answered correctly. For these items
the memory serves either the old value or a new item derived from records that arrived later. By the law
of total probability,
\begin{equation}
\mathrm{Update} \;=\; (1-m)\,r + m\,a .
\label{eq:decomp}
\end{equation}
Each term belongs to one part of a memory that a builder controls: $m$ to detection, $r$ to repair, and
$a$ to what is served without either. Compare two memories with the same repair, one that misses no
change ($m{=}0$) and one that never invalidates ($m{=}1$). Their Update differs by $r-a$, so this is
what detection is worth. Now take two kinds of repair, indexed 0 and 1. The difference between what
detection is worth under each,
\begin{equation}
(r_1-a_1)-(r_0-a_0),
\label{eq:interaction}
\end{equation}
is the \emph{interaction} between detection and repair. It is large when detection helps only with a
good repair.

\paragraph{Where a change comes from.} Update counts every change after the fork. A change is
\emph{caused by the shift} if the gold value at that checkpoint also differs from the gold value of the
same world continued without the shift. Otherwise the value would have changed anyway. Some changes come
from the clock alone: no record that the value depends on changed and the definition did not change, as
when a task becomes overdue or a meeting leaves a time window.

\subsection{Interface and Diagnostic Tests}
\label{sec:tests}

A system connects to \memshift{} through one interface. It ingests events, receives definitions and
consent revocations, sets the writer, answers probes, exports, imports, and reports its cost. A system
that offers only add and search runs through an adapter. The adapter ingests each event in time order
and returns the retrieved context to the shared reader. This is enough to measure carryover. A system
that also logs which items it invalidated and re-derived gets $m$, $r$, and $a$ for every item. Four
diagnostic tests with fixed thresholds check retention after a writer switch, carryover after an export,
accuracy before any shift, and serving cost (\cref{app:experiments}).

\begin{table}[t]
\caption{\memshift{} at a glance, and how it compares with the closest benchmarks.}
\label{tab:stats}
\centering\mtsetup
\setlength{\tabcolsep}{3pt}
\begin{mttabular}{@{}l r r r@{}}
\toprule
\multicolumn{2}{@{}l}{worlds (test + replication)} & \multicolumn{2}{r@{}}{\TestWorlds{} + \RepWorlds{}} \\
\multicolumn{2}{@{}l}{events per world} & \multicolumn{2}{r@{}}{\WorldEventsMin{}--\WorldEventsMax{}} \\
\multicolumn{2}{@{}l}{items per world (definition + other)} & \multicolumn{2}{r@{}}{6 + 42} \\
\multicolumn{2}{@{}l}{graded probes} & \multicolumn{2}{r@{}}{\MsQ{}} \\
\midrule
Changed items \mtmuted{(main study)} & Revision & Redefinition & Deletion \\
\cmidrule{1-4}
\mtin all & \KeyAuditChangedSrc{} & \KeyAuditChangedDef{} & \KeyAuditChangedOrphan{} \\
\mtin\mtin caused by the shift & \ExRevCauseMainSrcShiftAllItems{} & \ExRevCauseMainDefShiftAllItems{} &
  \ExRevCauseMainOrphanShiftAllItems{} \\
\mtin\mtin also without the shift & \ExRevCauseMainSrcDriftAllItems{} & \ExRevCauseMainDefDriftAllItems{} &
  \ExRevCauseMainOrphanDriftAllItems{} \\
\mtgap
\mtin clock alone & \ExRevClockMainSrcAllClockOnly{} & \ExRevClockMainDefAllClockOnly{} &
  \ExRevClockMainOrphanAllClockOnly{} \\
\midrule
& \multicolumn{2}{c}{Benchmarks (of \NbrTotal{})} & \\
\cmidrule(lr){2-3}
Property & Yes & Partly & \multicolumn{1}{c@{}}{\memshift{}} \\
\cmidrule{1-4}
\mtin gold fixed before text & \NbrGoldFirstYes{} & \NbrGoldFirstPart{} & \multicolumn{1}{c@{}}{\pass} \\
\mtin gold dependency sets & \NbrGoldDepYes{} & \NbrGoldDepPart{} & \multicolumn{1}{c@{}}{\pass} \\
\mtin redefinition & \NbrRedefYes{} & \NbrRedefPart{} & \multicolumn{1}{c@{}}{\pass} \\
\mtin writer switch & \NbrWriterYes{} & \NbrWriterPart{} & \multicolumn{1}{c@{}}{\pass} \\
\mtin export & \NbrImplYes{} & \NbrImplPart{} & \multicolumn{1}{c@{}}{$\sim$} \\
\mtin counterfactual branches & \NbrBranchYes{} & \NbrBranchPart{} & \multicolumn{1}{c@{}}{\pass} \\
\mtin no LLM judge & \NbrJudgeNone{} & \NbrJudgeMinor{} & \multicolumn{1}{c@{}}{\pass} \\
\bottomrule
\end{mttabular}
\mtnotes{Changed items: item-times, counted at each checkpoint. $\sim$:~partly, because the two
implementations share a data model. \Cref{app:related} compares them property by property.}
\end{table}

\paragraph{How \memshift{} compares.} \Cref{tab:stats} gives the size of \memshift{} and compares it
with the \NbrTotal{} closest benchmarks, each read from its own paper. None of them has all three
properties that the measures need: branches that differ only in the shift, gold dependencies that tell
which items a shift affects, and shifts that add no text. \memshift{} has all three and scores every
answer against gold.

%% file: sections/instrument.tex
\section{\holdmem{}: A Reference Memory}
\label{sec:instrument}

To test a measure of detection and repair, we need a memory whose detection and repair we control. Our
reference memory, \holdmem{}, derives each item from a 6,000-token BM25 selection of the records in the
item's sources. A stored item holds a one-sentence statement and up to six cited records with quotes, and
the reader sees only these. Two switches set how the memory detects and repairs. In the terms of
\cref{eq:decomp}, $B$ changes $m$ and $D$ changes $r$. Both also change $a$, because a memory that does
not invalidate an item may still derive a new one from records that arrived after its last write.

\paragraph{$B$ sets detection.} A $B{=}0$ memory only appends. It calls the writer only when new source
records arrive, derives new items from them, and never invalidates an item, so $m{=}1$. A $B{=}1$ memory
keeps a \emph{key} for each item. The key records how many events each input of the item has received,
the definition version, and the writer. When any of these changes, the memory invalidates the item and
re-derives it with the current writer. We test two keys. The \emph{source key} counts the events of each
source of the item. The \emph{citation key} counts the events of the record groups, such as a message
thread, that hold the records the writer cited. Every source receives events between checkpoints, so
the source key invalidates \KeyAuditFalseSourceSrcAll{}, \KeyAuditFalseSourceDefAll{}, and
\KeyAuditFalseSourceOrphanAll{} of the unchanged items after a source revision, a redefinition, and a
deletion. In these worlds it thus acts as a periodic full refresh.

\paragraph{$D$ sets repair.} At $D{=}0$ the writer re-derives an invalidated item from its BM25
selection. At $D{=}1$ the memory also checks the writer's value against the records
(\cref{app:holdmem}). If the check fails, \emph{exact recomputation} supplies the value: rules we wrote
read the record fields or, for a definition item, apply the codified definition as gold does. For 43 of
the \ExAttrWorldSlots{} items this computes exactly what gold computes, so $D{=}1$ is an upper bound on
repair. For the five free-text items, $D{=}1$ only checks the writer's value. A third setting,
the \emph{writer's programs} ($D_{\mathrm{m}}$), lets the writer itself supply the computation. The
writer writes one program per item and definition version, given the one-sentence definition and the
field names of the item's sources. A restricted interpreter runs the program on the memory's
\emph{typed records}, that is, the live records of those sources as typed fields instead of rendered
text.

\paragraph{Cells and controls.} We call each setting of $(D,B)$ a cell, written \holdmem{}$(D,B)$.
\holdmem{}(1,1) uses the source key unless we say otherwise. All cells are built from the same writer
outputs, and before any shift every cell gives the reader the same context for every probe. Keys never
depend on the repair, so $D$ cannot change detection. After a shift, the $D{=}1$ repair also differs
from the writer's re-derivation in how the value is presented and in which records it reads. Controls
vary one of these differences at a time (\cref{app:controls}).

%% file: sections/setup.tex
\section{Experimental Setup}
\label{sec:setup}

\paragraph{Models.} By default, Qwen3.5-9B \citep{qwen35_2026} serves as both writer and reader.
It runs at temperature~0 with constrained decoding. To test other model families, Llama-3.1-8B
\citep{arxiv2407_21783}, gemma-4-12B \citep{arxiv2607_02770}, and gpt-oss-20b \citep{arxiv2508_10925} each
replace it as writer and reader on \MTwoLlamaWorlds{} test worlds. To test stronger writers, Claude Sonnet
5.5 and Opus 5.5 \citep{anthropic_models_overview} write the definition items on 10 test worlds.

\paragraph{Baselines and native systems.} The main study includes three baselines. \emph{Plain RAG}
retrieves records with dense retrieval. The \emph{oracle context} gives the reader exactly the records
that the gold value depends on. \emph{No-store BM25} stores nothing and runs BM25
\citep{robertson2009bm25} over the live records at every query. We also run \MsNativeObs{} existing
memory systems, which we call native systems: Mem0 \citep{arxiv2504_19413}, A-Mem \citep{xu2025amem},
MemOS \citep{li2025memos}, LightRAG \citep{guo2025lightrag}, HippoRAG~2 \citep{gutierrez2025hipporag2},
and Cognee \citep{arxiv2505_24478}. Each ingests three-month worlds through its own interface and
answers through the shared reader, so only storage and retrieval differ. We read each system's $(D,B)$
setting from its documentation and code (\cref{tab:native-labels}). None checks a value against its
records, so all have $D{=}0$.

\paragraph{Experiments.} The main experiment runs six \holdmem{} cells after a source revision, a
redefinition, and a deletion on \TestWorlds{} twelve-month test worlds. The cells are the four $(D,B)$
cells with the source key and two $B{=}1$ cells with the citation key. The writer-switch experiment has
three tiers: from Qwen3.5-9B to MiMo-V2.6-Distill-Qwen-9B \citep{mimo_v2_6_2026} of the same size, from
Qwen3.5-9B up to Qwen3.8-27B \citep{qwen38_2026}, and from Qwen3.8-27B down to Qwen3.5-9B. Together with
an experiment over world lengths (horizons), these form the main study. Further experiments cover the
writer's programs, refresh cells, held-out definitions, the other writers, native systems, export, and
a replication on \RepWorlds{} fresh worlds (\cref{app:experiments}).

\paragraph{Statistics.} Every estimate has a 95\% world-bootstrap interval
\citep{efron1993bootstrap,field2007clustered,miller2024errorbars}. Differences between cells are paired
by item and checkpoint, and each of 2,000 bootstrap draws resamples the test worlds. The intervals thus
cover new worlds from the same generator. For held-out definitions they also resample the definitions.

%% file: sections/results.tex
\section{Results}
\label{sec:results}

Unless stated otherwise, the results are from the twelve-month test worlds of the main study, with
Qwen3.5-9B as writer and reader, and each contrast is over all changed items. \Cref{tab:overview} gives
Update and Stability after the three shifts that change the world. In its block of native systems, all
rows use the same worlds, but not all of them are scored on the same items.

\makeatletter
\providecommand{\mspct}[1]{\expandafter\ms@pct#1\ms@end}
\def\ms@pct#1\%#2\ms@end{#1}
\makeatother

\input{generated/tab_overview}

\begin{table*}[t]
\caption{Detection helps only as much as repair allows. (a)~The terms of \cref{eq:interaction} by item
class for Qwen3.5-9B. The writer re-derives few changed definition values correctly, and exact
recomputation nearly all. (b)~The main contrasts for four open-weight models, each as writer and
reader.}
\label{tab:main}\label{tab:families-main}
\centering\mtsetup
\setlength{\tabcolsep}{5pt}
\mtpairwidths{$+$00.0}{[$+$00.0,\,$+$00.0]}
\begin{mttabular*}{\linewidth}{@{}ll N{00.0} @{\hspace{14pt}} N{00.0}N{100.0} @{\hspace{14pt}} N{00.0}N{00.0} @{\hspace{14pt}} cc@{}}
\toprule
& & & \multicolumn{2}{c}{\mthead{Repair with a key}{\%}} & \multicolumn{2}{c}{\mthead{Without a key}{Update, \%}}
  & \multicolumn{2}{c@{}}{\mthead{$B{\times}D$ on Update}{pp}} \\
\cmidrule(lr){4-5}\cmidrule(lr){6-7}\cmidrule(l){8-9}
(a) Shift & Items & \multicolumn{1}{c}{Share} & \multicolumn{1}{c}{$r_0$} & \multicolumn{1}{c}{$r_1$}
  & \multicolumn{1}{c}{$a_0$} & \multicolumn{1}{c}{$a_1$} & caused by the shift
  & \multicolumn{1}{c@{}}{all} \\
\midrule
source revision & definition & \mspct\ExMixMainSrcDefinitionShare & \mspct\ExDetMainSrcDefinitionDzeroBoneRepairEst &
  \mspct\ExDetMainSrcDefinitionDoneBoneRepairEst & \mspct\ExDetMainSrcDefinitionDzeroBzeroUpdateEst &
  \mspct\ExDetMainSrcDefinitionDoneBzeroUpdateEst & \mtcipair\ExRevCauseMainSrcShiftDefinitionBxD &
  \mtcipair\ExAcctMainSrcDefinitionBxD \\
& other & \mspct\ExMixMainSrcOtherShare & \mspct\ExDetMainSrcOtherDzeroBoneRepairEst &
  \mspct\ExDetMainSrcOtherDoneBoneRepairEst & \mspct\ExDetMainSrcOtherDzeroBzeroUpdateEst &
  \mspct\ExDetMainSrcOtherDoneBzeroUpdateEst & \mtcipair\ExRevCauseMainSrcShiftOtherBxD &
  \mtcipair\ExAcctMainSrcOtherBxD \\
\addlinespace[2pt]
redefinition & definition & \mspct\ExMixMainDefDefinitionShare & \mspct\ExDetMainDefDefinitionDzeroBoneRepairEst &
  \mspct\ExDetMainDefDefinitionDoneBoneRepairEst & \mspct\ExDetMainDefDefinitionDzeroBzeroUpdateEst &
  \mspct\ExDetMainDefDefinitionDoneBzeroUpdateEst & \mtcipair\ExRevCauseMainDefShiftDefinitionBxD &
  \mtcipair\ExAcctMainDefDefinitionBxD \\
& other & \mspct\ExMixMainDefOtherShare & \mspct\ExDetMainDefOtherDzeroBoneRepairEst &
  \mspct\ExDetMainDefOtherDoneBoneRepairEst & \mspct\ExDetMainDefOtherDzeroBzeroUpdateEst &
  \mspct\ExDetMainDefOtherDoneBzeroUpdateEst & \mtna & \mtcipair\ExAcctMainDefOtherBxD \\
\addlinespace[2pt]
deletion & definition & \mspct\ExMixMainOrphanDefinitionShare & \mspct\ExDetMainOrphanDefinitionDzeroBoneRepairEst &
  \mspct\ExDetMainOrphanDefinitionDoneBoneRepairEst & \mspct\ExDetMainOrphanDefinitionDzeroBzeroUpdateEst &
  \mspct\ExDetMainOrphanDefinitionDoneBzeroUpdateEst & \mtcipair\ExRevCauseMainOrphanShiftDefinitionBxD &
  \mtcipair\ExAcctMainOrphanDefinitionBxD \\
& other & \mspct\ExMixMainOrphanOtherShare & \mspct\ExDetMainOrphanOtherDzeroBoneRepairEst &
  \mspct\ExDetMainOrphanOtherDoneBoneRepairEst & \mspct\ExDetMainOrphanOtherDzeroBzeroUpdateEst &
  \mspct\ExDetMainOrphanOtherDoneBzeroUpdateEst & \mtcipair\ExRevCauseMainOrphanShiftOtherBxD &
  \mtcipair\ExAcctMainOrphanOtherBxD \\
\bottomrule
\end{mttabular*}
\mtnotes[\linewidth]{Share: the class's share of the changed items (\%). $r_D$: repair correctness at
$D$ with a key. $a_D$: Update at $D$ without a key. $B{\times}D$: close to $(r_1{-}a_1)-(r_0{-}a_0)$.
---: a redefinition itself changes no other item.}
\par\medskip
\begin{mttabular}{@{}H{0.32\linewidth}{0.8em} c @{\hspace{10pt}} c @{\hspace{10pt}} c @{\hspace{10pt}} c@{}}
\toprule
(b) Contrast (pp) & Llama-3.1-8B & gemma-4-12B & gpt-oss-20b & \multicolumn{1}{c@{}}{Qwen3.5-9B} \\
\midrule
$B$ effect on Update, source revision & \mtcipair\ExClsLlamaSrcAllBPooled & \mtcipair\ExClsGemSrcAllBPooled
  & \mtcipair\ExClsGptSrcAllBPooled & \mtcipair\RendTemplateBUpdateSrc \\
$D$ effect on accuracy, source revision & \mtcipair\ExClsLlamaSrcAllDRepair & \mtcipair\ExClsGemSrcAllDRepair
  & \mtcipair\ExClsGptSrcAllDRepair & \mtcipair\RendTemplateDRepairSrc \\
$D$ effect on accuracy, redefinition & \mtcipair\ExClsLlamaDefAllDRepair & \mtcipair\ExClsGemDefAllDRepair
  & \mtcipair\ExClsGptDefAllDRepair & \mtcipair\RendTemplateDRepairDef \\
\addlinespace[2pt]
$B{\times}D$ on Update, source revision & \mtcipair\ExClsLlamaSrcAllBxD & \mtcipair\ExClsGemSrcAllBxD
  & \mtcipair\ExClsGptSrcAllBxD & \mtcipair\RendTemplateBxDSrc \\
$B{\times}D$ on Update, redefinition & \mtcipair\ExClsLlamaDefAllBxD & \mtcipair\ExClsGemDefAllBxD
  & \mtcipair\ExClsGptDefAllBxD & \mtcipair\RendTemplateBxDDef \\
$B{\times}D$ on Update, deletion$^\dagger$ & \mtcipair\ExClsLlamaOrphanAllBxD & \mtcipair\ExClsGemOrphanAllBxD
  & \mtcipair\ExClsGptOrphanAllBxD & \mtcipair\ExClsMainOrphanAllBxD \\
\addlinespace[2pt]
No-store BM25 $-$ \holdmem{}(1,1) & \mtcipair\MTwoLlamaZeroMinusHoldDef &
  \mtcipair\MTwoGemmaZeroMinusHoldDef & \mtcipair\MTwoGptOssZeroMinusHoldDef &
  \mtcipair\RendTemplateZeroMinusHoldDef \\
\bottomrule
\end{mttabular}
\mtnotes{The same \MTwoLlamaWorlds{} test worlds for every model. $^\dagger$For Qwen3.5-9B, the \TestWorlds{} worlds of
the main study. $B$ effect: source key, averaged over $D$. Accuracy: the share of re-derived items, changed or not, answered correctly. No-store BM25 $-$ \holdmem{}(1,1): after a redefinition. $D$'s
effect on missed invalidation was 0.0 for every model. Brackets: 95\% intervals.}
\end{table*}

\subsection{Detection Helps Only as Much as Repair Allows}
\label{sec:res-e1}

\paragraph{A key helps when the repair is good.} By \cref{eq:decomp}, a key changes $m$ and is worth
$r-a$. With Qwen3.5-9B as writer, a key raised Update after a redefinition by only
\EOneBEffectDzeroDefEst{} when the writer re-derived the item. It raised Update by
\ExClsDmDefAllBDoneEst{} when the writer's programs re-derived it, and by \EOneBEffectDoneDefEst{} with
exact recomputation, the upper bound. The same pattern holds for all four open-weight models
(\cref{tab:main}b). For each model, the key raised Update after a source revision, and $D{=}1$ raised the
accuracy on re-derived items without changing detection. The interaction $B{\times}D$ was positive after
a redefinition and after a deletion, with intervals above zero. After a source revision it was positive
for Qwen3.5-9B and gpt-oss-20b. A strong writer, in contrast, makes the key useful on its own. With
Claude writers of the definition items, the key alone raised Update on these items at $D{=}0$ by
\ExDetWriterDefinitionBDzeroMin{} to \ExDetWriterDefinitionBDzeroMax{} over the three shifts on 10
worlds.

\paragraph{The interaction comes from the definition items.} The source key missed at most
\ExDetMainSourceMissedMax{} of the changed items in any class, so the terms in \cref{tab:main}a explain
the interaction well. After a source revision, Qwen3.5-9B re-derived only
\ExDetMainSrcDefinitionDzeroBoneRepairEst{} of the changed definition items correctly, while exact
recomputation got \ExDetMainSrcDefinitionDoneBoneRepairEst{} right. All four models show this gap. On
changed definition items, each writer re-derived at most 25\% correctly after every shift, and exact
recomputation at least 91\% (\cref{tab:writers}). On the other items, the writer's re-derivation is
close to exact recomputation for every model except Llama-3.1-8B, and the interaction is small. On
definition items, the interaction was about the same for changes that the shift caused and for changes
that would have happened anyway (\cref{app:res-where}).

\paragraph{The source key works as a full refresh.} It re-derives nearly every item at every checkpoint.
On changed definition items, a refresh on the same schedule that keeps the old item next to the new one
does nearly as well. Removing stale items matters after a deletion, where such a refresh still serves
items whose record is gone (\cref{app:refresh-binding}). A key built from the gold dependencies
invalidates far fewer unchanged items and still keeps the interaction on definition items
(\cref{tab:oracle-key}).

\subsection{Repair Fails on Definition Items, and Programs Fix Most of It}
\label{sec:res-construct}

\begin{figure*}[t]
\centering
\includegraphics{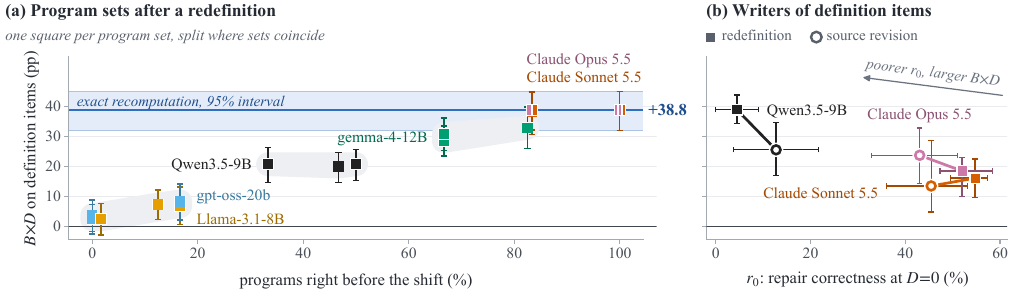}
\caption{$B{\times}D$ on definition items against the share of programs right before the shift (a) and
against $r_0$ (b). Computing the value helps most when the programs are right (a) and the writer's own
re-derivation is poor (b). (a)~The \RThreeGBDefSets{} program sets of the \RThreeGBWriters{} writers after a redefinition, one
square per set. Gray hulls enclose the sets of Qwen3.5-9B, of
gemma-4-12B, and of Llama-3.1-8B and gpt-oss-20b together. Blue line and band: exact recomputation and its
95\% interval. (b)~$B{\times}D$ with exact recomputation for three writers of definition items after a
redefinition (squares) and a source revision (circles). Bars: 95\% intervals.}
\label{fig:writers}
\end{figure*}

\paragraph{Programs written by the writer keep most of the interaction.} Repair ($D$) changes $r$. When
the programs of Qwen3.5-9B ran on the typed records with a key, they raised Update on changed definition
items after a source revision by \ExClsDmSrcDefinitionDBoneEst{} over the writer's own re-derivation,
compared with \ExClsMainSrcDefinitionDBoneEst{} for exact recomputation (\cref{tab:classsplit}). Over all items, the
interaction was \SynthCThreeDefEst{} after a redefinition and \SynthCThreeSrcEst{} after a source
revision. We also sampled \RThreeGBDefSets{} program sets from \RThreeGBWriters{} writers. Over these
sets, the interaction on definition items after a redefinition was \RThreeGBDefBxD{}, and it grew with
the share of programs that were correct before the shift (\cref{fig:writers}a). Two further tests show
where the programs fail. Returning a failing program's error message to the writer let more programs
run but repaired no more items, so the costly errors are wrong values, not crashes
(\cref{tab:feedback}). Programs that read fields extracted from model-written prose still gave a
positive interaction (\cref{tab:extract}). After a deletion, however, the programs re-derived the other
items less accurately than the writer did (\ExRThreeDelCOneOtherEst{}), mostly because they returned
values for deleted records (\cref{app:dm-deletion}).

\paragraph{A Claude writer that sees every record comes close to exact recomputation.} The $D{=}0$
writer sees only a 6,000-token selection. For active projects, overdue tasks, and working hours, this
selection always held every record that the definition ranges over. For close friends and VIP senders,
it never did. Where the selection was complete, Qwen3.5-9B still re-derived only
\ExRevCovQwenSrcCompleteRZeroEst{} of the changed definition values correctly after a source revision.
Claude Sonnet 5.5 and Opus 5.5 re-derived \ExRevCovSonnetSrcCompleteRZeroEst{} and
\ExRevCovOpusSrcCompleteRZeroEst{}, as much as exact recomputation on the same items. Over all
selections, the interaction after a redefinition was smaller with either Claude writer,
\RThreeHSonnetDefDefinitionDoneBxDEst{}, but still positive (\cref{fig:writers}b). When we gave Claude
Sonnet 5.5 every live record of the definition's sources as typed fields, it re-derived
\TypedInSonnetRTypedBothAllEst{} of the changed definition items correctly. This compares with
\TypedInSonnetRSelBothAllEst{} from its selection and \TypedInSonnetRAuthoredBothAllEst{} for exact
recomputation, at a cost of up to \TypedInTokensMaxAll{} tokens per re-derivation. Qwen3.5-9B reached
only \TypedInQwenRTypedBothAllEst{} with every record. This writer thus lacks the ability to compute
the value, not the evidence (\cref{tab:typed}).

\subsection{Keys Built from Citations Miss Records That Join Later}
\label{sec:res-keys}

\paragraph{The citation key notices every redefinition but misses later changes.} It missed
\ExDetMainSrcAllDoneBoneMissedCitedEst{} of the changed items after a source revision and
\ExDetMainDefAllDoneBoneMissedCitedEst{} after a redefinition. The key holds the definition version, so
it invalidated all \ExDetCitedDefShiftAllItems{} changes that the redefinition caused. It then missed a
later change of records or of the clock on \ExDetCitedDefShiftAllMissedLast{} of them. After a
redefinition, every wrong answer on changed definition items at $D{=}1$ came from such a miss
($r_1={}$\ExDetMainDefDefinitionDoneBoneRepairCitedEst{}), so this key fails at detection, not at
repair. Most misses were records that entered the item's scope after the write, as in the phantom
problem of predicate locks \citep{eswaran1976predicate}, or records that left a time window
(\cref{app:cited-misses}). Averaged over $D$, the citation key raised Update only after a deletion
(\EOneBEffectPooledCitedOrphanEst{}), because the writer's citations name the deleted record. Forcing
the memory to re-derive the definition items it missed after a source revision would have gained
\ExRevForcedDzeroSrcDefinitionGainEst{} at $D{=}0$ and \ExRevForcedDoneSrcDefinitionGainEst{} at $D{=}1$
(\cref{app:forced}).

\paragraph{The source key misses changes caused by the clock alone.} The key does not track time, so it
catches such a change only if an unrelated record of the same source arrives in the same interval. Such
stale values caused \ExRevClockMainShiftsErrorsOneOneClock{} of the \ExRevClockMainShiftsErrorsOneOne{}
wrong answers of \holdmem{}(1,1) on changed definition items (\cref{app:clock}).

\paragraph{An export that carries the keys keeps detection.}\label{sec:res-more} We exported
\holdmem{}(1,1) to Portable Memory with its keys and imported it into a second implementation. The import
matched the original (\ESixWithExtPostEst{}). Without the keys, the import detected no later change
(\cref{app:e7}).

\begin{table*}[t]
\caption{Design rules supported by the results, with their evidence. Writer and reader: Qwen3.5-9B unless
the evidence names other models.}
\label{tab:rules}
\centering\mtsetup
\setlength{\tabcolsep}{3pt}
\newcolumntype{Y}[1]{>{\raggedright\arraybackslash}p{#1}}
\begin{mttabular}{@{}H{5.3cm}{1em} Y{11.6cm}@{}}
\toprule
Rule & Evidence \\
\midrule
1. Key everything that can change after the write, and export the keys
& Citation keys miss \ExDetMainSrcAllDoneBoneMissedCitedEst{} and \ExDetMainDefAllDoneBoneMissedCitedEst{}
of the changed items after a source revision and a redefinition, mostly records that join later. An
import without the keys detects no change. \\
\addlinespace[2pt]
2. Compute definition items with a program over typed records, one per definition version
& Four open-weight writers re-derive at most 25\% of the changed definition items correctly, against at
least 91\% for exact recomputation. The programs of Qwen3.5-9B keep most of the interaction. \\
\addlinespace[2pt]
3. Remove the items whose record is deleted
& After a deletion, the key beats a refresh that keeps old items by
\ExRevRefreshDoneOrphanOtherBoneMinusRefreshEst{} on the other items at $D{=}1$. Invalidation keyed on
the source record raises Mem0's Update there by \RThreeXOrphanOtherDetEffectEst{}. \\
\addlinespace[2pt]
4. Check that the record a program names still exists
& After a deletion, the programs re-derive the other items less accurately than $D{=}0$
(\ExRThreeDelCOneOtherEst{}). Of their loss on deleted records, \ExRThreeDelLossShareVsDzeroAReplacementEst{}
is a value returned where the gold status is \texttt{abstain}. \\
\addlinespace[2pt]
5. Test a new writer on definition items
& At the three tiers, \ETwoExFailDefinition{} of the \ETwoExFailTotal{} items that fail a switch are
definition items. \\
\bottomrule
\end{mttabular}
\end{table*}

\subsection{Robustness and Native Systems}
\label{sec:res-families}

\paragraph{Fresh worlds, new definitions, and rewritten text.}\label{sec:res-rendered} On \RepWorlds{}
fresh worlds, every contrast with the source key replicated (\cref{app:replication}). We also replaced
the six definitions with \RThreeGADefDoneDefinitions{} held-out definitions from the same grammar on
\RThreeGAWorlds{} worlds. With intervals that resample both worlds and definitions, the interaction on
definition items after a redefinition was \RThreeGADefDoneBxD{}. When gemma-4-12B rewrote the message
bodies around unchanged gold facts, the effects of $B$ and $D$ and their interaction kept their sign
(\cref{app:rewritten-results}).

\paragraph{Writer switches fail mostly on definition items.}\label{sec:res-e2} At each of the three
tiers, at least 97\% of the answers that were correct before a switch stayed correct, and
\ETwoExFailDefinition{} of the \ETwoExFailTotal{} items that failed were definition items. Over all
\MOneSwitches{} switches, the new writer's own accuracy predicted which items stay correct at least as
well as the agreement between the old and new writers did (\cref{app:res-switch}).

\paragraph{Stored state pays off over time.}\label{sec:res-horizon} Before any shift, no-store BM25
scored \EThreeZeroMinusHoldTwelveEst{} compared with \holdmem{}(1,1) at twelve months, while the source key used
\ExCostModelMainTwelveHoldRatioZero{} times as many tokens (\cref{app:res-horizon}).

\paragraph{Documented settings match behavior but do not predict carryover.}\label{sec:res-native} For
each native system, the $B$ setting inferred from how many deleted items it still answered matched its
documentation. Yet most of their Update and Stability values fell outside the interval of the
\holdmem{} cell with the same setting (\cref{app:native}). The per-item terms explain more. The tested
Mem0 release never updates or deletes a memory, so $m{=}1$ for every item class, yet the class decides
which fix helps. A program run at query time on the live typed records, which Mem0 does not store,
raised its Update on changed definition items after a redefinition by
\RThreeXDefDefinitionRepEffectEst{}. Invalidation keyed on the source record raised its Update on the
other items after a deletion by \RThreeXOrphanOtherDetEffectEst{}.

%% file: generated/tab_overview.tex
\begin{table*}[t]
\caption{Update and Stability (\%) after the three shifts that change the world.}
\label{tab:overview}
\centering\mtsetup
\setlength{\tabcolsep}{6pt}
\begin{mttabular}{@{}l N{00.0}N{00.0} @{\hspace{24pt}} N{00.0}N{00.0} @{\hspace{24pt}} N{00.0}N{00.0}@{}}
\toprule
& \multicolumn{2}{c}{Source revision} & \multicolumn{2}{c}{Redefinition} & \multicolumn{2}{c@{}}{Deletion} \\
\cmidrule(lr){2-3}\cmidrule(lr){4-5}\cmidrule(l){6-7}
Memory & \multicolumn{1}{c}{Update} & \multicolumn{1}{c}{Stability} & \multicolumn{1}{c}{Update} & \multicolumn{1}{c}{Stability} & \multicolumn{1}{c}{Update} & \multicolumn{1}{c@{}}{Stability} \\
\midrule
\mtgroup{7}{Main study: 20 twelve-month test worlds, Qwen3.5-9B as writer and reader}
\mtin\holdmem{}(0,0) & 62.8 & 72.7 & 26.9 & 77.8 & 9.0 & 78.3 \\
\mtin\holdmem{}(0,1) & 68.3 & 90.5 & 27.1 & 91.8 & 58.2 & 84.8 \\
\mtin\holdmem{}(1,0) & 80.1 & 72.8 & 64.7 & 77.8 & 18.8 & 79.6 \\
\mtin\holdmem{}(1,1) & 93.1 & 93.1 & 91.6 & 93.0 & 73.1 & 88.4 \\
\mtgap
\mtin\holdmem{}(0,1), citation key & 63.2 & 90.3 & 21.9 & 91.8 & 54.7 & 84.3 \\
\mtin\holdmem{}(1,1), citation key & 79.6 & 91.2 & 67.2 & 91.8 & 58.1 & 85.0 \\
\mtgap
\mtin\holdmem{}($D_{\mathrm{m}}$,0) & 68.5 & 69.2 & 45.0 & 78.0 & 16.8 & 78.0 \\
\mtin\holdmem{}($D_{\mathrm{m}}$,1) & 85.0 & 88.4 & 70.4 & 90.7 & 45.2 & 82.7 \\
\mtgap
\mtin no-store BM25 & 61.0 & 69.3 & 25.3 & 67.4 & 71.3 & 64.8 \\
\mtin plain RAG & 63.1 & 66.1 & 25.6 & 64.7 & 69.7 & 64.4 \\
\mtin oracle context & 74.1 & 87.0 & 45.9 & 87.3 & 86.9 & 87.3 \\
\midrule
\mtgroup{7}{Native systems: 15 three-month worlds, scored on the half of the items not used to infer the settings}
\mtin\holdmem{}(0,0) & 57.6 & 66.6 & 47.2 & 69.1 & 44.0 & 74.1 \\
\mtin\holdmem{}(0,1), citation key & 77.8 & 98.7 & 65.5 & 97.6 & 70.1 & 97.2 \\
\mtin\holdmem{}(1,1) & 98.2 & 99.0 & 97.4 & 99.0 & 97.9 & 99.2 \\
\mtgap
\mtin no-store BM25 & 71.1 & 70.8 & 58.3 & 70.1 & 60.3 & 72.1 \\
\mtin plain RAG & 73.4 & 66.5 & 59.1 & 69.7 & 58.1 & 67.6 \\
\mtin oracle context & 93.8 & 89.2 & 91.1 & 92.0 & 95.7 & 91.6 \\
\mtgap
\mtin Mem0 & 70.6 & 67.5 & 50.6 & 69.7 & 44.7 & 70.0 \\
\mtin A-Mem & 69.3 & 80.2 & 50.6 & 81.1 & 34.2 & 80.9 \\
\mtin MemOS & 70.4 & 68.7 & 57.4 & 67.6 & 42.1 & 68.6 \\
\mtin LightRAG & 64.9 & 85.2 & 50.6 & 80.1 & 31.6 & 80.9 \\
\mtin HippoRAG 2 & 74.3 & 82.6 & 58.3 & 83.8 & 47.4 & 82.7 \\
\mtin Cognee & 68.8 & 95.1 & 56.2 & 95.0 & 42.1 & 95.7 \\
\bottomrule
\end{mttabular}
\mtnotes{\holdmem{}($D$,$B$): $D{=}1$ exact recomputation, $D_{\mathrm{m}}$ the writer's programs, $B{=}1$ a key (the source key unless the citation key is named). $n$ per main-study \holdmem{} cell: \ExDetMainSrcUpdateItems{}, \ExDetMainDefUpdateItems{}, and \ExDetMainOrphanUpdateItems{} changed and \ExDetMainSrcStabilityItems{}, \ExDetMainDefStabilityItems{}, and \ExDetMainOrphanStabilityItems{} unchanged items.}
\end{table*}

%% file: sections/discussion.tex
\section{Discussion}
\label{sec:discussion}

\paragraph{Detection is worth as much as the repair that follows.} With a language model as writer, the
re-derivation is not exact, and it is worst for values that aggregate many records under a person's
definition. A key helps there mainly with exact recomputation or with a program over typed records. A
strong writer that sees every record comes close, but it must read every record at each re-derivation.

\paragraph{A final-answer score hides which step failed.} The diagnosis names the term that failed and
the item class it failed on, and the class tells which fix helps, as the Mem0 interventions show. On
three-month worlds, the ranking of the six \holdmem{} cells by Update after a source revision nearly
fixes their ranking after a redefinition (Spearman $\rho={}$\RankExRhoSrcDefHoldMem{}), but among the
six native systems it does not ($\rho={}$\RankExRhoSrcDefNative{}).

\paragraph{Recommendations.} We recommend reporting $m$ and $r$ by item class, and \cref{tab:rules} lists
design rules for memories.

%% file: sections/conclusion.tex
\section{Conclusion}
\label{sec:conclusion}\label{sec:concl}
We proposed \memshift{}, a benchmark for evaluating LLMs' performance in derived personal memory.
\memshift{} scores each item that a shift changes on detection and repair, against gold fixed before
any text. Detection helps only as much as repair allows. Repair fails on the items computed by the
person's definitions, and programs over typed records close most of this gap. Keys built from the
writer's citations miss records that join an item later. A memory's accuracy after a change measures
its detection and its repair together. Future work can test memories in which a language model decides
invalidation, and real records with vaguer definitions.

%% file: sections/statements.tex
\input{statements/limits}
\input{statements/ethics}
\input{statements/reproduce}
\input{statements/ack}

%% file: statements/limits.tex
\section*{Limitations}

\paragraph{Synthetic worlds and codified definitions.} The worlds are synthetic and rendered from
templates, and each definition is a one-sentence rule over structured fields. This lets an
evaluator compute gold values and dependencies exactly, without annotators, but real records are messier
and real definitions vaguer. Only a small share of the message text was rewritten by a model, field
extraction was tested only on model-written prose, and the held-out definitions come from one grammar.
Programs written from definitions in the user's own words still give a positive interaction
(\cref{app:res-repair}).

\paragraph{Reference memory and native systems.} Every cell with a known $(D,B)$ setting is a
\holdmem{} cell that we designed. $D{=}1$ runs the same evaluator as gold, so it is an upper bound on
repair, not a system under test. The native systems ran on three-month worlds at $D{=}0$ through our
adapters, and their ingestion is noisy. The tested Mem0 release only adds memories, and Graphiti
finished no world, so no native run measures invalidation decided by a language model.

\paragraph{Models, readers, and serving format.} The main writers and readers are open-weight models
with 8--27B parameters. Claude Sonnet 5.5 and Opus 5.5 wrote only definition items, and only Sonnet
received every record of a definition. Update also depends on the reader and on how the store is
presented. Serving the same $D{=}1$ store as JSON records changes Update (\cref{tab:json-arm}), so the
signs of the effects generalize better than their absolute levels.

\paragraph{Statistics.} The main-study intervals rest on \TestWorlds{} worlds, which are few clusters
for a bootstrap \citep{cameron2008bootstrap}. The replication repeats every contrast on \RepWorlds{}
fresh worlds. Intervals are per contrast. Holm and Benjamini--Hochberg corrections over the main
contrasts keep \MultExOrigHolm{} and \MultExOrigBh{} of the \MultExOrigInterval{} intervals that exclude
zero (\cref{app:multiple}).

%% file: statements/ethics.tex
\section*{Ethics Statement}
This work concerns systems that store what AI assistants know about individual people. Measuring
whether this knowledge stays correct after models, sources, or products change can make assistants
more reliable and make it easier for people to take their data with them. The same capabilities,
persistent and portable personal profiles, raise privacy risks. All worlds in \memshift{} are
synthetic, no personal data was collected, and no human subjects were involved. Derived profiles also
hold facts about third parties, such as the employers and cities of a person's contacts, and exported
provenance reveals which sources and models produced each item. The benchmark scores consent
revocation as an outcome, and every stored derived memory we measured still leaks after a revocation
(\cref{app:e7}). The writer's programs run only on synthetic records, in a separate process with a time
limit and a memory cap. A restricted interpreter accepts only listed syntax and built-in functions
(\cref{app:programs}).

%% file: statements/reproduce.tex
\section*{Reproducibility Statement}
Worlds, shifts, and probes are generated deterministically from seeds, and a deterministic scorer
checks every answer without an LLM judge. Open-weight models run at temperature~0 with constrained
decoding, except for the sampled program sets (temperature 0.7). Claude Sonnet 5.5 and Opus 5.5 ran
through Anthropic's API with the settings in \cref{app:models}. Every run records its code version,
world seeds, model revisions, and output hashes in a manifest. The released repository holds
the code, generator, worlds, gold labels, all model responses, and run manifests, and it regenerates
every number in the paper.

%% file: statements/ack.tex
\section*{Acknowledgements}
This work was supported by JST CREST Grant Number JPMJCR22M2 and JSPS KAKENHI Grant Numbers JP23K17456, JP23K28096, JP25H01117, JP25K21207, and JP26K03246.

%% file: sections/appendix.tex
\appendix

\section*{Appendix}

\input{appendix/app_related}
\input{appendix/app_benchmark}
\input{appendix/app_systems}
\input{appendix/app_prompts}
\input{appendix/app_settings}
\input{appendix/app_results}
\input{appendix/app_errors}
\input{appendix/app_datasheet}

%% file: appendix/app_related.tex
\section{Extended Related Work}
\label{app:related}

This section describes the closest work of \cref{sec:related} in more detail. It then groups further
work by the question it answers for \memshift{}.

\paragraph{Memory systems and the closest work.} The agent memories in the main text are built for
recall. We ask what happens to the values they derived when their inputs change. StateMemBench
\citep{arxiv2608_19652} uses scripted dialogues rendered by an LLM and an LLM judge that scores whether
an answer is current or outdated. The scored answers include derived values that are not recomputed
after an input changes. It runs eight memory and graph-RAG systems, but every change arrives as dialogue
text, and it scores a stale answer, not whether the change was noticed. Its method, StateMem, records
derived-from links between memory entries at ingestion, flags the entries that depend on a changed entry
without calling a model, and lets an LLM recompute a flagged value at answer time. In our terms, its
detection is a key recorded at write time, close to our citation key, and its repair is $D{=}0$ at
query time. Such links miss records that join a value's scope after it was written
(\cref{app:cited-misses}). Moreover, Qwen3.5-9B re-derived fewer than half of the changed definition
items correctly even from complete evidence (\cref{app:coverage}). MemTrace \citep{arxiv2606_17328}
holds one typed fact fixed and varies its age, the question type, and the evidence, with an LLM judge.
We instead hold the probes fixed and change what lies behind them. PLANFENCE \citep{arxiv2609_03340}
binds a derived plan to the declared versions of its inputs and checks them before an action. Its repair
is always an LLM replan, so it does not vary how a value is recomputed. RECON \citep{arya2026recon} asks
which derived conclusions lose support after an invalidation, but it never compares a stale store with
its repaired version. MemoRepair \citep{arxiv2605_07242} withdraws everything derived from an
invalidated source and rebuilds it by deterministic recomputation, regeneration, or removal. It assumes
complete provenance. In our terms, its detection is a key over complete dependencies, like our oracle
key built from the gold dependencies (\cref{app:controls}). Its deterministic recomputation is $D{=}1$,
and its regeneration is $D{=}0$. We measure each part when the dependencies are recorded at write time
and a language model re-derives. \citet{arxiv2609_05339} switch the writer or the embedder over a stored
state in four idealized formats while the history stays fixed. They find that the representation
decides what survives.

\paragraph{Stale values in agent memory.} Supersede trains a small model to avoid stale values in a
notes memory that the agent rewrites itself \citep{arxiv2606_27472}. RD-Forget suppresses superseded
facts through slots that an LLM curator assigns \citep{arxiv2609_10263}. Reliable post-retrieval
assembly uses code to pick the newest version among candidates that an LLM extracted
\citep{arxiv2606_01435}. Supersede and the assembly rule replace single facts, so neither derives a
value from several records, and RD-Forget's slots are judged by a model rather than keyed to source
versions. Selective QA over conflicting personal sources scores one fixed snapshot
\citep{arxiv2605_30087}, the setting of our conflict probe. Sleep-time compute precomputes quantities
from stored context \citep{lin2025sleeptime}, and carryover asks whether such quantities stay correct
when the context changes. Close to our writer switch, agents keep one identity across models and hosts
\citep{arxiv2609_00546}, and readers adopt stale memory values \citep{arxiv2609_01852}. Labels extracted
from clinical summaries change with the size of the extraction model \citep{arxiv2606_05970}.
Comparisons of memory designs find no universal winner \citep{arxiv2606_24775,arxiv2608_15008} and find
that the backend matters more than the reader's size \citep{arxiv2608_02613}. Checking drafts against
stored state repairs implicit stale dependencies \citep{arxiv2608_01619}. Temporal question answering
\citep{chen2021timeqa,liska2022streamingqa,kasai2023realtimeqa,vu2024freshllms} and work on knowledge
conflicts \citep{xu2024conflicts} study facts that change over time and model knowledge that goes out of
date.

\input{generated/tab_neighbours_full}

\paragraph{Knowledge editing.} Knowledge editing asks whether a model's answers, including multi-hop
answers and ripple effects, follow an edit to its weights
\citep{meng2022rome,meng2023memit,zhong2023mquake,cohen2024ripple}. Memory-based editors
\citep{mitchell2022serac,hartvigsen2023grace} record the edits but not what a stored value was derived
from.

\paragraph{Closest benchmarks compared property by property.} \Cref{tab:neighbors-full} compares
\memshift{} with the \NbrTotal{} benchmarks of our audit, and the counts in the main text come from it.
Every entry is read from the benchmark's paper, and the supporting quotes with their pages are released
with the code. StateMemBench, MemTrace, PLANFENCE, and MemoRepair are discussed above. The table uses the
following definitions. A benchmark has \emph{derived items} when its graded items aggregate several
records (counts, sets, sums, multi-hop, or temporal aggregation) instead of restating one stored fact.
\memshift{} has them only partly: a writer derives every item, but of a world's 48 items only the six
definition items aggregate records, the two appointments pick the earliest of several, and the rest
are attributes or latest values. An \emph{implicit conflict} is conflicting information without an explicit replacement, which
the system must detect. \emph{Consent} is revoked when the user withdraws consent or asks to forget, and
the memory must then stop using or delete the data. A benchmark is \emph{counterfactual} when it runs the
same history with and without the change (\memshift{} has byte-identical branches). \emph{Native
systems} are the existing memory systems that a paper evaluates, grouped as the paper groups them,
without long-context and plain RAG baselines, the paper's own systems, or oracles. Histories (worlds,
users, or trajectories) and questions are counted as the paper reports them. For MemoryArena the
questions are its tasks, and for EvoArena the counts are those of its PersonaMem-Evo subset.

\paragraph{Checking derived values.} LLM extraction and semantic operators build structured views of
documents
\citep{arora2023evaporate,patel2025lotus,shankar2025docetl,tang2026documentdatabase,hajidehi2026cracking,structure_query2026}.
Data-quality systems check and repair such outputs with declarative constraints
\citep{schelter2018deequ,rekatsinas2017holoclean,lee2025sic} or synthesized assertions
\citep{shankar2024spade}, including incremental checks \citep{kaminsky2025incrementaldc}. ChatDB keeps
memory as SQL tables that the model queries and updates \citep{hu2023chatdb}. Our check of a stored value
against its records is a per-item constraint. In our results it repairs only when a program also
recomputes the value from the memory's typed records (\cref{app:res-repair}).

\paragraph{Dependency tracking.} Our key is standard dependency tracking: the memory compares a stored
key with the current one to decide whether to rebuild \citep{mokhov2018build}. Self-adjusting
computation and incremental view maintenance recompute only what a change affects
\citep{acar2002adaptive,hammer2014adapton,gupta1995materialized,differential2013,noria2018,dbsp2023}, and
a redefinition corresponds to a view redefinition \citep{gupta1995adapting}. Deletion propagation asks the
reverse question, which source tuples to delete so that a given view tuple disappears
\citep{buneman2002deletion}. Truth maintenance withdraws beliefs whose
justifications fail \citep{doyle1979tms}, and assumption-based truth maintenance labels
each belief with the assumption sets it rests on \citep{dekleer1986atms}. A value defined by a predicate
depends on records outside any read set, which is what predicate locks capture
\citep{eswaran1976predicate}. Build systems and view maintenance gain from invalidation because their rebuild is
deterministic. When a language model rebuilds, the result is lossy in a way that depends on the item
class and the writer, and a read set or citation list recorded at write time is not a sound dependency
set. Dependency-guided rollback repair scores the invalidation of claims separately from recovery
\citep{arxiv2608_10502}. Database semantics for agent state include versioned resources with isolation
\citep{arxiv2608_05412}, transactional belief commits \citep{arxiv2607_23929}, bitemporal operators
\citep{arxiv2606_06240,snodgrass1999temporal}, and dependency masks \citep{arxiv2609_24090}. Temporal
memories retire or revoke superseded values \citep{arxiv2606_26511,arxiv2608_07429}.

\paragraph{Provenance and freshness.} BeliefBank's constraints relate beliefs to one another
\citep{kassner2021beliefbank}, not a derived value to the source versions it was built from. Provenance
models \citep{whywhere2001,semirings2007,curatedprovenance2006,prov_dm,cheney2009provenance} and
provenance systems for agents and LLM processing \citep{provagent2025,blip2026,wang2026trace} record or
infer what a result was derived from. Our key is provenance reduced to what invalidation needs: the
event counts of the sources, the definition version, and the writer. View maintenance
\citep{colby1996deferred,zhuge1995warehouse}, session guarantees \citep{terry1994session}, and stream
processing \citep{dataflow2015,streams2024,dynamic_tables2025} state which input changes a derived
result must reflect. For derived items, Update measures how often a read after a shift reflects the
shift. Our sync simulators apply the matching contracts of consumer APIs to a personal memory
(\cref{app:sync-sim}).

\paragraph{Portability, control, and consent.} Tool protocols such as MCP \citep{mcp20251125} and
personal agents over a user-controlled repository \citep{cupa2025} let different models reach a person's
applications and data. For this reason, the derived memory, not the model, has to carry state across a
change of model. Portability is a legal right \citep{gdpr2016} and an engineering practice
\citep{dtp2018}. Portable Memory \citep{portablememory_spec,portablememory2026paper}, MIF
\citep{mif_spec}, and memorywire \citep{arxiv2606_01138} carry provenance, but none records the
definition versions that a derivation depended on. Only MIF can record the version of a source
document, as an optional field with no invalidation rule. We make such a derivation extension a
controlled variable. MIF records the version of the agent that created an item, which matches the
writer component of our key. Personal knowledge graphs, personal data stores, local-first software,
enclaves, and user-owned agents keep a person's data under the person's control
\citep{pkgsurvey2024,mansour2016solid,kleppmann2019localfirst,arxiv2604_02522,arxiv2607_05363}. Usage
control, authentication and authorization, information-flow control for agents, and data-use policies
enforce consent over records
\citep{ucon2004,zanzibar2019,airgapagent2024,camel2025,fides2025,datalawyer2015,thoth2016,odrl22,conseca2025,solidwac2024,solidoidc2022}.
Our failure-mode probes show that revoking consent over derived items needs the dependency set, which a
recorded read set only approximates (\cref{app:e7}).

\paragraph{Agent benchmarks.} Interactive benchmarks place agents among a person's apps and tools or
with simulated users \citep{appworld2024,toolsandbox2025,taubench2025,tau2bench2025}, and they evaluate
what an agent does in one episode. Carryover instead evaluates what a memory still knows after its inputs
have changed.

%% file: generated/tab_neighbours_full.tex
\begin{table*}[t]
\caption{\memshift{} against the closest benchmarks on every property of the audit, each read from the benchmark's own paper.}
\label{tab:neighbors-full}
\centering\mtsetup
\setlength{\tabcolsep}{3pt}
\begin{mttabular}{@{}l cccc ccccccc cc rrr@{}}
\toprule
& \multicolumn{4}{c}{Data} & \multicolumn{7}{c}{Changes tested} & \multicolumn{2}{c}{Protocol} & \multicolumn{3}{c@{}}{Scale} \\
\cmidrule(lr){2-5}\cmidrule(lr){6-12}\cmidrule(lr){13-14}\cmidrule(l){15-17}
Benchmark & \rotatebox{90}{World text} & \rotatebox{90}{\begin{tabular}{@{}l@{}}Gold fixed\\before text\end{tabular}} & \rotatebox{90}{\begin{tabular}{@{}l@{}}Gold\\dependency sets\end{tabular}} & \rotatebox{90}{Derived items} & \rotatebox{90}{Source revision} & \rotatebox{90}{Deletion} & \rotatebox{90}{Implicit conflict} & \rotatebox{90}{Redefinition} & \rotatebox{90}{Writer switch} & \rotatebox{90}{Export} & \rotatebox{90}{Consent revoked} & \rotatebox{90}{\begin{tabular}{@{}l@{}}Counterfactual\\branches\end{tabular}} & \rotatebox{90}{No LLM judge} & \multicolumn{1}{c}{\rotatebox{90}{Native systems}} & \multicolumn{1}{c}{\rotatebox{90}{Histories}} & \multicolumn{1}{c@{}}{\rotatebox{90}{Questions}} \\
\midrule
LoCoMo~\citep{maharana2024locomo} & L+H & $\sim$ & $\sim$ & $\sim$ & \textcolor{cgrey}{$\times$} & \textcolor{cgrey}{$\times$} & \textcolor{cgrey}{$\times$} & \textcolor{cgrey}{$\times$} & \textcolor{cgrey}{$\times$} & \textcolor{cgrey}{$\times$} & \textcolor{cgrey}{$\times$} & \textcolor{cgrey}{$\times$} & \checkmark & 0 & 50 & 7,512 \\
LongMemEval~\citep{wu2025longmemeval} & L+H & \checkmark & $\sim$ & $\sim$ & \checkmark & \textcolor{cgrey}{$\times$} & $\sim$ & \textcolor{cgrey}{$\times$} & $\sim$ & \textcolor{cgrey}{$\times$} & \textcolor{cgrey}{$\times$} & \textcolor{cgrey}{$\times$} & \textcolor{cgrey}{$\times$} & 2 & 500 & 500 \\
MemoryAgentBench~\citep{hu2025memoryagentbench} & M & $\sim$ & \textcolor{cgrey}{$\times$} & $\sim$ & \checkmark & \textcolor{cgrey}{$\times$} & $\sim$ & \textcolor{cgrey}{$\times$} & $\sim$ & \textcolor{cgrey}{$\times$} & \textcolor{cgrey}{$\times$} & \textcolor{cgrey}{$\times$} & $\sim$ & 10 & 130 & 2,071 \\
HaluMem~\citep{chen2025halumem} & L+H & $\sim$ & \checkmark & $\sim$ & \checkmark & $\sim$ & $\sim$ & \textcolor{cgrey}{$\times$} & \textcolor{cgrey}{$\times$} & \textcolor{cgrey}{$\times$} & \textcolor{cgrey}{$\times$} & $\sim$ & \textcolor{cgrey}{$\times$} & 6 & 20 & 3,467 \\
MemoryArena~\citep{he2026memoryarena} & M & $\sim$ & $\sim$ & \checkmark & $\sim$ & \textcolor{cgrey}{$\times$} & \textcolor{cgrey}{$\times$} & \textcolor{cgrey}{$\times$} & \textcolor{cgrey}{$\times$} & \textcolor{cgrey}{$\times$} & \textcolor{cgrey}{$\times$} & \textcolor{cgrey}{$\times$} & \textcolor{cgrey}{n/s} & 5 & \textcolor{cgrey}{n/s} & 701 \\
\mtgap
AMA-Bench~\citep{zhao2026amabench} & M & $\sim$ & $\sim$ & $\sim$ & \checkmark & \textcolor{cgrey}{$\times$} & $\sim$ & \textcolor{cgrey}{$\times$} & $\sim$ & \textcolor{cgrey}{$\times$} & \textcolor{cgrey}{$\times$} & \textcolor{cgrey}{$\times$} & \textcolor{cgrey}{$\times$} & 11 & 208+ & 3,696 \\
EvoArena~\citep{xu2026evoarena} & M & $\sim$ & $\sim$ & $\sim$ & \checkmark & $\sim$ & \checkmark & \textcolor{cgrey}{$\times$} & $\sim$ & \textcolor{cgrey}{$\times$} & $\sim$ & \textcolor{cgrey}{$\times$} & \checkmark & 1 & 10 & 505 \\
STALE~\citep{arxiv2605_06527} & L+H & \checkmark & \checkmark & $\sim$ & \checkmark & \textcolor{cgrey}{$\times$} & \checkmark & \textcolor{cgrey}{$\times$} & \textcolor{cgrey}{$\times$} & \textcolor{cgrey}{$\times$} & \textcolor{cgrey}{$\times$} & \textcolor{cgrey}{$\times$} & \textcolor{cgrey}{$\times$} & 5 & 400 & 1,200 \\
RECON~\citep{arya2026recon} & L & \checkmark & \checkmark & $\sim$ & $\sim$ & \checkmark & \checkmark & \textcolor{cgrey}{$\times$} & \textcolor{cgrey}{$\times$} & \textcolor{cgrey}{$\times$} & \textcolor{cgrey}{$\times$} & \textcolor{cgrey}{$\times$} & $\sim$ & 4 & 24 & 1,604 \\
DynamicMem~\citep{arxiv2606_22877} & L & \checkmark & $\sim$ & $\sim$ & \checkmark & \textcolor{cgrey}{$\times$} & \checkmark & \textcolor{cgrey}{$\times$} & \textcolor{cgrey}{$\times$} & \textcolor{cgrey}{$\times$} & \textcolor{cgrey}{$\times$} & \textcolor{cgrey}{$\times$} & \textcolor{cgrey}{$\times$} & 4 & 10 & 3,634 \\
\mtgap
Ground Truth First~\citep{arxiv2607_21962} & L & \checkmark & $\sim$ & $\sim$ & \checkmark & \textcolor{cgrey}{$\times$} & $\sim$ & \textcolor{cgrey}{$\times$} & \textcolor{cgrey}{$\times$} & \textcolor{cgrey}{$\times$} & \textcolor{cgrey}{$\times$} & $\sim$ & \textcolor{cgrey}{$\times$} & 0 & 20 & 383 \\
\citet{arxiv2609_05339} & \textcolor{cgrey}{n/s} & \checkmark$^\ddagger$ & $\sim$ & $\sim$ & \checkmark & \textcolor{cgrey}{$\times$} & $\sim$ & \textcolor{cgrey}{$\times$} & \checkmark & $\sim$ & \textcolor{cgrey}{$\times$} & \checkmark & \checkmark & 0 & 48 & 7,680 \\
\citet{arxiv2607_16019} & R & \textcolor{cgrey}{$\times$}$^\ddagger$ & \checkmark & $\sim$ & \checkmark & $\sim$ & \checkmark & \textcolor{cgrey}{$\times$} & \textcolor{cgrey}{$\times$} & \textcolor{cgrey}{$\times$} & \textcolor{cgrey}{$\times$} & $\sim$ & \textcolor{cgrey}{$\times$} & 0 & \textcolor{cgrey}{n/s} & 2,907 \\
MemSecBench~\citep{chen2026memsecbench} & L+H & $\sim$ & $\sim$ & \textcolor{cgrey}{$\times$} & $\sim$ & $\sim$ & $\sim$ & \textcolor{cgrey}{$\times$} & $\sim$ & \textcolor{cgrey}{$\times$} & \textcolor{cgrey}{$\times$} & $\sim$ & \textcolor{cgrey}{$\times$} & 4 & \textcolor{cgrey}{n/s} & 930 \\
GateMem~\citep{ren2026gatemem} & L+H & $\sim$ & $\sim$ & $\sim$ & \checkmark & \checkmark & \textcolor{cgrey}{$\times$} & \textcolor{cgrey}{$\times$} & $\sim$ & \textcolor{cgrey}{$\times$} & \checkmark & \textcolor{cgrey}{$\times$} & \textcolor{cgrey}{$\times$} & 3 & 91 & 2,218 \\
\midrule
\textbf{\memshift{}} & T\,(+L) & \checkmark & \checkmark & $\sim$ & \checkmark & \checkmark & $\sim$ & \checkmark & \checkmark & $\sim$ & \checkmark & \checkmark & \checkmark & 6 & 40 & 19,562 \\
\bottomrule
\end{mttabular}
\mtnotes{\checkmark{}~yes, $\sim$~partial, $\times$~no such element, n/s~not stated, $^\ddagger$~inferred by the audit, not stated in the paper. World text: T~template-rendered, L~written by an LLM, L+H~by an LLM with human editing or review, R~real records, M~mixture of these or of existing datasets, (+L)~plus an experiment in which an LLM rewrote the message bodies. \memshift{}: test worlds and graded probes over 1,920 knowledge items.}
\end{table*}

%% file: appendix/app_benchmark.tex
\section{Benchmark Construction}
\label{app:worlds}

This section specifies the worlds, items, gold labels, branches, and scoring of \cref{sec:benchmark}. It
follows one item of a released world through a redefinition. It also describes the variants used in the
further experiments: held-out definitions, definitions in the user's own words, unnamed probes,
rewritten message bodies, and simulated sync. Generation is deterministic, so a seed and a generator
version reproduce a world byte for byte.

\paragraph{A world is a log of versioned source objects.} We sample the true changes of a world before
any text exists. Each change inserts, updates, or deletes a source object (called a record in the main
text). Each source object has an object id and a revision number that only increases, and a deletion
leaves a tombstone. Deterministic templates render the objects as text. Background traffic adds
distracting text: newsletters that mention the same organizations and cities, small talk, routine work
mail, and one-off meetings. A fidelity checker confirms that every gold value can be found in the text
of its dependencies. It succeeded on \WorldFidelity{} of the sampled items and times in the development
worlds and on \TestFidelity{} gold facts over all branches of the test worlds. A rendered world that fails
the check is not used.

\paragraph{Forty-eight items.} Each world has six attributes of the person (employer, manager, home
city, time zone, diet, and commute) and an employer and a city for each of twelve contacts. It also has
four recurring series, the next dentist and doctor appointments, and six task due dates. The six
definition items are important contacts, close friends, active projects, working hours, VIP senders, and
overdue tasks.

\paragraph{Gold is computed from the live source objects.} For every item and time, an evaluator returns
the gold value, the gold status, and the dependency set, that is, the source objects with their
revisions. A person attribute follows the world's script, and a contact attribute is read from the
contact card. A series or a due date is read from its object, and an appointment is the earliest
upcoming one of its type. A definition item applies the definition version in effect. An item with no
current value, because its object is deleted or no appointment lies ahead, has the gold status
\texttt{abstain}. Each definition has a version v1 and a version v2 over structured fields of the
sources: calendar entries, contact cards, tasks, message and e-mail headers, and a working-hours note.
For example, v1 of important contacts counts one-off meetings over a longer window than v2 and requires
more of them. No annotator labels an item.

\paragraph{Branches differ only in the shift.} A world forks at the end of month nine. Each shift draws
from its own random stream, so all branches of a world share every other event byte for byte. A source
revision ($\Ssrc$) revises, through ordinary events, the sources behind a sampled share of the items.
These events are a job change (with a new manager) or a reorganization, a relocation (with a new time
zone), a new diet or commute, an edited contact card, a rescheduled or canceled series, a booked, moved,
or canceled appointment, and a new due date. For a definition item, they are new meetings, messages,
e-mails, tasks, or a new working-hours note. A redefinition ($\Sdef$) moves half of the definitions to v2
at the fork and adds no source event, and gold is evaluated under v2. A deletion ($\Sdel$) deletes the
source object behind half of the items read from a series, task, or contact card. A consent revocation
($\Scons$) revokes one scope of one agent. Probes asked by that agent whose dependencies touch the scope
have the gold status \texttt{blocked}. The conflict branch of the failure-mode probes adds, on the same
day, an update to a contact card and a message from that contact. The two live sources state different
values for the contact's city or employer, and gold is \texttt{abstain}. A writer switch ($\Swr$) and an
export ($\Simpl$) change the memory, not the world, so they need no branch of their own. The unshifted
branch has the same events as the shifted branches up to the fork and no shift after it.

\paragraph{An unstated redefinition limits carryover.} If the person does not state a redefinition, no
memory can observe it. In the failure-mode probes, carryover after an unstated redefinition is thus at
most \ESevenImplicitCeiling{} of \ESevenImplicitItems{} items, the share whose old answer equals the new
gold.

\paragraph{Scoring is typed matching.} No LLM judge scores an answer. Dates, times, and weekdays are
parsed and compared as values. Entities are compared after normalizing case, punctuation, and
whitespace. Diet and commute values use a fixed table of synonyms, and sets are compared as sets. An
answer is correct only if its status also equals the gold status.

\input{generated/example_world}
\paragraph{A worked example.} \Cref{tab:example} follows one definition item of a released test world
through a redefinition. It shows the definition in effect, the records that decide the gold value, the
item that every \holdmem{} cell stored before the fork, and the answers of three cells before and after
the redefinition.

\begin{table*}[t]
\caption{$B{\times}D$ on Update per definition: (a) the six definitions, (b) the \RThreeGADefDoneDefinitions{}
held-out definitions in three draws of six, each on its own changed item-times with exact recomputation
as $D{=}1$ (\RThreeGAWorlds{} worlds, point estimates).}
\label{tab:perdef}\label{tab:heldout}
\centering\mtsetup
\setlength{\tabcolsep}{4pt}
\input{generated/tab_perdef}
\par\medskip
\input{generated/tab_heldout}
\end{table*}
\paragraph{Held-out definitions come from a grammar.}\label{app:heldout-grammar} A grammar samples a
definition as the records of one source that meet a condition, an aggregate over them (count, members,
or span), a window, and a threshold. Its pairs of source and condition differ from those of the six
fixed definitions. The second version changes the window or the threshold, and, as for the six, a
redefinition moves half of a world's definitions. A filter computed from gold alone rejects candidates
whose gold value rarely changes or is mostly empty. It would also reject two of the six fixed
definitions. The windows and thresholds of the grammar were set from gold statistics alone, and no
model output enters the draws. Each of three draws of six definitions (\cref{tab:heldout}) replaces the
six fixed definitions on ten test worlds.

\paragraph{Definitions in the user's own words keep the codified content.}\label{app:userwords} The
program request of the writer's programs states each definition in a codified form. For this experiment,
we restated each of the six definitions as the user would say it, with the same sources, thresholds,
windows, and conditions. For important contacts, the first version reads ``Count someone as an
important contact if they've been at five or more of my one-off meetings in the past 90 days. Recurring
series don't count.'' The record fields, the item's sources, and the task text, which form the memory's
own data model, stay as they are.

\paragraph{Unnamed probes leave out the definition.} A probe about a definition item ends with a sentence
that names the definition in effect. An unnamed probe leaves out this sentence and still asks for the
value under the user's current definition.

\paragraph{Rewritten worlds keep every gold fact verbatim.}\label{app:rewritten} gemma-4-12B rewrites the
bodies of e-mails, chat messages, and notes at temperature~0. Headers and every field read by gold, the
evaluators, and the shift scripts keep their template values, so gold is the same as in the template
world. Structured records, delete events, and newsletter digests stay templated. The digests hold no
gold fact and make up \RendDigestTokenShareMin{} to \RendDigestTokenShareMax{} of the body text. A
rewritten body is accepted only if the reply is complete JSON and every gold fact that gold reads from
the record appears verbatim in one of its surface forms. The rewrite must also add no number, month, or weekday name, and no
person, company, or city from the generator's pools. A failing body is retried once and then kept as
the template. A rewritten world is used only if the fidelity checker finds every gold value (or as many
as in the template world).

\paragraph{Sync simulators follow documented contracts.}\label{app:sync-sim} The sync experiment
replaces ideal delivery with the documented incremental-sync contract of a widely used API for each
source \citep{google_calendar_sync,sys_google_people,sys_gmail}. These are Google Calendar
\texttt{events.list} (sync token, full resync on \texttt{410}), People \texttt{connections.list} (sync
token, 7-day expiry), Gmail \texttt{history.list} (history id, full resync on \texttt{404}), Google Tasks
\texttt{tasks.list} (\texttt{updatedMin}), Google Drive \texttt{changes.list} (page token), and Slack
\texttt{conversations.history} (\texttt{oldest}). Deletions arrive as tombstones except in chat. WebDAV
sync \citep{rfc6578} (used for CalDAV \citep{rfc4791}), JMAP \citep{sys_jmap}, and Microsoft Graph delta
queries \citep{sys_microsoft_graph_delta} follow the same pattern. A client polls every 15 minutes in
pages of 50 with a visibility lag $L$, so cursors never expire. The only full resyncs are the calendar's
documented invalidations of its sync token (probability 0.001 per poll).

\paragraph{Worlds are seeded and versioned.}\label{app:reuse} In the two closest generators, events are
text without object identity \citep{arxiv2607_21962}, or they come from a language model called at
every stage with an unseeded random number generator \citep{arxiv2606_22877}. Both judge answers with a
language model. Our generator is seeded and versioned, and its events are typed operations on versioned
objects, with object revisions, deletions, versioned definitions, and per-agent consent.

%% file: generated/example_world.tex
\begin{table*}[t]
\caption{A worked example: the definition item of the user's active projects in one released test world, before the fork and one week after a redefinition, in the main experiment with Qwen3.5-9B as writer and reader.}
\label{tab:example}
\centering\mtsetup
\setlength{\tabcolsep}{4pt}
\newcolumntype{P}[1]{>{\raggedright\arraybackslash}p{#1}}
\newcolumntype{C}[1]{>{\centering\arraybackslash}p{#1}}
\def\exG{\dimexpr(\linewidth-12.9cm-8\tabcolsep)/2\relax}
\begin{mttabular}{@{}P{2.0cm} P{2.3cm}P{\dimexpr3.6cm-2\tabcolsep\relax} P{1.6cm}P{\dimexpr3.4cm-2\tabcolsep\relax} C{\exG}C{\exG}@{}}
\toprule
question & \multicolumn{6}{P{\dimexpr\linewidth-2.0cm-2\tabcolsep\relax}@{}}{Which projects are the user's active projects under the user's current definition? List project names.} \\
\midrule
 & \multicolumn{2}{P{5.9cm}}{Before the fork, probe on Sep 30, 2026} & \multicolumn{4}{P{\dimexpr\linewidth-2.0cm-5.9cm-4\tabcolsep\relax}@{}}{One week after the redefinition, probe on Oct 8, 2026} \\
\cmidrule(l){2-7}
definition & \multicolumn{2}{P{5.9cm}}{v1: An active project is a project with at least one open task.} & \multicolumn{4}{P{\dimexpr\linewidth-2.0cm-5.9cm-4\tabcolsep\relax}@{}}{v2: An active project is a project with at least one task created or updated in the last 14 days (any status).} \\
\midrule
 & Project & Task & Status & Last updated & Gold v1 & \multicolumn{1}{c@{}}{Gold v2} \\
\cmidrule(l){2-7}
records & Beacon & Prepare design doc & open & Jun 28 & $\bullet$ &  \\
 & Delta Migration & Present customer survey & open & Aug 17 & $\bullet$ &  \\
 & Horizon Audit & Test design doc & open & Sep 29 & $\bullet$ & $\bullet$ \\
 & Horizon Audit & Update hiring plan & open & Oct 5 &  & $\bullet$ \\
\midrule
 & \multicolumn{2}{P{5.9cm}}{Before the fork} & \multicolumn{2}{P{5.0cm}}{After the redefinition} & \multicolumn{2}{P{\dimexpr2\exG+2\tabcolsep\relax}@{}}{What the memory did} \\
\cmidrule(l){2-7}
gold & \multicolumn{2}{P{5.9cm}}{Beacon, Delta Migration, Horizon Audit} & \multicolumn{2}{P{5.0cm}}{\textbf{Horizon Audit}} & \multicolumn{2}{P{\dimexpr2\exG+2\tabcolsep\relax}@{}}{} \\
stored item & \multicolumn{2}{P{5.9cm}}{``As of 2026-09-30, the user's active projects are Beacon, Delta Migration, and Horizon Audit, as these are the only projects with at least one open task.''} & \multicolumn{2}{P{5.0cm}}{} & \multicolumn{2}{P{\dimexpr2\exG+2\tabcolsep\relax}@{}}{} \\
\addlinespace[2pt]
\holdmem{}(0,0) & \multicolumn{2}{P{5.9cm}}{Beacon, Delta Migration, Horizon Audit \pass} & \multicolumn{2}{P{5.0cm}}{abstains \fail} & \multicolumn{2}{P{\dimexpr2\exG+2\tabcolsep\relax}@{}}{kept} \\
\holdmem{}(0,1) & \multicolumn{2}{P{5.9cm}}{Beacon, Delta Migration, Horizon Audit \pass} & \multicolumn{2}{P{5.0cm}}{Garnet, Beacon, Delta Migration, Horizon Audit \fail} & \multicolumn{2}{P{\dimexpr2\exG+2\tabcolsep\relax}@{}}{invalidated, re-derived} \\
\holdmem{}(1,1) & \multicolumn{2}{P{5.9cm}}{Beacon, Delta Migration, Horizon Audit \pass} & \multicolumn{2}{P{5.0cm}}{\textbf{Horizon Audit} \pass} & \multicolumn{2}{P{\dimexpr2\exG+2\tabcolsep\relax}@{}}{invalidated, re-derived} \\
\bottomrule
\end{mttabular}
\mtnotes{Gold v1, v2: $\bullet$ when the record is in the dependency set of the gold value under that version. \pass{}~or~\fail{}: status and value match gold or not. Bold: the changed gold value and the answer that matches it.}
\end{table*}

%% file: generated/tab_perdef.tex
\begin{mttabular}{@{}l r@{\,}l N{000} @{\hspace{14pt}} r@{\,}l N{000} @{\hspace{14pt}} r@{\,}l N{000}@{}}
\toprule
& \multicolumn{3}{c}{$\Ssrc$} & \multicolumn{3}{c}{$\Sdef$} & \multicolumn{3}{c@{}}{$\Sdel$} \\
\cmidrule(r){2-4}\cmidrule(r){5-7}\cmidrule{8-10}
(a) Definition & \multicolumn{2}{c}{$B{\times}D$} & \multicolumn{1}{c}{$n$} & \multicolumn{2}{c}{$B{\times}D$} & \multicolumn{1}{c}{$n$} & \multicolumn{2}{c}{$B{\times}D$} & \multicolumn{1}{c@{}}{$n$} \\
\midrule
important contacts & $+$75.4 & \mtint{[$+$60.3,\,$+$88.1]} & 61 & $+$60.3 & \mtint{[$+$48.5,\,$+$73.8]} & 63 & $+$72.4 & \mtint{[$+$57.4,\,$+$85.2]} & 58 \\
close friends & $+$35.4 & \mtint{[$+$26.6,\,$+$44.3]} & 79 & $+$35.4 & \mtint{[$+$28.2,\,$+$42.5]} & 79 & $+$42.3 & \mtint{[$+$34.2,\,$+$50.0]} & 78 \\
working hours & 0.0 & & 22 & $+$30.9 & \mtint{[$+$18.6,\,$+$43.6]} & 55 & 0.0 & & 6 \\
active projects & $-$20.0 & \mtint{[$-$55.3,\,$+$10.4]} & 40 & $+$59.1 & \mtint{[$+$29.5,\,$+$82.6]} & 44 & $+$2.9 & \mtint{[$-$31.0,\,$+$34.2]} & 35 \\
VIP senders & $+$4.3 & \mtint{[0.0,\,$+$9.4]} & 70 & $+$2.7 & \mtint{[0.0,\,$+$6.7]} & 74 & $+$2.8 & \mtint{[$-$2.8,\,$+$10.1]} & 72 \\
overdue tasks & $+$36.1 & \mtint{[$+$23.2,\,$+$49.3]} & 61 & $+$54.2 & \mtint{[$+$37.5,\,$+$69.0]} & 72 & $+$52.5 & \mtint{[$+$34.5,\,$+$69.0]} & 59 \\
\midrule
leave one out & \multicolumn{3}{c}{$+$16.5 to $+$33.8} & \multicolumn{3}{c}{$+$34.6 to $+$47.3} & \multicolumn{3}{c@{}}{$+$26.8 to $+$45.3} \\
\bottomrule
\end{mttabular}
\mtnotes{$n$: the definition's changed item-times. Leave one out: the smallest and the largest $B{\times}D$ of the definition class with one definition removed. The intervals resample worlds, not definitions.}

%% file: generated/tab_heldout.tex
\begin{mttabular}{@{}c l l l l r@{\;$\to$\;}l N{$+$000.0}N{$+$000.0}N{$+$000.0}@{}}
\toprule
& & \multicolumn{2}{c}{Records} & & \multicolumn{2}{c}{} & \multicolumn{3}{c@{}}{$B{\times}D$} \\
\cmidrule(lr){3-4}\cmidrule(l){8-10}
(b) Draw & Definition & Source & Condition & Aggregate & \multicolumn{1}{r@{\;$\to$\;}}{v1} & v2 & \multicolumn{1}{c}{$\Ssrc$} & \multicolumn{1}{c}{$\Sdef$} & \multicolumn{1}{c@{}}{$\Sdel$} \\
\midrule
1 & priority project & tasks & high priority task & count & ${\ge}$1 in 30\,d & 45\,d & $+$73.3 & $+$107.7 & $+$106.2 \\
 & social hours & calendar & social meeting & span & 21\,d & 14\,d & $+$26.9 & $+$28.6 & $+$29.2 \\
 & recently completed task & tasks & done task & members & 7\,d & 14\,d & $+$33.3 & $+$43.8 & $+$53.1 \\
 & evening chat partner & messages & evening message & count & ${\ge}$1 in 21\,d & ${\ge}$3 & $+$2.5 & $+$7.5 & $+$2.5 \\
 & after hours sender & email & after hours email & count & ${\ge}$1 in 90\,d & ${\ge}$2 & $+$4.3 & $+$3.6 & 0.0 \\
 & team regular & calendar & group meeting & count & ${\ge}$1 in 90\,d & ${\ge}$2 & $+$13.3 & $+$35.3 & $+$6.7 \\
\mtgap
2 & frequent recipient & messages & sent message & count & ${\ge}$2 in 60\,d & ${\ge}$4 & 0.0 & $+$5.4 & 0.0 \\
 & deep dive partner & calendar & long meeting & count & ${\ge}$2 in 60\,d & 90\,d & $+$36.4 & $+$37.9 & $+$39.4 \\
 & after hours sender & email & after hours email & count & ${\ge}$3 in 90\,d & ${\ge}$2 & 0.0 & 0.0 & 0.0 \\
 & productive project & tasks & done task & count & ${\ge}$1 in 30\,d & 60\,d & $+$51.5 & $+$50.0 & $+$80.0 \\
 & team regular & calendar & group meeting & count & ${\ge}$2 in 45\,d & ${\ge}$1 & $+$40.7 & $+$41.0 & $+$44.4 \\
 & priority project & tasks & high priority task & count & ${\ge}$1 in 45\,d & ${\ge}$2 & $+$64.3 & $+$68.2 & $+$93.8 \\
\mtgap
3 & frequent recipient & messages & sent message & count & ${\ge}$3 in 21\,d & ${\ge}$2 & $+$20.5 & $+$10.3 & $+$25.6 \\
 & social hours & calendar & social meeting & span & 45\,d & 14\,d & $+$11.1 & $+$25.0 & $+$6.2 \\
 & upcoming task & tasks & due ahead & members & 7\,d & 21\,d & $+$53.3 & $+$63.2 & $+$65.5 \\
 & recently completed task & tasks & done task & members & 14\,d & 21\,d & $+$30.8 & $+$50.0 & $+$42.1 \\
 & hot task & tasks & high priority task & members & 7\,d & 60\,d & $+$20.0 & $+$29.4 & $+$18.2 \\
 & close relative & messages & family message & count & ${\ge}$8 in 30\,d & 60\,d & $+$33.3 & $+$38.5 & $+$45.5 \\
\bottomrule
\end{mttabular}
\mtnotes{Records: the source and the condition the records meet. Count: the people or projects with at least the threshold of these records within the window. Members: the tasks. Span: the earliest start to the latest end of the meetings, in whole hours. v2 changes the window or the threshold. $B{\times}D$: point estimates. A per-item double difference lies between $-$200 and $+$200\,pp.}

%% file: appendix/app_systems.tex
\section{Memory Systems, Baselines, and Models}
\label{app:systems}

This section describes the reference memory \holdmem{} and its cells, the baselines, the writer's
programs, the refresh and upsert cells, and the controls. It then describes the native systems, the
placement of real memory systems on the $(D,B)$ grid, and the models.

\subsection{\holdmem{} Cells}
\label{app:holdmem}

\paragraph{Four cells and two keys.} The cells are $(D,B)\in\{0,1\}^2$. A $B{=}1$ key is either a source
key or a citation key, which gives the six cells of the main experiment. The source key holds the number
of events that each of the item's sources has received, the definition version, and the writer. The
citation key holds the same count for each record group, such as a message thread or a contact card,
that contains a record the writer cited, together with the definition version and the writer. An item
is the tuple
$\{\texttt{item\_id},\allowbreak \texttt{statement},\allowbreak \texttt{key}\,|\,\emptyset,\allowbreak \texttt{typed}\,|\,\emptyset,\allowbreak \texttt{status},
\allowbreak \texttt{history}\}$. The key is empty at $B{=}0$. The typed value with its located spans (and
the writer's program at $D_{\mathrm{m}}$) is empty at $D{=}0$.

\paragraph{Writing and maintaining an item.} To write an item, the writer receives a BM25 selection of
the records of the item's sources (6,000-token budget, in time order). It returns JSON with a typed
value, a one-sentence statement, and up to six cited records with exact quotes (\cref{app:prompts}). A
$D{=}1$ cell also keeps the typed value with the located spans of its quotes. A checker verifies the
value against the live records at write time and at repair time. If the check fails, exact
recomputation supplies the value. Exact recomputation has no rule for the five free-text items, so for
them $D{=}1$ only checks, and an item that fails the check is marked unverifiable and not served. During
maintenance, a $B{=}1$ item whose key changed is invalidated and re-derived by the current writer. A
$B{=}0$ cell derives new items from the delta, that is, the records that arrived since its last write. It
skips the writer when the delta holds no source record, and it never invalidates. After a writer switch
it re-derives nothing and keeps every answer, so we leave it out of the retention test,
which requires that at least 97\% of the correct answers survive a switch (\cref{app:experiments}). With
exact recomputation, the stored statement names the definition version it evaluated. The template
control measures what this presentation adds (\cref{app:res-repair}).

\paragraph{Every cell starts from the same reader context.} All cells are built from the same writer
outputs. Before any intervention, the reader context of every probe is hashed in every cell, and a run
stops unless all hashes match. After a shift, the cells thus differ only in what the key
invalidates and in how an invalidated item is re-derived.

\paragraph{Why the citation key counts only cited record groups.}\label{app:granularity} A key over
every record group that a derivation read would keep firing on busy groups that read sets include, such
as newsletter threads. The citation key thus counts only the groups of the records the writer
cited, in both $D$ arms. On the test worlds, the source key invalidated \EOneGranFalseSrc{} more of the
unchanged items than the citation key after a source revision. \Cref{app:res-keys} shows what the
citation key misses and why.

\paragraph{Baselines.} The main study includes no-store BM25, plain RAG, and the oracle context.
No-store BM25 keeps nothing and runs BM25 over the live records at every query. Plain RAG retrieves
records with dense embeddings. The oracle context gives the reader the records that the gold label
depends on. The JSON arm serves the same $D{=}1$ store as JSON records instead of statements.

\subsection{The Writer's Programs and the Check-Only Cell}
\label{app:programs}

The $D{=}1$ cell repairs a failed span check with exact recomputation. For definition items, both the
check and the repair call the evaluator that computes gold. The $D_{\mathrm{m}}$ cell keeps the checker
but takes the replacement value from a program. The writer model writes this program in one extra
request per item and definition version (\cref{app:prompts}), and the program runs on the typed records
of the item's sources. The cells are $D_{\mathrm{m}}$ with $B{=}0$, with the source key, and with the
citation key, on the 20 test worlds in the three branches. They are built from the writer outputs of the
main experiment, and the keys come only from these outputs, so $D$ cannot affect detection. The program
request gives the item, the definition in effect with its version, and the typed fields of the item's
sources. It holds no world data, so identical items share one program. A source change runs the program
again, and a definition change requests a new program.

Programs run in a separate process with a memory cap and a 2\,s limit on CPU time. A restricted
interpreter accepts only listed syntax and built-in functions, with no imports, file access, reflection,
or wall clock. A program that raises an exception, times out, or returns the wrong type has failed.
Definition items are never checked against the gold evaluator. The writer's value passes if it equals
the output of its own program. Otherwise the program's output replaces it. Other items keep the span
check of $D{=}1$, and if the check fails, the program's output replaces the value. An item whose program
failed when its output was needed is unverifiable. The check-only cell keeps the checker of $D{=}1$ but
derives no replacement value.

\subsection{Refresh and Upsert Cells}
\label{app:refresh-cells}

At each checkpoint, a refresh cell re-derives every item whose source key changed, with the request of
the $B{=}1$ cell, and appends the new item. Nothing is invalidated. An upsert cell sends the requests of
the append-only $B{=}0$ cell, and a new item for a slot (its \texttt{item\_id}) retires the slot's older
items. Both reuse the writes of the existing cells, and both run at $D{=}0$, $D{=}1$, and
$D_{\mathrm{m}}$ on the 20 test worlds. The add-only refresh control below is the same refresh at
$D{=}1$ on \CtlWorlds{} worlds.

\subsection{Controls}
\label{app:controls}
\begin{table*}[t]
\caption{The keys. (a) What each key invalidates on the test worlds, audited without a model. (b) The
oracle key against the source key as cells on \ExDetKeyWorlds{} test worlds: the effect of $B$ on Update
and $B{\times}D$ (pp).}
\label{tab:keyaudit}\label{tab:oracle-key}
\centering\mtsetup
\setlength{\tabcolsep}{4pt}
\input{generated/tab_keyaudit}
\par\medskip
\begin{mttabular}{@{}ll r@{\,}l r@{\,}l r@{\,}l@{\hspace{14pt}}r@{\,}l@{}}
\toprule
& & \multicolumn{6}{c}{Definition items} & \multicolumn{2}{c@{}}{All items} \\
\cmidrule(lr){3-8}\cmidrule(l){9-10}
(b) Shift & Key & \multicolumn{2}{c}{$B$ at $D{=}0$} & \multicolumn{2}{c}{$B$ at $D{=}1$}
  & \multicolumn{2}{c}{$B{\times}D$} & \multicolumn{2}{c@{}}{$B{\times}D$} \\
\midrule
$\Ssrc$ & oracle & \mtci{\ExDetKeyOracleSrcDefinitionBDzero} & \mtci{\ExDetKeyOracleSrcDefinitionBDone}
  & \mtci{\ExDetKeyOracleSrcDefinitionBxD} & \mtci{\ExDetKeyOracleSrcAllBxD} \\
 & source & \mtci{\ExDetKeySourceSrcDefinitionBDzero} & \mtci{\ExDetKeySourceSrcDefinitionBDone}
  & \mtci{\ExDetKeySourceSrcDefinitionBxD} & \mtci{\ExDetKeySourceSrcAllBxD} \\
\mtgap
$\Sdef$ & oracle & \mtci{\ExDetKeyOracleDefDefinitionBDzero} & \mtci{\ExDetKeyOracleDefDefinitionBDone}
  & \mtci{\ExDetKeyOracleDefDefinitionBxD} & \mtci{\ExDetKeyOracleDefAllBxD} \\
 & source & \mtci{\ExDetKeySourceDefDefinitionBDzero} & \mtci{\ExDetKeySourceDefDefinitionBDone}
  & \mtci{\ExDetKeySourceDefDefinitionBxD} & \mtci{\ExDetKeySourceDefAllBxD} \\
\mtgap
$\Sdel$ & oracle & \mtci{\ExDetKeyOracleOrphanDefinitionBDzero} & \mtci{\ExDetKeyOracleOrphanDefinitionBDone}
  & \mtci{\ExDetKeyOracleOrphanDefinitionBxD} & \mtci{\ExDetKeyOracleOrphanAllBxD} \\
 & source & \mtci{\ExDetKeySourceOrphanDefinitionBDzero} & \mtci{\ExDetKeySourceOrphanDefinitionBDone}
  & \mtci{\ExDetKeySourceOrphanDefinitionBxD} & \mtci{\ExDetKeySourceOrphanAllBxD} \\
\bottomrule
\end{mttabular}
\end{table*}

Six controls vary one confound at a time. The oracle key is built from the gold dependencies. The
scope key is built from each definition item's definition scope, the records its definition ranges
over. The add-only refresh adds the new derivation without invalidating the old one. The template
control renders the $D{=}0$ value in the $D{=}1$ template. The query-time view is no-store BM25 plus the
writer's program, run on the live records at query time. The controls ran on \CtlWorlds{} test worlds.
The oracle, scope, and source keys are functions of the world, so we audit their detection offline,
without a model (\cref{tab:keyaudit}).

\paragraph{The full-access writer.}\label{app:full-access} This $D{=}0$ writer reads every live record
of the item's sources, not only its definition scope, up to \CtlFullLimit{} tokens. When the input is too
long, we keep the newest records of the largest source. This cut affects only close friends and VIP
senders, in \CtlFullTruncatedShareAll{} of the re-derivations. For active projects and overdue tasks,
whose records the $D{=}0$ selection already held in full, its request equals the $D{=}0$ request
(\cref{app:coverage}).

\subsection{Native Systems}
\label{app:native-design}

\paragraph{Systems and integration.} We ask whether the $(D,B)$ setting of a native memory system, read
from its documentation and code, predicts how the system does after a shift. The systems are Mem0
\citep{arxiv2504_19413} and five others: A-Mem \citep{xu2025amem}, MemOS \citep{li2025memos}, LightRAG
\citep{guo2025lightrag}, HippoRAG~2 \citep{gutierrez2025hipporag2}, and Cognee \citep{arxiv2505_24478}.
MemOS runs its server-free \texttt{general\_text} memory, because \texttt{tree\_text} needs a graph
server. With its reorganizer off, which is the default, \texttt{tree\_text} updates in the same way. Each
system uses the main writer, with thinking off, as its LLM. Where a system accepts an embedder, it uses
Qwen3-Embedding-0.6B \citep{arxiv2506_05176}. We use only each system's retrieval context. This context
is packed into the 4,096-token evidence region, and the shared reader answers from it, so only storage
and retrieval differ. Ingestion failures count as system behavior. Graphiti \citep{arxiv2501_13956},
whose LLM invalidates contradicted facts, made \EFiveCostGraphitiCalls{} LLM calls per event with the
same model and stopped on invalid JSON after \EFiveCostGraphitiUnits{} events. Since the tested Mem0
release only adds (see below), no native run has an LLM decide to invalidate a stored value.

\paragraph{Update contracts.} The adapter passes each change of the event stream, in time order, through
the system's own API. Document systems (LightRAG, HippoRAG~2, and Cognee) keep one document per source
object. A revision replaces the document through a delete and an insert, and a deletion removes it.
Mem0, A-Mem, and MemOS receive each revision through their add path and each deletion as tombstone
text. Outside the Mem0 invalidation arm (\cref{app:native}), the adapter never calls Mem0's update or
delete API, and the benchmark never invalidates on a system's behalf. The scores thus describe these
integrations.

\paragraph{The tested Mem0 release only adds.} Its add path with inference makes one extraction call,
which can only add memories. It never runs the step of the Mem0 paper in which the LLM decides to add,
update, or delete a memory, and all its history tables hold additions only. Mem0's $m{=}1$ thus
describes this release, not the design in the Mem0 paper. Mem0 ran without its optional spaCy and
fastembed extras, so its search is dense only, without the entity boost.

\input{generated/audit_table}

\paragraph{Worlds, reference cells, and settings.} Each of the fifteen three-month test worlds is
ingested up to one minute before the fork, and each branch is ingested from that snapshot. Probes are
one, two, four, and eight weeks after the fork (one week after a deletion). The reference cells are the
four source-key cells on the same worlds plus the citation-key cells $(0,1)$ and $(1,1)$, whose
granularity is closest to a key per document. \Cref{tab:audit} lists the documented settings together with
the audited systems of \cref{app:audit}. We infer the setting on one half of the items. A system gets
$B{=}1$ if, after the items' dependency objects are deleted in the way the adapter delivers a deletion
(a delete call or a tombstone), it still gives the old value with probability at most 0.5. This needs at
least 20 valid samples, and otherwise the setting is unidentifiable. $D{=}1$ would need a structured-repair
precision of at least 0.98 and an export with typed values and locatable source spans. On the
\holdmem{} cells, whose settings are known, Cohen's $\kappa$ between the inferred and the known setting was
\EFiveKappaBDev{} and \EFiveKappaBShort{} for $B$ (development and three-month worlds) and
\EFiveKappaDDev{} and \EFiveKappaDShort{} for $D$. Both values for $D$ are below 0.7, so we read $D$ from
the export structure instead. Carryover is measured on the other half of the items. Two diagnostic tests
(\cref{app:experiments}), the utility floor (accuracy before any shift at least that of the oracle
context minus 3\,pp) and the cost ceiling (a serving cost per query of at most half a full re-read), are
computed as in the horizon experiment, with the oracle context of these worlds as reference.

\input{generated/tab_models}

\subsection{Memory Systems on the \texorpdfstring{$(D,B)$}{(D,B)} Grid}
\label{app:audit}

\Cref{tab:audit} places fifteen real memory systems and plain RAG on the $(D,B)$ grid, together with the
four further systems of the native-system experiment and our own cells. Each placement rests on code, a
paper, documentation, or indirect evidence (what the documentation implies without stating it).
Documents are as of 2026-09-28, and the settings refer to the releases current on that date. The released
configuration pins the exact version of each system we run, and the released audit record gives the
reasons and pointers for every placement.
\begin{figure*}[t]
\centering
\includegraphics[width=\linewidth]{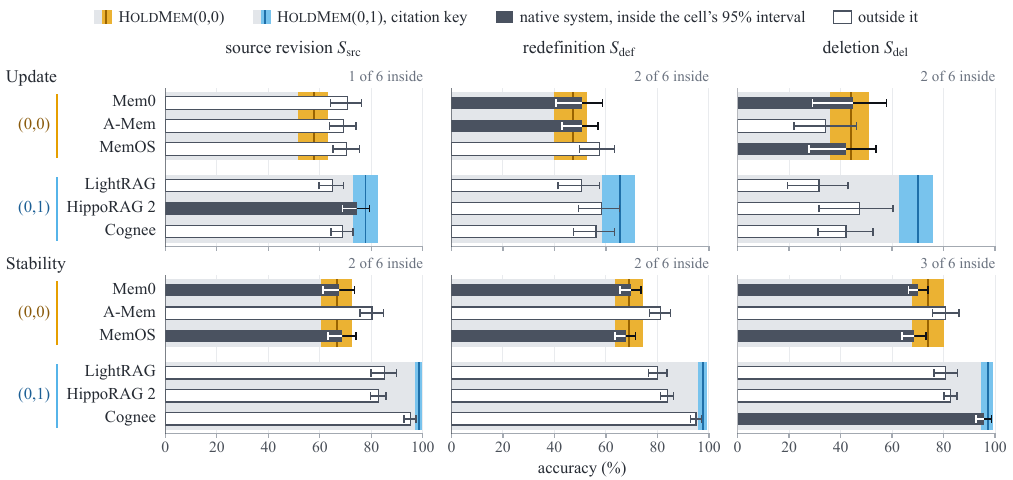}
\caption{Update and Stability of each native memory system and of the \holdmem{} cell with its inferred
setting, after each shift. Wide gray bars: the cell's estimate, with its 95\% interval as a band in the
color of its $B$ arm. Narrow bars: the native system, filled if its estimate lies inside the cell's
interval and outlined if not. Error bars: 95\% intervals.}
\label{fig:native-systems}
\end{figure*}

The audit reads $D$ and $B$ by one rubric. Partial $D$ means typed structure or citations without a
span-level check. A key that lacks one of its components (source event counts or read set, definition
version, or writer) counts as $B{=}0$, and temporal validity means invalidation only on contradicting
information. The native-system experiment instead assigns a system its setting at the document level and infers the
setting from measurements (\cref{fig:native-systems}), so the two settings can differ, as they do for Cognee.
The six systems we run all have $D{=}0$, because no export keeps a typed value with a source span.

\subsection{Models and Serving}
\label{app:models}

\Cref{tab:models} lists every model that the paper uses. Writer and reader are Qwen3.5-9B unless stated
otherwise. The writer-switch tiers use MiMo-V2.6-Distill-Qwen-9B and Qwen3.8-27B. Ministral-3-14B is the
replacement reader in the failure-mode probes, and bge-m3 replaces Qwen3-Embedding-0.6B in the embedding
swap. The other model families serve as new writers, as writer and reader, and as readers. The reasoning
of gpt-oss is removed from its reply before parsing. The released run records pin the exact model
revisions and software versions.

Claude Sonnet 5.5 and Claude Opus 5.5 run through Anthropic's API, and all their responses are released
with the code. The API accepts neither a temperature nor a thinking budget for these models, so they use
its default sampling. As reader, Sonnet returns its answer under a JSON schema. As writers of definition
items, both think adaptively at high effort with at most 17,000 output tokens. Their program samples use
the default sampling, while the open-weight writers sample at temperature 0.7.

%% file: generated/tab_keyaudit.tex
\begin{mttabular}{@{}l N{0,000}N{0,000} l N{100.0}@{\hspace{11pt}}N{100.0}@{\hspace{11pt}}N{100.0}@{\hspace{11pt}}N{100.0} @{\hspace{14pt}} N{0,000}N{000}N{0,000}N{00.0}@{}}
\toprule
& \multicolumn{2}{c}{Item-times} & & \multicolumn{4}{c}{Unchanged item-times invalidated} & \multicolumn{4}{c@{}}{Re-derivations} \\
\cmidrule(lr){2-3}\cmidrule(lr){5-8}\cmidrule(l){9-12}
(a) Shift & \multicolumn{1}{c}{Changed} & \multicolumn{1}{c}{Unchanged} & Key & \multicolumn{1}{c}{\mthead{All}{(\%)}} & \multicolumn{1}{c}{\mthead{Def.}{(\%)}} & \multicolumn{1}{c}{\mthead{Field}{(\%)}} & \multicolumn{1}{c}{\mthead{Free}{(\%)}} & \multicolumn{1}{c}{All} & \multicolumn{1}{c}{Def.} & \multicolumn{1}{c}{Field} & \multicolumn{1}{c@{}}{\mthead{Def.\ share}{(\%)}} \\
\midrule
$\Ssrc$ & 1,168 & 2,672 & source key & 99.7 & 98.6 & 99.7 & 100.0 & 3,800 & 470 & 2,930 & 97.9 \\
 &  &  & oracle & 16.8 & 50.3 & 16.4 & 1.5 & 764 & 377 & 344 & 78.5 \\
 &  &  & scope & \mtna & 53.1 & \mtna & \mtna & \mtna & 382 & \mtna & 79.6 \\
\mtgap
$\Sdef$ & 558 & 3,282 & source key & 99.3 & 95.7 & 99.3 & 100.0 & 3,792 & 469 & 2,923 & 97.7 \\
 &  &  & oracle & 4.3 & 61.3 & 2.9 & 1.1 & 553 & 415 & 124 & 86.5 \\
 &  &  & scope & \mtna & 64.5 & \mtna & \mtna & \mtna & 429 & \mtna & 89.4 \\
\mtgap
$\Sdel$ & 1,767 & 2,073 & source key & 100.0 & 100.0 & 100.0 & 100.0 & 3,808 & 472 & 2,936 & 98.3 \\
 &  &  & oracle & 7.0 & 49.4 & 3.7 & 1.1 & 805 & 373 & 418 & 77.7 \\
 &  &  & scope & \mtna & 52.3 & \mtna & \mtna & \mtna & 379 & \mtna & 79.0 \\
\bottomrule
\end{mttabular}
\mtnotes{Counted since the fork, no key missed a changed item-time (the largest share not invalidated is 0.0\%). Re-derivations: invalidations over the four checkpoints, one writer request each, summed over the 20 worlds. Def.\ share: share of the 480 definition item-times of a shift that were re-derived. ---: the scope key is built for definition items only.}

%% file: generated/audit_table.tex
\begin{table*}[t]
\caption{Memory systems on the $(D,B)$ grid: the audited systems, the further systems that we run natively, and our cells.}
\label{tab:audit}\label{tab:native-labels}
\centering\mtsetup
\setlength{\tabcolsep}{4pt}
\begin{mttabular}{@{}l l l c H{5.6cm}{1em}@{}}
\toprule
System & $D$ & $B$ & \begin{tabular}[b]{@{}c@{}}Run\\natively\end{tabular} & Basis \\
\midrule
\mtgroup{5}{Products}
\mtin Apple semantic index & partial & 0 &  & indirect evidence \\
\mtin ChatGPT memory & 0 & 0 &  & documentation, indirect evidence \\
\mtin Claude API memory tool & 0 & 0 &  & indirect evidence \\
\mtin Claude memory & 0 & 0 &  & documentation, indirect evidence \\
\mtin Gemini memory & 0 & 0 &  & documentation, indirect evidence (low confidence) \\
\midrule
\mtgroup{5}{Memory libraries}
\mtin A-Mem \citep{xu2025amem} & 0 & 0 & \checkmark & every event becomes a new note, and nothing supersedes or deletes a note \\
\mtin Cognee \citep{arxiv2505_24478} & partial & 0$^{a}$ & \checkmark & delete purges the item's chunks, summaries and the entities and edges only it produced \\
\mtin Graphiti / Zep \citep{arxiv2501_13956} & partial & temporal validity only &  & code, paper, documentation \\
\mtin HippoRAG~2 \citep{gutierrez2025hipporag2} & 0 & 1 & \checkmark & delete removes the passage and the triples and phrases no other passage supports \\
\mtin Honcho & partial & 0 &  & code, documentation \\
\mtin Khoj & 0 & 0 &  & code \\
\mtin Letta (.af) \citep{packer2023memgpt} & 0 & 0 &  & code, documentation \\
\mtin LightRAG \citep{guo2025lightrag} & 0 & 1 & \checkmark & delete purges the document's chunks, vectors and entity and relation contributions \\
\mtin Mem0 (self-hosted) \citep{arxiv2504_19413} & 0 & 0 & \checkmark & the add path with inference emits only new memories, and updates and deletions need explicit calls \\
\mtin MemOS \citep{li2025memos} & 0 & 0 & \checkmark & \texttt{add} never reads existing memories or their stored source and revision \\
\mtin Supermemory & partial & temporal validity only &  & indirect evidence \\
\midrule
\mtgroup{5}{Research systems and data stores}
\mtin Opal \citep{arxiv2604_02522} & partial & 0 &  & paper \\
\mtin Solid pod \citep{mansour2016solid} & --- & --- &  & documentation \\
\mtin User as Code \citep{arxiv2606_16707} & partial & 0 &  & code, paper \\
\midrule
\mtgroup{5}{Our baselines and cells}
\mtin plain RAG \citep{lewis2020rag} & --- & --- &  & code \\
\mtin no-store BM25 & --- & --- &  & code \\
\mtin \holdmem{}(0,0), (0,1), (1,0), (1,1) & \multicolumn{2}{l}{as named} &  & code \\
\bottomrule
\end{mttabular}
\mtnotes{\checkmark: run natively in the native-system experiment. ---: the system stores no derived item. Basis: what the placement rests on, and for a system we run, the property of its code that sets $B$. $^{a}$Measured at document level, as the native-system experiment runs Cognee: \NatCogneeLabel{}.}
\end{table*}

%% file: generated/tab_models.tex
\begin{table*}[t]
\caption{Models, with the context each was served with (tokens) and its roles in the runs.}
\label{tab:models}\label{tab:family-models}
\centering\mtsetup
\setlength{\tabcolsep}{4pt}
\begin{mttabular}{@{}H{6.4cm}{1em} r r H{7.0cm}{1em}@{}}
\toprule
Model & Size & Context & Roles \\
\midrule
\mbox{Qwen/Qwen3.5-9B} \citep{qwen35_2026,hf_qwen3_5_9b} & 9B & 16,384--65,536 & writer, reader, program writer, record rewriter, field extractor, native systems' LLM \\
\mbox{Claude Sonnet 5.5} \citep{anthropic_models_overview} & \mtna & 1,000,000 & reader, definition-item writer, program writer \\
\mbox{Claude Opus 5.5} \citep{anthropic_models_overview} & \mtna & 1,000,000 & definition-item writer, program writer \\
\mbox{google/gemma-4-12B-it} \citep{arxiv2607_02770} & 12B & 16,384 & writer, reader, program writer, world rewriter \\
\mbox{meta-llama/Llama-3.1-8B-Instruct} \citep{arxiv2407_21783} & 8B & 16,384 & writer, reader, program writer \\
\mbox{openai/gpt-oss-20b} \citep{arxiv2508_10925} & 20B & 16,384 & writer, reader, program writer \\
\mbox{mistralai/Ministral-3-14B-Instruct-2512-BF16} \citep{arxiv2601_08584,hf_ministral_3_14b_instruct_2512} & 14B & 16,384 & writer, reader \\
\mbox{Qwen/Qwen3.8-27B} \citep{qwen38_2026,hf_qwen3_8_27b} & 27B & 16,384 & writer \\
\mbox{XiaomiMiMo/MiMo-V2.6-Distill-Qwen-9B} \citep{mimo_v2_6_2026,hf_mimo_v2_6_distill_qwen_9b} & 9B & 16,384 & writer \\
\mbox{Qwen/Qwen3-8B} \citep{arxiv2505_09388} & 8B & 16,384 & writer \\
\mbox{Qwen/Qwen3-Embedding-0.6B} \citep{arxiv2506_05176,hf_qwen3_embedding_0_6b} & 0.6B & 8,192 & embedder \\
\mbox{BAAI/bge-m3} \citep{m3embedding2024} & 0.57B & 8,192 & embedder \\
\bottomrule
\end{mttabular}
\mtnotes{Open-weight models run under vLLM \citep{kwon2023vllm} in bfloat16 (gpt-oss-20b in MXFP4) at temperature~0 with JSON-schema decoding. Reasoning is off except in Claude Sonnet 5.5 (effort high for definition items), Claude Opus 5.5 (effort high for definition items, low for programs), and gpt-oss-20b (effort low). The Claude models think adaptively. ---: not disclosed.}
\end{table*}

%% file: appendix/app_prompts.tex
\section{Prompts and Templates}
\label{app:prompts}

This section gives the writer, program, feedback, and reader requests, and the templates of the stored
statements, verbatim from the released code. The other requests (typed input, the rewrites of records
and message bodies, and field extraction) are described with their experiments, and every request is
released with the code. Angle brackets mark placeholders, which are filled in for each item, time, and
world. The examples use the important-contacts item under its first version. Each request below, except
the feedback request, consists of a system message and one user message. The open-weight models answer
with constrained decoding under a JSON schema.

\makeatletter
\newenvironment{promptbox}[1]{%
  \par\smallskip\noindent\textbf{#1}\par\nobreak\@beginparpenalty=\@M
  \begin{list}{}{\leftmargin=1.2em\rightmargin=0pt\topsep=2pt\itemsep=0pt\parsep=2pt}\item\relax
  \footnotesize\ttfamily\frenchspacing\raggedright\hyphenpenalty=10000\exhyphenpenalty=10000}%
  {\end{list}\par\smallskip}
\makeatother
\newcommand{\ph}[1]{\textrm{\textit{$\langle$#1$\rangle$}}}
\newcommand{\bq}{\textasciigrave}

\paragraph{The writer request.} The writer derives one item from its source records. A $B{=}0$ cell that
writes from the delta adds the sentence about the records that arrived since its last write.

\begin{promptbox}{System}
You maintain a personal knowledge store for one person (called 'the user'). You write one knowledge item
at a time from the user's source records, citing the records you used.
\end{promptbox}

\begin{promptbox}{User}
Knowledge item: the set of the user's important contacts (full names)

Value format: a JSON array of full names ([] if none)

As of: \ph{YYYY-MM-DD HH:MM}

Definition in effect (important\_contact v1): An important contact is a person who attended at least 5
of the user's one-off calendar meetings (not recurring series) in the last 90 days.

\ph{$B{=}0$ delta writes only:} The records below are only those added or changed since
\ph{YYYY-MM-DD HH:MM}.

Source records (each starts with its id and revision):

<record id=\ph{object id} rev=\ph{revision}>\\
\ph{the record rendered as text}\\
</record>

\ph{further records, in time order, up to the writer's token budget}

Task: determine the value of the knowledge item as of the date above, relying on the most recent
applicable evidence. \bq{}value\bq{} must be only the value in the format given above (not a sentence); if the
records do not establish a current value, set value to null. Write \bq{}statement\bq{} as one or two
self-contained sentences stating the value (name the item and the date). Give up to 6 citations, each
with the record id, revision and a short quote (at most 25 words) copied exactly from that record. Reply
with JSON only.
\end{promptbox}

The reply follows a schema with three keys: \texttt{value} (a string, a list of strings, or null),
\texttt{statement}, and \texttt{citations} (at most six, each with \texttt{object\_id},
\texttt{revision}, and \texttt{quote}).

\paragraph{The stored statement.} The reader sees only the statement $s_k$ of an item. For a value written
by the writer, $s_k$ is rendered from the writer's statement and citations.

\begin{promptbox}{Statement of a written item}
[the set of the user's important contacts (full names) | written \ph{YYYY-MM-DD}] \ph{the writer's
statement} Evidence: \ph{object id} r\ph{revision}: "\ph{quote}"; \ph{up to six citations}
\end{promptbox}

\paragraph{The $D{=}1$ statement template.} When the check fails, exact recomputation (or, in the
$D_{\mathrm{m}}$ cell, the writer's program) supplies the value, and $s_k$ follows a fixed template. For a
definition item, the template names the definition version used to compute the value.

\begin{promptbox}{Statement of a recomputed item}
[the set of the user's important contacts (full names) | written \ph{YYYY-MM-DD}] As of
\ph{YYYY-MM-DD}, the set of the user's important contacts (full names) under the definition in effect
(important\_contact v1: An important contact is a person who attended at least 5 of the user's one-off
calendar meetings (not recurring series) in the last 90 days): \ph{the names, comma-separated, or
none}. (\ph{programmatic extraction, or: computed by the writer's own program} from \ph{object id}
r\ph{revision}; \ph{up to six source objects})
\end{promptbox}

If there is no current value, the body reads ``As of \ph{date}, there is no current value for \ph{item}
\ph{under the definition in effect}.''

\paragraph{The program request.} In the $D_{\mathrm{m}}$ cell, the writer also writes a program for each
item. The request contains no world data. The experiment with definitions in the user's own words
replaces only the statement after ``Definition in effect''.

\begin{promptbox}{System}
You maintain a personal knowledge store for one person (called 'the user'). You write small,
deterministic Python functions that compute one knowledge item from the user's structured source
records.
\end{promptbox}

\begin{promptbox}{User}
Knowledge item: the set of the user's important contacts (full names)

Value format: a JSON array of full names ([] if none)

Definition in effect (important\_contact v1): An important contact is a person who attended at least 5
of the user's one-off calendar meetings (not recurring series) in the last 90 days.

Declared sources: calendar

Every record is a dict. Besides the source-specific keys below it has: object\_id (str), source (str),
revision (int), updated (str "YYYY-MM-DD HH:MM", time of the last change), cluster (str, grouping id)
and text (str, the record rendered as plain text).

\ph{the fields of each declared source, here the four kinds of calendar record: settings, series,
appointment, and meeting, each with its typed fields}

Task: write a Python function \bq{}def derive(records, as\_of):\bq{} that computes the value of the knowledge
item as of \bq{}as\_of\bq{} (a string "YYYY-MM-DD HH:MM") from \bq{}records\bq{}, the live records of the declared
sources at that time (grouped by source in the order listed above, oldest \bq{}updated\bq{} first within a
source). It must return a Python list of strings in the value format above (an empty list if there are
none). \ph{the allowed built-in functions, exception classes, and helpers, and the limits of the
interpreter: no imports, no file, or network access, no wall clock, a time limit} Reply with JSON only:
\{"code": "<the Python source of the function and any helpers>"\}.
\end{promptbox}

\paragraph{The feedback turn.} With execution feedback, a program that fails on the records of a write
is sent back to the writer. The request repeats the program request and adds the failed reply as the
assistant's turn, followed by this user turn.

\begin{promptbox}{User}
Your function failed in the restricted interpreter on the live records of the declared sources:
\ph{the interpreter's message, with the returned value if it has the wrong type, at most 600
characters}\\
Write a corrected function. Reply with JSON only, in the same format: \{"code": "<the Python source of
the function and any helpers>"\}.
\end{promptbox}

\paragraph{The reader request.} The shared reader answers every probe of every system from the evidence
items that the system returns, within the evidence budget. For \holdmem{}, these items are the stored
statements. An unnamed probe leaves out the last sentence of the question.

\begin{promptbox}{System}
You answer questions about a person (called 'the user') using only the evidence provided. Evidence items
may disagree or be outdated; prefer the most recent information that applies at the question's date. If
the evidence does not determine the answer, abstain.
\end{promptbox}

\begin{promptbox}{User}
Evidence:

{}[1] \ph{evidence item}

{}[2] \ph{evidence item}

\ph{further items, up to the evidence budget}

Question: As of \ph{YYYY-MM-DD HH:MM}: Who are the user's important contacts under the user's current
definition? List full names. Definition in effect: An important contact is a person who attended at
least 5 of the user's one-off calendar meetings (not recurring series) in the last 90 days.

Reply with JSON \{"status": "answer" or "abstain", "value": ...\}. When answering, give only the value
in the requested format: a string, or a JSON array of strings for lists (use [] for an empty list). When
abstaining, set value to null.
\end{promptbox}

%% file: appendix/app_settings.tex
\section{Setup, Contrasts, and Statistics}
\label{app:experiments}

\begin{table*}[t]
\caption{Experiments: what each varies, the cells or systems it runs, its worlds (test worlds unless
stated), and the section that reports it.}
\label{tab:experiments}
\centering\mtsetup
\setlength{\tabcolsep}{3pt}\fontsize{8}{9}\selectfont
\newcolumntype{Y}[1]{>{\raggedright\arraybackslash}p{#1}}
\begin{mttabular}{@{}H{2.6cm}{1em} H{6.2cm}{0.7em} H{5.2cm}{0.7em} Y{1.25cm} Y{1.05cm}@{}}
\toprule
Experiment & What varies & Cells or systems & Worlds & Section \\
\midrule
\mtgroup{5}{Main study}
\mtin main experiment & $\Ssrc$, $\Sdef$, and $\Sdel$, probed 1--8 weeks after the fork & six \holdmem{} cells,
check-only cell, JSON arm, no-store BM25, plain RAG, oracle context & \EOneWorlds{} &
\ref{app:res-main}--\ref{app:res-keys} \\
\mtin writer switch & $\Swr$ at three tiers & \holdmem{} cells, calibrated on disjoint worlds & \TestWorlds{},
and 10 calibration & \ref{app:res-switch} \\
\mtin horizon & horizons of 1, 3, and 12 months & no-store BM25, plain RAG, \holdmem{}(1,1), oracle context,
with a five-part cost & \TestWorlds{} & \ref{app:res-horizon} \\
\midrule
\mtgroup{5}{Where the interaction lies, and the reader}
\mtin unshifted branch & the same world without the shift & four cells & \RThreeNSrcAllWorlds{} &
\ref{app:res-where} \\
\mtin unnamed probes & definition-item probes that do not name the definition & six cells, $\Sdef$ &
\RThreeQWorlds{} & \ref{app:res-where} \\
\mtin held-out definitions & 18 definitions from a grammar & \holdmem{} and $D_{\mathrm{m}}$ cells &
\RThreeGAWorlds{} & \ref{app:res-where} \\
\mtin Claude reader & Claude Sonnet 5.5 reads the main study's stores & six cells, JSON arm & \TestWorlds{} &
\ref{app:res-reader} \\
\midrule
\mtgroup{5}{What repairs, keys and refresh}
\mtin writer's programs & the writer's program replaces exact recomputation ($D_{\mathrm{m}}$) &
$D_{\mathrm{m}}$ with $B{=}0$, source and citation keys & \SynthWorlds{} &
\ref{app:res-repair} \\
\mtin Claude writers & Claude Sonnet 5.5 and Opus 5.5 write the definition items & four cells and two
$D_{\mathrm{m}}$ cells & \RThreeHSonnetWorlds{} & \ref{app:res-repair} \\
\mtin definitions in the user's own words & definitions as the user would state them, programs from three
writers & two $D_{\mathrm{m}}$ cells & \TestWorlds{} & \ref{app:res-repair} \\
\mtin program sets from six writers & three program sets from each of six writers, sampled & two
$D_{\mathrm{m}}$ cells & \RThreeGBWorlds{} & \ref{app:res-repair} \\
\mtin every record as typed fields & the writer re-derives definition items from every live record of
their sources, as typed fields, after $\Ssrc$ and $\Sdef$ & $D{=}0$, $D{=}1$, and the typed $D{=}0$
cell with a key, Qwen3.5-9B and Claude Sonnet 5.5 as writers of the definition items &
\TypedInWorlds{} & \ref{app:res-repair} \\
\mtin execution feedback & a failing program goes back to the writer with the interpreter's message,
for at most two rounds & $D{=}0$, $D_{\mathrm{m}}$, and $D_{\mathrm{m}}$ with
feedback, each with and without a key & \ExecFbWorlds{} & \ref{app:res-repair} \\
\mtin extracted fields & the programs read fields that the writer extracted from prose rewritten from the
records, after $\Ssrc$ and $\Sdef$ & $D{=}0$, $D_{\mathrm{m}}$, and $D_{\mathrm{m}}$ on extracted fields,
each with and without a key & \FieldExWorlds{} & \ref{app:res-repair} \\
\mtin controls & six arms that vary one confound at a time & \holdmem{} and $D_{\mathrm{m}}$ cells &
\CtlWorlds{} & \ref{app:res-repair}, \ref{app:res-keys} \\
\mtin refresh and upsert & refresh and slot-keyed upsert & at $D{=}0$, $D{=}1$, and $D_{\mathrm{m}}$ &
\RThreeUWorlds{} & \ref{app:res-keys} \\
\midrule
\mtgroup{5}{Generality}
\mtin replication & the main study on other worlds & the cells and baselines of the main study &
\RepWorlds{} replication & \ref{app:res-generality} \\
\mtin other writers & five other writers as the new writer (four other families and Qwen3-8B) &
writer-switch cells & \MOneExWriterWorldsPooled{} & \ref{app:res-switch} \\
\mtin other families & three other families as writer and reader & main-experiment cells &
\MTwoLlamaWorlds{} & \ref{app:res-generality} \\
\mtin other readers & three other families as reader & main-experiment cells & \MThreeLlamaWorlds{} &
\ref{app:res-generality} \\
\mtin rewritten worlds & a model rewrites message bodies around verbatim gold facts & main-experiment cells &
\RendWorlds{} & \ref{app:res-generality} \\
\midrule
\mtgroup{5}{Native systems, export, failure modes and sync}
\mtin native systems & native ingestion, black-box $(D,B)$ inference & Mem0 and five further systems, with
\holdmem{} cells and baselines on the same worlds & \EFiveMemWorlds{} three-month & \ref{app:native} \\
\mtin Mem0 interventions & source-keyed invalidation, query-time view & Mem0 & \RThreeXWorlds{} three-month &
\ref{app:native} \\
\mtin export & $\Simpl$: export to Portable Memory (\texttt{.mem}) with and without the derivation extension &
\holdmem{}(1,1) and a second implementation & \TestWorlds{} & \ref{app:e7} \\
\mtin failure-mode probes & conflicts, consent, an unstated redefinition, definition items under citation
keys, reader swap, embedding swap & \holdmem{} cells, no-store BM25, plain RAG & \ESevenWorlds{} &
\ref{app:res-generality}, \ref{app:e7} \\
\mtin sync & documented incremental-sync contracts instead of ideal delivery & \holdmem{} ($B\in\{0,1\}$,
$D{=}1$), no-store BM25 & \EFourXWorlds{} & \ref{app:sync} \\
\bottomrule
\end{mttabular}
\end{table*}

This section lists every experiment with the section that reports it (\cref{tab:experiments}). All
experiments use the same reader template and the same 4,096-token evidence budget. The section then
gives the standard report, the four diagnostic tests with their thresholds, the contrasts that separate
detection from repair, the statistics, and the compute. \Cref{app:models} gives the models and how they
were served.

\paragraph{The standard report.}\label{app:report} Every system implements one interface: ingest,
define, consent, set writer, query, export and import, and cost. For each shift, the report gives
Update and Stability with world-bootstrap intervals. Where a system reports which items are valid, it
also gives the missed invalidations, the invalidations of unchanged items (items invalidated although their
gold value and status did not change), and the accuracy on re-derived items (the share of re-derived items that the reader
answers correctly). Finally, it gives the outcomes of the four diagnostic tests below. A system that
offers only add and search runs through an adapter that ingests each event in time order and retrieves
context without generating an answer (\cref{app:native-design}).

\paragraph{Diagnostic tests and their thresholds.} The \emph{retention test} applies to a switch between
comparable writers (the three tiers). It requires that the retention, the share of answers correct
before the switch that stay correct, is at least 97\%. It is a one-sided non-inferiority test with a
90\% interval, and we report the net change in accuracy next to it. Switches to other model families use
the same threshold. The \emph{leaving test} applies after an export to a third-party format with a
derivation extension that carries the keys, followed by an import into another implementation. It
requires that carryover equals that of the original within $\pm3$\,pp, tested with two one-sided tests
(TOST, \citealp{schuirmann1987tost}). The \emph{utility floor} requires an accuracy before any shift of at
least that of the oracle context minus 3\,pp. The \emph{cost ceiling} requires a serving cost per query of
at most half a full re-read. It bounds serving, not maintenance. Three tests pass by construction in some
settings: the retention test for cells that never re-derive, the leaving test for two implementations
of one data model, and the cost ceiling for a retrieval-based system.

\subsection{Contrasts and Thresholds}
\label{app:contrasts}

By construction, after a source revision, a redefinition, or a deletion, only $B$ and the granularity of
the key can change detection, and only $D$ can change the accuracy on re-derived items. After a writer
switch, a $B{=}1$ memory invalidates every item and a $B{=}0$ memory none, so only $D$ can change
accuracy, and only at $B{=}1$. After an export, $B$ and $D$ are kept if and only if the derivation
extension is exported. Three quantities test this division directly, each against a threshold we set.
First, the effect of $B$ on Update after a source revision, averaged over $D$, is compared with
$+2$\,pp. This is the gain over flat retrieval that \citet{arxiv2607_16019} found for a fine-grained
revision ledger on values that were changed and then restored, once the presentation of evidence is
matched. Second, $D$'s effect on missed invalidation is compared with a threshold of 2\,pp in absolute
value in each of six contrasts (two keys and three shifts). Third, writer agreement, the share of items on
which two writers are both right or both wrong, should predict \emph{survival}, that is, whether an item
correct before a switch stays correct. We regress $D{=}0$, $B{=}1$ survival on the agreement of the two
writers. Its calibration slope is compared with 0 (no relation), and its AUROC with 0.60, the weakest
discrimination we count as useful, both with 90\% intervals.

\subsection{Statistics and Compute}
\label{app:statistics}

\paragraph{Statistics.} Intervals are 95\% world-bootstrap intervals over differences paired by item and
checkpoint, with 2,000 draws \citep{efron1993bootstrap,field2007clustered}. Equivalence tests and writer
agreement as a predictor use 90\% intervals, and equivalence uses TOST with the margin stated for each
test \citep{schuirmann1987tost,lakens2017equivalence}. We summarize calibration by the logistic
calibration slope \citep{cox1958binary,vancalster2019calibration} and discrimination by AUROC. For the held-out definitions
and the program sets, the bootstrap draws worlds and definitions (or program sets) independently and
takes the product sample. Intervals are per contrast, and we analyze each model family and each native
system separately.

\paragraph{Multiple contrasts.}\label{app:limitations}\label{app:multiple} The \MultExOrigContrasts{}
contrasts of $B$, $D$, and $B{\times}D$ on the three branches that can differ from zero by construction
(\cref{app:contrasts}) form one family per test set, with one-sided bootstrap p-values. The Holm
correction \citep{holm1979simple} keeps \MultExOrigHolm{} of the \MultExOrigInterval{} intervals that
exclude zero on the test worlds and \MultExRepHolm{} of \MultExRepInterval{} on the replication.
Benjamini--Hochberg at $q{=}0.05$ \citep{benjamini1995fdr} keeps \MultExOrigBh{} and \MultExRepBh{}. The
one interval that Holm drops is the $B{\times}D$ of the citation key after a redefinition
(\RepBxDInteractionUpdateCitedDefOrig{}), which Holm keeps on the replication
(\RepBxDInteractionUpdateCitedDef{}). The detection contrasts sit at the lowest p-value the bootstrap can
give. Without them, Holm with \ExHolmMainDraws{} draws over the remaining \ExHolmMainFamily{} contrasts
keeps \ExHolmMainSurvive{} of those that exclude zero on the test worlds and \ExHolmRepSurvive{} on the
replication.

\paragraph{Compute.} The open-weight models were served on \GpuModel{} GPUs, which the model servers held
for about \GpuHoursTotal{} GPU-hours in total. The native-system runs used \GpuHoursNativeSystems{} of
them, all other runs \GpuHoursOtherRuns{}, and the servers were idle for the rest. Claude Sonnet 5.5
(\SonnetModelId) and Claude Opus 5.5 (\OpusModelId) were called through Anthropic's API from
\ClaudeCallsFrom{} to \ClaudeCallsTo{}, with \ClaudeTokensTotal{} input and output tokens in total at a
cost of \ClaudeSpendUSD{}~US dollars.

%% file: appendix/app_results.tex
\section{Extended Results}
\label{app:results}

This section gives the full results, ordered by the question they answer. Unless a caption says
otherwise, a result comes from the main study: \TestWorlds{} test worlds, Qwen3.5-9B as writer and
reader, and the source key for $B{=}1$. Brackets give 95\% world-bootstrap intervals. In a table, an estimate
shown without an interval has an interval of zero width at the precision shown, unless the notes of the
table call it a point estimate.

\input{generated/tab_store_reader}
\begin{table*}[t]
\caption{Evidence coverage of the $D{=}0$ writer's 6,000-token selection, and the size of the complete typed
input.}
\label{tab:coverage}\label{tab:input-sizes}
\centering\mtsetup
\setlength{\tabcolsep}{3pt}
\begin{mttabular*}{\linewidth}{l N{000} >{\mtpairwidths{100.0}{[00.0, 00.0]}}c>{\mtpairwidths{100.0}{[00.0, 00.0]}}c N{000} N{100.0} >{\mtpairwidths{100.0}{[00.0, 00.0]}}c N{00,000}N{00,000}N{100.0}@{}}
\toprule
& & \multicolumn{2}{c}{Definition scope in the selection (\%)} & \multicolumn{3}{c}{Full-access control}
  & \multicolumn{3}{c@{}}{Typed input (tokens)} \\
\cmidrule(lr){3-4}\cmidrule(lr){5-7}\cmidrule(l){8-10}
(a) Definition & \multicolumn{1}{c}{$n$} & Held & Complete & \multicolumn{1}{c}{Re-derivations} & \multicolumn{1}{c}{Cut (\%)} & Same request (\%)
  & \multicolumn{1}{c}{Median} & \multicolumn{1}{c}{Max} & \multicolumn{1}{c@{}}{Fit (\%)} \\
\midrule
important contacts & \ExRevCovImportantContactItemTimes & \mtcipair{\ExRevCovImportantContactScopeShare} & \mtcipair{\ExRevCovImportantContactCompleteShare} & \ExRevFullImportantContactRederived & \mspct\ExRevFullImportantContactTruncatedShareEst & \mtcipair{\ExRevFullImportantContactSameRequestShare} & \ExRevSizeMainImportantContactColumnsNoBodyMedian & \ExRevSizeMainImportantContactColumnsNoBodyMax & \mspct\ExRevSizeMainImportantContactColumnsNoBodyFit \\
close friends & \ExRevCovCloseFriendItemTimes & \mtcipair{\ExRevCovCloseFriendScopeShare} & \mtcipair{\ExRevCovCloseFriendCompleteShareEst} & \ExRevFullCloseFriendRederived & \mspct\ExRevFullCloseFriendTruncatedShareEst & \mtcipair{\ExRevFullCloseFriendSameRequestShareEst} & \ExRevSizeMainCloseFriendColumnsNoBodyMedian & \ExRevSizeMainCloseFriendColumnsNoBodyMax & \mspct\ExRevSizeMainCloseFriendColumnsNoBodyFit \\
working hours & \ExRevCovWorkingHoursItemTimes & \mtcipair{\ExRevCovWorkingHoursScopeShareEst} & \mtcipair{\ExRevCovWorkingHoursCompleteShareEst} & \ExRevFullWorkingHoursRederived & \mspct\ExRevFullWorkingHoursTruncatedShareEst & \mtcipair{\ExRevFullWorkingHoursSameRequestShareEst} & \ExRevSizeMainWorkingHoursColumnsNoBodyMedian & \ExRevSizeMainWorkingHoursColumnsNoBodyMax & \mspct\ExRevSizeMainWorkingHoursColumnsNoBodyFit \\
active projects & \ExRevCovActiveProjectItemTimes & \mtcipair{\ExRevCovActiveProjectScopeShareEst} & \mtcipair{\ExRevCovActiveProjectCompleteShareEst} & \ExRevFullActiveProjectRederived & \mspct\ExRevFullActiveProjectTruncatedShareEst & \mtcipair{\ExRevFullActiveProjectSameRequestShareEst} & \ExRevSizeMainActiveProjectColumnsNoBodyMedian & \ExRevSizeMainActiveProjectColumnsNoBodyMax & \mspct\ExRevSizeMainActiveProjectColumnsNoBodyFit \\
VIP senders & \ExRevCovVipSenderItemTimes & \mtcipair{\ExRevCovVipSenderScopeShare} & \mtcipair{\ExRevCovVipSenderCompleteShareEst} & \ExRevFullVipSenderRederived & \mspct\ExRevFullVipSenderTruncatedShareEst & \mtcipair{\ExRevFullVipSenderSameRequestShareEst} & \ExRevSizeMainVipSenderColumnsNoBodyMedian & \ExRevSizeMainVipSenderColumnsNoBodyMax & \mspct\ExRevSizeMainVipSenderColumnsNoBodyFit \\
overdue tasks & \ExRevCovOverdueTaskItemTimes & \mtcipair{\ExRevCovOverdueTaskScopeShareEst} & \mtcipair{\ExRevCovOverdueTaskCompleteShareEst} & \ExRevFullOverdueTaskRederived & \mspct\ExRevFullOverdueTaskTruncatedShareEst & \mtcipair{\ExRevFullOverdueTaskSameRequestShareEst} & \ExRevSizeMainOverdueTaskColumnsNoBodyMedian & \ExRevSizeMainOverdueTaskColumnsNoBodyMax & \mspct\ExRevSizeMainOverdueTaskColumnsNoBodyFit \\
\end{mttabular*}\nointerlineskip
\begin{mttabular*}{\linewidth}{l l N{000} >{\mtpairwidths{100.0}{[00.0, 100.0]}}c>{\mtpairwidths{100.0}{[00.0, 100.0]}}c>{\mtpairwidths{$+$00.0}{[$+$00.0, $+$00.0]}}c@{}}
\midrule
(b) Shift & Definition scope & \multicolumn{1}{c}{$n$} & $r_0$ (\%) & $r_1$ (\%) & \multicolumn{1}{c@{}}{$B{\times}D$ (pp)} \\
\midrule
\mtgroup{6}{Qwen3.5-9B, \TestWorlds{} worlds}
\mtin $\Ssrc$ & complete & \ExRevCovQwenSrcCompleteItemsZero & \mtcipair{\ExRevCovQwenSrcCompleteRZero} & \mtcipair{\ExRevCovQwenSrcCompleteROne} & \mtcipair{\ExRevCovQwenSrcCompleteDoneBxD} \\
 & incomplete & \ExRevCovQwenSrcIncompleteItemsZero & \mtcipair{\ExRevCovQwenSrcIncompleteRZero} & \mtcipair{\ExRevCovQwenSrcIncompleteROneEst} & \mtcipair{\ExRevCovQwenSrcIncompleteDoneBxD} \\
\mtin $\Sdef$ & complete & \ExRevCovQwenDefCompleteItemsZero & \mtcipair{\ExRevCovQwenDefCompleteRZero} & \mtcipair{\ExRevCovQwenDefCompleteROne} & \mtcipair{\ExRevCovQwenDefCompleteDoneBxD} \\
 & incomplete & \ExRevCovQwenDefIncompleteItemsZero & \mtcipair{\ExRevCovQwenDefIncompleteRZero} & \mtcipair{\ExRevCovQwenDefIncompleteROneEst} & \mtcipair{\ExRevCovQwenDefIncompleteDoneBxD} \\
\mtin $\Sdel$ & complete & \ExRevCovQwenOrphanCompleteItemsZero & \mtcipair{\ExRevCovQwenOrphanCompleteRZero} & \mtcipair{\ExRevCovQwenOrphanCompleteROne} & \mtcipair{\ExRevCovQwenOrphanCompleteDoneBxD} \\
 & incomplete & \ExRevCovQwenOrphanIncompleteItemsZero & \mtcipair{\ExRevCovQwenOrphanIncompleteRZero} & \mtcipair{\ExRevCovQwenOrphanIncompleteROneEst} & \mtcipair{\ExRevCovQwenOrphanIncompleteDoneBxD} \\
\midrule
\mtgroup{6}{Claude Sonnet 5.5}
\mtin $\Ssrc$ & complete & \ExRevCovSonnetSrcCompleteItemsZero & \mtcipair{\ExRevCovSonnetSrcCompleteRZero} & \mtcipair{\ExRevCovSonnetSrcCompleteROne} & \mtcipair{\ExRevCovSonnetSrcCompleteDoneBxD} \\
 & incomplete & \ExRevCovSonnetSrcIncompleteItemsZero & \mtcipair{\ExRevCovSonnetSrcIncompleteRZero} & \mtcipair{\ExRevCovSonnetSrcIncompleteROne} & \mtcipair{\ExRevCovSonnetSrcIncompleteDoneBxD} \\
\mtin $\Sdef$ & complete & \ExRevCovSonnetDefCompleteItemsZero & \mtcipair{\ExRevCovSonnetDefCompleteRZero} & \mtcipair{\ExRevCovSonnetDefCompleteROne} & \mtcipair{\ExRevCovSonnetDefCompleteDoneBxD} \\
 & incomplete & \ExRevCovSonnetDefIncompleteItemsZero & \mtcipair{\ExRevCovSonnetDefIncompleteRZero} & \mtcipair{\ExRevCovSonnetDefIncompleteROne} & \mtcipair{\ExRevCovSonnetDefIncompleteDoneBxD} \\
\mtin $\Sdel$ & complete & \ExRevCovSonnetOrphanCompleteItemsZero & \mtcipair{\ExRevCovSonnetOrphanCompleteRZero} & \mtcipair{\ExRevCovSonnetOrphanCompleteROne} & \mtcipair{\ExRevCovSonnetOrphanCompleteDoneBxD} \\
 & incomplete & \ExRevCovSonnetOrphanIncompleteItemsZero & \mtcipair{\ExRevCovSonnetOrphanIncompleteRZero} & \mtcipair{\ExRevCovSonnetOrphanIncompleteROne} & \mtcipair{\ExRevCovSonnetOrphanIncompleteDoneBxD} \\
\midrule
\mtgroup{6}{Claude Opus 5.5}
\mtin $\Ssrc$ & complete & \ExRevCovOpusSrcCompleteItemsZero & \mtcipair{\ExRevCovOpusSrcCompleteRZero} & \mtcipair{\ExRevCovOpusSrcCompleteROne} & \mtcipair{\ExRevCovOpusSrcCompleteDoneBxD} \\
 & incomplete & \ExRevCovOpusSrcIncompleteItemsZero & \mtcipair{\ExRevCovOpusSrcIncompleteRZero} & \mtcipair{\ExRevCovOpusSrcIncompleteROne} & \mtcipair{\ExRevCovOpusSrcIncompleteDoneBxD} \\
\mtin $\Sdef$ & complete & \ExRevCovOpusDefCompleteItemsZero & \mtcipair{\ExRevCovOpusDefCompleteRZero} & \mtcipair{\ExRevCovOpusDefCompleteROne} & \mtcipair{\ExRevCovOpusDefCompleteDoneBxD} \\
 & incomplete & \ExRevCovOpusDefIncompleteItemsZero & \mtcipair{\ExRevCovOpusDefIncompleteRZero} & \mtcipair{\ExRevCovOpusDefIncompleteROne} & \mtcipair{\ExRevCovOpusDefIncompleteDoneBxD} \\
\mtin $\Sdel$ & complete & \ExRevCovOpusOrphanCompleteItemsZero & \mtcipair{\ExRevCovOpusOrphanCompleteRZero} & \mtcipair{\ExRevCovOpusOrphanCompleteROne} & \mtcipair{\ExRevCovOpusOrphanCompleteDoneBxD} \\
 & incomplete & \ExRevCovOpusOrphanIncompleteItemsZero & \mtcipair{\ExRevCovOpusOrphanIncompleteRZero} & \mtcipair{\ExRevCovOpusOrphanIncompleteROne} & \mtcipair{\ExRevCovOpusOrphanIncompleteDoneBxD} \\
\bottomrule
\end{mttabular*}
\mtnotes{(a) $n$: the definition item-times after the three shifts. Held: the definition-scope
records in the selection. Complete: the item-times with all of them. Cut: the full-access control's
requests cut at \CtlFullLimit{} tokens (\CtlWorlds{} worlds, $\Ssrc$ and $\Sdef$). Same request: those
equal to the $D{=}0$ request. Over all six definitions the control re-derived \ExRevFullAllRederived{}
times, \ExRevFullAllSameRequestShare{} with the same request. Typed input: a $D_{\mathrm{m}}$ program's
complete typed input per item-time, as tables without bodies, in Qwen3.5-9B tokens. Fit: the share at
or below 31,000 tokens. (b) On the changed definition item-times, $B{\times}D$ at $D{=}1$.}
\end{table*}
\input{generated/tab_typed}
\input{generated/tab_writers}
\begin{table*}[t]
\caption{Changed item-times that the citation key did not invalidate since the fork: (a) by the cause of
the change ($D{=}1$ cell), (b) Update on them.}
\label{tab:cited-misses}\label{tab:forced}\label{tab:misses}
\centering\mtsetup
\setlength{\tabcolsep}{4pt}
\begin{mttabular}{@{}ll N{0,000}N{000} r@{\,}l@{\hspace{14pt}} r@{\,}l@{\hspace{14pt}} r@{\,}l@{}}
\toprule
& & \multicolumn{4}{c}{All changed item-times} & \multicolumn{2}{c}{Caused by the shift}
  & \multicolumn{2}{c@{}}{Also without the shift} \\
\cmidrule(lr){3-6}\cmidrule(lr){7-8}\cmidrule(l){9-10}
(a) Shift & Class & \multicolumn{1}{c}{$n$} & \multicolumn{1}{c}{Missed} & \multicolumn{2}{c}{Share (\%)}
  & \multicolumn{2}{c}{Share (\%)} & \multicolumn{2}{c@{}}{Share (\%)} \\
\midrule
$\Ssrc$ & definition & \ExRevCitedSrcPooledDefinitionItems{} & \ExRevCitedSrcPooledDefinitionMissed{} & \mtci{\ExRevCitedSrcPooledDefinitionMissShare} & \mtci{\ExRevCitedSrcShiftDefinitionMissShare} & \mtci{\ExRevCitedSrcDriftDefinitionMissShare} \\
 & other & \ExRevCitedSrcPooledOtherItems{} & \ExRevCitedSrcPooledOtherMissed{} & \mtci{\ExRevCitedSrcPooledOtherMissShare} & \mtci{\ExRevCitedSrcShiftOtherMissShare} & \mtci{\ExRevCitedSrcDriftOtherMissShare} \\
 & all & \ExRevCitedSrcPooledAllItems{} & \ExRevCitedSrcPooledAllMissed{} & \mtci{\ExRevCitedSrcPooledAllMissShare} & \mtci{\ExRevCitedSrcShiftAllMissShare} & \mtci{\ExRevCitedSrcDriftAllMissShare} \\
\mtgap
$\Sdef$ & definition & \ExRevCitedDefPooledDefinitionItems{} & \ExRevCitedDefPooledDefinitionMissed{} & \mtci{\ExRevCitedDefPooledDefinitionMissShare} & \mtci{\ExRevCitedDefShiftDefinitionMissShareEst} & \mtci{\ExRevCitedDefDriftDefinitionMissShare} \\
 & other & \ExRevCitedDefPooledOtherItems{} & \ExRevCitedDefPooledOtherMissed{} & \mtci{\ExRevCitedDefPooledOtherMissShare} & \multicolumn{2}{c}{---} & \mtci{\ExRevCitedDefDriftOtherMissShare} \\
 & all & \ExRevCitedDefPooledAllItems{} & \ExRevCitedDefPooledAllMissed{} & \mtci{\ExRevCitedDefPooledAllMissShare} & \mtci{\ExRevCitedDefShiftAllMissShareEst} & \mtci{\ExRevCitedDefDriftAllMissShare} \\
\mtgap
$\Sdel$ & definition & \ExRevCitedOrphanPooledDefinitionItems{} & \ExRevCitedOrphanPooledDefinitionMissed{} & \mtci{\ExRevCitedOrphanPooledDefinitionMissShare} & \mtci{\ExRevCitedOrphanShiftDefinitionMissShare} & \mtci{\ExRevCitedOrphanDriftDefinitionMissShare} \\
 & other & \ExRevCitedOrphanPooledOtherItems{} & \ExRevCitedOrphanPooledOtherMissed{} & \mtci{\ExRevCitedOrphanPooledOtherMissShare} & \mtci{\ExRevCitedOrphanShiftOtherMissShareEst} & \mtci{\ExRevCitedOrphanDriftOtherMissShare} \\
 & all & \ExRevCitedOrphanPooledAllItems{} & \ExRevCitedOrphanPooledAllMissed{} & \mtci{\ExRevCitedOrphanPooledAllMissShare} & \mtci{\ExRevCitedOrphanShiftAllMissShare} & \mtci{\ExRevCitedOrphanDriftAllMissShare} \\
\bottomrule
\end{mttabular}
\mtnotes{$n$: changed item-times. Missed: not invalidated. Share: missed over the changed item-times of
the part (---: no item-time).}
\par\medskip
\begin{mttabular}{@{}ll N{100.0}N{100.0}N{100.0} @{\hspace{14pt}} r@{\,}l r@{\,}l@{}}
\toprule
& & \multicolumn{3}{c}{Update at $D{=}1$ (\%)} & \multicolumn{4}{c@{}}{Gain (pp)} \\
\cmidrule(lr){3-5}\cmidrule(l){6-9}
(b) Shift & Class & \multicolumn{1}{c}{Cited} & \multicolumn{1}{c}{Forced} & \multicolumn{1}{c}{$B{=}0$}
  & \multicolumn{2}{c}{$D{=}0$} & \multicolumn{2}{c@{}}{$D{=}1$} \\
\midrule
$\Ssrc$ & definition & \mspct\ExRevForcedDoneSrcDefinitionCitedEst & \mspct\ExRevForcedDoneSrcDefinitionForcedEst & \mspct\ExRevForcedDoneSrcDefinitionBzeroEst & \mtci{\ExRevForcedDzeroSrcDefinitionGain} & \mtci{\ExRevForcedDoneSrcDefinitionGain} \\
 & other & \mspct\ExRevForcedDoneSrcOtherCitedEst & \mspct\ExRevForcedDoneSrcOtherForcedEst & \mspct\ExRevForcedDoneSrcOtherBzeroEst & \mtci{\ExRevForcedDzeroSrcOtherGain} & \mtci{\ExRevForcedDoneSrcOtherGain} \\
\mtgap
$\Sdef$ & definition & \mspct\ExRevForcedDoneDefDefinitionCitedEst & \mspct\ExRevForcedDoneDefDefinitionForcedEst & \mspct\ExRevForcedDoneDefDefinitionBzeroEst & \mtci{\ExRevForcedDzeroDefDefinitionGain} & \mtci{\ExRevForcedDoneDefDefinitionGain} \\
 & other & \mspct\ExRevForcedDoneDefOtherCitedEst & \mspct\ExRevForcedDoneDefOtherForcedEst & \mspct\ExRevForcedDoneDefOtherBzeroEst & \mtci{\ExRevForcedDzeroDefOtherGain} & \mtci{\ExRevForcedDoneDefOtherGain} \\
\mtgap
$\Sdel$ & definition & \mspct\ExRevForcedDoneOrphanDefinitionCitedEst & \mspct\ExRevForcedDoneOrphanDefinitionForcedEst & \mspct\ExRevForcedDoneOrphanDefinitionBzeroEst & \mtci{\ExRevForcedDzeroOrphanDefinitionGain} & \mtci{\ExRevForcedDoneOrphanDefinitionGain} \\
 & other & \mspct\ExRevForcedDoneOrphanOtherCitedEst & \mspct\ExRevForcedDoneOrphanOtherForcedEst & \mspct\ExRevForcedDoneOrphanOtherBzeroEst & \mtci{\ExRevForcedDzeroOrphanOtherGain} & \mtci{\ExRevForcedDoneOrphanOtherGain} \\
\bottomrule
\end{mttabular}
\mtnotes{On the missed item-times of (a). Cited: the citation-key cell, which served the stale item.
Forced: the source-key cell, which re-derived the same item-times with the same request. $B{=}0$: the
append-only cell. Gain: forced minus cited. Update: point estimates.}
\end{table*}
\input{generated/tab_replication}
\input{generated/tab_mem0x}
\begin{figure*}[t]
\centering
\includegraphics{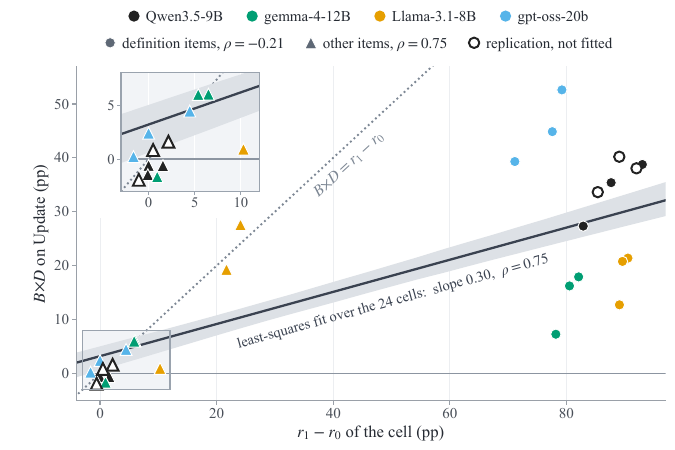}
\caption{$B{\times}D$ on Update against $r_1-r_0$ of the same cell, one marker per
writer, item class, and shift (source key). Hollow: the replication, held out of the least-squares fit
over the four writers. Band: the fit's 95\% bootstrap band. $\rho$ in the key: Spearman correlation within
the item class. Inset: the framed region near the origin. A definition-item marker lies below the diagonal
by the effect of $D$ at $B{=}0$.}
\label{fig:account}
\end{figure*}
\begin{figure*}[t]
\centering
\includegraphics[width=\linewidth]{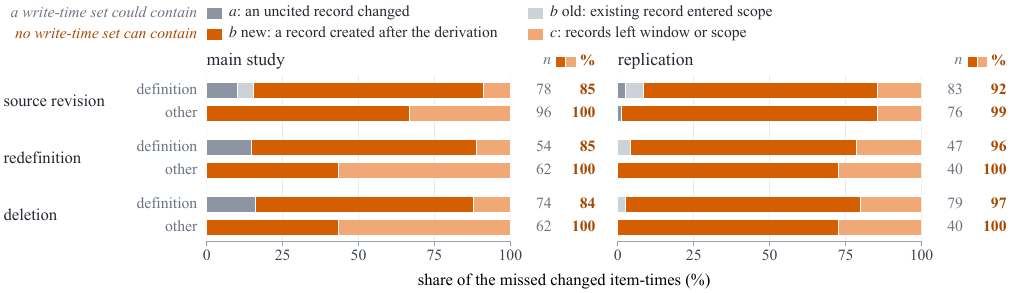}
\caption{The changed item-times the citation key missed since the fork, by cause, per shift and item
class, in the
main study (\TestWorlds{} worlds) and the replication (\RepWorlds{} worlds). Most come from records that
no set recorded at write time can contain. Orange: a record created after the derivation ($b$ new), or records that
left a time window or a narrowed scope ($c$). Gray: a record of the dependency set that changed and that
the writer had not cited ($a$), or an existing record that entered the scope ($b$ old). The classes are
exclusive. $n$: the number missed. \%: the orange share.}
\label{fig:misses}
\end{figure*}
\paragraph{Symbols.}\label{app:notation} The shifts are a source revision $\Ssrc$, a redefinition
$\Sdef$, a deletion $\Sdel$, a writer switch $\Swr$, an export $\Simpl$, and, in the failure-mode probes,
a consent revocation $\Scons$. \holdmem{}$(D,B)$ is a cell of the reference memory. At $B{=}0$ there is
no key and the memory only appends. At $B{=}1$ a key invalidates an item, and the current writer
re-derives it. At $D{=}0$ the writer re-derives, at $D{=}1$ exact recomputation repairs, and at
$D_{\mathrm{m}}$ the writer's programs repair. An \emph{item-time} is one item at one checkpoint, and
every count and share below is over item-times. An item-time is changed when its gold value or status
differs from that at the fork. On the changed item-times, $m$ is the share not invalidated after the
latest change of the gold. An invalidation counts only if it happens after the checkpoint before the
interval in which the gold last changed. $r$ is the share of the re-derived item-times that are answered
correctly, and $a$ is the share answered correctly among those not invalidated. $r_D$ and $a_D$ are $r$
and $a$ at a given $D$, $r_D$ with a key and $a_D$ without. The accuracy on re-derived items is the share
of re-derived item-times, changed or not, that the reader answers correctly (or, where a table says so,
whose re-derived value equals gold). Effects are in percentage points (pp). The $B$ effect is $B{=}1$
minus $B{=}0$ at a given $D$. The $D$ effect is $D{=}1$ minus $D{=}0$ at a given $B$. $B{\times}D$ is the
$B$ effect at $D{=}1$ (or $D_{\mathrm{m}}$) minus the $B$ effect at $D{=}0$. A change is caused by the
shift when its gold differs from the gold of the same world at the same time without the shift (the
unshifted branch). Otherwise, it would also have happened without the shift. If we counted $m$ from the
fork, an item-time that was re-derived after an earlier change and is stale after a later one would
count as a failed repair. For the source key the two ways of counting differ on at most
\ExDetMainSourceMissedMax{} of a class's changed item-times (\ExDetRepSourceMissedMax{} in the
replication), and for the citation key on far more (\cref{app:clock}). Some analyses count detection
from the fork: evidence coverage, typed inputs, the Claude writers' $r_0$, the controls, the account
across writers, the key audit, and the citation key's misses
(\cref{tab:store-reader,tab:coverage,tab:typed,tab:writers,tab:keyaudit,tab:cited-misses,tab:replication,tab:mem0x}
and \cref{fig:writers,fig:account,fig:misses}). There the source key misses no changed item-time, so its
$r$ equals its Update.

\paragraph{Which item-times are scored.} At each checkpoint after the fork, the reader is asked about
every changed item-time. For each world, shift, and checkpoint, it is also asked about a random sample of
the unchanged item-times of the same size, drawn with a fixed seed (or all of them if there are fewer).
Every \holdmem{} cell is asked about the same item-times. Update is computed on the changed item-times
and Stability on the sample, while the invalidation shares of the key audit (\cref{tab:keyaudit}) count
every item-time. In the main study, each cell is thus scored on \ExDetMainSrcUpdateItems{},
\ExDetMainDefUpdateItems{}, and \ExDetMainOrphanUpdateItems{} changed and
\ExDetMainSrcStabilityItems{}, \ExDetMainDefStabilityItems{}, and \ExDetMainOrphanStabilityItems{}
unchanged item-times after a source revision, a redefinition, and a deletion. In the replication, the
numbers are \ExDetRepSrcUpdateItems{}, \ExDetRepDefUpdateItems{}, and \ExDetRepOrphanUpdateItems{}
changed and \ExDetRepSrcStabilityItems{}, \ExDetRepDefStabilityItems{}, and
\ExDetRepOrphanStabilityItems{} unchanged item-times.

\subsection{Main Contrasts}
\label{app:res-main}

\input{generated/tab_classsplit}

\begin{figure*}[t]
\centering
\includegraphics[width=\linewidth]{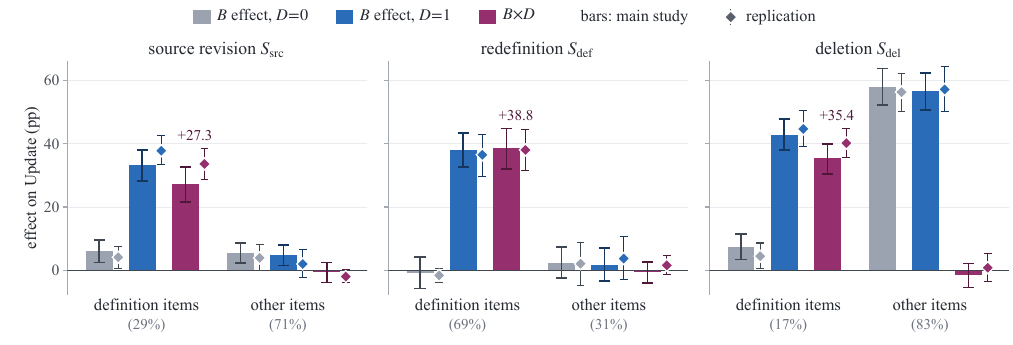}
\caption{Effect of $B$ (source key) on Update at $D{=}0$ and $D{=}1$, and $B{\times}D$, by shift and
item class, with Qwen3.5-9B as writer and reader. On definition items, the key helps mainly when exact
recomputation repairs them. Bars: main study (\TestWorlds{} worlds), with its $B{\times}D$ on definition
items printed above. Diamonds: replication on \RepWorlds{} fresh worlds. Error bars: 95\% intervals over
worlds. In brackets: the class's share of the changed item-times in the main study.}
\label{fig:classsplit}
\end{figure*}

\paragraph{The key raises Update, and $D$ does not change detection.} \Cref{tab:replication} gives
every contrast of the main study next to its estimate in the replication (\cref{app:res-generality}).
The effect of $B$ on Update after a source revision, averaged over $D$, was \EOneBEffectPooledSrc{},
above the threshold of $+2$\,pp. After a redefinition and a deletion it was \EOneBEffectPooledDef{} and
\EOneBEffectPooledOrphan{}. The keys do not depend on $D$, so $D$'s effect on missed invalidation was
\GatePathTwoMaxAbs{} for both keys after every shift, and an import without the keys misses every change
(\cref{app:e7}). By design, the source key's effects on detection are near 100\,pp, apart from the
changes that come from the clock alone (\cref{app:clock}). With either key, $D$ raised the accuracy on
re-derived items after every shift, whether we score the reader's answer or the re-derived value.

\subsection{Where the Interaction Lies}
\label{app:res-where}

\paragraph{The class split.}\label{app:classsplit} \Cref{fig:classsplit} shows the pattern. On definition
items, the key helps mainly with exact recomputation (\ExClsMainSrcDefinitionBDone{} at $D{=}1$ against
\ExClsMainSrcDefinitionBDzero{} at $D{=}0$ after a source revision). On the other items, the interaction
is near zero with this reader in both sets of worlds. \Cref{tab:classsplit} gives every contrast by
shift and item class, with exact recomputation in the main study and the replication, and with the
writer's programs.

\begin{table*}[t]
\caption{The $B$ effects and $B{\times}D$ on Update by the cause of the change (a), and $B{\times}D$ without
the shift on the changed item-times that the shift caused (b).}
\label{tab:cause}\label{tab:natural}
\centering\mtsetup
\setlength{\tabcolsep}{4pt}
\mtpairwidths{$-$00.0}{[$-$00.0,\,$-$00.0]}
\begin{mttabular*}{\linewidth}{@{}lll N{0,000} cccc@{}}
\toprule
& & & & \multicolumn{2}{c}{$B$ effect} & \multicolumn{2}{c@{}}{$B{\times}D$} \\
\cmidrule(lr){5-6}\cmidrule(l){7-8}
Shift & Class & Cause & \multicolumn{1}{c}{$n$} & $D{=}0$ & $D{=}1$ & $D{=}1$ & \multicolumn{1}{c@{}}{$D_{\mathrm{m}}$} \\
\midrule
$\Ssrc$ & definition & caused by the shift & \ExRevCauseMainSrcShiftDefinitionItems & \mtcipair{\ExAttrSrcShiftDefinitionBDzero} & \mtcipair{\ExAttrSrcShiftDefinitionBDone} & \mtcipair{\ExRevCauseMainSrcShiftDefinitionBxD} & \mtcipair{\ExRevCauseDmSrcShiftDefinitionBxD} \\
 &  & also without the shift & \ExRevCauseMainSrcDriftDefinitionItems & \mtcipair{\ExAttrSrcDriftDefinitionBDzero} & \mtcipair{\ExAttrSrcDriftDefinitionBDone} & \mtcipair{\ExRevCauseMainSrcDriftDefinitionBxD} & \mtcipair{\ExRevCauseDmSrcDriftDefinitionBxD} \\
\mtgap
 & other & caused by the shift & \ExRevCauseMainSrcShiftOtherItems & \mtcipair{\ExAttrSrcShiftOtherBDzero} & \mtcipair{\ExAttrSrcShiftOtherBDone} & \mtcipair{\ExRevCauseMainSrcShiftOtherBxD} & \mtcipair{\ExRevCauseDmSrcShiftOtherBxD} \\
 &  & also without the shift & \ExRevCauseMainSrcDriftOtherItems & \mtcipair{\ExAttrSrcDriftOtherBDzero} & \mtcipair{\ExAttrSrcDriftOtherBDone} & \mtcipair{\ExRevCauseMainSrcDriftOtherBxD} & \mtcipair{\ExRevCauseDmSrcDriftOtherBxD} \\
\mtgap
 & all & caused by the shift & \ExRevCauseMainSrcShiftAllItems & \mtcipair{\ExAttrSrcShiftAllBDzero} & \mtcipair{\ExAttrSrcShiftAllBDone} & \mtcipair{\ExRevCauseMainSrcShiftAllBxD} & \mtcipair{\ExRevCauseDmSrcShiftAllBxD} \\
 &  & also without the shift & \ExRevCauseMainSrcDriftAllItems & \mtcipair{\ExAttrSrcDriftAllBDzero} & \mtcipair{\ExAttrSrcDriftAllBDone} & \mtcipair{\ExRevCauseMainSrcDriftAllBxD} & \mtcipair{\ExRevCauseDmSrcDriftAllBxD} \\
\midrule
$\Sdef$ & definition & caused by the shift & \ExRevCauseMainDefShiftDefinitionItems & \mtcipair{\ExAttrDefShiftDefinitionBDzero} & \mtcipair{\ExAttrDefShiftDefinitionBDone} & \mtcipair{\ExRevCauseMainDefShiftDefinitionBxD} & \mtcipair{\ExRevCauseDmDefShiftDefinitionBxD} \\
 &  & also without the shift & \ExRevCauseMainDefDriftDefinitionItems & \mtcipair{\ExAttrDefDriftDefinitionBDzero} & \mtcipair{\ExAttrDefDriftDefinitionBDone} & \mtcipair{\ExRevCauseMainDefDriftDefinitionBxD} & \mtcipair{\ExRevCauseDmDefDriftDefinitionBxD} \\
 &  & new version & \ExAttrDefKappaRedefinedItems & \mtcipair{\ExAttrDefKappaRedefinedBDzero} & \mtcipair{\ExAttrDefKappaRedefinedBDone} & \mtcipair{\ExAttrDefKappaRedefinedBxD} & \multicolumn{1}{c@{}}{---} \\
 &  & same version & \ExAttrDefKappaKeptItems & \mtcipair{\ExAttrDefKappaKeptBDzero} & \mtcipair{\ExAttrDefKappaKeptBDone} & \mtcipair{\ExAttrDefKappaKeptBxD} & \multicolumn{1}{c@{}}{---} \\
\mtgap
 & other & also without the shift & \ExRevCauseMainDefDriftOtherItems & \mtcipair{\ExAttrDefDriftOtherBDzero} & \mtcipair{\ExAttrDefDriftOtherBDone} & \mtcipair{\ExRevCauseMainDefDriftOtherBxD} & \mtcipair{\ExRevCauseDmDefDriftOtherBxD} \\
\mtgap
 & all & also without the shift & \ExRevCauseMainDefDriftAllItems & \mtcipair{\ExAttrDefDriftAllBDzero} & \mtcipair{\ExAttrDefDriftAllBDone} & \mtcipair{\ExRevCauseMainDefDriftAllBxD} & \mtcipair{\ExRevCauseDmDefDriftAllBxD} \\
\midrule
$\Sdel$ & definition & caused by the shift & \ExRevCauseMainOrphanShiftDefinitionItems & \mtcipair{\ExAttrOrphanShiftDefinitionBDzero} & \mtcipair{\ExAttrOrphanShiftDefinitionBDone} & \mtcipair{\ExRevCauseMainOrphanShiftDefinitionBxD} & \mtcipair{\ExRevCauseDmOrphanShiftDefinitionBxD} \\
 &  & also without the shift & \ExRevCauseMainOrphanDriftDefinitionItems & \mtcipair{\ExAttrOrphanDriftDefinitionBDzero} & \mtcipair{\ExAttrOrphanDriftDefinitionBDone} & \mtcipair{\ExRevCauseMainOrphanDriftDefinitionBxD} & \mtcipair{\ExRevCauseDmOrphanDriftDefinitionBxD} \\
\mtgap
 & other & caused by the shift & \ExRevCauseMainOrphanShiftOtherItems & \mtcipair{\ExAttrOrphanShiftOtherBDzero} & \mtcipair{\ExAttrOrphanShiftOtherBDone} & \mtcipair{\ExRevCauseMainOrphanShiftOtherBxD} & \mtcipair{\ExRevCauseDmOrphanShiftOtherBxD} \\
 &  & also without the shift & \ExRevCauseMainOrphanDriftOtherItems & \mtcipair{\ExAttrOrphanDriftOtherBDzero} & \mtcipair{\ExAttrOrphanDriftOtherBDone} & \mtcipair{\ExRevCauseMainOrphanDriftOtherBxD} & \mtcipair{\ExRevCauseDmOrphanDriftOtherBxD} \\
\mtgap
 & all & caused by the shift & \ExRevCauseMainOrphanShiftAllItems & \mtcipair{\ExAttrOrphanShiftAllBDzero} & \mtcipair{\ExAttrOrphanShiftAllBDone} & \mtcipair{\ExRevCauseMainOrphanShiftAllBxD} & \mtcipair{\ExRevCauseDmOrphanShiftAllBxD} \\
 &  & also without the shift & \ExRevCauseMainOrphanDriftAllItems & \mtcipair{\ExAttrOrphanDriftAllBDzero} & \mtcipair{\ExAttrOrphanDriftAllBDone} & \mtcipair{\ExRevCauseMainOrphanDriftAllBxD} & \mtcipair{\ExRevCauseDmOrphanDriftAllBxD} \\
\bottomrule
\end{mttabular*}
\mtnotes{$n$: changed item-times. New or same version:
whether the definition item's version moved after $\Sdef$, which changes only definition items.
$D_{\mathrm{m}}$: the writer's programs as the $D{=}1$ arm, not split by version (---). Of the
\ExAttrSrcItemTimes{} item-times after each shift (\ExAttrWorldSlots{} items at four checkpoints in
\TestWorlds{} worlds), \ExAttrSrcChangedShareOfAll{}, \ExAttrDefChangedShareOfAll{}, and
\ExAttrOrphanChangedShareOfAll{} changed after $\Ssrc$, $\Sdef$, and $\Sdel$: \ExAttrSrcShiftShareOfAll{},
\ExAttrDefShiftShareOfAll{}, and \ExAttrOrphanShiftShareOfAll{} caused by the shift and
\ExAttrSrcDriftShareOfAll{}, \ExAttrDefDriftShareOfAll{}, and \ExAttrOrphanDriftShareOfAll{} also
without it. $\Ssrc$ revises the sources of \ExAttrSrcTargetSlots{} of the \ExAttrWorldSlots{} items
(\ExAttrSrcTargetSlotShare{}).}
\par\medskip
\begin{mttabular}{@{}ll r@{\,}l @{\hspace{14pt}} r@{\,}l @{\hspace{14pt}} r@{\,}l@{}}
\toprule
(b) Shift & Class & \multicolumn{2}{c}{$B{\times}D$, unshifted (pp)} & \multicolumn{2}{c}{Shifted $-$ unshifted (pp)}
  & \multicolumn{2}{c@{}}{Moved (\%)} \\
\midrule
$\Ssrc$ & definition & \mtci{\RThreeNSrcDefinitionNatural} & \mtci{\RThreeNSrcDefinitionDiff} & \mtci{\RThreeNSrcDefinitionNaturalMoved} \\
 & other & \mtci{\RThreeNSrcOtherNatural} & \mtci{\RThreeNSrcOtherDiff} & \mtci{\RThreeNSrcOtherNaturalMoved} \\
\mtgap
$\Sdef$ & definition & \mtci{\RThreeNDefDefinitionNatural} & \mtci{\RThreeNDefDefinitionDiff} & \mtci{\RThreeNDefDefinitionNaturalMoved} \\
\mtgap
$\Sdel$ & definition & \mtci{\RThreeNOrphanDefinitionNatural} & \mtci{\RThreeNOrphanDefinitionDiff} & \mtci{\RThreeNOrphanDefinitionNaturalMoved} \\
 & other & \mtci{\RThreeNOrphanOtherNatural} & \mtci{\RThreeNOrphanOtherDiff} & \mtci{\RThreeNOrphanOtherNaturalMoved} \\
\bottomrule
\end{mttabular}
\mtnotes{Unshifted: on the unshifted branch, which has the same events up to the fork and no shift.
Shifted $-$ unshifted: the paired difference from $B{\times}D$ at $D{=}1$ on the same item-times with the
shift, in (a). Moved: the share of these item-times whose value also changed on the unshifted branch.}
\end{table*}

\paragraph{With exact recomputation, the interaction depends mainly on the item class.} Update counts every
item-time whose gold differs from its value at the fork. This includes changes after the fork that would
have happened without the shift. \Cref{tab:cause} separates the changes that the shift caused from the
others. It gives the $B$ effects and $B{\times}D$ on each part, by item class, with exact recomputation
and with the writer's programs. On definition items, the interaction was the same whether the
redefinition changed the value or the value would have changed anyway (\ExAttrDefShiftDefinitionBxDEst{}
and \ExAttrDefDriftDefinitionBxDEst{}).

\paragraph{The interaction by cause.}\label{app:cause} Over all items, the interaction after a source
revision was \ExRevCauseMainSrcPooledAllBxD{} on all changes and \ExRevCauseMainSrcShiftAllBxD{} on the
changes the shift caused. After a redefinition the two values were \ExRevCauseMainDefPooledAllBxD{} and
\ExRevCauseMainDefShiftAllBxD{}, and after a deletion \ExRevCauseMainOrphanPooledAllBxD{} and
\ExRevCauseMainOrphanShiftAllBxD{}. A source revision and a deletion change mostly other items, where the
interaction is near zero, so their all-item values on the changes they caused are small. On definition
items, the interaction after every shift was about as large on the changes the shift caused as on the
others. With the writer's programs, it was \ExRevCauseDmDefShiftAllBxD{} on the changes that a
redefinition caused.

\paragraph{Small all-item values on shift-caused changes come from the mix of item classes.} The all-item
interaction on a set of item-times is a mean, so it equals the sum over classes of each class's share
times its value. On the item-times that a source revision itself changed, definition items make up
\ExRThreeNatSrcShiftDefinitionShare{} and contribute \ExRThreeNatSrcShiftContributionDefinition{}, and
the other items contribute \ExRThreeNatSrcShiftContributionOther{}. Together they give the all-item
value of \ExAttrSrcShiftAllBxD{}. After a deletion, the definition share is
\ExRThreeNatOrphanShiftDefinitionShare{}, and the two contributions are
\ExRThreeNatOrphanShiftContributionDefinition{} and \ExRThreeNatOrphanShiftContributionOther{}. If we
re-weight to the class mix of all changed item-times, the changes a source revision caused give
\ExRThreeNatSrcShiftAtPooledMix{} and the rest give \ExRThreeNatSrcDriftAtPooledMix{} (after a deletion,
\ExRThreeNatOrphanShiftAtPooledMixEst{} and \ExRThreeNatOrphanDriftAtPooledMixEst{}). The two parts are
different item-times, so this comparison is not paired. The unshifted branch gives a paired comparison.

\paragraph{After a source revision, the interaction on definition items also appears without the
shift.} The unshifted branch has the same events as the shifted branches up to the fork and no shift
after it. Four cells answered every item at every checkpoint on it. \Cref{tab:natural} gives the
interaction without the shift on the item-times that a shift changed, and the difference from the
interaction with the shift. This difference, shifted minus unshifted, was \RThreeNDefDefinitionDiff{}
after a redefinition and \RThreeNSrcDefinitionDiff{} after a source revision. After a source revision,
\RThreeNSrcDefinitionNaturalMovedEst{} of these item-times also changed without the shift, so the
interaction there is not specific to the shift. After a deletion, the difference on definition items was
\RThreeNOrphanDefinitionDiff{}. On the other items, no shift changes the interaction measurably.

\paragraph{The six definitions differ.} \Cref{tab:perdef} gives the interaction for each definition. It
varies widely, so at this level the six definitions are the effective sample. After a source revision it
ranges from \ExPerDefMainSrcActiveProjectBxDEst{} (active projects) to
\ExPerDefMainSrcImportantContactBxDEst{} (important contacts). Leaving one definition out moves the
class value between \ExPerDefLooMainSrcBxDMin{} and \ExPerDefLooMainSrcBxDMax{} after a source revision,
and between \ExPerDefLooMainDefBxDMin{} and \ExPerDefLooMainDefBxDMax{} after a redefinition.

\paragraph{The interaction holds on held-out definitions.} Each draw of six held-out definitions
(\cref{app:heldout-grammar}) replaced the six fixed definitions on ten test worlds, with Qwen3.5-9B as
writer and reader and every other request unchanged. After the fork, only definition items were asked.
The bootstrap draws worlds and definitions independently and takes the product sample. The interaction
is positive at both $D{=}1$ and $D_{\mathrm{m}}$: $B{\times}D$ was \RThreeGADefDoneBxD{} and
\RThreeGADefDmBxD{} after a redefinition, \RThreeGASrcDoneBxD{} and \RThreeGASrcDmBxD{} after a source
revision, and \RThreeGAOrphanDoneBxD{} and \RThreeGAOrphanDmBxD{} after a deletion. It was positive for
\RThreeGADefDonePositive{} of the 18 definitions after a redefinition (\cref{tab:heldout}).

\paragraph{Naming the definition in the probe changes little.} On all 20 worlds after a redefinition,
the six cells answered again with unnamed probes (\cref{app:worlds}). Items, keys, and writes were
unchanged. The effect of $B$ at $D{=}1$ was \RThreeQBEffectUnnamed{} without the name and
\RThreeQBEffectNamed{} with it (difference \RThreeQOneDiff{}). $B{\times}D$ was \RThreeQDoneBxDUnnamed{}
without the name, and \RThreeQDmBxDUnnamed{} with the writer's programs. An append-only memory that holds
items under both versions cannot choose between them without the name. However, the key helps mostly
where the append-only memory holds no item for the slot derived after the fork
(\RThreeUDeltaDoneDefDefinitionNoneEffectShare{} of its effect at $D{=}1$ after a redefinition,
\cref{app:res-keys}). There, the name has nothing to choose between.

\input{generated/tab_reader}

\subsection{Stored Value and Reader}
\label{app:res-reader}

\input{generated/tab_dm_deletion}
\input{generated/tab_json_arm}

\paragraph{With Qwen3.5-9B as reader, the interaction on definition items comes from the stored
value.}\label{app:store-reader} The terms of \cref{eq:decomp} are based on the reader's answers, so $r$
mixes two things: whether the re-derivation stored the right value, and whether the reader used it. For
each changed item-time, we read the item's stored state at the probe time (at $B{=}0$, the newest served
item) and judge it with the scorer's matcher. An item that is not served stores no value. Let
$r_{\mathrm{store}}$ be the share of right stores, $u$ the share of right stores that the reader answers
correctly, and $l$ the share of wrong stores that it still answers correctly. Then
$r=r_{\mathrm{store}}u+(1{-}r_{\mathrm{store}})l$, and the $B{\times}D$ interaction splits exactly into a
store part and a reader part (\cref{tab:store-reader-bxd,tab:store-reader}). On definition items, the
reader used a right store in \ExRThreeSRMainSrcDefinitionDzeroBoneUEst{} of cases at $D{=}0$ and
\ExRThreeSRMainSrcDefinitionDoneBoneUEst{} at $D{=}1$ after a source revision. The reader part of the
interaction was \ExRThreeSRMainSrcDefinitionReaderPart{}, \ExRThreeSRMainDefDefinitionReaderPart{},
and \ExRThreeSRMainOrphanDefinitionReaderPart{} after the three shifts. A failed re-derivation of these
items is thus a wrong stored value, not a reading error.

\begin{figure*}[t]
\centering
\includegraphics[width=\linewidth]{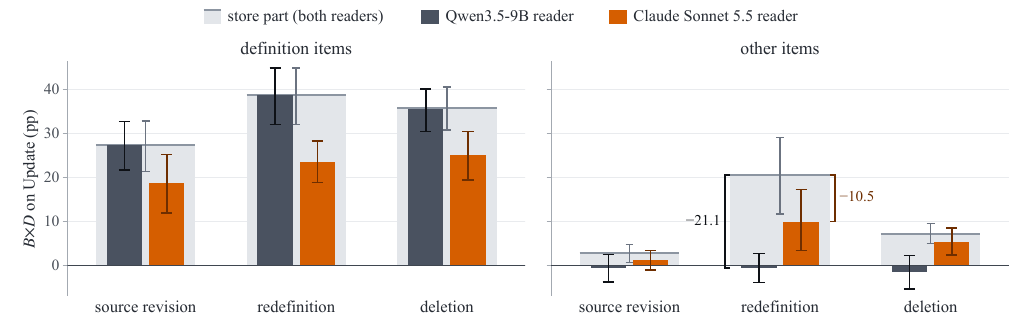}
\caption{$B{\times}D$ on Update with each reader of the same stores, and its store part, by item class
and shift (exact recomputation as $D{=}1$, source key). The stored values carry the interaction. Qwen3.5-9B
loses it on the other items, and Claude Sonnet 5.5 keeps part of it. Wide gray bars: the store part, the same
for both readers, with each reader's bar in front of it. The gap between the top of the gray bar and a
reader's bar is what that reader adds or loses, printed for the other items after a redefinition. Error
bars: 95\% intervals, the store part's between the two readers.}
\label{fig:reader}
\end{figure*}

\paragraph{On the other items, the reader loses what $D{=}1$ stores.} Here $D{=}1$ raised the share of
right stores, after a redefinition from \ExRThreeSRMainDefOtherDzeroBoneRStoreEst{} to
\ExRThreeSRMainDefOtherDoneBoneRStoreEst{}. When the right store was a value, the reader answered
correctly in \ExRThreeSRMainDefOtherDoneBoneUAnswerEst{} of cases. When the right store was ``no value'',
it answered correctly in only \ExRThreeSRMainDefOtherDoneBoneUAbstain{}
(\ExRThreeSRMainOrphanOtherDoneBoneUAbstain{} after a deletion). On deleted records,
\ExRThreeDelDoneDNoValueCount{} of the \ExRThreeDelDoneDCount{} reader errors of the $D{=}1$ cell after
a deletion answer ``there is no current value'' but with the status \texttt{answer}
(\cref{tab:dm-deletion}). The near-zero interaction on these items is thus a gain in the store that the
reader loses, not a re-derivation that already succeeds at $D{=}0$. When the $D{=}1$ store is served as
JSON records, the stored state is the same by construction. The reader then used a right store on
definition items in \ExRThreeSRMainSrcDefinitionJsonBoneUEst{},
\ExRThreeSRMainDefDefinitionJsonBoneUEst{}, and \ExRThreeSRMainOrphanDefinitionJsonBoneUEst{} of cases
after the three shifts, so the loss from the JSON format (\cref{tab:json-arm}) lies entirely in the
reader. The writer's programs store the right definition value less often
(\ExRThreeSRDmSrcDefinitionDoneBoneRStoreEst{} after a source revision), and their interaction there
again comes from the store. After a deletion, their interaction on the other items has a negative store
part (\ExRThreeSRDmOrphanOtherStorePart{}) and a negative reader part
(\ExRThreeSRDmOrphanOtherReaderPart{}, \cref{app:dm-deletion}). Note that $u$ includes finding the
statement among all served statements, which the answer logs cannot separate from reading.

\paragraph{A stronger reader keeps more of what the store holds.} Claude Sonnet 5.5 read every probe of
the six cells and the JSON arm in the main study's stores, with the same reader request and no thinking
before the answer. None of its \RThreeYAnswers{} answers was unreadable or cut off. With this reader, the
interaction on the other items is positive after a redefinition and after a deletion
(\RThreeYDefOtherDoneBxD{} and \RThreeYOrphanOtherDoneBxD{}). The stores are the same, so the store part
of $B{\times}D$ is unchanged, and the difference between readers lies in the reader part
(\cref{tab:reader,fig:reader}). After a redefinition, the reader part on the other items is
\RThreeYSRMainDefOtherReaderPart{}, compared with \ExRThreeSRMainDefOtherReaderPart{} for Qwen3.5-9B. Of
the \RThreeYSRMainShiftsOtherRightAbstain{} right ``no value'' stores on the other items at $D{=}1$, this
reader misread \RThreeYSRMainShiftsOtherLostAbstain{}, compared with
\ExRThreeSRMainShiftsOtherDoneBoneLostAbstain{} for Qwen3.5-9B. After a source revision, the interaction
on the other items stays near zero (\RThreeYSrcOtherDoneBxD{}). With the writer's programs, it is
\RThreeYDefOtherDmBxD{} after a redefinition and \RThreeYOrphanOtherDmBxD{} after a deletion, where the
programs' stored values cause the loss.

\paragraph{On definition items, the gap between readers comes from re-derivation without exact
recomputation.} On definition items, the interaction with this reader is lower than with Qwen3.5-9B
(\RThreeYSrcDefinitionDoneDiff{} after a source revision and \RThreeYDefDefinitionDoneDiff{} after a
redefinition). The gap arises at $D{=}0$ with a key. There, this reader answered
\RThreeYSRMainSrcDefinitionDzeroBoneREst{} of the changed item-times correctly after a source revision,
although only \RThreeYSRMainSrcDefinitionDzeroBoneRStoreEst{} stored a correct value. It recomputed
overdue tasks and active projects itself from the records quoted in the stored statement.

\subsection{What Repairs}
\label{app:res-repair}

\paragraph{Programs over the typed records repair much of what exact recomputation repairs.} Four
contrasts describe the writer's programs. On the accuracy on re-derived items, $D_{\mathrm{m}}$ minus
$D{=}0$ (source key) was \SynthCOneSrc{} after a source revision and \SynthCOneDef{} after a
redefinition. $D{=}1$ minus $D_{\mathrm{m}}$ was \SynthCTwoSrc{} and \SynthCTwoDef{}, so
\SynthCTwoShareSrcEst{} and \SynthCTwoShareDefEst{} of the effect of $D$ need exact recomputation. With
$D_{\mathrm{m}}$ as the $D{=}1$ arm, the interaction over all items was \SynthCThreeDef{} after a
redefinition and \SynthCThreeSrc{} after a source revision. $D_{\mathrm{m}}$'s effect on missed
invalidation was \SynthCFourSrcEst{} after both shifts. Before any shift, the programs, all written by
Qwen3.5-9B to run on typed records, run for \SynthExecAllEst{} of \SynthItemsAll{} items and agree with
gold on \SynthAgreeAllEst{}. Two definition programs fail. The program for working hours has a syntax
error and never runs (\SynthExecKappaWorkingHoursEst{} of items, \cref{app:errors}). The program for VIP
senders runs but agrees with gold on \SynthAgreeKappaVipSenderEst{} of items. After a deletion,
$D_{\mathrm{m}}$ has a lower accuracy on re-derived items than $D{=}0$ (\SynthExCOneOrphan{}), and the
interaction over all items becomes negative (\SynthExCThreeOrphan{}, \cref{app:dm-deletion}).

\begin{table*}[t]
\caption{The writer's programs ($D_{\mathrm{m}}$) on the definition items: (a) without and with execution
feedback (\ExecFbWorlds{} worlds), (b) on typed and on extracted fields (\FieldExWorlds{} worlds), (c)
written from the definitions in the user's own words (three program sets per writer).}
\label{tab:feedback}\label{tab:extract}\label{tab:userwords}
\centering\mtsetup
\setlength{\tabcolsep}{4pt}
\input{generated/tab_programs}
\par\medskip
\input{generated/tab_userwords}
\end{table*}

\paragraph{The writer's programs keep the interaction on definition items.} The class split in
\cref{tab:classsplit} also uses $D_{\mathrm{m}}$ as the $D{=}1$ arm. On definition items, the interaction
with the writer's programs was \ExClsDmSrcDefinitionBxDEst{} after a source revision and
\ExClsDmDefDefinitionBxDEst{} after a redefinition. After a redefinition, this is close to the
\ExClsMainDefDefinitionBxDEst{} of exact recomputation, although the level of Update is lower. On changed
definition items after a source revision, the programs raise Update with a key over $D{=}0$ by
\ExClsDmSrcDefinitionDBone{}, compared with \ExClsMainSrcDefinitionDBone{} for exact recomputation. After
a deletion at $B{=}1$, the programs change Update on the other items by \ExClsDmOrphanOtherDBone{}
compared with $D{=}0$, and the interaction there is \ExClsDmOrphanOtherBxD{}. The definition items rely
on at most twelve programs (six definitions with two versions each), shared by both $B$ arms and every
world. Here and in the feedback and extraction cells below, the intervals thus resample worlds,
not programs.

\paragraph{Execution feedback lets more programs run but repairs no more items.}\label{app:feedback} Two
further $D_{\mathrm m}$ cells, with and without a key, send a failing program back to the writer
(\ExecFbWorlds{} test worlds, all three shifts). A program fails when it does not end with a valid typed
value. This happens with a reply without code or without a function to call, a syntax error, an
exception, a timeout, a sandbox violation, or a value of the wrong type. The writer then receives the
program request, its own reply, and the interpreter's message (at most 600 characters), and it may
rewrite the program, for at most two rounds. The message contains no gold value and no expected value.
The first program that runs is used and kept for the item's later writes. If all rounds fail, the
outcome is that of $D_{\mathrm m}$. Of all \ExecFbMessages{} feedback requests, before and after the
fork, \ExecFbMessagesWorldData{} contained a string value of the records, so the corrections do not
depend on the world.

\Cref{tab:feedback} gives the results. Of the program runs for definition items after the fork,
\ExecFbValidWithoutAll{} ended with a valid value without feedback and \ExecFbValidWithAll{} with it,
with \ExecFbRequestsAll{} feedback requests over \ExecFbRunsAll{} runs. Only the working-hours programs
failed after the fork (\cref{app:errors}), and feedback made \ExecFbValidWithWorkingHoursEst{} of their
runs valid. It corrected the program that raised an exception, but not the program with a syntax error,
which failed again in both rounds. Repair did not change. On the changed definition item-times, $r$
moved by \ExecFbRDiffSrc{} after a source revision and \ExecFbRDiffDef{} after a redefinition. Update on
definition items after a deletion moved by \ExecFbDelUpdateDiffBone{} with a key, and $B{\times}D$ on
definition items by \ExecFbBxDDiffSrcDefItemsEst{}, \ExecFbBxDDiffDefDefItemsEst{}, and
\ExecFbBxDDiffDelDefItemsEst{} after the three shifts. On the other items, feedback raised the share of
valid runs from \ExecFbValidWithoutOtherItemsEst{} to \ExecFbValidWithOtherItemsEst{}, and $B{\times}D$
moved by \ExecFbBxDDiffSrcOtherItemsEst{}, \ExecFbBxDDiffDefOtherItemsEst{}, and
\ExecFbBxDDiffDelOtherItemsEst{}. The errors that cost repair are thus wrong values from programs that
run, such as the values returned for deleted records (\cref{app:dm-deletion}), which the interpreter
does not report.

\paragraph{Programs on fields that the writer extracted from prose keep the
interaction.}\label{app:extract} Two further $D_{\mathrm m}$ cells, with and without a key, run the same
programs on fields that the writer extracted from prose, instead of on the generator's typed fields
(\FieldExWorlds{} test worlds, source revision and redefinition). Qwen3.5-9B at temperature~0 first
rewrote every version of every record of the program sources (contacts, calendar, tasks, e-mail, chat
messages, and notes) from its typed fields into one to three sentences. Newsletter e-mails, on which no
item depends, stay typed. The prompt lists each value as the exact string that the prose must contain,
with fixed words for enumerations and booleans (such as ``starred'' or ``done''), and it allows no lists
or ``field: value'' lines. A check requires every value verbatim. A rewrite that misses a value is
retried once with the missing values named. If it still misses one, it is kept and counted as missed.
The writer then extracts each source's fields from the prose with a JSON schema for that source, given
the field description of the program requests. The programs read the extracted fields together with the
record metadata that the store keeps (identifier, source, revision, update time, and cluster). The
writer's own requests, the stores of the other cells, and the reader see the original records, so the
contrast changes only what the programs read.

\Cref{tab:extract} gives the results. Of the \FieldExRecordsAllSources{} distinct record contents,
\FieldExRecordCorrectAllSources{} were extracted with every field equal to the typed field after the
scorer's string normalization (case, punctuation, and whitespace, with attendees as a set). The share was
\FieldExRecordCorrectContactsEst{} for contacts, \FieldExRecordCorrectCalendarEst{} for calendar
records, \FieldExRecordCorrectMessagesEst{} for chat messages, \FieldExRecordCorrectNotesEst{} for notes,
\FieldExRecordCorrectEmailEst{} for e-mails, and \FieldExRecordCorrectTasksEst{} for tasks, where most
errors were in the title. A value was missing from \FieldExRewriteMissedAllSources{} of the rewrites and
from \FieldExRewriteMissedEmail{} of the e-mail rewrites. On the definition items, $B{\times}D$ on Update
stayed positive, \FieldExBxDExtractedSrcDefItems{} after a source revision and
\FieldExBxDExtractedDefDefItems{} after a redefinition. The paired differences from typed fields were
\FieldExBxDDiffSrcDefItemsEst{} and \FieldExBxDDiffDefDefItemsEst{}. Repair correctness on the changed
definition item-times fell from \FieldExRTypedSrcEst{} to \FieldExRExtractedSrcEst{} after a source
revision and from \FieldExRTypedDefEst{} to \FieldExRExtractedDefEst{} after a redefinition. Whenever the
value of a definition item differed between the two inputs, its write had read a record with an
extraction error. On the other items, $B{\times}D$ did not move (\FieldExBxDDiffSrcOtherItems{} and
\FieldExBxDDiffDefOtherItems{}). The writer's programs thus keep the interaction when they read fields
that the writer itself extracted from rewritten prose.

\paragraph{The interaction holds for definitions in the user's own words.} Nine program sets, written
from the statements of \cref{app:userwords}, ran in the two $D_{\mathrm{m}}$ cells on all 20 worlds.
They are three samples each from Qwen3.5-9B at temperature 0.7 and from Claude Sonnet 5.5 and Opus 5.5
with the settings of \cref{app:models}. After the fork, only definition items were asked. With intervals
over worlds and program sets, $B{\times}D$ on definition items was \RThreeVDefBxD{} after a redefinition,
\RThreeVSrcBxD{} after a source revision, and \RThreeVOrphanBxD{} after a deletion. The design rule to
compute definition items with a program over typed records (\cref{tab:rules}) thus also applies to a
definition that the user states in their own words, given the memory's data model. We also compared each
program set with the same writer's programs written from the codified statement, using the same samples.
The user's own words changed Update on changed definition items with a key by
\RThreeVQwenUpdatePooledDiff{} for Qwen3.5-9B, \RThreeVSonnetUpdatePooledDiff{} for Sonnet, and
\RThreeVOpusUpdatePooledDiff{} for Opus. They changed agreement with gold before the shift by
\RThreeVQwenAgreeDiffEst{}, \RThreeVSonnetAgreeDiffEst{}, and \RThreeVOpusAgreeDiffEst{}
(\cref{tab:userwords}).

\paragraph{Over programs from six writers, the interaction stays positive.} Eighteen program sets, written
from the codified statements, ran in the two $D_{\mathrm{m}}$ cells on all 20 worlds. There are three
samples from each of six writers. Qwen3.5-9B, gemma-4-12B, Llama-3.1-8B, and gpt-oss-20b sampled at
temperature 0.7, and Claude Sonnet 5.5 and Opus 5.5 at the API's default sampling. After the fork, only
definition items were asked. With intervals over worlds and program sets, $B{\times}D$ on definition
items was \RThreeGBDefBxD{} after a redefinition, \RThreeGBSrcBxD{} after a source revision, and
\RThreeGBOrphanBxD{} after a deletion. After a redefinition, it was \RThreeGBQwenDefBxDEst{} with the
programs of Qwen3.5-9B, \RThreeGBSonnetDefBxDEst{} with Sonnet, \RThreeGBOpusDefBxDEst{} with Opus, and
\RThreeGBGemmaDefBxDEst{} with gemma-4-12B, but only \RThreeGBLlamaDefBxD{} with Llama-3.1-8B and
\RThreeGBGptOssDefBxD{} with gpt-oss-20b. The writer explains almost all of the variance of this
interaction across the 18 program sets ($\eta^2=\RThreeGBDefWriterEtaSq{}$).

\paragraph{Claude writers re-derive definition items far better than Qwen3.5-9B.} On ten test worlds,
Claude Sonnet 5.5 wrote the definition items after the shift in all six cells, with the main writer's
request text. It also wrote the programs of the $D_{\mathrm{m}}$ cells. The main study's Qwen3.5-9B
outputs supply every other item and the store before the shift. Claude Opus 5.5 ran the same design with
the same settings, except that it wrote programs at low effort, because its thinking cannot be switched
off. None of the \RThreeHSonnetWrites{} distinct writes of either model stopped at the output limit or
was unreadable (the longest used \RThreeHSonnetTokensMax{} and \RThreeHOpusTokensMax{} of 17,000
tokens). $B{\times}D$ on definition items at $D{=}1$ was \RThreeHSonnetSrcDefinitionDoneBxD{},
\RThreeHSonnetDefDefinitionDoneBxD{}, and \RThreeHSonnetOrphanDefinitionDoneBxD{} after the three shifts
with Sonnet, and \RThreeHOpusSrcDefinitionDoneBxD{}, \RThreeHOpusDefDefinitionDoneBxD{}, and
\RThreeHOpusOrphanDefinitionDoneBxD{} with Opus. At $D{=}0$, the key alone raised Update on changed
definition items by \ExDetWriterSonnetSrcDefinitionBDzero{}, \ExDetWriterSonnetDefDefinitionBDzero{},
and \ExDetWriterSonnetOrphanDefinitionBDzero{} with Sonnet, and by \ExDetWriterOpusSrcDefinitionBDzero{},
\ExDetWriterOpusDefDefinitionBDzero{}, and \ExDetWriterOpusOrphanDefinitionBDzero{} with Opus. With
Qwen3.5-9B on 20 worlds, the values were \ExClsMainSrcDefinitionBDzeroEst{},
\ExClsMainDefDefinitionBDzeroEst{}, and \ExClsMainOrphanDefinitionBDzeroEst{} (\cref{tab:classsplit}).
With each writer's own programs, $B{\times}D$ after a redefinition was \RThreeHSonnetDefDefinitionDmBxD{}
and \RThreeHOpusDefDefinitionDmBxD{}, and Opus's programs answered as exact recomputation did. After a
source revision, $r_0$ on all changed definition item-times was \RThreeHSonnetRZero{} for Sonnet and
\RThreeHOpusRZero{} for Opus, compared with \RThreeHSonnetRZeroQwen{} for Qwen3.5-9B on the same
item-times, a gain of \RThreeHSonnetRZeroDiff{} and \RThreeHOpusRZeroDiff{}. The remaining interaction
comes from the item-times whose full definition scope was missing from the writer's selection, including
every close-friends and VIP-sender item-time (\cref{tab:coverage}).

\paragraph{The writer's selection never held the full definition scope of two
definitions.}\label{app:coverage} An item's \emph{definition scope} is the set of records that its
definition ranges over. The $D{=}0$ writer reads a 6,000-token BM25 selection of the records of the
item's sources (\cref{app:holdmem}), while a $D_{\mathrm{m}}$ program reads every typed record of these
sources. For each definition, \cref{tab:coverage}a gives the share of the definition-scope records that
the selection held, and the share of item-times at which it held all of them (complete definition
scope), over the definition item-times after the shift. The selection held the full definition scope of
working hours, active projects, and overdue tasks at every item-time, and of important contacts at
\ExRevCovImportantContactCompleteShare{} of them. It never held the full definition scope of close
friends or VIP senders, and it held \ExRevCovCloseFriendScopeShare{} and \ExRevCovVipSenderScopeShare{}
of their definition-scope records.

\paragraph{With the full definition scope, the Claude writers re-derived as well as exact
recomputation.} \Cref{tab:coverage}b splits $r_0$, $r_1$, and $B{\times}D$ on changed definition
item-times by complete or incomplete definition scope. Where the scope was complete, Claude Sonnet 5.5
and Opus 5.5 reached an $r_0$ of \ExRevCovSonnetDefCompleteRZero{} and \ExRevCovOpusDefCompleteRZero{}
after a redefinition, equal to their $r_1$. $B{\times}D$ was then negative
(\ExRevCovSonnetDefCompleteDoneBxD{} and \ExRevCovOpusDefCompleteDoneBxD{}). With equal repair, the
interaction is $a_0-a_1$, and the append-only memory serves better values at $D{=}1$. Where the scope
was incomplete, $r_0$ fell to \ExRevCovSonnetDefIncompleteRZero{} and \ExRevCovOpusDefIncompleteRZero{},
and $B{\times}D$ rose to \ExRevCovSonnetDefIncompleteDoneBxD{} and \ExRevCovOpusDefIncompleteDoneBxD{}.
Qwen3.5-9B stayed far below $r_1$ even with the complete definition scope (\ExRevCovQwenDefCompleteRZero{}
after a redefinition and \ExRevCovQwenSrcCompleteRZero{} after a source revision). For this writer, the
gap at $D{=}0$ is thus a failure to compute the value. For the Claude writers, the gap lies where the
selection was incomplete. Completeness goes with the definition: close friends and VIP senders are never
complete, and three other definitions always are. This split thus cannot separate completeness from
the definition itself. Giving the writer every record does (\cref{tab:typed}).

\paragraph{The full-access control added uncut records for only two definitions.} The full-access writer
reads every live record of the item's sources, up to \CtlFullLimit{} tokens (\cref{app:full-access}).
\Cref{tab:coverage}a also gives its re-derivations per definition, the share cut at the limit, and the
share whose request equaled the $D{=}0$ request. Every close-friends and VIP-sender request was cut,
\ExRevFullAllTruncatedShare{} of all its re-derivations. For active projects and overdue tasks, the
request always equaled the $D{=}0$ request, because the selection already held every record of their
sources. The control thus tested more records without a cut only on important contacts and working
hours, and only with Qwen3.5-9B as writer.

\paragraph{The complete typed inputs of two definitions exceed the writer's
context.}\label{app:input-sizes} A $D_{\mathrm{m}}$ program receives every typed source record of its
item. \Cref{tab:input-sizes} gives the size of this input per definition item-time after the shift, in
Qwen3.5-9B tokens, in the most compact of three forms. The first form writes each record as a JSON
object. The second leaves out the bodies of e-mails and messages, which no definition reads. The third
also leaves out the bodies and writes one table per source, with the column names once and one row per
record. The fit column gives the share of item-times at or below 31,000 tokens, which is the writer's
32,768-token context minus the request and the answer. As tables without bodies, the inputs of close
friends and VIP senders reached \ExRevSizeMainCloseFriendColumnsNoBodyMax{} and
\ExRevSizeMainVipSenderColumnsNoBodyMax{} tokens, and none of them fit. On the held-out definitions,
\ExRevSizeHeldoutAllColumnsNoBodyFit{} of the item-times fit in this form.

\paragraph{Given every record as typed fields, Claude Sonnet 5.5 comes close to exact recomputation,
and Qwen3.5-9B does not.}\label{app:typed} A further $D{=}0$ cell with a key re-derives the definition
items written after the fork from every live record of the definition's sources, instead of from the
writer's 6,000-token selection (\TypedInWorlds{} test worlds, source revision and redefinition). The
records are the typed fields that the $D_{\mathrm m}$ programs read, as one table per source without
message bodies (the third form of \cref{tab:input-sizes}). They are uncut and follow the field
description of the program requests. The rest of the request is the $D{=}0$ request. Qwen3.5-9B and
Claude Sonnet 5.5 each wrote the definition items with the settings of \cref{app:models}, and the main
study's Qwen3.5-9B outputs supply every other item and the store before the shift. The requests had a
median of \TypedInTokensMedianAll{}, a 95th percentile of \TypedInTokensPninetyfiveAll{}, and a maximum
of \TypedInTokensMaxAll{} Qwen3.5-9B tokens, with medians of \TypedInTokensMedianCloseFriend{} for close
friends and \TypedInTokensMedianVipSender{} for VIP senders. Qwen3.5-9B was thus served with a
65,536-token context here. Of the \TypedInQwenWrites{} typed writes of each writer,
\TypedInQwenUnreadableShare{} of Qwen3.5-9B's and \TypedInSonnetUnreadableShare{} of Sonnet's were
unreadable, and \TypedInQwenAtLimitShare{} and \TypedInSonnetAtLimitShare{} stopped at the output limit.

\Cref{tab:typed} gives $r$ on the changed definition item-times, pooled over the two shifts. Claude
Sonnet 5.5 re-derived \TypedInSonnetRTypedBothAll{} correctly with every record, compared with
\TypedInSonnetRSelBothAll{} from its selection and \TypedInSonnetRAuthoredBothAll{} for exact
recomputation (exact minus typed \TypedInSonnetRAuthoredMinusTypedBothAll{}). Where its selection had
been complete, every record changed $r$ by \TypedInSonnetRTypedMinusSelBothAllSelComplete{}, so the
typed format itself costs no measurable accuracy. Where it had been incomplete, $r$ rose from
\TypedInSonnetRSelBothAllSelIncompleteEst{} to \TypedInSonnetRTypedBothAllSelIncompleteEst{}. Close
friends and VIP senders, whose records the selection never held in full, reached
\TypedInSonnetRTypedBothCloseFriendEst{} and \TypedInSonnetRTypedBothVipSenderEst{}. For this writer,
their low $r_0$ thus came from the selection, not from the definitions (\cref{tab:coverage}). Important
contacts stayed below exact recomputation (\TypedInSonnetRTypedBothImportantContactEst{} against
\TypedInSonnetRAuthoredBothImportantContactEst{}). Qwen3.5-9B re-derived \TypedInQwenRTypedBothAll{}
correctly with every record and \TypedInQwenRSelBothAll{} with its selection, compared with
\TypedInQwenRAuthoredBothAll{} for exact recomputation. Before the fork, its typed derivation equaled
gold for \TypedInQwenTzeroCorrectAll{} of the definition items, and Sonnet's for
\TypedInSonnetTzeroCorrectAll{}. What the 9B writer lacks is thus the ability to compute the value, not
the evidence. Claude Sonnet 5.5 with every record comes close to exact recomputation, at the cost of
reading every record in each re-derivation.

\paragraph{A check alone does not repair.} The check-only cell had a lower accuracy on re-derived
definition items than $D{=}0$ (\EOneExCheckOnlyRepairDefinitionSrcEst{} after a source revision and
\EOneExCheckOnlyRepairDefinitionDefEst{} after a redefinition). Over all items, adding exact
recomputation to the checker raised the accuracy on re-derived items by \EOneExFallbackRepairSrcEst{}
and \EOneExFallbackRepairDefEst{}.

\paragraph{The controls.} Each control keeps the main study's reader context, probe set, and answers
before the shift. The scope cell uses the citation key on the other items.

\input{generated/tab_view}

\paragraph{More records did not help Qwen3.5-9B, and the template does not replace exact
recomputation.} A $D{=}0$ Qwen3.5-9B writer given every source record gained
\CtlFGainRepSrcDefinitionEst{} and \CtlFGainRepDefDefinitionEst{} in accuracy on re-derived definition
items after a source revision and a redefinition. Exact recomputation was still ahead by
\CtlFRepSrcDefinitionEst{} and \CtlFRepDefDefinitionEst{}. The control added uncut records only for
important contacts and working hours (\cref{tab:coverage}). Qwen3.5-9B's re-derivation also fails where
its selection held the full definition scope, so for this writer the advantage of $D$ comes from
computation. The control did not run with the Claude writers, but the typed-input cell (\cref{tab:typed})
gives every record uncut to both Qwen3.5-9B and Claude Sonnet 5.5. Rendering the $D{=}0$ value in the
$D{=}1$ template changed Update on definition items by \CtlTUpdSrcDefinition{}, \CtlTUpdDefDefinition{},
and \CtlTUpdOrphanDefinition{} after a source revision, a redefinition, and a deletion. Over all items
after a deletion, it changed Update by \CtlTUpdOrphanAll{}. The template is applied to every item
written after the shift, while $D{=}1$ applies only to items whose check failed.

\paragraph{On definition items, a program run at query time comes within 10\,pp of the stored
program.}\label{app:view} On Update of definition items, the query-time view minus the stored programs
was \CtlZVsDonesSrcDefinition{} after a source revision and \CtlZVsDonesDefDefinition{} after a
redefinition, both within $\pm$10\,pp. On the other items, the view fell below no-store BM25 after a
source revision (\CtlZUpdSrcOtherEst{}).

\paragraph{On the other items, the stored programs beat the same programs run at query time.}
\Cref{tab:view} compares the view and the stored cell on every item class and on Stability. The
differences here are the view minus the stored $D_{\mathrm{m}}$ cell with a key. Pooled over the three
shifts, they were \ExRThreeViewPooledAllUpdateVsDm{} on Update and \ExRThreeViewPooledAllStabilityVsDm{}
on Stability. On the other items, they were \ExRThreeViewSrcOtherUpdateVsDmEst{} and
\ExRThreeViewDefOtherUpdateVsDmEst{} on Update after a source revision and a redefinition, with a tie
after a deletion. On Stability they ranged from \ExRThreeViewOrphanOtherStabilityVsDmEst{} to
\ExRThreeViewDefOtherStabilityVsDmEst{} after every shift. On definition items, the stored cell also won
after a redefinition (\ExRThreeViewDefDefinitionUpdateVsDm{}), and Update tied after a source revision
and a deletion. On the other items, the stored cell keeps the writer's span-checked value and uses the
program's output only when the check fails, while the view uses the program's value first. The losses
fit a wrong program value that displaces correct live evidence. The view was ahead only on Stability of
definition items (\ExRThreeViewPooledDefinitionStabilityVsDm{},
\ExRThreeViewPooledDefinitionStabilityItems{} item-times), and only because of working hours, whose
program never runs (\cref{app:errors}). Without working hours, the difference on Stability of definition
items was \ExRThreeViewWithoutWorkingHoursDefinitionStabilityVsDm{}.

\paragraph{The query-time view costs more to serve and less to write.} Serving the view cost
\ExRThreeViewServePooledVsDm{} reader tokens per query more than serving the stored cell (paired). On the
write side, the view requests one program per asked item and definition version, which took
\ExRThreeViewWriteViewSrcIngest{} tokens per world after a source revision. The stored cell writes all 48
items, with \ExRThreeViewWriteDmSrcIngest{} tokens of ingestion and \ExRThreeViewWriteDmSrcMaintain{}
of maintenance. Probes coincide with maintenance, so the two cells are equally fresh, and the comparison
is between a stored value and its program.

\paragraph{Across writers, $B{\times}D$ tracks $r_1-r_0$ on the other items but not on definition
items.} \Cref{fig:account} plots $B{\times}D$ against $r_1-r_0$ with the source key for the main study,
the replication, and the three other model families as writer and reader, and \cref{tab:writers} gives
the terms. Over \ExAcctFitTwoCells{} combinations of four writers, two classes, and three shifts, with
the replication held out, the interaction tracks $r_1-r_0$ ($\rho={}$\ExAcctFitTwoSpearman{},
least-squares slope \ExAcctFitTwoSlope{}, and intercept \ExAcctFitTwoIntercept{}). It also does so within
the other items ($\rho={}$\ExAcctFitTwoOtherSpearman{}), but not within definition items
($\rho={}$\ExAcctFitTwoDefinitionSpearman{}). There, $r_1-r_0$ is large for every writer, but the
interaction after a source revision ranges from \ExAcctGemSrcDefinitionBxDEst{} (gemma-4-12B) to
\ExAcctGptSrcDefinitionBxDEst{} (gpt-oss-20b). The source key missed no changed item-time (with detection
counted from the fork), so \cref{eq:interaction} becomes $B{\times}D=(r_1-r_0)-(a_1-a_0)$. On definition
items, the effect of $D$ at $B{=}0$, $a_1-a_0$, is large and differs by writer, because writers differ in
what the append-only memory derives from the delta. For Qwen3.5-9B it is
\ExClsMainSrcDefinitionDBzeroEst{}, \ExClsMainDefDefinitionDBzeroEst{}, and
\ExClsMainOrphanDefinitionDBzeroEst{} after the three shifts (\cref{tab:classsplit}). For Llama-3.1-8B,
re-derivation also fails on the other items after a deletion (interaction
\ExAcctLlamaOrphanOtherBxDEst{}). After a redefinition, its interaction on the other items is
\ExAcctLlamaDefOtherBxD{}.

\subsection{Keys and Refresh}
\label{app:res-keys}

\begin{table*}[t]
\caption{The key ($B{=}1$) against the refresh and the upsert, two $B{=}0$ variants, on the same changed
item-times.}
\label{tab:refresh-binding}\label{tab:refresh-upsert}
\centering\mtsetup
\setlength{\tabcolsep}{4pt}
\begin{mttabular}{@{}ll N{00.0}N{00.0}N{00.0}N{00.0} @{\hspace{14pt}} r@{\,}l r@{\,}l @{\hspace{14pt}} r@{\,}l r@{\,}l@{}}
\toprule
& & \multicolumn{4}{c}{Update, definition items (\%)} & \multicolumn{4}{c}{$R$, definition items}
  & \multicolumn{4}{c@{}}{$B{=}1$ $-$ refresh (pp)} \\
\cmidrule(lr){3-6}\cmidrule(lr){7-10}\cmidrule(l){11-14}
Shift & $D$ & \multicolumn{1}{c}{$B{=}0$} & \multicolumn{1}{c}{Refresh} & \multicolumn{1}{c}{Upsert}
  & \multicolumn{1}{c}{$B{=}1$} & \multicolumn{2}{c}{Refresh} & \multicolumn{2}{c}{Upsert}
  & \multicolumn{2}{c}{definition} & \multicolumn{2}{c@{}}{other} \\
\midrule
$\Ssrc$ & 0 & \mspct\RThreeUDzeroSrcDefinitionUpdateBzeroEst & \mspct\RThreeUDzeroSrcDefinitionUpdateRefreshEst & \mspct\RThreeUDzeroSrcDefinitionUpdateUpsertEst & \mspct\RThreeUDzeroSrcDefinitionUpdateBoneEst & \mtci{\RThreeUDzeroSrcDefinitionR} & \mtci{\RThreeUDzeroSrcDefinitionUpsertR}
  & \mtci{\ExRevRefreshDzeroSrcDefinitionBoneMinusRefresh} & \mtci{\ExRevRefreshDzeroSrcOtherBoneMinusRefresh} \\
 & 1 & \mspct\RThreeUDoneSrcDefinitionUpdateBzeroEst & \mspct\RThreeUDoneSrcDefinitionUpdateRefreshEst & \mspct\RThreeUDoneSrcDefinitionUpdateUpsertEst & \mspct\RThreeUDoneSrcDefinitionUpdateBoneEst & \mtci{\RThreeUDoneSrcDefinitionR} & \mtci{\RThreeUDoneSrcDefinitionUpsertR}
  & \mtci{\ExRevRefreshDoneSrcDefinitionBoneMinusRefresh} & \mtci{\ExRevRefreshDoneSrcOtherBoneMinusRefresh} \\
 & $D_{\mathrm{m}}$ & \mspct\RThreeUDmSrcDefinitionUpdateBzeroEst & \mspct\RThreeUDmSrcDefinitionUpdateRefreshEst & \mspct\RThreeUDmSrcDefinitionUpdateUpsertEst & \mspct\RThreeUDmSrcDefinitionUpdateBoneEst & \mtci{\RThreeUDmSrcDefinitionR} & \mtci{\RThreeUDmSrcDefinitionUpsertR}
  & \mtci{\ExRevRefreshDmSrcDefinitionBoneMinusRefresh} & \mtci{\ExRevRefreshDmSrcOtherBoneMinusRefresh} \\
\mtgap
$\Sdef$ & 0 & \mspct\RThreeUDzeroDefDefinitionUpdateBzeroEst & \mspct\RThreeUDzeroDefDefinitionUpdateRefreshEst & \mspct\RThreeUDzeroDefDefinitionUpdateUpsertEst & \mspct\RThreeUDzeroDefDefinitionUpdateBoneEst & \mtci{\RThreeUDzeroDefDefinitionR} & \mtci{\RThreeUDzeroDefDefinitionUpsertR}
  & \mtci{\ExRevRefreshDzeroDefDefinitionBoneMinusRefresh} & \mtci{\ExRevRefreshDzeroDefOtherBoneMinusRefresh} \\
 & 1 & \mspct\RThreeUDoneDefDefinitionUpdateBzeroEst & \mspct\RThreeUDoneDefDefinitionUpdateRefreshEst & \mspct\RThreeUDoneDefDefinitionUpdateUpsertEst & \mspct\RThreeUDoneDefDefinitionUpdateBoneEst & \mtci{\RThreeUDoneDefDefinitionR} & \mtci{\RThreeUDoneDefDefinitionUpsertR}
  & \mtci{\ExRevRefreshDoneDefDefinitionBoneMinusRefresh} & \mtci{\ExRevRefreshDoneDefOtherBoneMinusRefresh} \\
 & $D_{\mathrm{m}}$ & \mspct\RThreeUDmDefDefinitionUpdateBzeroEst & \mspct\RThreeUDmDefDefinitionUpdateRefreshEst & \mspct\RThreeUDmDefDefinitionUpdateUpsertEst & \mspct\RThreeUDmDefDefinitionUpdateBoneEst & \mtci{\RThreeUDmDefDefinitionR} & \mtci{\RThreeUDmDefDefinitionUpsertR}
  & \mtci{\ExRevRefreshDmDefDefinitionBoneMinusRefresh} & \mtci{\ExRevRefreshDmDefOtherBoneMinusRefresh} \\
\mtgap
$\Sdel$ & 0 & \mspct\RThreeUDzeroOrphanDefinitionUpdateBzeroEst & \mspct\RThreeUDzeroOrphanDefinitionUpdateRefreshEst & \mspct\RThreeUDzeroOrphanDefinitionUpdateUpsertEst & \mspct\RThreeUDzeroOrphanDefinitionUpdateBoneEst & \mtci{\RThreeUDzeroOrphanDefinitionR} & \mtci{\RThreeUDzeroOrphanDefinitionUpsertR}
  & \mtci{\ExRevRefreshDzeroOrphanDefinitionBoneMinusRefresh} & \mtci{\ExRevRefreshDzeroOrphanOtherBoneMinusRefresh} \\
 & 1 & \mspct\RThreeUDoneOrphanDefinitionUpdateBzeroEst & \mspct\RThreeUDoneOrphanDefinitionUpdateRefreshEst & \mspct\RThreeUDoneOrphanDefinitionUpdateUpsertEst & \mspct\RThreeUDoneOrphanDefinitionUpdateBoneEst & \mtci{\RThreeUDoneOrphanDefinitionR} & \mtci{\RThreeUDoneOrphanDefinitionUpsertR}
  & \mtci{\ExRevRefreshDoneOrphanDefinitionBoneMinusRefresh} & \mtci{\ExRevRefreshDoneOrphanOtherBoneMinusRefresh} \\
 & $D_{\mathrm{m}}$ & \mspct\RThreeUDmOrphanDefinitionUpdateBzeroEst & \mspct\RThreeUDmOrphanDefinitionUpdateRefreshEst & \mspct\RThreeUDmOrphanDefinitionUpdateUpsertEst & \mspct\RThreeUDmOrphanDefinitionUpdateBoneEst & \mtci{\RThreeUDmOrphanDefinitionR} & \mtci{\RThreeUDmOrphanDefinitionUpsertR}
  & \mtci{\ExRevRefreshDmOrphanDefinitionBoneMinusRefresh} & \mtci{\ExRevRefreshDmOrphanOtherBoneMinusRefresh} \\
\bottomrule
\end{mttabular}
\mtnotes{Update: point estimates. $R$: the share of the effect of $B$ on Update that a $B{=}0$ variant
recovers, $(x-B_0)/(B_1-B_0)$, not evaluated (---) when $B_1-B_0<5$\,pp. $B{=}1$ $-$ refresh: Update of
$B{=}1$ minus that of the refresh at each $D$, paired per item-time.}
\end{table*}

\paragraph{The oracle key keeps the effect of $B$ but not Stability.} \Cref{tab:keyaudit} audits the
source, oracle, and scope keys (\cref{app:controls}) on the test worlds without a model. The scope key
covers definition items only. Offline, the oracle key misses no change since the fork and invalidates
only \KeyAuditFalseOracleSrcAll{}, \KeyAuditFalseOracleDefAll{}, and \KeyAuditFalseOracleOrphanAll{} of
the unchanged item-times after the three shifts. Over the \KeyAuditWorlds{} worlds, it re-derives
single-field items \KeyAuditRederivOracleSrcRule{} times after a source revision, compared with
\KeyAuditRederivSourceSrcRule{} for the source key. It re-derives definition items at
\KeyAuditDefShareOracleMin{} to \KeyAuditDefShareOracleMax{} of their item-times, because the time
windows of definitions slide, and it invalidates about half of the unchanged definition item-times
(\KeyAuditFalseOracleOrphanDefinition{} to \KeyAuditFalseOracleDefDefinition{}). Run as cells, it keeps,
and slightly exceeds, the source key's Update with fewer re-derivations (\CtlGPoolAll{}). This estimate
is the mean of per-item paired differences pooled over the six contrasts of $D$ and branch on
\CtlWorlds{} worlds. Each contrast is thus weighted by its changed item-times, and the two deletion
contrasts carry half the weight. On definition items, the oracle key keeps the interaction of the source
key (\cref{tab:oracle-key}). Its $B{\times}D$ was \ExDetKeyOracleSrcDefinitionBxD{},
\ExDetKeyOracleDefDefinitionBxD{}, and \ExDetKeyOracleOrphanDefinitionBxD{} after the three shifts,
compared with \ExDetKeySourceSrcDefinitionBxDEst{}, \ExDetKeySourceDefDefinitionBxDEst{}, and
\ExDetKeySourceOrphanDefinitionBxDEst{} for the source key on the same worlds, and its effect at
$D{=}0$ stayed between \ExDetKeyOracleDefinitionBDzeroMin{} and \ExDetKeyOracleDefinitionBDzeroMax{}.
Like the source key, it helps mostly through repair. The oracle key loses Stability
(\CtlGStabDoneSrcDefinition{} on definition items at $D{=}1$ after a source revision). The source key's
Stability comes from re-deriving unchanged items whose value before the shift the writer got wrong, not
from detection.

\paragraph{On definition items, the key works by adding the new derivation.} On these items, the scope
key matches the source key on Update (\CtlKUpdSrcDefinition{} after a source revision) and beats the
citation key by \CtlKCitedSrcDefinition{}. On changed definition items at $D{=}0$ and $D{=}1$, the key
and the refresh give nearly the same Update. The key beats the refresh mainly on the other items after a
deletion, where it removes the stale item (\cref{tab:refresh-binding}). On the other items after a
deletion, the scope cell uses the citation key, so the two cells detect the same changes, and the scope
cell minus the citation-key cell on Update is \CtlKCitedOrphanOther{}.

\paragraph{The key compared with a refresh.}\label{app:refresh-binding} In these worlds, the source key
acts as a periodic full refresh. Some record of an item's sources arrives between almost any two
checkpoints, so the key invalidates nearly every unchanged item-time (\cref{tab:keyaudit}). The add-only
refresh (\cref{app:refresh-cells}) re-derives the same items with the same request but keeps the old
item. \Cref{tab:refresh-binding} compares the two on the same item-times. On changed definition items,
the key minus the refresh was \ExRevRefreshDzeroSrcDefinitionBoneMinusRefresh{},
\ExRevRefreshDzeroDefDefinitionBoneMinusRefresh{}, and
\ExRevRefreshDzeroOrphanDefinitionBoneMinusRefresh{} at $D{=}0$ after the three shifts, and
\ExRevRefreshDoneSrcDefinitionBoneMinusRefresh{} to \ExRevRefreshDoneDefDefinitionBoneMinusRefresh{} at
$D{=}1$. With the writer's programs, the key led by \ExRevRefreshDmSrcDefinitionBoneMinusRefresh{} to
\ExRevRefreshDmOrphanDefinitionBoneMinusRefresh{}. With the refresh as the $B$ arm, $B{\times}D$ on
definition items after a redefinition was \ExRevRefreshDoneDefDefinitionBxDRefresh{}, compared with
\ExRevCauseMainDefPooledDefinitionBxD{} with a key. Removing the stale item matters most on the other
items after a deletion. There, the key led the refresh by \ExRevRefreshDoneOrphanOtherBoneMinusRefresh{}
at $D{=}1$ and \ExRevRefreshDzeroOrphanOtherBoneMinusRefresh{} at $D{=}0$, because the refresh keeps the
item derived from the deleted record next to the new one.

\paragraph{A refresh recovers the effect of $B$, and an upsert does not.} The refresh and upsert cells
(\cref{app:refresh-cells}) extend the add-only control to $D{=}0$, $D_{\mathrm{m}}$, and
\RThreeUWorlds{} worlds (\cref{tab:refresh-upsert}). On definition items, the refresh recovers
\RThreeUDoneSrcDefinitionR{} and \RThreeUDoneDefDefinitionR{} of the effect of $B$ at $D{=}1$ after a
source revision and a redefinition, and \RThreeUDmSrcDefinitionR{} and \RThreeUDmDefDefinitionR{} at
$D_{\mathrm{m}}$. At $D{=}0$, the effect of $B$ after a redefinition is
\RThreeUDzeroDefDefinitionBEffect{}, too small for the ratio. After a source revision, the refresh
recovers all of it (\RThreeUTwoSrc{}). The upsert recovers \RThreeUThreeR{} of the effect of $B$ over the
same four combinations, so the effect comes from the key's re-derivation schedule. After a deletion at
$D{=}1$, the upsert's Update over all items is \RThreeUFourUpdateUpsertEst{}, compared with
\RThreeUFourUpdateBzeroEst{} for the append-only cell, a difference of \RThreeUFourDiff{}. Three other
model families also served as writer and reader on \MTwoLlamaWorlds{} worlds. Pooled over $D{=}1$ after a
source revision and a redefinition, the upsert recovered \RThreeUFiveGemmaR{} of the effect of $B$ on
definition items with gemma-4-12B, \RThreeUFiveLlamaR{} with Llama-3.1-8B, and \RThreeUFiveGptOssR{} with
gpt-oss-20b. In all three families, the effect thus comes from the key's schedule, as with the main
writer.

\paragraph{The key helps where the delta yields no new item, which an upsert cannot reach.} The upsert
differs from the append-only cell only on slots for which the delta yields a new item. At $D{=}1$, the
delta had yielded one by the checkpoint on \RThreeUDeltaDoneSrcDefinitionYieldedShare{} and
\RThreeUDeltaDoneDefDefinitionYieldedShare{} of the changed definition item-times after a source revision
and a redefinition, and there the append-only cell already answered
\RThreeUDeltaDoneSrcDefinitionYieldedUpdateBzeroEst{} and
\RThreeUDeltaDoneDefDefinitionYieldedUpdateBzeroEst{} correctly. On the rest, both $B{=}0$ cells were
right on at most \RThreeUDeltaDoneSrcDefinitionNoneUpdateUpsertEst{}, and the $B{=}1$ cell on
\RThreeUDeltaDoneSrcDefinitionNoneUpdateBoneEst{} and \RThreeUDeltaDoneDefDefinitionNoneUpdateBoneEst{}.
These item-times carried \RThreeUDeltaDoneSrcDefinitionNoneEffectShare{} and
\RThreeUDeltaDoneDefDefinitionNoneEffectShare{} of the effect of $B$ at $D{=}1$. The key earns this
effect through its re-derivation schedule, not by overwriting or removing the stale item. Over the three
shifts, the upsert changed \RThreeUDeltaDoneShiftsDefinitionMoved{} answers on definition items and
\RThreeUDeltaDoneShiftsOtherMoved{} on the other items. After a deletion, the delta yielded no new item
on \RThreeUDeltaDoneOrphanOtherNoneShare{} of the changed other item-times, so the upsert could not
replace their stale items.

\paragraph{The citation key mostly misses records that could not have been cited.} The citation key
never invalidated \ExIntMainSrcAllMOneCitedEst{}, \ExIntMainDefAllMOneCitedEst{}, and
\ExIntMainOrphanAllMOneCitedEst{} of the changed item-times after the fork, for a source revision, a
redefinition, and a deletion. Counted from the latest change, it missed
\ExDetMainSrcAllDoneBoneMissedCitedEst{}, \ExDetMainDefAllDoneBoneMissedCitedEst{}, and
\ExDetMainOrphanAllDoneBoneMissedCitedEst{}. By item class, counted from the fork, it missed
\ExIntMainSrcDefinitionMOneCitedEst{} of the changed definition item-times and
\ExIntMainSrcOtherMOneCitedEst{} of the other changed item-times after a source revision, and
\ExIntMainDefDefinitionMOneCitedEst{} and \ExIntMainDefOtherMOneCitedEst{} after a redefinition.
\Cref{fig:misses} classifies every changed item-time that the key missed by the state of the gold
dependency set and the writer's citations at derivation time. Either a record of the set changed that
the writer had not cited ($a$), or else a record entered the set ($b$), or else records left a time
window or a narrowed scope ($c$). Classes $b$ and $c$, records that entered or left the item's scope
after the item was derived, make up nearly all misses. A miss of class $a$ can also involve a record
that entered the set. This was the case for \ExMissClassMainSrcDefinitionAAndBEst{},
\ExMissClassMainDefDefinitionAAndBEst{}, and \ExMissClassMainOrphanDefinitionAAndBEst{} of the missed
definition item-times in the main study after the three shifts.

\paragraph{Most missed changes would also have happened without the shift.}\label{app:cited-misses}
\Cref{tab:cited-misses} splits the changed item-times that the citation key did not invalidate since the
fork by the cause of the change. Because the key holds the definition version, it missed none of the
\ExRevCitedDefShiftDefinitionItems{} definition item-times that a redefinition changed. After a
redefinition, every change it missed would thus also have happened without the shift: it missed
\ExRevCitedDefDriftAllMissed{} of these \ExRevCitedDefDriftAllItems{} item-times
(\ExRevCitedDefDriftAllMissShare{}). Counted from the latest change, it missed
\ExDetCitedDefShiftAllMissedLast{} of the \ExDetCitedDefShiftAllItems{} item-times
(\ExDetCitedDefShiftAllMissShareLast{}) whose gold changed again after the redefinition, through records
or the clock, and \ExDetCitedDefDriftAllMissedLast{} of the \ExDetCitedDefDriftAllItems{} others
(\ExDetCitedDefDriftAllMissShareLast{}). After a source revision, it missed
\ExRevCitedSrcShiftAllMissShare{} of the changes caused by the shift and
\ExRevCitedSrcDriftAllMissShare{} of the others. After a deletion, the shares were
\ExRevCitedOrphanShiftAllMissShare{} and \ExRevCitedOrphanDriftAllMissShare{}. Nearly all its misses come
from records that entered or left an item's scope after the item was derived, which no citation
recorded at write time can contain.

\paragraph{The citation key does not raise Update after a source revision or a redefinition.} Its effect
on Update, averaged over $D$, was \EOneBEffectPooledCitedSrc{} after a source revision and
\EOneBEffectPooledCitedDef{} after a redefinition. After a deletion, where the citations name the deleted
record, it helped (\EOneBEffectPooledCitedOrphan{}).

\paragraph{A missed item is served stale.} \Cref{tab:forced} gives the accuracies behind the citation
key's effect. On the other items at $D{=}1$ after a source revision, items that the citation key missed
are correct in \ExMissedMainSrcDoneOtherACitedEst{} of cases. The append-only cell, which derives a new
item from the delta, is correct on \ExMissedMainSrcDoneOtherAZeroOnMissedEst{} of the same items. The
key's average effect after a source revision thus hides a split. At $D{=}1$, its Update over $B{=}0$ is
\ExMissedMainSrcDoneDefinitionBEffect{} on definition items and \ExMissedMainSrcDoneOtherBEffect{} on the
other items.

\paragraph{Forcing a re-derivation of the missed items helps only where repair succeeds.}\label{app:forced}
On the changed item-times that the citation-key cell did not invalidate, the source-key cell re-derived
the same items with the same writer request. Its answers thus show what a forced re-derivation of the
missed items would serve (\cref{tab:forced}). At $D{=}0$, forcing gained little on definition items
(\ExRevForcedDzeroSrcDefinitionGain{}, \ExRevForcedDzeroDefDefinitionGain{}, and
\ExRevForcedDzeroOrphanDefinitionGain{} after the three shifts), because the writer's own re-derivation
of these items fails. At $D{=}1$, it gained \ExRevForcedDoneDefDefinitionGain{} to
\ExRevForcedDoneSrcDefinitionGain{}. On the other items, forcing gained at both levels of $D$
(\ExRevForcedDzeroSrcOtherGain{} at $D{=}0$ after a source revision), and the append-only cell served
about as many right answers on these item-times (\ExRevForcedDzeroSrcOtherBzero{}).

\paragraph{Changes that come from the clock.}\label{app:clock} The source key counts events per source and
does not track time. A change that comes from the clock alone, such as a task that becomes overdue or a
meeting that leaves a time window, changes no event count. The key catches it only when an unrelated
record of the item's sources arrives in the same interval between checkpoints. Over the three shifts of
the main study, the last gold change came from the clock alone for
\ExRevClockMainShiftsDefinitionClockOnly{} of the \ExRevClockMainShiftsDefinitionChanged{} changed
definition item-times (\ExRevClockMainShiftsDefinitionClockOnlyShare{}), and the key could not see
\ExRevClockMainShiftsDefinitionClockSilent{} of them. Of the \ExRevClockMainShiftsErrorsOneOne{} wrong
answers of \holdmem{}(1,1) on changed definition item-times, \ExRevClockMainShiftsErrorsOneOneClock{}
are such stale values (\ExRevClockRepShiftsErrorsOneOneClock{} of \ExRevClockRepShiftsErrorsOneOne{} in
the replication). If missed invalidation were counted from the fork, these errors would count as failed
repairs. Counted from the latest change, as in the main text, they are missed detections. The source key
then missed \ExRevClockMainSrcDefinitionDzeroBoneMissedLast{},
\ExRevClockMainDefDefinitionDzeroBoneMissedLast{}, and
\ExRevClockMainOrphanDefinitionDzeroBoneMissedLast{} of the changed definition item-times after the
three shifts, and $r_1$ stayed at \ExRevClockMainSrcDefinitionDoneBoneRepairLast{} or above. The citation
key missed \ExDetMainSrcDefinitionDzeroBoneMissedCitedEst{},
\ExDetMainDefDefinitionDzeroBoneMissedCitedEst{}, and \ExDetMainOrphanDefinitionDzeroBoneMissedCitedEst{}
of them, and its $r_1$ was \ExDetMainSrcDefinitionDoneBoneRepairCitedEst{} or above. With exact
recomputation, nearly every wrong answer of this key is thus a missed change. Counted from the fork, its
$r_1$ after a redefinition would be \ExDecMainDefDefinitionDoneBoneRepairCitedEst{}.

\subsection{Horizons, Cost, and No-Store BM25}
\label{app:res-horizon}

\begin{figure*}[t]
\centering
\includegraphics[width=\linewidth]{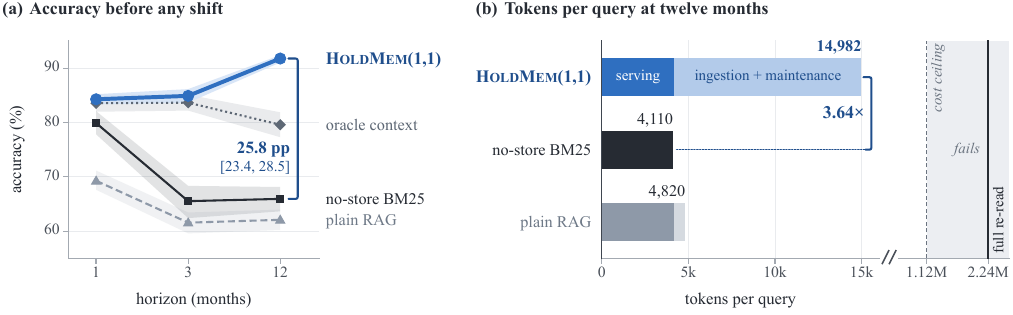}
\caption{No-store BM25 against \holdmem{}(1,1) over horizons. (a)~Accuracy before any shift, with
95\% intervals as bands. The bracket gives the paired difference at twelve months with its 95\% interval.
(b)~Tokens per query at twelve months, split into serving (dark) and amortized ingestion and
maintenance (light). The bracket gives the ratio of the two totals. Oracle context was not costed and
is not in (b). The axis breaks before the cost ceiling (dashed), which bounds serving at half of one full
re-read, and the gray region beyond it fails the ceiling.}
\label{fig:zero}
\end{figure*}
\paragraph{Horizons and cost.} \Cref{fig:zero} gives the accuracy by horizon and the tokens per query.
The cost model replays the maintenance schedule at the measured tokens per write. No-store BM25 and plain
RAG failed the utility floor. On the three-month worlds, the key at $D{=}0$ changed Update after a
redefinition by \ExClsShortDefAllBDzero{}, compared with \EOneBEffectDzeroDef{} on the twelve-month
worlds, and $D$ raised repair correctness on changed definition items after a redefinition by
\ExClsShortDefDefinitionDRepairChanged{}, compared with \ExClsMainDefDefinitionDRepairChanged{}.

\input{generated/tab_diagnostic_tests}

\input{generated/tab_zero_after}

\paragraph{At one and three months, \holdmem{}(1,1) passes the utility floor because of the definition
items.} The definition items contribute \EThreeExTThreeHoldDefinitionOneContribution{},
\EThreeExTThreeHoldDefinitionThreeContribution{}, and \EThreeExTThreeHoldDefinitionTwelveContribution{}
to the margin of \holdmem{}(1,1) over the utility floor at one, three, and twelve months. Without them,
\holdmem{}(1,1) passes the floor at \EThreeExNoDefTThreePassHold{} horizons instead of
\EThreeTThreePassHold{}.

\begin{figure*}[t]
\centering
\includegraphics[width=\linewidth]{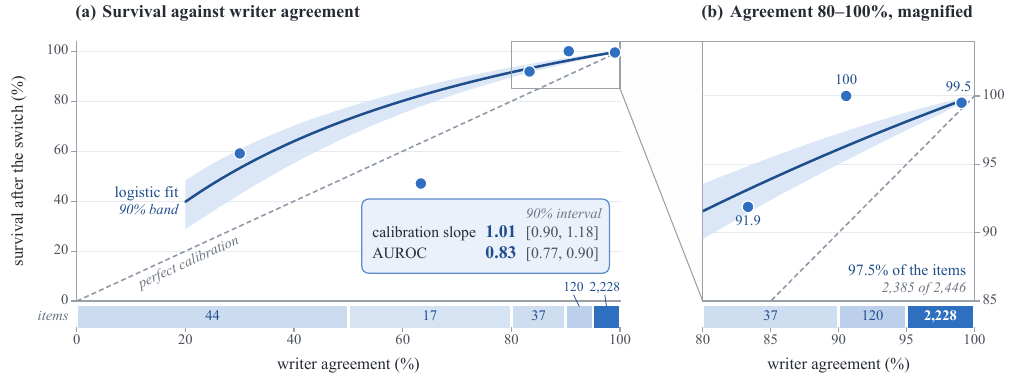}
\caption{Survival of a switch against writer agreement, for the \ETwoPathThreeN{} $D{=}0$, $B{=}1$ items
pooled over the three tiers. Points: the survival of each agreement bin (edges at 50, 80, 90, and 95\%)
at its mean agreement. The strip under each axis gives the items per bin, and its tint deepens with the
bin's share of the items. Line and band: the pooled logistic fit and its 90\% band. Dashed: perfect
calibration. The card in (a) gives the calibration slope of the fit and the AUROC of agreement with
90\% intervals. (b)~magnifies agreement from 80 to 100\%, where nearly all items lie, and prints the
survival of each bin.}
\label{fig:calibration}
\end{figure*}

\paragraph{Without a shift, stored state pays off over time.}\label{sec:res-e3} At twelve months, the
accuracy of no-store BM25 minus that of \holdmem{}(1,1) was \EThreeZeroMinusHoldTwelve{}. In a model of
its maintenance, the source key cost \ExCostModelMainTwelveHoldRatioZero{} times the tokens of no-store
BM25. In the main study, \holdmem{}(1,1) passed the three diagnostic tests that can fail, with the
retention test at the three tiers (\cref{tab:diagnostic-tests}). The fourth, the leaving test, cannot
fail between two implementations of one data model (\cref{app:experiments}).

\paragraph{After a deletion, no-store BM25 ties overall.} \Cref{tab:zero-after} gives Update and
Stability of no-store BM25, \holdmem{}(1,1), and the append-only \holdmem{}(0,0) by item class after each
shift. On Update over all items, no-store BM25 minus \holdmem{}(1,1) was
\ExZeroMainSrcAllUpdateDiffHoldOneOne{} after a source revision and
\ExZeroMainDefAllUpdateDiffHoldOneOne{} after a redefinition. After a deletion, no-store BM25 tied
\holdmem{}(1,1) on Update over all items (\ExZeroMainOrphanAllUpdateZeroEst{} against
\ExZeroMainOrphanAllUpdateHoldOneOneEst{}, a paired difference of
\ExZeroMainOrphanAllUpdateDiffHoldOneOne{}) and won on the other items
(\ExZeroMainOrphanOtherUpdateZeroEst{} against \ExZeroMainOrphanOtherUpdateHoldOneOneEst{}), because
no-store BM25 needs no key. \holdmem{}(1,1) won on Stability and on definition items after each shift of
the main experiment. The \EOneBEffectPooledOrphanEst{} that the key gains after a deletion is measured
against an append-only store that keeps the stale item: \holdmem{}(0,0) answered
\ExZeroMainOrphanOtherUpdateHoldZeroZeroEst{} of the changed other item-times correctly. The key thus
repairs a failure that stored state itself creates. An upsert does not get this benefit, because a
deletion yields no new item to replace the stale one.

\paragraph{The four diagnostic tests.} \Cref{tab:diagnostic-tests} gives the results of the retention
test, the leaving test, the utility floor, and the cost ceiling for each tier, switch, horizon, and
system, in the main study and the replication.

\subsection{Writer Switches}
\label{app:res-switch}

\input{generated/tab_switches}

\paragraph{Design of the eight switches.} On the \TestWorlds{} test worlds, the old writer (Qwen3.5-9B,
or Qwen3.8-27B for the downgrade) builds the four source-key cells one minute before the fork, with the
sources frozen. Each new writer then replaces it in turn and re-derives every $B{=}1$ item. Retention is
$P(\text{correct after}\mid\text{correct before})$. The agreement rate of an item type is the share of
items on which both writers are correct or both are wrong, measured on the calibration worlds. For writer
agreement as a predictor, we pool the eight switches (the three main-study tiers and the five other
writers), regress $D{=}0$, $B{=}1$ survival on agreement, and resample by world. In every switch, we also
compare the item-paired survival of $D{=}1$ and $D{=}0$ items at $B{=}1$, and the retention test applies
to each switch.

\paragraph{Retention stays above 97\% at all three tiers.} In the worst case, retention changed by
\ETwoTOneSame{} (90\% interval) relative to full retention, at the same tier. Pooled over the three tiers,
writer agreement predicted which of the \ETwoPathThreeN{} items survive the switches on the test worlds,
with a calibration slope of \ETwoExWriterAgreePooledSlope{} (90\% interval) and an AUROC of
\ETwoExWriterAgreePooledAurocEst{}. \ETwoExFailTotal{} items failed, and \ETwoExFailDefinition{} of them
were definition items.

\paragraph{Survival of a switch exceeds the checker pass rate.} At every tier, the $D{=}1$ survival of a
writer switch was higher than the checker pass rate: \ETwoDoneSurvivalSame{} against
\ETwoCheckerPassSame{} at the same tier, \ETwoDoneSurvivalUp{} against \ETwoCheckerPassUp{} for the
upgrade, and \ETwoDoneSurvivalDown{} against \ETwoCheckerPassDown{} for the downgrade. Exact
recomputation recomputes the items that the checker rejects.

\paragraph{The new writer's own accuracy predicts survival at least as well as agreement.} The predictor
is the per-type agreement of the old and the new writer on the calibration worlds
(\cref{fig:calibration}). Pooled over the three tiers, the new writer's per-type accuracy reaches an
AUROC of \ETwoExWriterNewAccPooledAurocEst{}, the old writer's \ETwoExWriterOldAccPooledAurocEst{}, and
an indicator of definition items alone \ETwoExWriterNonDefPooledAurocEst{}. Outside the definition items,
where few items fail, agreement does not predict survival (\ETwoExWriterNoDefAgreePooledAurocEst{}). Over
the eight switches, the new writer's accuracy reaches \MOneExWriterNewAccPooledAurocEst{} against
\MOneExWriterAgreePooledAurocEst{} for agreement (difference \MOneExWriterNewAccPooledAurocDiffType{},
with item types resampled). The old writer's accuracy (\MOneExWriterOldAccPooledAurocEst{}) and the
indicator (\MOneExWriterNonDefPooledAurocEst{}) fall behind both.

\paragraph{Pooled over the eight switches, agreement predicts survival.} In \cref{tab:switches}, the
pooled slope was \MOneExWriterAgreePooledSlope{} (90\% interval) and the AUROC
\MOneExWriterAgreePooledAurocEst{}. For \MOneCaveats{}, almost no item failed, so their slopes are not
usable. By point estimate, $D{=}1$ items survived at least as often as $D{=}0$ items in
\MOneMBHolding{} switches, and slightly less often after gemma-4-12B (\MOneMBGemmaEst{}) and gpt-oss-20b
(\MOneMBGptOssEst{}). Llama-3.1-8B, gpt-oss-20b, and Qwen3-8B did not pass the retention test.

\paragraph{After a same-tier switch, retention falls mainly on free text.} A same-tier switch makes
\ETwoExGainsDoneBoneSame{} items of \holdmem{}(1,1) right and \ETwoExLossesDoneBoneSame{} wrong, so net
accuracy rises by \ETwoExNetDoneBoneSameEst{}. The new writer's $D{=}1$ build recomputes items that the
old writer had wrong. Retention falls mainly on free text. There, the $D{=}1$ cell keeps
\ETwoExRetentionDoneBoneFreetextSameEst{} compared with \ETwoExRetentionDzeroBoneFreetextSameEst{} at
$D{=}0$, because the span check rejects a paraphrase.

\subsection{Generality}
\label{app:res-generality}

\paragraph{Design of the replication.} The replication repeats the main study on \RepWorlds{} fresh test
worlds that are disjoint from every other split (\cref{app:datasheet}). It runs the main experiment with
every cell and baseline in the three branches. It also runs the writer-switch experiment with the main
study's calibration agreement rates (not recomputed) and the horizon experiment at one, three, and
twelve months. The diagnostic tests apply to each tier, system, and horizon on the fresh worlds as well.

\paragraph{Every contrast with the source key replicates.}\label{app:replication} On the fresh worlds,
every contrast with the source key in \cref{tab:replication} has the main study's sign and a 95\%
interval that excludes zero. The exception is $D$'s effect on missed invalidation, which stays at zero
as in the main study. With the source key, the effect of $B$ on detection is near 100\,pp by design.
With the citation key, the effect of $B$ on Update after a source revision and after a redefinition has
an interval that includes zero in both studies (averaged over $D$: \RepBEffectUpdatePooledCitedSrcOrig{}
and \RepBEffectUpdatePooledCitedDefOrig{} in the main study, \RepBEffectUpdatePooledCitedSrc{} and
\RepBEffectUpdatePooledCitedDef{} on the fresh worlds). Its interaction after a source revision moves
from \RepBxDInteractionUpdateCitedSrcOrig{} to \RepBxDInteractionUpdateCitedSrc{}. On the fresh worlds,
the effect of $B$ on Update after a source revision, averaged over $D$, was \RepPathOneEstimate{}. $D$'s
effect on missed invalidation was \RepPathTwoMaxAbs{}, and writer agreement as a predictor of survival
had a slope of \RepPathThreeSlope{} and an AUROC of \RepPathThreeAuroc{}. The retention test, the
utility floor, and the cost ceiling give the same pass counts as in the main study
(\cref{tab:diagnostic-tests}).

\input{generated/tab_families}

\paragraph{With another family as writer and reader, the main contrasts keep their
direction.}\label{app:families} Other model families (\cref{tab:family-models}) replace Qwen3.5-9B in
three roles: as the new writer at a switch (\cref{app:res-switch}), as writer and reader of the main
experiment, and as reader only. The main writer's requests are unchanged, so the reader swaps and the
writer's programs reuse the main study's writer outputs. Each of Llama-3.1-8B, gemma-4-12B, and
gpt-oss-20b is writer and reader of the main experiment on \MTwoLlamaWorlds{} test worlds (all branches,
cells, and baselines), next to Qwen3.5-9B on the same worlds (\cref{tab:main}b). In all three families,
$D$'s effect on missed invalidation is zero. The following contrasts have 95\% intervals that exclude
zero in the main study's direction: the $B$ effect on Update after a source revision, the $D$ effect on
the accuracy on re-derived items, $B{\times}D$ on Update after a redefinition and after a deletion, and
the gap of no-store BM25 below \holdmem{}(1,1). \Cref{tab:writers} splits the terms by item class.

\paragraph{Other readers keep the signs of the main contrasts.}\label{app:robust} Each of the same
families also replaces only the reader, on \MThreeLlamaWorlds{} test worlds after a source revision and
a redefinition. Every swapped reader keeps the sign of the $B$ effect on Update, the $D$ effect on the
accuracy on re-derived items, and $B{\times}D$ in both branches (\cref{tab:families}), and $D$'s effect on
missed invalidation stays zero.

\paragraph{Swapping the reader or the embedder changes levels but keeps the signs of the effects of $B$
and $D$.} In the experiment with the probes on conflicts and consent (\cref{app:e7}), on
\ESevenThetaRWorlds{} test worlds, Ministral-3-14B replaces Qwen3.5-9B as reader. Its $B$ effect on Update
was \ESevenThetaRBUpdateSwapSrc{} and \ESevenThetaRBUpdateSwapDef{} after a source revision and a
redefinition (Qwen3.5-9B: \ESevenThetaRBUpdateMainSrc{} and \ESevenThetaRBUpdateMainDef{}). Its $D$
effect on the accuracy on re-derived items was \ESevenThetaRDRepairSwapSrc{} and
\ESevenThetaRDRepairSwapDef{} (Qwen3.5-9B: \ESevenThetaRDRepairMainSrc{} and
\ESevenThetaRDRepairMainDef{}). After a redefinition, plain RAG overtook no-store BM25 with this reader
(\ESevenThetaRRagSwapDef{} against \ESevenThetaRZeroSwapDef{}, while Qwen3.5-9B gave
\ESevenThetaRRagMainDef{} against \ESevenThetaRZeroMainDef{}). With bge-m3 as the embedding model of
plain RAG, its accuracy after the shift moved from \ESevenThetaERagMainSrc{} to \ESevenThetaERagSwapSrc{}
after a source revision and from \ESevenThetaERagMainDef{} to \ESevenThetaERagSwapDef{} after a
redefinition.

\paragraph{With rewritten message bodies, the five contrasts keep their direction, and the effect of $D$
on the accuracy on re-derived items shrinks.}\label{app:rewritten-results} We repeat the main experiment
on worlds in which gemma-4-12B rewrote the bodies of e-mails, chat messages, and notes
(\cref{app:rewritten}). These bodies are \RendRewrittenTokenShare{} of the body text, and each keeps its
gold surface forms (\RendFidelityMin{} found at the worst world and branch). Newsletter digests, which
hold no gold fact, stay templated and make up the rest of the body text. On the \RendWorlds{} template
worlds of the other model families, the experiment repeats their cells, branches, and probes with the
main writer and reader. Both conditions ask identical probes, so each contrast is paired by item
(rewritten minus template), with world-bootstrap intervals. On the \RendWorlds{} rewritten worlds, the
five contrasts have the same direction as on their template twins (\RendBUpdateSrc{}, \RendDRepairSrc{},
\RendDRepairDef{}, \RendBxDDef{}, and \RendZeroMinusHoldDef{}). The values for the template twins are in
the Qwen3.5-9B column of \cref{tab:main}b. The interaction moves by \RendDiffBxDDef{} after a
redefinition and by \RendDiffBxDSrc{} after a source revision. The $D$ effect on the accuracy on
re-derived items changes by \RendDiffDRepairSrc{} after a source revision and \RendDiffDRepairDef{} after
a redefinition. Definition items come from structured fields that the rewrite does not touch, so the
rewrite can affect only the other items.

\subsection{Native Systems}
\label{app:native}

\input{generated/tab_native_null}

\paragraph{Mem0 falls outside the interval of its cell only for Update after a source revision.} Mem0
kept \EFiveExMemBRateEst{} of the deleted items. By point estimate, this gives its documented setting
$(0,0)$. The interval of this share includes the 50\% threshold, so the setting is not identified by the
interval (\cref{tab:native-null}). Mem0 lies outside the interval of its cell only for Update after a
source revision (\EFiveMemUpdateSrc{} against \EFiveMemUpdatePredSrc{}). There, Mem0 lies inside the
intervals of no-store BM25 and plain RAG (\cref{tab:native-diff}), as do A-Mem and MemOS, the other
systems with its setting. Of its extraction outputs, \EFiveMemParseErrors{} (\EFiveMemParseErrorRate{})
broke off near Mem0's default output limit and were dropped, which counts as system behavior. A second
ingestion of \EFiveMemNoiseWorlds{} worlds with identical requests changed \EFiveMemNoiseDifferOrphan{} to
\EFiveMemNoiseDifferSrc{} of the answers after the three shifts and moved Update by
\EFiveMemNoiseUpdateSrc{} to \EFiveMemNoiseUpdateOrphan{}. Before the fork, Mem0's margin over the
oracle context was \EFiveMemTThreeEst{}, so it does not pass the utility floor, like no-store BM25
(\EFiveZeroTThreeEst{}) and plain RAG (\EFiveRagTThreeEst{}). Serving costs \EFiveMemTFour{} of a full
re-read. The retention test does not apply to Mem0, and its leaving test was not run.

\paragraph{How the native values are compared.} For the five further systems, we compare the setting
inferred from behavior with the documented one. We compare each system's Update and Stability after the
three shifts with the 95\% interval of the \holdmem{} cell that has its inferred setting, on the same worlds
and branch. For $B{=}1$, this is the citation-key cell for a document-level system and the source-key
cell otherwise. No-store BM25 and plain RAG, which store no derived item, are compared with the same
intervals, and we compare mean Stability after a deletion between systems with $B{=}1$ and $B{=}0$. A
native value says something about its setting only where it also lies outside the intervals of no-store
BM25 and plain RAG. Every system ingested \EFiveMemNoiseWorlds{} of the worlds a second time, and
\cref{tab:native-diff} widens its intervals by the variance between the two ingestions.

\paragraph{The settings are recovered from behavior, but they do not predict how a system does after a
shift.} By point estimate, the inferred setting equals the documented one for \NatNOneAgree{} systems.
\NatNThreeInside{} comparisons (\NatNThreeShare{}), Mem0 included, fall inside the interval of the cell
with the inferred setting (\cref{fig:native-systems}), and \cref{tab:native-null} gives the count for each
system. Mean Stability after a deletion was \NatNFourStabilityBOne{} for $B{=}1$ systems and
\NatNFourStabilityBZero{} for $B{=}0$ systems. $B$ is inferred from the deletion branch, so this last
comparison is not independent of the setting. The persistence intervals of Mem0 and MemOS include 0.5, so
\NatExNullBIdentified{} settings are identified by interval when Mem0 is counted.

\paragraph{After a source revision, all native systems except LightRAG update like plain RAG, whatever
their documented setting.} On Update after a source revision, every native system except LightRAG lies
inside the intervals of no-store BM25 and plain RAG, and its paired difference from no-store BM25 has a
95\% interval that includes zero (\cref{tab:native-diff}). On the comparisons that were not used to infer
the setting, \NatExNullInsideOosOf{} native values lie inside the interval of the cell with their setting,
and \NatExNullRagInsideOosOf{} values of plain RAG, which stores no derived item, lie inside the same
intervals. Per event, the five further systems made \NatMemosCostCalls{} to \NatLightRagCostCalls{} LLM
calls and used \NatHippoRagCostRatio{} to \NatLightRagCostRatio{} of the prompt and completion tokens of
Mem0 (\NatMemCostTokens{}).

\paragraph{The ranking after a source revision nearly fixes the ranking after a redefinition among the
\holdmem{} cells, but not among the native systems.} The rank correlation in \cref{sec:discussion} covers
every system measured on these worlds except the JSON arm. It also includes the redefinition and
deletion branches of the oracle context, no-store BM25, and plain RAG. Over all \RankExSystems{}
systems, the ranking by Update after a source revision correlates with the ranking after a redefinition
at Spearman $\rho={}$\RankExRhoSrcDef{} and with the ranking by Stability after a deletion at
\RankExRhoSrcOrphan{}. The first correlation comes mostly from the \holdmem{} cells, which differ widely
in Update. Among the \RankExSrcDefHoldMemSystems{} cells it is \RankExRhoSrcDefHoldMem{}, among the
\RankExSrcDefNativeSystems{} native systems \RankExRhoSrcDefNative{}, and among the native systems with
no-store BM25 and plain RAG \RankExRhoSrcDefNativeBase{}. Each interval comes from its own world
bootstrap. The native systems lie close together after both shifts, so their ranking after a source
revision does not settle their ranking after a redefinition.

\paragraph{Deleting a stale memory does not derive the new value.} Mem0 ran as in the native-system
experiment on \RThreeXWorlds{} of its three-month worlds, from its stores at the fork, with two
interventions. Source-keyed invalidation deletes the memories of a record's earlier revision before the
new revision or the deletion is ingested. The adapter selects these memories by their source record,
without gold, and deletes them through Mem0's own delete call, so Mem0's LLM decides none of the
deletions. A query-time view puts the value of the writer's program for a definition item, run at query
time on the live typed records, first in the evidence for that item. The view thus adds both an evidence
path, the typed records that Mem0 does not store, and the computation. Mem0 logged
\RThreeXNativeMutations{} updates or deletions after the fork, so $m{=}1$ after every shift and in every
item class, and Mem0's Update equals $a$, the accuracy of what it serves without invalidation. The item
class, not $m$, tells which intervention helps (\cref{tab:mem0x}). Mem0 answered
\RThreeXSrcDefinitionAEst{}, \RThreeXDefDefinitionAEst{}, and \RThreeXOrphanDefinitionAEst{} of the
changed definition item-times correctly after the three shifts. The query-time view raised Update on
them by \RThreeXSrcDefinitionRepEffect{}, \RThreeXDefDefinitionRepEffect{}, and
\RThreeXOrphanDefinitionRepEffect{}. Invalidation changed it by \RThreeXSrcDefinitionDetEffect{} to
\RThreeXOrphanDefinitionDetEffect{}, with intervals that include zero. Of the definition item-times whose
stale memories were deleted by invalidation, Mem0 then answered only \RThreeXDefDefinitionDetREst{} to
\RThreeXOrphanDefinitionDetREst{} correctly. On the other items, which the view does not touch,
invalidation raised Update by \RThreeXOrphanOtherDetEffect{} after a deletion, where records lose their
source. It changed Update by \RThreeXDefOtherDetEffect{} after a redefinition.

\subsection{Export, Consent, Conflicts, and Sync}
\label{app:e7}\label{app:sync}

The probes on conflicts and consent ran on \ESevenWorlds{} test worlds.

\paragraph{Moving to another implementation needs the keys.} Exported to Portable Memory with its keys
and imported by our second implementation, \holdmem{}(1,1) matched the original (\ESixWithExtPost{}) and
passed the leaving test. Without the keys, the import detected no later change, and its missed
invalidation rose by \ESixNoExtMissed{}. Both outcomes follow from the design (\cref{app:experiments}).

\paragraph{A key does not detect conflicts.} For \ESevenConflictItems{} contact attributes, we made two
live sources disagree, so the gold status is \texttt{abstain}. No \holdmem{} cell abstained. The share of
these probes answered with \texttt{abstain} was \ESevenAbstainDoneBone{} for \holdmem{}(1,1) and
\ESevenAbstainZero{} for no-store BM25. A key detects a changed source, not a disagreement, and both
values pass the span check.

\paragraph{Consent is applied through the read set.} After one agent's scope was revoked, $B{=}0$ cells
answered \ESevenLeakDzeroBzero{} of its blocked probes. $B{=}1$ cells revoke the items whose recorded
read set touches the scope. They still answered \ESevenLeakDoneBone{} of the blocked probes and blocked
\ESevenOverBlockDoneBone{} of the answerable ones. No-store BM25 filters the raw sources at query time and
leaked \ESevenLeakZero{}. Revoking by read set blocks the right items only when the read set equals the
dependency set.

\paragraph{Documented sync contracts cost accuracy with or without a key.} With the simulators of
\cref{app:sync-sim}, the accuracy on calendar probes changed by \EFourXCalendarDropBone{} with $B{=}1$,
\EFourXCalendarDropBzero{} with $B{=}0$, and \EFourXCalendarDropZero{} for no-store BM25. The sync
contracts and the key did not interact measurably on calendar probes (\EFourXCalendar{},
\EFourXCalendarN{} probes on \EFourXWorlds{} test worlds). Pooled over six sources, the interaction was
\EFourXPooled{}. The sync worlds sample only events that touch an item, and each event is probed on
every item it touches and on equally many items it does not touch.

%% file: generated/tab_store_reader.tex
\begin{table*}[t]
\caption{The stored value and the reader behind $r$ and $a$ of \cref{eq:decomp}, in percent (point estimates).}
\label{tab:store-reader}
\centering\mtsetup
\setlength{\tabcolsep}{6pt}
\begin{mttabular}{@{}lll N{0,000} N{00.0}N{100.0}N{100.0}N{00.0}N{00.0}N{00.0} @{\hspace{14pt}} N{00.0}N{00.0}@{}}
\toprule
& & & & \multicolumn{6}{c}{$B{=}1$: re-derived} & \multicolumn{2}{c@{}}{$B{=}0$} \\
\cmidrule(lr){5-10}\cmidrule(l){11-12}
Shift & Class & Cell & \multicolumn{1}{c}{$n$} & \multicolumn{1}{c}{$r_{\mathrm{store}}$} & \multicolumn{1}{c}{$u$} & \multicolumn{1}{c}{$u$, answer} & \multicolumn{1}{c}{$u$, abstain} & \multicolumn{1}{c}{$l$} & \multicolumn{1}{c}{$r$} & \multicolumn{1}{c}{$a_{\mathrm{store}}$} & \multicolumn{1}{c@{}}{$a$} \\
\midrule
$\Ssrc$ & definition & $D{=}0$ & 333 & 16.2 & 98.1 & 98.1 & \mtna{} & 0.4 & 16.2 & 10.2 & 10.2 \\
 &  & $D{=}1$ &  & 99.4 & 99.7 & 99.7 & \mtna{} & 0.0 & 99.1 & 66.1 & 65.8 \\
 &  & $D{=}1$, JSON &  & 99.4 & 70.1 & 70.1 & \mtna{} & 0.0 & 69.7 & \mtna{} & \multicolumn{1}{c@{}}{---} \\
 &  & $D_{\mathrm{m}}$ &  & 71.8 & 99.6 & 99.6 & \mtna{} & 7.4 & 73.6 & 38.4 & 40.2 \\
\mtgap
 & other & $D{=}0$ & 835 & 92.0 & 96.7 & 99.9 & 17.2 & 1.5 & 89.1 & 84.2 & 83.7 \\
 &  & $D{=}1$ &  & 95.4 & 94.9 & 99.7 & 40.9 & 2.6 & 90.7 & 84.9 & 85.9 \\
 &  & $D{=}1$, JSON &  & 95.4 & 93.6 & 100.0 & 22.7 & 0.0 & 89.3 & \mtna{} & \multicolumn{1}{c@{}}{---} \\
 &  & $D_{\mathrm{m}}$ &  & 93.5 & 95.5 & 100.0 & 39.7 & 3.7 & 89.6 & 78.7 & 79.8 \\
\midrule
$\Sdef$ & definition & $D{=}0$ & 387 & 6.7 & 100.0 & 100.0 & \mtna{} & 0.0 & 6.7 & 8.8 & 7.5 \\
 &  & $D{=}1$ &  & 99.7 & 100.0 & 100.0 & \mtna{} & 0.0 & 99.7 & 63.0 & 61.8 \\
 &  & $D{=}1$, JSON &  & 99.7 & 54.4 & 54.4 & \mtna{} & 0.0 & 54.3 & \mtna{} & \multicolumn{1}{c@{}}{---} \\
 &  & $D_{\mathrm{m}}$ &  & 66.4 & 100.0 & 100.0 & \mtna{} & 8.5 & 69.3 & 33.6 & 35.1 \\
\mtgap
 & other & $D{=}0$ & 171 & 73.7 & 98.4 & 100.0 & 33.3 & 2.2 & 73.1 & 70.2 & 70.8 \\
 &  & $D{=}1$ &  & 96.5 & 75.8 & 100.0 & 7.0 & 0.0 & 73.1 & 72.5 & 71.3 \\
 &  & $D{=}1$, JSON &  & 96.5 & 74.5 & 100.0 & 2.3 & 0.0 & 71.9 & \mtna{} & \multicolumn{1}{c@{}}{---} \\
 &  & $D_{\mathrm{m}}$ &  & 91.8 & 79.0 & 100.0 & 5.7 & 7.1 & 73.1 & 67.3 & 67.3 \\
\midrule
$\Sdel$ & definition & $D{=}0$ & 308 & 11.4 & 100.0 & 100.0 & \mtna{} & 0.4 & 11.7 & 4.9 & 4.2 \\
 &  & $D{=}1$ &  & 99.4 & 100.0 & 100.0 & \mtna{} & 0.0 & 99.4 & 57.1 & 56.5 \\
 &  & $D{=}1$, JSON &  & 99.4 & 67.0 & 67.0 & \mtna{} & 0.0 & 66.6 & \mtna{} & \multicolumn{1}{c@{}}{---} \\
 &  & $D_{\mathrm{m}}$ &  & 74.0 & 100.0 & 100.0 & \mtna{} & 15.0 & 77.9 & 31.8 & 34.7 \\
\mtgap
 & other & $D{=}0$ & 1,459 & 73.0 & 92.4 & 100.0 & 91.8 & 2.3 & 68.1 & 6.2 & 10.0 \\
 &  & $D{=}1$ &  & 81.1 & 82.9 & 100.0 & 81.7 & 1.4 & 67.5 & 7.1 & 10.9 \\
 &  & $D{=}1$, JSON &  & 81.1 & 19.9 & 100.0 & 13.9 & 0.0 & 16.1 & \mtna{} & \multicolumn{1}{c@{}}{---} \\
 &  & $D_{\mathrm{m}}$ &  & 52.9 & 71.8 & 100.0 & 68.4 & 0.7 & 38.3 & 6.4 & 13.0 \\
\bottomrule
\end{mttabular}
\mtnotes{$n$: changed item-times. $r_{\mathrm{store}}$: the share of re-derived item-times whose stored state (status and typed value) is the gold. $u$, $l$: accuracy given a right or a wrong store, so $r=r_{\mathrm{store}}u+(1{-}r_{\mathrm{store}})l$. $u$ is also given by gold status: answer, or abstain where the right state is no value. $a_{\mathrm{store}}$, $a$: the same at $B{=}0$, where the store is the newest served item. JSON: the $D{=}1$ store served as JSON records. ---: not defined, as every changed definition item-time has a value as its right state and the JSON store has no $B{=}0$ cell.}
\end{table*}

%% file: generated/tab_typed.tex
\begin{table*}[t]
\caption{Given every live record of the definition's sources as typed fields, Claude Sonnet 5.5 raises $r$ on the changed definition item-times close to that of exact recomputation, while Qwen3.5-9B stays far below it.}
\label{tab:typed}
\centering\mtsetup
\setlength{\tabcolsep}{2pt}
\begin{mttabular*}{\linewidth}{l N{000.0} N{000}@{\hspace{6pt}}>{\mtpairwidths{100.0}{[00.0,\,100.0]}}c>{\mtpairwidths{100.0}{[00.0,\,100.0]}}c>{\mtpairwidths{100.0}{[00.0,\,100.0]}}c>{\mtpairwidths{$+$00.0}{[$+$00.0,\,$+$100.0]}}c>{\mtpairwidths{$+$100.0}{[$+$00.0,\,$+$100.0]}}c@{}}
\toprule
& & & \multicolumn{3}{c}{$r$ (\%)} & \multicolumn{2}{c@{}}{Difference (pp)} \\
\cmidrule(lr){4-6}\cmidrule(l){7-8}
& \multicolumn{1}{c}{\mthead{Complete}{(\%)}} & \multicolumn{1}{c}{$n$} & Selection & Typed & Exact & Typed $-$ selection & \multicolumn{1}{c@{}}{Exact $-$ typed} \\
\midrule
\mtgroup{8}{(a) Qwen3.5-9B}
important contacts & 50.0 & 67 & \mtpair{0.0}{} & \mtpair{0.0}{} & \mtpair{100.0}{} & \mtpair{0.0}{} & \mtpair{$+$100.0}{} \\
close friends & 0.0 & 79 & \mtpair{0.0}{} & \mtpair{0.0}{} & \mtpair{100.0}{} & \mtpair{0.0}{} & \mtpair{$+$100.0}{} \\
working hours & 100.0 & 39 & \mtpair{20.5}{[0.0,\,34.7]} & \mtpair{20.5}{[0.0,\,34.7]} & \mtpair{100.0}{} & \mtpair{0.0}{} & \mtpair{$+$79.5}{[$+$65.3,\,$+$100.0]} \\
active projects & 100.0 & 40 & \mtpair{42.5}{[12.5,\,72.7]} & \mtpair{12.5}{[0.0,\,37.5]} & \mtpair{100.0}{} & \mtpair{$-$30.0}{[$-$62.5,\,0.0]} & \mtpair{$+$87.5}{[$+$62.5,\,$+$100.0]} \\
VIP senders & 0.0 & 72 & \mtpair{0.0}{} & \mtpair{0.0}{} & \mtpair{100.0}{} & \mtpair{0.0}{} & \mtpair{$+$100.0}{} \\
overdue tasks & 100.0 & 69 & \mtpair{7.2}{[0.0,\,17.8]} & \mtpair{18.8}{[9.0,\,29.7]} & \mtpair{98.6}{[95.7,\,100.0]} & \mtpair{$+$11.6}{[$+$3.0,\,$+$22.2]} & \mtpair{$+$79.7}{[$+$69.1,\,$+$89.5]} \\
\addlinespace[2pt]
all six & 58.3 & 366 & \mtpair{8.2}{[3.3,\,14.0]} & \mtpair{7.1}{[4.2,\,10.3]} & \mtpair{99.7}{[99.2,\,100.0]} & \mtpair{$-$1.1}{[$-$5.8,\,$+$3.2]} & \mtpair{$+$92.6}{[$+$89.3,\,$+$95.6]} \\
\mtin after $\Ssrc$ &  & 165 & \mtpair{12.7}{[3.8,\,21.7]} & \mtpair{9.1}{[2.6,\,15.4]} & \mtpair{99.4}{[98.2,\,100.0]} & \mtpair{$-$3.6}{[$-$10.1,\,$+$2.5]} & \mtpair{$+$90.3}{[$+$83.7,\,$+$97.1]} \\
\mtin after $\Sdef$ &  & 201 & \mtpair{4.5}{[0.0,\,9.0]} & \mtpair{5.5}{[3.2,\,8.0]} & \mtpair{100.0}{} & \mtpair{$+$1.0}{[$-$3.4,\,$+$4.9]} & \mtpair{$+$94.5}{[$+$92.0,\,$+$96.8]} \\
\mtin selection complete &  & 176 & \mtpair{17.0}{[7.1,\,28.5]} & \mtpair{14.8}{[8.8,\,21.0]} & \mtpair{99.4}{[98.3,\,100.0]} & \mtpair{$-$2.3}{[$-$12.0,\,$+$6.8]} & \mtpair{$+$84.7}{[$+$78.1,\,$+$91.0]} \\
\mtin selection incomplete &  & 190 & \mtpair{0.0}{} & \mtpair{0.0}{} & \mtpair{100.0}{} & \mtpair{0.0}{} & \mtpair{$+$100.0}{} \\
\midrule
\mtgroup{8}{(b) Claude Sonnet 5.5}
important contacts & 50.0 & 67 & \mtpair{56.7}{[35.8,\,75.8]} & \mtpair{65.7}{[53.1,\,76.7]} & \mtpair{88.1}{[75.4,\,97.0]} & \mtpair{$+$9.0}{[$-$2.9,\,$+$19.5]} & \mtpair{$+$22.4}{[$+$9.0,\,$+$35.2]} \\
close friends & 0.0 & 79 & \mtpair{1.3}{[0.0,\,3.8]} & \mtpair{97.5}{[93.8,\,100.0]} & \mtpair{100.0}{} & \mtpair{$+$96.2}{[$+$92.4,\,$+$100.0]} & \mtpair{$+$2.5}{[0.0,\,$+$6.2]} \\
working hours & 100.0 & 39 & \mtpair{100.0}{} & \mtpair{100.0}{} & \mtpair{97.4}{[90.6,\,100.0]} & \mtpair{0.0}{} & \mtpair{$-$2.6}{[$-$9.4,\,0.0]} \\
active projects & 100.0 & 40 & \mtpair{92.5}{[83.3,\,100.0]} & \mtpair{90.0}{[77.3,\,100.0]} & \mtpair{97.5}{[93.2,\,100.0]} & \mtpair{$-$2.5}{[$-$6.8,\,0.0]} & \mtpair{$+$7.5}{[0.0,\,$+$16.7]} \\
VIP senders & 0.0 & 72 & \mtpair{0.0}{} & \mtpair{98.6}{[95.6,\,100.0]} & \mtpair{100.0}{} & \mtpair{$+$98.6}{[$+$95.6,\,$+$100.0]} & \mtpair{$+$1.4}{[0.0,\,$+$4.4]} \\
overdue tasks & 100.0 & 69 & \mtpair{98.6}{[95.6,\,100.0]} & \mtpair{100.0}{} & \mtpair{100.0}{} & \mtpair{$+$1.4}{[0.0,\,$+$4.3]} & \mtpair{0.0}{} \\
\addlinespace[2pt]
all six & 58.3 & 366 & \mtpair{50.0}{[45.2,\,54.5]} & \mtpair{91.8}{[89.8,\,93.7]} & \mtpair{97.3}{[95.4,\,98.9]} & \mtpair{$+$41.8}{[$+$38.0,\,$+$46.1]} & \mtpair{$+$5.5}{[$+$2.7,\,$+$8.4]} \\
\mtin after $\Ssrc$ &  & 165 & \mtpair{45.5}{[36.0,\,53.1]} & \mtpair{93.3}{[90.9,\,95.4]} & \mtpair{97.0}{[95.3,\,98.7]} & \mtpair{$+$47.9}{[$+$41.3,\,$+$55.6]} & \mtpair{$+$3.6}{[$+$1.1,\,$+$6.2]} \\
\mtin after $\Sdef$ &  & 201 & \mtpair{53.7}{[49.5,\,57.3]} & \mtpair{90.5}{[86.7,\,93.7]} & \mtpair{97.5}{[95.3,\,99.5]} & \mtpair{$+$36.8}{[$+$34.0,\,$+$40.0]} & \mtpair{$+$7.0}{[$+$3.1,\,$+$11.8]} \\
\mtin selection complete &  & 176 & \mtpair{95.5}{[91.5,\,98.9]} & \mtpair{94.9}{[91.7,\,98.6]} & \mtpair{95.5}{[91.7,\,98.8]} & \mtpair{$-$0.6}{[$-$3.4,\,$+$2.4]} & \mtpair{$+$0.6}{[$-$1.7,\,$+$3.1]} \\
\mtin selection incomplete &  & 190 & \mtpair{7.9}{[4.5,\,10.7]} & \mtpair{88.9}{[84.9,\,93.1]} & \mtpair{98.9}{[96.8,\,100.0]} & \mtpair{$+$81.1}{[$+$75.1,\,$+$86.9]} & \mtpair{$+$10.0}{[$+$5.0,\,$+$14.6]} \\
\bottomrule
\end{mttabular*}
\mtnotes{$r$ on the changed definition item-times after $\Ssrc$ and $\Sdef$ (10 worlds) when the writer re-derives the item from its own selection ($D{=}0$) or from every live record of the definition's sources as typed fields, and with exact recomputation ($D{=}1$). Complete: the share of item-times whose selection held every record the definition ranges over (point estimates). In (b), Claude Sonnet 5.5 writes the definition items.}
\end{table*}

%% file: generated/tab_writers.tex
\begin{table*}[t]
\caption{The account by writer and item class: repair correctness $r_D$ at $D$ (\%) and $B{\times}D$ on Update (pp).}
\label{tab:writers}
\centering\mtsetup
\setlength{\tabcolsep}{4pt}
\begin{mttabular}{@{}ll N{100}N{100}r@{\,}l @{\hspace{14pt}} N{100}N{100}r@{\,}l @{\hspace{14pt}} N{100}N{100}r@{\,}l@{}}
\toprule
& & \multicolumn{4}{c}{$\Ssrc$} & \multicolumn{4}{c}{$\Sdef$} & \multicolumn{4}{c@{}}{$\Sdel$} \\
\cmidrule(lr){3-6}\cmidrule(lr){7-10}\cmidrule(l){11-14}
Writer & Class & \multicolumn{1}{c}{$r_0$} & \multicolumn{1}{c}{$r_1$} & \multicolumn{2}{c}{$B{\times}D$} & \multicolumn{1}{c}{$r_0$} & \multicolumn{1}{c}{$r_1$} & \multicolumn{2}{c}{$B{\times}D$} & \multicolumn{1}{c}{$r_0$} & \multicolumn{1}{c}{$r_1$} & \multicolumn{2}{c@{}}{$B{\times}D$} \\
\midrule
Qwen3.5-9B & definition & 16 & 99 & $+$27.3 & \mtint{[$+$21.6,\,$+$32.6]} & 7 & 100 & $+$38.8 & \mtint{[$+$32.0,\,$+$44.8]} & 12 & 99 & $+$35.4 & \mtint{[$+$30.4,\,$+$40.0]} \\
\mtmuted{main study} & other & 89 & 91 & $-$0.6 & \mtint{[$-$3.8,\,$+$2.5]} & 73 & 73 & $-$0.6 & \mtint{[$-$4.0,\,$+$2.7]} & 68 & 68 & $-$1.4 & \mtint{[$-$5.4,\,$+$2.3]} \\
 & \mtin field & 90 & 93 & $+$1.3 & \mtint{[$-$1.8,\,$+$4.4]} & 71 & 72 & 0.0 & \mtint{[$-$3.7,\,$+$3.3]} & 68 & 67 & $-$1.5 & \mtint{[$-$5.5,\,$+$2.3]} \\
 & \mtin free text & 85 & 76 & $-$10.7 & \mtint{[$-$21.3,\,$-$1.3]} & 86 & 82 & $-$4.5 & \mtint{[$-$15.0,\,0.0]} & 86 & 82 & 0.0 & \\
\mtgap
Qwen3.5-9B & definition & 14 & 100 & $+$33.6 & \mtint{[$+$28.7,\,$+$38.5]} & 7 & 99 & $+$38.0 & \mtint{[$+$31.5,\,$+$44.5]} & 11 & 100 & $+$40.2 & \mtint{[$+$35.6,\,$+$44.9]} \\
\mtmuted{replication} & other & 90 & 90 & $-$1.9 & \mtint{[$-$3.9,\,$+$0.2]} & 84 & 86 & $+$1.6 & \mtint{[$-$1.2,\,$+$4.8]} & 66 & 67 & $+$0.8 & \mtint{[$-$3.6,\,$+$5.2]} \\
 & \mtin field & 93 & 94 & $-$1.6 & \mtint{[$-$3.5,\,$+$0.3]} & 84 & 87 & $+$3.2 & \mtint{[$+$0.6,\,$+$6.6]} & 66 & 66 & $+$0.9 & \mtint{[$-$3.6,\,$+$5.4]} \\
 & \mtin free text & 74 & 66 & $-$3.8 & \mtint{[$-$12.1,\,$+$4.2]} & 82 & 82 & $-$7.1 & \mtint{[$-$17.2,\,0.0]} & 82 & 82 & 0.0 & \\
\mtgap
gemma-4-12B & definition & 19 & 97 & $+$7.3 & \mtint{[$-$2.2,\,$+$16.1]} & 16 & 99 & $+$17.9 & \mtint{[$+$9.0,\,$+$26.3]} & 19 & 99 & $+$16.2 & \mtint{[$+$7.9,\,$+$25.5]} \\
 & other & 95 & 96 & $-$1.6 & \mtint{[$-$4.4,\,$+$0.7]} & 89 & 95 & $+$6.0 & \mtint{[$-$1.7,\,$+$13.9]} & 79 & 85 & $+$6.0 & \mtint{[$+$2.8,\,$+$9.2]} \\
 & \mtin field & 97 & 98 & $-$0.8 & \mtint{[$-$4.2,\,$+$2.4]} & 89 & 96 & $+$6.8 & \mtint{[$-$2.1,\,$+$15.2]} & 79 & 85 & $+$6.1 & \mtint{[$+$2.8,\,$+$9.3]} \\
 & \mtin free text & 84 & 84 & $-$6.9 & \mtint{[$-$12.5,\,0.0]} & 90 & 90 & 0.0 & & 90 & 90 & 0.0 & \\
\mtgap
Llama-3.1-8B & definition & 8 & 98 & $+$12.7 & \mtint{[$+$5.9,\,$+$18.8]} & 1 & 92 & $+$21.4 & \mtint{[$+$15.3,\,$+$28.5]} & 5 & 94 & $+$20.8 & \mtint{[$+$11.4,\,$+$28.7]} \\
 & other & 76 & 86 & $+$0.9 & \mtint{[$-$9.3,\,$+$11.6]} & 71 & 93 & $+$19.3 & \mtint{[$-$1.4,\,$+$41.0]} & 40 & 64 & $+$27.6 & \mtint{[$+$21.2,\,$+$34.2]} \\
 & \mtin field & 76 & 92 & $+$6.8 & \mtint{[$-$3.9,\,$+$18.6]} & 68 & 95 & $+$20.5 & \mtint{[$-$3.0,\,$+$43.8]} & 39 & 63 & $+$27.8 & \mtint{[$+$21.3,\,$+$34.7]} \\
 & \mtin free text & 74 & 47 & $-$36.2 & \mtint{[$-$55.4,\,$-$12.5]} & 90 & 80 & $+$10.0 & \mtint{[0.0,\,$+$50.0]} & 90 & 80 & $+$10.0 & \mtint{[$-$18.8,\,$+$62.5]} \\
\mtgap
gpt-oss-20b & definition & 19 & 97 & $+$44.8 & \mtint{[$+$34.7,\,$+$55.2]} & 25 & 96 & $+$39.3 & \mtint{[$+$34.3,\,$+$44.0]} & 19 & 99 & $+$52.6 & \mtint{[$+$44.6,\,$+$61.1]} \\
 & other & 96 & 94 & $+$0.2 & \mtint{[$-$3.3,\,$+$4.2]} & 90 & 90 & $+$2.4 & \mtint{[$-$9.7,\,$+$12.1]} & 79 & 83 & $+$4.5 & \mtint{[$+$0.7,\,$+$7.8]} \\
 & \mtin field & 98 & 99 & $+$1.4 & \mtint{[$-$0.5,\,$+$3.2]} & 90 & 97 & $+$6.8 & \mtint{[$-$1.7,\,$+$15.4]} & 78 & 84 & $+$4.9 & \mtint{[$+$0.8,\,$+$8.5]} \\
 & \mtin free text & 81 & 62 & $-$6.9 & \mtint{[$-$23.6,\,$+$18.3]} & 90 & 40 & $-$30.0 & \mtint{[$-$66.7,\,0.0]} & 90 & 40 & $-$30.0 & \mtint{[$-$66.7,\,0.0]} \\
\bottomrule
\end{mttabular}
\mtnotes{Field: single-field items. Free text: free-text items. Other: both. $r_0$, $r_1$: point estimates. Qwen3.5-9B ran on 20 worlds in the main study and in the replication, the other writers on ten.}
\end{table*}

%% file: generated/tab_replication.tex
\begin{table*}[t]
\caption{Every contrast of the main experiment in the main study and in the replication (20 test worlds each).}
\label{tab:replication}
\centering\mtsetup
\setlength{\tabcolsep}{4pt}
\begin{mttabular}{@{}ll r@{\,}l @{\hspace{14pt}} r@{\,}l @{\hspace{14pt}} r@{\,}l@{}}
\toprule
Contrast (pp) & Study & \multicolumn{2}{c}{$\Ssrc$} & \multicolumn{2}{c}{$\Sdef$} & \multicolumn{2}{c@{}}{$\Sdel$} \\
\midrule
\mtgroup{8}{Source key}
\mtin $B$ effect, detection (either $D$) & main study & $+$100.0 &  & $+$100.0 &  & $+$100.0 &  \\
 & replication & $+$99.9 & \mtint{[$+$99.7,\,$+$100.0]} & $+$99.8 & \mtint{[$+$99.5,\,$+$100.0]} & $+$100.0 &  \\
\addlinespace[2pt]
\mtin $B$ effect, Update (averaged over $D$) & main study & $+$9.2 & \mtint{[$+$6.7,\,$+$11.9]} & $+$13.5 & \mtint{[$+$10.6,\,$+$16.5]} & $+$51.7 & \mtint{[$+$47.3,\,$+$56.0]} \\
 & replication & $+$8.2 & \mtint{[$+$5.3,\,$+$11.2]} & $+$12.8 & \mtint{[$+$9.1,\,$+$16.5]} & $+$51.0 & \mtint{[$+$45.8,\,$+$56.2]} \\
\addlinespace[2pt]
\mtin $D$ effect, accuracy on re-derived items & main study & $+$13.6 & \mtint{[$+$12.4,\,$+$14.9]} & $+$33.1 & \mtint{[$+$30.5,\,$+$35.6]} & $+$9.4 & \mtint{[$+$7.6,\,$+$11.3]} \\
 & replication & $+$13.1 & \mtint{[$+$11.9,\,$+$14.6]} & $+$31.6 & \mtint{[$+$29.3,\,$+$34.2]} & $+$9.7 & \mtint{[$+$7.6,\,$+$11.8]} \\
\addlinespace[2pt]
\mtin $D$ effect, re-derived value equals gold & main study & $+$12.2 & \mtint{[$+$11.6,\,$+$12.8]} & $+$14.0 & \mtint{[$+$13.3,\,$+$14.7]} & $+$14.2 & \mtint{[$+$13.2,\,$+$15.2]} \\
 & replication & $+$11.4 & \mtint{[$+$11.0,\,$+$11.9]} & $+$13.0 & \mtint{[$+$12.4,\,$+$13.7]} & $+$13.4 & \mtint{[$+$12.6,\,$+$14.2]} \\
\midrule
\mtgroup{8}{Citation key}
\mtin $B$ effect, detection (either $D$) & main study & $+$85.1 & \mtint{[$+$82.1,\,$+$88.3]} & $+$79.2 & \mtint{[$+$75.1,\,$+$83.7]} & $+$92.3 & \mtint{[$+$90.8,\,$+$93.8]} \\
 & replication & $+$86.4 & \mtint{[$+$84.2,\,$+$88.4]} & $+$84.9 & \mtint{[$+$80.9,\,$+$88.7]} & $+$93.2 & \mtint{[$+$92.4,\,$+$93.9]} \\
\addlinespace[2pt]
\mtin $B$ effect, Update ($D{=}1$) & main study & $-$0.5 & \mtint{[$-$3.9,\,$+$2.9]} & $+$2.5 & \mtint{[$-$2.7,\,$+$7.4]} & $+$39.3 & \mtint{[$+$33.8,\,$+$44.3]} \\
 & replication & $-$0.7 & \mtint{[$-$3.7,\,$+$2.5]} & $+$1.7 & \mtint{[$-$4.3,\,$+$8.0]} & $+$40.2 & \mtint{[$+$34.5,\,$+$46.3]} \\
\addlinespace[2pt]
\mtin $B$ effect, Update (averaged over $D$) & main study & 0.0 & \mtint{[$-$3.0,\,$+$3.1]} & $-$1.3 & \mtint{[$-$4.8,\,$+$2.0]} & $+$42.5 & \mtint{[$+$37.8,\,$+$47.0]} \\
 & replication & $-$0.9 & \mtint{[$-$3.9,\,$+$2.0]} & $-$1.7 & \mtint{[$-$5.9,\,$+$2.7]} & $+$42.3 & \mtint{[$+$37.4,\,$+$47.4]} \\
\addlinespace[2pt]
\mtin $D$ effect, accuracy on re-derived items & main study & $+$21.8 & \mtint{[$+$18.7,\,$+$25.0]} & $+$54.6 & \mtint{[$+$49.3,\,$+$60.1]} & $+$10.7 & \mtint{[$+$7.9,\,$+$13.6]} \\
 & replication & $+$21.2 & \mtint{[$+$18.7,\,$+$23.9]} & $+$47.5 & \mtint{[$+$42.7,\,$+$53.2]} & $+$9.3 & \mtint{[$+$7.3,\,$+$11.5]} \\
\addlinespace[2pt]
\mtin $D$ effect, re-derived value equals gold & main study & $+$18.1 & \mtint{[$+$16.5,\,$+$20.0]} & $+$25.6 & \mtint{[$+$22.6,\,$+$28.9]} & $+$16.3 & \mtint{[$+$14.8,\,$+$18.0]} \\
 & replication & $+$17.4 & \mtint{[$+$15.8,\,$+$19.1]} & $+$24.1 & \mtint{[$+$21.6,\,$+$26.9]} & $+$16.1 & \mtint{[$+$14.6,\,$+$17.7]} \\
\addlinespace[2pt]
\mtin $B{\times}D$ interaction, Update & main study & $-$0.9 & \mtint{[$-$3.6,\,$+$1.6]} & $+$7.5 & \mtint{[$+$1.1,\,$+$14.0]} & $-$6.4 & \mtint{[$-$10.1,\,$-$3.0]} \\
 & replication & $+$0.5 & \mtint{[$-$1.4,\,$+$2.7]} & $+$6.8 & \mtint{[$+$2.4,\,$+$11.6]} & $-$4.1 & \mtint{[$-$7.5,\,$-$0.5]} \\
\bottomrule
\end{mttabular}
\mtnotes{Detection: share of the changed item-times the cell invalidated since the fork. Value equals gold: share of the re-derived item-times whose value equals gold. $D$'s effect on missed invalidation was 0.0 with both keys after every shift in both studies.}
\end{table*}

%% file: generated/tab_mem0x.tex
\begin{table*}[t]
\caption{Two interventions on Mem0 on 10 of its three-month worlds: the terms of $\mathrm{Update}=(1-m)\,r+m\,a$ on changed item-times (\%) and each intervention's effect on Update.}
\label{tab:mem0x}
\centering\mtsetup
\setlength{\tabcolsep}{4pt}
\begin{mttabular}{@{}ll N{000} @{\hspace{12pt}} N{00.0} @{\hspace{12pt}} N{00.0}N{00.0} r@{\,}l @{\hspace{12pt}} r@{\,}l@{}}
\toprule
& & & \multicolumn{1}{c}{Mem0} & \multicolumn{4}{c}{Invalidation} & \multicolumn{2}{c@{}}{Query-time view} \\
\cmidrule(lr){4-4}\cmidrule(lr){5-8}\cmidrule(l){9-10}
Shift & Class & \multicolumn{1}{c}{$n$} & \multicolumn{1}{c}{$a$} & \multicolumn{1}{c}{$m$} & \multicolumn{1}{c}{$r$} & \multicolumn{2}{c}{Effect (pp)} & \multicolumn{2}{c@{}}{Effect (pp)} \\
\midrule
$\Ssrc$ & definition & 164 & 12.2 & 54.3 & 28.0 & $+$0.6 & \mtint{[$-$3.5,\,$+$4.2]} & $+$72.6 & \mtint{[$+$63.1,\,$+$81.2]} \\
 & other & 507 & 88.0 & 13.0 & 87.3 & $-$3.4 & \mtint{[$-$7.9,\,$+$1.4]} & 0.0 &  \\
\mtgap
$\Sdef$ & definition & 191 & 8.9 & 66.5 & 25.0 & $+$1.6 & \mtint{[$-$0.6,\,$+$4.1]} & $+$64.4 & \mtint{[$+$57.4,\,$+$71.0]} \\
 & other & 188 & 90.4 & 13.3 & 83.4 & $-$8.5 & \mtint{[$-$15.1,\,$-$1.1]} & 0.0 &  \\
\mtgap
$\Sdel$ & definition & 28 & 10.7 & 53.6 & 30.8 & $+$3.6 & \mtint{[0.0,\,$+$10.7]} & $+$71.4 & \mtint{[$+$52.0,\,$+$88.6]} \\
 & other & 184 & 47.3 & 2.7 & 83.2 & $+$35.9 & \mtint{[$+$26.1,\,$+$44.8]} & 0.0 &  \\
\bottomrule
\end{mttabular}
\mtnotes{Mem0 invalidates nothing ($m{=}1$ in every row), so its $r$ is undefined and its Update is $a$. $a$, $m$, $r$: point estimates. Effect: Update minus Mem0's, paired per item-time.}
\end{table*}

%% file: generated/tab_classsplit.tex
\begin{table*}[t]
\caption{The contrasts on Update (pp) by shift and item class, with exact recomputation and with the writer's programs as the $D{=}1$ arm.}
\label{tab:classsplit}
\centering\mtsetup
\setlength{\tabcolsep}{3.2pt}
\begin{mttabular}{@{}lll r@{\,}l r@{\,}l @{\hspace{14pt}} r@{\,}l r@{\,}l @{\hspace{14pt}} r@{\,}l@{}}
\toprule
& & & \multicolumn{4}{c}{$B$ effect} & \multicolumn{4}{c}{$D$ effect} & & \\
\cmidrule(lr){4-7}\cmidrule(lr){8-11}
Shift & Class & Arm & \multicolumn{2}{c}{$D{=}0$} & \multicolumn{2}{c}{$D{=}1$} & \multicolumn{2}{c}{$B{=}0$} & \multicolumn{2}{c}{$B{=}1$} & \multicolumn{2}{c@{}}{$B{\times}D$} \\
\midrule
\mtgroup{13}{Main study: 20 test worlds}
$\Ssrc$ & definition & exact & $+$6.0 & \mtint{[$+$2.5,\,$+$9.6]} & $+$33.3 & \mtint{[$+$28.2,\,$+$38.0]} & $+$55.6 & \mtint{[$+$50.1,\,$+$61.1]} & $+$82.9 & \mtint{[$+$77.0,\,$+$89.1]} & $+$27.3 & \mtint{[$+$21.6,\,$+$32.6]} \\
 &  & programs & $+$6.0 & \mtint{[$+$2.5,\,$+$9.6]} & $+$33.3 & \mtint{[$+$28.4,\,$+$38.0]} & $+$30.0 & \mtint{[$+$21.6,\,$+$38.2]} & $+$57.4 & \mtint{[$+$49.0,\,$+$65.7]} & $+$27.3 & \mtint{[$+$21.8,\,$+$32.7]} \\
\mtgap
 & other & exact & $+$5.4 & \mtint{[$+$2.3,\,$+$8.7]} & $+$4.8 & \mtint{[$+$1.6,\,$+$8.0]} & $+$2.2 & \mtint{[$+$0.6,\,$+$3.8]} & $+$1.6 & \mtint{[$-$1.0,\,$+$4.2]} & $-$0.6 & \mtint{[$-$3.8,\,$+$2.5]} \\
 &  & programs & $+$5.4 & \mtint{[$+$2.3,\,$+$8.7]} & $+$9.8 & \mtint{[$+$6.0,\,$+$13.8]} & $-$4.0 & \mtint{[$-$6.8,\,$-$1.2]} & $+$0.5 & \mtint{[$-$1.9,\,$+$3.0]} & $+$4.4 & \mtint{[$+$0.3,\,$+$8.6]} \\
\mtgap
 & all & exact & $+$5.6 & \mtint{[$+$2.8,\,$+$8.3]} & $+$12.9 & \mtint{[$+$10.1,\,$+$15.9]} & $+$17.4 & \mtint{[$+$15.4,\,$+$19.7]} & $+$24.7 & \mtint{[$+$22.7,\,$+$26.7]} & $+$7.4 & \mtint{[$+$5.0,\,$+$9.8]} \\
 &  & programs & $+$5.6 & \mtint{[$+$2.8,\,$+$8.3]} & $+$16.5 & \mtint{[$+$13.1,\,$+$20.0]} & $+$5.7 & \mtint{[$+$3.0,\,$+$8.3]} & $+$16.7 & \mtint{[$+$14.3,\,$+$19.3]} & $+$11.0 & \mtint{[$+$8.0,\,$+$14.0]} \\
\midrule
$\Sdef$ & definition & exact & $-$0.8 & \mtint{[$-$5.7,\,$+$4.2]} & $+$38.0 & \mtint{[$+$32.6,\,$+$43.4]} & $+$54.3 & \mtint{[$+$48.9,\,$+$59.7]} & $+$93.0 & \mtint{[$+$89.0,\,$+$96.7]} & $+$38.8 & \mtint{[$+$32.0,\,$+$44.8]} \\
 &  & programs & $-$0.8 & \mtint{[$-$5.7,\,$+$4.2]} & $+$34.1 & \mtint{[$+$27.5,\,$+$39.8]} & $+$27.6 & \mtint{[$+$21.7,\,$+$34.1]} & $+$62.5 & \mtint{[$+$57.5,\,$+$67.3]} & $+$34.9 & \mtint{[$+$27.5,\,$+$41.9]} \\
\mtgap
 & other & exact & $+$2.3 & \mtint{[$-$2.4,\,$+$7.4]} & $+$1.8 & \mtint{[$-$3.4,\,$+$7.1]} & $+$0.6 & \mtint{[$-$1.9,\,$+$3.2]} & 0.0 & \mtint{[$-$3.4,\,$+$2.3]} & $-$0.6 & \mtint{[$-$4.0,\,$+$2.7]} \\
 &  & programs & $+$2.3 & \mtint{[$-$2.4,\,$+$7.4]} & $+$5.8 & \mtint{[$-$1.9,\,$+$13.9]} & $-$3.5 & \mtint{[$-$10.6,\,$+$2.5]} & 0.0 & \mtint{[$-$1.9,\,$+$1.4]} & $+$3.5 & \mtint{[$-$3.3,\,$+$10.8]} \\
\mtgap
 & all & exact & $+$0.2 & \mtint{[$-$2.9,\,$+$3.4]} & $+$26.9 & \mtint{[$+$22.3,\,$+$31.6]} & $+$37.8 & \mtint{[$+$33.6,\,$+$42.1]} & $+$64.5 & \mtint{[$+$59.9,\,$+$69.1]} & $+$26.7 & \mtint{[$+$21.2,\,$+$31.8]} \\
 &  & programs & $+$0.2 & \mtint{[$-$2.9,\,$+$3.4]} & $+$25.4 & \mtint{[$+$19.1,\,$+$31.5]} & $+$18.1 & \mtint{[$+$13.0,\,$+$23.5]} & $+$43.4 & \mtint{[$+$39.3,\,$+$47.8]} & $+$25.3 & \mtint{[$+$18.8,\,$+$31.7]} \\
\midrule
$\Sdel$ & definition & exact & $+$7.5 & \mtint{[$+$3.4,\,$+$11.5]} & $+$42.9 & \mtint{[$+$38.0,\,$+$47.8]} & $+$52.3 & \mtint{[$+$47.0,\,$+$57.3]} & $+$87.7 & \mtint{[$+$83.8,\,$+$91.5]} & $+$35.4 & \mtint{[$+$30.4,\,$+$40.0]} \\
 &  & programs & $+$7.5 & \mtint{[$+$3.4,\,$+$11.5]} & $+$43.2 & \mtint{[$+$37.1,\,$+$49.2]} & $+$30.5 & \mtint{[$+$24.4,\,$+$36.0]} & $+$66.2 & \mtint{[$+$61.4,\,$+$71.0]} & $+$35.7 & \mtint{[$+$30.1,\,$+$40.9]} \\
\mtgap
 & other & exact & $+$58.1 & \mtint{[$+$52.2,\,$+$63.7]} & $+$56.6 & \mtint{[$+$50.6,\,$+$62.3]} & $+$0.9 & \mtint{[$-$0.5,\,$+$2.4]} & $-$0.5 & \mtint{[$-$4.5,\,$+$3.2]} & $-$1.4 & \mtint{[$-$5.4,\,$+$2.3]} \\
 &  & programs & $+$58.1 & \mtint{[$+$52.2,\,$+$63.7]} & $+$25.3 & \mtint{[$+$21.0,\,$+$29.9]} & $+$3.0 & \mtint{[$+$1.2,\,$+$4.8]} & $-$29.7 & \mtint{[$-$33.9,\,$-$25.2]} & $-$32.8 & \mtint{[$-$37.5,\,$-$27.7]} \\
\mtgap
 & all & exact & $+$49.2 & \mtint{[$+$44.6,\,$+$54.0]} & $+$54.2 & \mtint{[$+$49.4,\,$+$58.8]} & $+$9.8 & \mtint{[$+$8.1,\,$+$11.5]} & $+$14.8 & \mtint{[$+$11.3,\,$+$18.3]} & $+$5.0 & \mtint{[$+$1.7,\,$+$8.1]} \\
 &  & programs & $+$49.2 & \mtint{[$+$44.6,\,$+$54.0]} & $+$28.4 & \mtint{[$+$24.9,\,$+$32.3]} & $+$7.8 & \mtint{[$+$6.2,\,$+$9.5]} & $-$13.0 & \mtint{[$-$16.9,\,$-$8.8]} & $-$20.8 & \mtint{[$-$25.1,\,$-$16.4]} \\
\midrule
\mtgroup{13}{Replication: 20 fresh worlds}
$\Ssrc$ & definition & exact & $+$4.2 & \mtint{[$+$0.6,\,$+$7.6]} & $+$37.8 & \mtint{[$+$33.4,\,$+$42.6]} & $+$51.5 & \mtint{[$+$46.0,\,$+$56.4]} & $+$85.1 & \mtint{[$+$80.7,\,$+$89.8]} & $+$33.6 & \mtint{[$+$28.7,\,$+$38.5]} \\
 & other & exact & $+$4.0 & \mtint{[0.0,\,$+$8.1]} & $+$2.0 & \mtint{[$-$2.2,\,$+$6.6]} & $+$1.3 & \mtint{[$-$0.2,\,$+$2.8]} & $-$0.6 & \mtint{[$-$2.3,\,$+$1.1]} & $-$1.9 & \mtint{[$-$3.9,\,$+$0.2]} \\
 & all & exact & $+$4.0 & \mtint{[$+$0.8,\,$+$7.0]} & $+$12.3 & \mtint{[$+$9.4,\,$+$15.7]} & $+$15.7 & \mtint{[$+$13.5,\,$+$17.9]} & $+$24.0 & \mtint{[$+$22.0,\,$+$26.2]} & $+$8.3 & \mtint{[$+$6.3,\,$+$10.4]} \\
\midrule
$\Sdef$ & definition & exact & $-$1.5 & \mtint{[$-$3.9,\,$+$0.5]} & $+$36.5 & \mtint{[$+$29.8,\,$+$43.0]} & $+$53.7 & \mtint{[$+$47.5,\,$+$60.4]} & $+$91.8 & \mtint{[$+$88.2,\,$+$94.9]} & $+$38.0 & \mtint{[$+$31.5,\,$+$44.5]} \\
 & other & exact & $+$2.2 & \mtint{[$-$4.7,\,$+$8.9]} & $+$3.8 & \mtint{[$-$2.8,\,$+$10.8]} & $+$0.5 & \mtint{[$-$1.2,\,$+$2.3]} & $+$2.2 & \mtint{[0.0,\,$+$4.9]} & $+$1.6 & \mtint{[$-$1.2,\,$+$4.8]} \\
 & all & exact & $-$0.3 & \mtint{[$-$3.2,\,$+$2.4]} & $+$25.9 & \mtint{[$+$20.5,\,$+$31.9]} & $+$36.5 & \mtint{[$+$32.6,\,$+$41.1]} & $+$62.8 & \mtint{[$+$58.2,\,$+$67.9]} & $+$26.3 & \mtint{[$+$21.5,\,$+$31.7]} \\
\midrule
$\Sdel$ & definition & exact & $+$4.5 & \mtint{[$+$0.6,\,$+$8.7]} & $+$44.7 & \mtint{[$+$39.0,\,$+$50.4]} & $+$48.9 & \mtint{[$+$43.4,\,$+$54.2]} & $+$89.1 & \mtint{[$+$84.6,\,$+$93.4]} & $+$40.2 & \mtint{[$+$35.6,\,$+$44.9]} \\
 & other & exact & $+$56.3 & \mtint{[$+$50.2,\,$+$62.1]} & $+$57.1 & \mtint{[$+$50.1,\,$+$64.3]} & $-$0.4 & \mtint{[$-$1.2,\,$+$0.5]} & $+$0.5 & \mtint{[$-$4.2,\,$+$5.2]} & $+$0.8 & \mtint{[$-$3.6,\,$+$5.2]} \\
 & all & exact & $+$47.0 & \mtint{[$+$42.0,\,$+$51.8]} & $+$54.9 & \mtint{[$+$48.7,\,$+$61.4]} & $+$8.5 & \mtint{[$+$7.2,\,$+$9.8]} & $+$16.3 & \mtint{[$+$12.2,\,$+$20.5]} & $+$7.9 & \mtint{[$+$3.7,\,$+$12.1]} \\
\bottomrule
\end{mttabular}
\mtnotes{Exact: exact recomputation as the $D{=}1$ arm. Programs: the writer's programs ($D_{\mathrm{m}}$) as the $D{=}1$ arm, so in their rows $D{=}1$ is $D_{\mathrm{m}}$. The $D{=}0$ cells of the main study serve both of its arms.}
\end{table*}

%% file: generated/tab_reader.tex
\begin{table*}[t]
\caption{The $B{\times}D$ interaction on Update split into a store part and a reader part, with two readers of the same stores.}
\label{tab:store-reader-bxd}\label{tab:reader}
\centering\mtsetup
\setlength{\tabcolsep}{4pt}
\begin{mttabular}{@{}lll r@{\,}l r@{\,}l r@{\,}l r@{\,}l r@{\,}l@{}}
\toprule
& & & & & \multicolumn{2}{c}{Qwen3.5-9B} & \multicolumn{6}{c@{}}{Claude Sonnet 5.5} \\
\cmidrule(lr){6-7}\cmidrule(l){8-13}
Shift & Class & Arm & \multicolumn{2}{c}{Store part} & \multicolumn{2}{c}{Reader part} & \multicolumn{2}{c}{Reader part} & \multicolumn{2}{c}{$B{\times}D$} & \multicolumn{2}{c@{}}{Sonnet $-$ Qwen} \\
\midrule
$\Ssrc$ & definition & $D{=}1$ & $+$27.3 & \mtint{[$+$21.3,\,$+$32.8]} & 0.0 & \mtint{[$-$0.9,\,$+$0.9]} & $-$8.7 & \mtint{[$-$13.4,\,$-$4.7]} & $+$18.6 & \mtint{[$+$11.9,\,$+$25.1]} & $-$8.7 & \mtint{[$-$13.4,\,$-$4.6]} \\
 &  & $D_{\mathrm{m}}$ & $+$27.3 & \mtint{[$+$21.3,\,$+$32.8]} & 0.0 & \mtint{[$-$1.5,\,$+$1.4]} & $-$8.7 & \mtint{[$-$14.2,\,$-$4.1]} & $+$18.6 & \mtint{[$+$11.1,\,$+$25.6]} & $-$8.7 & \mtint{[$-$14.1,\,$-$3.8]} \\
\mtgap
 & other & $D{=}1$ & $+$2.8 & \mtint{[$+$0.6,\,$+$4.7]} & $-$3.4 & \mtint{[$-$6.6,\,0.0]} & $-$1.6 & \mtint{[$-$3.4,\,$+$0.5]} & $+$1.2 & \mtint{[$-$1.1,\,$+$3.4]} & $+$1.8 & \mtint{[$-$1.9,\,$+$6.1]} \\
 &  & $D_{\mathrm{m}}$ & $+$7.1 & \mtint{[$+$4.5,\,$+$9.8]} & $-$2.6 & \mtint{[$-$5.5,\,$+$0.3]} & $-$2.0 & \mtint{[$-$3.7,\,$-$0.4]} & $+$5.0 & \mtint{[$+$2.0,\,$+$8.2]} & $+$0.6 & \mtint{[$-$2.5,\,$+$4.2]} \\
\mtgap
 & all & $D{=}1$ & $+$9.8 & \mtint{[$+$8.0,\,$+$11.2]} & $-$2.4 & \mtint{[$-$4.7,\,0.0]} & $-$3.6 & \mtint{[$-$5.4,\,$-$1.9]} & $+$6.2 & \mtint{[$+$3.8,\,$+$8.3]} & $-$1.2 & \mtint{[$-$3.9,\,$+$1.8]} \\
 &  & $D_{\mathrm{m}}$ & $+$12.8 & \mtint{[$+$10.7,\,$+$15.0]} & $-$1.9 & \mtint{[$-$3.9,\,0.0]} & $-$3.9 & \mtint{[$-$6.0,\,$-$2.0]} & $+$8.9 & \mtint{[$+$6.2,\,$+$11.5]} & $-$2.1 & \mtint{[$-$4.5,\,$+$0.9]} \\
\midrule
$\Sdef$ & definition & $D{=}1$ & $+$38.8 & \mtint{[$+$32.0,\,$+$44.8]} & 0.0 & & $-$15.2 & \mtint{[$-$19.1,\,$-$11.0]} & $+$23.5 & \mtint{[$+$18.8,\,$+$28.2]} & $-$15.2 & \mtint{[$-$19.1,\,$-$11.0]} \\
 &  & $D_{\mathrm{m}}$ & $+$34.9 & \mtint{[$+$28.9,\,$+$41.0]} & 0.0 & \mtint{[$-$3.6,\,$+$3.2]} & $-$15.0 & \mtint{[$-$18.6,\,$-$10.9]} & $+$19.9 & \mtint{[$+$14.9,\,$+$24.7]} & $-$15.0 & \mtint{[$-$20.1,\,$-$9.5]} \\
\mtgap
 & other & $D{=}1$ & $+$20.5 & \mtint{[$+$11.7,\,$+$29.1]} & $-$21.1 & \mtint{[$-$31.7,\,$-$10.8]} & $-$10.5 & \mtint{[$-$16.6,\,$-$4.6]} & $+$9.9 & \mtint{[$+$3.4,\,$+$17.3]} & $+$10.5 & \mtint{[$+$3.1,\,$+$18.9]} \\
 &  & $D_{\mathrm{m}}$ & $+$21.1 & \mtint{[$+$13.8,\,$+$29.1]} & $-$17.5 & \mtint{[$-$28.2,\,$-$8.1]} & $-$12.3 & \mtint{[$-$18.6,\,$-$6.2]} & $+$8.8 & \mtint{[$-$0.6,\,$+$18.8]} & $+$5.3 & \mtint{[$-$4.1,\,$+$15.8]} \\
\mtgap
 & all & $D{=}1$ & $+$33.2 & \mtint{[$+$28.1,\,$+$37.9]} & $-$6.5 & \mtint{[$-$9.8,\,$-$3.3]} & $-$13.8 & \mtint{[$-$17.1,\,$-$10.5]} & $+$19.4 & \mtint{[$+$15.7,\,$+$23.1]} & $-$7.3 & \mtint{[$-$11.4,\,$-$3.2]} \\
 &  & $D_{\mathrm{m}}$ & $+$30.6 & \mtint{[$+$26.0,\,$+$35.2]} & $-$5.4 & \mtint{[$-$10.4,\,$-$1.2]} & $-$14.2 & \mtint{[$-$17.7,\,$-$10.6]} & $+$16.5 & \mtint{[$+$11.7,\,$+$21.4]} & $-$8.8 & \mtint{[$-$14.2,\,$-$2.6]} \\
\midrule
$\Sdel$ & definition & $D{=}1$ & $+$35.7 & \mtint{[$+$30.7,\,$+$40.4]} & $-$0.3 & \mtint{[$-$1.0,\,0.0]} & $-$10.7 & \mtint{[$-$15.8,\,$-$5.9]} & $+$25.0 & \mtint{[$+$19.3,\,$+$30.4]} & $-$10.4 & \mtint{[$-$15.5,\,$-$5.5]} \\
 &  & $D_{\mathrm{m}}$ & $+$35.7 & \mtint{[$+$30.7,\,$+$40.4]} & 0.0 & \mtint{[$-$2.4,\,$+$2.3]} & $-$11.4 & \mtint{[$-$16.3,\,$-$6.6]} & $+$24.4 & \mtint{[$+$18.6,\,$+$29.7]} & $-$11.4 & \mtint{[$-$17.0,\,$-$6.0]} \\
\mtgap
 & other & $D{=}1$ & $+$7.2 & \mtint{[$+$5.0,\,$+$9.5]} & $-$8.6 & \mtint{[$-$11.2,\,$-$6.1]} & $-$1.9 & \mtint{[$-$3.2,\,$-$0.6]} & $+$5.3 & \mtint{[$+$2.3,\,$+$8.4]} & $+$6.7 & \mtint{[$+$3.7,\,$+$9.6]} \\
 &  & $D_{\mathrm{m}}$ & $-$20.4 & \mtint{[$-$24.0,\,$-$16.7]} & $-$12.4 & \mtint{[$-$16.2,\,$-$7.9]} & $-$2.3 & \mtint{[$-$3.7,\,$-$0.8]} & $-$22.6 & \mtint{[$-$26.4,\,$-$18.7]} & $+$10.1 & \mtint{[$+$5.2,\,$+$14.3]} \\
\mtgap
 & all & $D{=}1$ & $+$12.2 & \mtint{[$+$10.0,\,$+$14.7]} & $-$7.2 & \mtint{[$-$9.3,\,$-$5.1]} & $-$3.5 & \mtint{[$-$4.5,\,$-$2.5]} & $+$8.7 & \mtint{[$+$6.2,\,$+$11.6]} & $+$3.7 & \mtint{[$+$1.4,\,$+$6.1]} \\
 &  & $D_{\mathrm{m}}$ & $-$10.6 & \mtint{[$-$14.3,\,$-$6.7]} & $-$10.2 & \mtint{[$-$13.3,\,$-$6.6]} & $-$3.8 & \mtint{[$-$5.0,\,$-$2.7]} & $-$14.4 & \mtint{[$-$18.3,\,$-$10.3]} & $+$6.4 & \mtint{[$+$2.5,\,$+$9.6]} \\
\bottomrule
\end{mttabular}
\mtnotes{Store part: the same contrast of the shares of item-times whose stored value is right ($r_{\mathrm{store}}$ and $a_{\mathrm{store}}$ of \cref{tab:store-reader}), the same for both readers of the same stores. Reader part: the rest. $B{\times}D$ with Qwen3.5-9B is the store part plus its reader part (\cref{tab:classsplit}). $D_{\mathrm{m}}$: the writer's programs as the $D{=}1$ arm. Sonnet $-$ Qwen: $B{\times}D$ with Claude Sonnet 5.5 minus that with Qwen3.5-9B, paired per item-time.}
\end{table*}

%% file: generated/tab_dm_deletion.tex
\begin{table*}[t]
\caption{Why the writer's programs ($D_{\mathrm{m}}$) repair worse than $D{=}0$ after a deletion: the 1,365 changed item-times whose record was deleted (gold \texttt{abstain}) by what the cell stored and how the reader answered, and what $D_{\mathrm{m}}$ loses against $D{=}0$ there.}
\label{tab:dm-deletion}
\centering\mtsetup
\setlength{\tabcolsep}{4pt}
\begin{mttabular}{@{}l N{00.0}N{00.0}N{00.0} @{\hspace{14pt}} r@{\,}l r@{\,}l@{}}
\toprule
& \multicolumn{3}{c}{Item-times (\%)} & \multicolumn{4}{c@{}}{$D_{\mathrm{m}}$ against $D{=}0$} \\
\cmidrule(lr){2-4}\cmidrule(l){5-8}
Class & \multicolumn{1}{c}{$D{=}0$} & \multicolumn{1}{c}{$D{=}1$} & \multicolumn{1}{c}{$D_{\mathrm{m}}$} & \multicolumn{2}{c}{Loss (pp)} & \multicolumn{2}{c@{}}{Share of the loss (\%)} \\
\midrule
correct & 66.7 & 66.2 & 34.9 & & & & \\
value stored & 27.3 & 19.0 & 48.4 & 24.7 & \mtint{[20.5,\,28.7]} & 68.9 & \mtint{[61.1,\,76.6]} \\
\mtin the program's value & & & & \multicolumn{2}{c}{---} & 67.7 & \mtint{[60.0,\,75.3]} \\
reader & 5.4 & 14.1 & 15.9 & 10.9 & \mtint{[7.8,\,14.1]} & 30.5 & \mtint{[22.9,\,38.2]} \\
failed & 0.0 & 0.0 & 0.0 & 0.0 &  & 0.0 &  \\
not re-derived & 0.6 & 0.8 & 0.8 & 0.2 & \mtint{[0.0,\,0.4]} & 0.6 & \mtint{[0.0,\,1.3]} \\
\mtgap
lost &  &  &  & 35.8 & \mtint{[31.2,\,39.9]} & 100.0 &  \\
gained &  &  &  & 4.0 & \mtint{[2.6,\,5.5]} & \multicolumn{2}{c@{}}{---} \\
\bottomrule
\end{mttabular}
\mtnotes{Value stored: a value stored where gold is no value. Reader: the stored state is right but the reader's answer is wrong. Failed: the replacement failed, so nothing is served. Not re-derived: not re-derived at the checkpoint. Item-times: point estimates. Loss: the item-times $D{=}0$ answers right and $D_{\mathrm{m}}$ wrong, by the class $D_{\mathrm{m}}$ falls in. ---: not defined.}
\end{table*}

%% file: generated/tab_json_arm.tex
\begin{table*}[t]
\caption{The same $D{=}1$ store served as JSON records instead of statements ($s_k$), in the main study and the replication.}
\label{tab:json-arm}
\centering\mtsetup
\setlength{\tabcolsep}{3.5pt}
\begin{mttabular}{@{}ll N{100}N{100} r@{\,}l r@{\,}l @{\hspace{14pt}} N{100}N{100} r@{\,}l r@{\,}l@{}}
\toprule
& & \multicolumn{6}{c}{Main study} & \multicolumn{6}{c@{}}{Replication} \\
\cmidrule(lr){3-8}\cmidrule(l){9-14}
& & \multicolumn{2}{c}{Update (\%)} & \multicolumn{4}{c}{JSON $-$ $s_k$ (pp)} & \multicolumn{2}{c}{Update (\%)} & \multicolumn{4}{c@{}}{JSON $-$ $s_k$ (pp)} \\
\cmidrule(lr){3-4}\cmidrule(lr){5-8}\cmidrule(lr){9-10}\cmidrule(l){11-14}
Shift & Class & \multicolumn{1}{c}{$s_k$} & \multicolumn{1}{c}{JSON} & \multicolumn{2}{c}{Update} & \multicolumn{2}{c}{Stability} & \multicolumn{1}{c}{$s_k$} & \multicolumn{1}{c}{JSON} & \multicolumn{2}{c}{Update} & \multicolumn{2}{c@{}}{Stability} \\
\midrule
$\Ssrc$ & definition & 99 & 70 & $-$29.4 & \mtint{[$-$33.6,\,$-$25.2]} & $-$38.7 & \mtint{[$-$50.0,\,$-$27.6]} & 99 & 67 & $-$32.1 & \mtint{[$-$35.5,\,$-$28.7]} & $-$36.4 & \mtint{[$-$49.4,\,$-$22.2]} \\
 & other & 91 & 89 & $-$1.3 & \mtint{[$-$2.8,\,$-$0.1]} & $-$0.6 & \mtint{[$-$1.5,\,$+$0.1]} & 90 & 90 & $+$0.1 & \mtint{[$-$1.3,\,$+$1.6]} & $-$0.4 & \mtint{[$-$1.2,\,$+$0.4]} \\
 & all & 93 & 84 & $-$9.3 & \mtint{[$-$10.9,\,$-$7.7]} & $-$2.7 & \mtint{[$-$3.9,\,$-$1.6]} & 92 & 83 & $-$9.1 & \mtint{[$-$10.6,\,$-$7.7]} & $-$2.4 & \mtint{[$-$4.0,\,$-$1.0]} \\
\mtgap
$\Sdef$ & definition & 100 & 54 & $-$45.5 & \mtint{[$-$48.9,\,$-$42.2]} & $-$46.2 & \mtint{[$-$80.0,\,$-$20.0]} & 99 & 56 & $-$43.4 & \mtint{[$-$48.1,\,$-$38.8]} & $-$30.8 & \mtint{[$-$54.5,\,$-$8.3]} \\
 & other & 73 & 72 & $-$1.2 & \mtint{[$-$3.9,\,0.0]} & 0.0 & \mtint{[$-$0.8,\,$+$0.7]} & 86 & 83 & $-$2.7 & \mtint{[$-$5.8,\,0.0]} & $-$0.2 & \mtint{[$-$0.8,\,$+$0.4]} \\
 & all & 92 & 60 & $-$31.9 & \mtint{[$-$35.4,\,$-$28.6]} & $-$1.1 & \mtint{[$-$2.3,\,0.0]} & 95 & 65 & $-$30.3 & \mtint{[$-$33.5,\,$-$27.1]} & $-$0.9 & \mtint{[$-$1.9,\,0.0]} \\
\mtgap
$\Sdel$ & definition & 99 & 67 & $-$32.8 & \mtint{[$-$36.3,\,$-$29.0]} & $-$33.3 & \mtint{[$-$39.2,\,$-$27.2]} & 100 & 64 & $-$36.0 & \mtint{[$-$39.9,\,$-$32.0]} & $-$30.2 & \mtint{[$-$37.9,\,$-$23.1]} \\
 & other & 68 & 16 & $-$51.4 & \mtint{[$-$56.8,\,$-$45.1]} & $-$0.3 & \mtint{[$-$1.0,\,$+$0.2]} & 67 & 16 & $-$50.4 & \mtint{[$-$55.5,\,$-$45.1]} & $-$0.3 & \mtint{[$-$0.7,\,0.0]} \\
 & all & 73 & 25 & $-$48.2 & \mtint{[$-$52.7,\,$-$43.0]} & $-$2.9 & \mtint{[$-$3.8,\,$-$2.1]} & 73 & 25 & $-$47.8 & \mtint{[$-$51.9,\,$-$43.5]} & $-$2.6 & \mtint{[$-$3.5,\,$-$1.8]} \\
\bottomrule
\end{mttabular}
\mtnotes{Update on the changed item-times in percent, and the paired differences JSON $-$ $s_k$ in Update and in Stability on the unchanged item-times. Update levels: point estimates.}
\end{table*}

%% file: generated/tab_programs.tex
\mtpairwidths{$+$00.0}{[$+$00.0,\,$+$00.0]}
\begin{mttabular}{@{}ll N{0,000} ccc@{}}
\toprule
(a) & & \multicolumn{1}{c}{$n$} & Without feedback & With feedback & \multicolumn{1}{c@{}}{Difference} \\
\midrule
$B{\times}D$ on Update (pp), definition items & $\Ssrc$ & 333 & \mtpair{$+$27.3}{[$+$21.8,\,$+$32.7]} & \mtpair{$+$28.2}{[$+$22.6,\,$+$33.9]} & \mtpair{$+$0.9}{[$-$0.3,\,$+$2.5]} \\
 & $\Sdef$ & 387 & \mtpair{$+$34.9}{[$+$27.5,\,$+$41.9]} & \mtpair{$+$34.9}{[$+$27.9,\,$+$41.5]} & \mtpair{0.0}{[$-$2.3,\,$+$2.0]} \\
 & $\Sdel$ & 308 & \mtpair{$+$35.7}{[$+$30.1,\,$+$40.9]} & \mtpair{$+$35.7}{[$+$29.8,\,$+$41.3]} & \mtpair{0.0}{[$-$1.6,\,$+$1.6]} \\
\addlinespace[2pt]
$r$ (\%), definition items & $\Ssrc$ & 333 & \mtpair{73.6}{[69.3,\,77.9]} & \mtpair{74.5}{[70.3,\,78.8]} & \mtpair{$+$0.9}{[$-$0.3,\,$+$2.5]} \\
 & $\Sdef$ & 387 & \mtpair{69.3}{[66.6,\,72.0]} & \mtpair{69.0}{[66.2,\,71.9]} & \mtpair{$-$0.3}{[$-$1.7,\,$+$1.0]} \\
\addlinespace[2pt]
valid runs (\%), definition items &  & 2,085 & \mtpair{85.8}{[84.8,\,86.7]} & \mtpair{90.4}{[89.5,\,91.2]} &  \\
\midrule
(b) & & \multicolumn{1}{c}{$n$} & Typed fields & Extracted fields & \multicolumn{1}{c@{}}{Difference} \\
\midrule
$B{\times}D$ on Update (pp), definition items & $\Ssrc$ & 165 & \mtpair{$+$24.2}{[$+$16.4,\,$+$32.5]} & \mtpair{$+$18.2}{[$+$7.7,\,$+$29.7]} & \mtpair{$-$6.1}{[$-$9.8,\,$-$1.4]} \\
 & $\Sdef$ & 201 & \mtpair{$+$35.8}{[$+$28.0,\,$+$42.4]} & \mtpair{$+$25.4}{[$+$18.6,\,$+$32.4]} & \mtpair{$-$10.4}{[$-$16.7,\,$-$2.9]} \\
\addlinespace[2pt]
$r$ (\%), definition items & $\Ssrc$ & 165 & \mtpair{77.0}{[71.6,\,82.9]} & \mtpair{55.8}{[50.6,\,60.9]} & \mtpair{$-$21.2}{[$-$26.4,\,$-$16.5]} \\
 & $\Sdef$ & 201 & \mtpair{69.2}{[64.3,\,73.7]} & \mtpair{52.7}{[47.3,\,58.3]} & \mtpair{$-$16.4}{[$-$21.8,\,$-$10.6]} \\
\bottomrule
\end{mttabular}
\mtnotes{(a) $B{\times}D$ and $r$ ($n$: changed item-times), and the program runs after the fork that end with a valid typed value ($n$: runs), without and with feedback. (b) $B{\times}D$ and $r$ with programs on typed and on extracted fields. Values differed between the two inputs on 50 of 198 and 30 of 206 definition item-times after $\Ssrc$ and $\Sdef$ at $B{=}0$, and on 58 of 198 and 53 of 206 at $B{=}1$. Of these, 100.0\% had read a record with an extraction error.}

%% file: generated/tab_userwords.tex
\begin{mttabular}{@{}l N{$+$00.0}N{$+$00.0}N{$+$00.0} @{\hspace{14pt}} r@{\,}l @{\hspace{14pt}} r@{\,}l@{}}
\toprule
& \multicolumn{3}{c}{$B{\times}D$ (pp)} & \multicolumn{4}{c@{}}{Words $-$ codified statement (pp)} \\
\cmidrule(lr){2-4}\cmidrule(l){5-8}
(c) Writer & \multicolumn{1}{c}{$\Ssrc$} & \multicolumn{1}{c}{$\Sdef$} & \multicolumn{1}{c}{$\Sdel$} & \multicolumn{2}{c}{Update} & \multicolumn{2}{c@{}}{Agreement} \\
\midrule
Qwen3.5-9B & $+$12.7 & $+$22.0 & $+$16.0 & $+$1.4 & \mtint{[$-$0.8,\,$+$3.7]} & $-$4.4 & \mtint{[$-$5.3,\,$-$3.3]} \\
Claude Sonnet 5.5 & $+$27.0 & $+$38.2 & $+$35.4 & $-$7.3 & \mtint{[$-$8.3,\,$-$6.2]} & $-$5.4 & \mtint{[$-$5.6,\,$-$5.1]} \\
Claude Opus 5.5 & $+$27.5 & $+$38.2 & $+$35.5 & $-$18.1 & \mtint{[$-$20.5,\,$-$15.7]} & $-$16.0 & \mtint{[$-$16.4,\,$-$15.6]} \\
\bottomrule
\end{mttabular}
\mtnotes{$B{\times}D$: the interaction at $D_{\mathrm{m}}$ on definition items (point estimates). Words $-$ codified statement: the programs from the user's words minus the same writer's programs from the codified statement, same samples, paired per item. Update: on the changed definition item-times at $D_{\mathrm{m}}$ with a key ($B{=}1$), the three shifts pooled. Agreement: the programs' values before the shift that equal gold.}

%% file: generated/tab_view.tex
\begin{table*}[t]
\caption{The query-time view against the cell with the stored programs on every item class (10 worlds, the same probes).}
\label{tab:view}
\centering\mtsetup
\setlength{\tabcolsep}{4pt}
\begin{mttabular}{@{}ll N{0,000} @{\hspace{10pt}} N{100}N{100}N{100}N{100} @{\hspace{10pt}} >{\mtpairwidths{$+$00.0}{[$-$00.0,\,$-$00.0]}}c>{\mtpairwidths{$+$00.0}{[$-$00.0,\,$-$00.0]}}c@{}}
\toprule
& & & \multicolumn{4}{c}{Level (\%)} & \multicolumn{2}{c@{}}{Paired difference (pp)} \\
\cmidrule(lr){4-7}\cmidrule(l){8-9}
Shift & Class & \multicolumn{1}{c}{$n$} & \multicolumn{1}{c}{View} & \multicolumn{1}{c}{$D_{\mathrm{m}}$} & \multicolumn{1}{c}{$B{=}0$} & \multicolumn{1}{c}{No-store} & View $-$ $D_{\mathrm{m}}$ & \multicolumn{1}{c@{}}{View $-$ no-store} \\
\midrule
\mtgroup{9}{Update (changed item-times)}
\mtin $\Ssrc$ & definition & 165 & 77 & 77 & 48 & 6 & \mtpair{0.0}{[$-$6.4,\,$+$5.5]} & \mtpair{$+$70.9}{[$+$65.8,\,$+$75.3]} \\
 & other & 428 & 70 & 92 & 81 & 88 & \mtpair{$-$22.2}{[$-$28.3,\,$-$16.0]} & \mtpair{$-$18.2}{[$-$24.9,\,$-$11.0]} \\
 & all & 593 & 72 & 88 & 72 & 65 & \mtpair{$-$16.0}{[$-$19.7,\,$-$12.6]} & \mtpair{$+$6.6}{[$+$1.6,\,$+$12.0]} \\
\mtgap
\mtin $\Sdef$ & definition & 201 & 63 & 69 & 35 & 4 & \mtpair{$-$6.0}{[$-$8.3,\,$-$4.0]} & \mtpair{$+$59.2}{[$+$54.9,\,$+$63.6]} \\
 & other & 83 & 69 & 81 & 77 & 81 & \mtpair{$-$12.0}{[$-$22.8,\,$-$2.9]} & \mtpair{$-$12.0}{[$-$21.9,\,$-$4.9]} \\
 & all & 284 & 65 & 73 & 47 & 26 & \mtpair{$-$7.7}{[$-$10.7,\,$-$4.8]} & \mtpair{$+$38.4}{[$+$34.1,\,$+$43.5]} \\
\mtgap
\mtin $\Sdel$ & definition & 154 & 76 & 79 & 36 & 3 & \mtpair{$-$3.2}{[$-$7.6,\,$+$1.1]} & \mtpair{$+$73.4}{[$+$67.2,\,$+$79.0]} \\
 & other & 718 & 43 & 42 & 15 & 88 & \mtpair{$+$1.0}{[$-$5.5,\,$+$7.6]} & \mtpair{$-$45.0}{[$-$49.8,\,$-$39.7]} \\
 & all & 872 & 49 & 49 & 18 & 73 & \mtpair{$+$0.2}{[$-$4.9,\,$+$5.4]} & \mtpair{$-$24.1}{[$-$28.4,\,$-$19.2]} \\
\mtgap
\mtin pooled & definition & 520 & 71 & 75 & 39 & 4 & \mtpair{$-$3.3}{[$-$6.6,\,$-$0.5]} & \mtpair{$+$67.1}{[$+$64.0,\,$+$70.6]} \\
 & other & 1,229 & 54 & 62 & 42 & 88 & \mtpair{$-$8.0}{[$-$12.5,\,$-$3.9]} & \mtpair{$-$33.4}{[$-$37.9,\,$-$29.0]} \\
 & all & 1,749 & 59 & 66 & 41 & 63 & \mtpair{$-$6.6}{[$-$9.2,\,$-$4.2]} & \mtpair{$-$3.5}{[$-$7.0,\,$+$0.3]} \\
\midrule
\mtgroup{9}{Stability (unchanged item-times)}
\mtin pooled & definition & 100 & 92 & 49 & 72 & 50 & \mtpair{$+$43.0}{[$+$33.3,\,$+$55.6]} & \mtpair{$+$42.0}{[$+$29.5,\,$+$53.6]} \\
 & other & 1,609 & 62 & 88 & 74 & 70 & \mtpair{$-$25.3}{[$-$32.1,\,$-$18.9]} & \mtpair{$-$7.1}{[$-$13.7,\,$+$1.0]} \\
 & all & 1,709 & 64 & 85 & 74 & 68 & \mtpair{$-$21.3}{[$-$28.0,\,$-$14.8]} & \mtpair{$-$4.2}{[$-$11.4,\,$+$4.3]} \\
\bottomrule
\end{mttabular}
\mtnotes{View: no-store BM25 plus the writer's program run on the live records at the probe's time, with nothing stored. $D_{\mathrm{m}}$: \holdmem{} with the same programs, stored with the source key. $B{=}0$: the same without a key. No-store: no-store BM25. $n$: paired item-times. Pooled: the three shifts together. Levels: point estimates, rounded to whole percent.}
\end{table*}

%% file: generated/tab_diagnostic_tests.tex
\begin{table*}[t]
\caption{The four diagnostic tests (\cref{app:experiments}) per tier, switch, horizon, and system, in the main study and the replication.}
\label{tab:diagnostic-tests}
\centering\mtsetup
\setlength{\tabcolsep}{4pt}
\begin{mttabular}{@{}l >{\raggedright\arraybackslash}p{5cm} c c @{\hspace{14pt}} c c@{}}
\toprule
& & \multicolumn{2}{c}{Main study} & \multicolumn{2}{c@{}}{Replication} \\
\cmidrule(lr){3-4}\cmidrule(l){5-6}
Test & Systems or units & Estimate & Passes & Estimate & Passes \\
\midrule
Retention test & \holdmem{}(1,1), three tiers & $-$1.4 to $-$0.4 & 3/3 & $-$1.9 to $-$0.1 & 3/3 \\
 & \holdmem{}(1,1), five other writers & $-$5.2 to $-$0.1 & 2/5 & \mtna & \mtna \\
\addlinespace[2pt]
Leaving test & \holdmem{}(1,1) with its keys & 0.0 & 1/1 & \mtna & \mtna \\
\addlinespace[2pt]
Utility floor & \mbox{\holdmem{}(1,1)}~$\cdot$ \mbox{no-store BM25}~$\cdot$ \mbox{plain RAG}, three horizons & $-$22.1 to $+$12.2 & 3/3\,$\cdot$\,0/3\,$\cdot$\,0/3 & $-$23.4 to $+$13.3 & 3/3\,$\cdot$\,0/3\,$\cdot$\,0/3 \\
Cost ceiling & the same systems and horizons & at most 2.1\% & 9/9 & at most 2.0\% & 9/9 \\
\addlinespace[2pt]
Utility floor & native systems and the two baselines & $-$21.7 to $-$1.8 & 1/8 & \mtna & \mtna \\
Cost ceiling & the same systems & at most 1.1\% & 8/8 & \mtna & \mtna \\
\bottomrule
\end{mttabular}
\mtnotes{Estimate: point estimates of the change in retention, the leaving test's difference in carryover and the margin over the utility floor in pp, and of the serving cost as a share of a full re-read, a range over the units. Passes: passing units, per system where a row names several. Thresholds: retention test 90\% interval above $-$3\,pp, leaving test 90\% interval within $\pm$3\,pp, utility floor $\ge-3$\,pp, cost ceiling at most half. Passing: gemma-4-12B and Ministral-3-14B of the five other writers; Cognee of the native systems and the two baselines. ---: not run in the replication.}
\end{table*}

%% file: generated/tab_zero_after.tex
\begin{table*}[t]
\caption{Update and Stability of no-store BM25 (no-store), \holdmem{}(1,1) and \holdmem{}(0,0) by item class after each shift, and no-store BM25 minus \holdmem{}(1,1) on Update, paired.}
\label{tab:zero-after}
\centering\mtsetup
\setlength{\tabcolsep}{4pt}
\begin{mttabular}{@{}ll N{100}N{100}N{100} r@{\,}l @{\hspace{14pt}} N{100}N{100}N{100}@{}}
\toprule
& & \multicolumn{5}{c}{Update (\%)} & \multicolumn{3}{c@{}}{Stability (\%)} \\
\cmidrule(lr){3-7}\cmidrule(l){8-10}
Shift & Class & \multicolumn{1}{c}{No-store} & \multicolumn{1}{c}{(1,1)} & \multicolumn{1}{c}{(0,0)} & \multicolumn{2}{c}{No-store $-$ (1,1) (pp)} & \multicolumn{1}{c}{No-store} & \multicolumn{1}{c}{(1,1)} & \multicolumn{1}{c@{}}{(0,0)} \\
\midrule
\mtgroup{10}{Main study}
\mtin$\Ssrc$ & definition & 7 & 99 & 10 & $-$91.9 & \mtint{[$-$96.0,\,$-$87.9]} & 45 & 100 & 48 \\
 & other & 83 & 91 & 84 & $-$8.1 & \mtint{[$-$11.3,\,$-$4.9]} & 71 & 93 & 74 \\
 & all & 61 & 93 & 63 & $-$32.0 & \mtint{[$-$34.7,\,$-$29.5]} & 69 & 93 & 73 \\
\mtgap
\mtin$\Sdef$ & definition & 5 & 100 & 7 & $-$94.3 & \mtint{[$-$97.2,\,$-$91.2]} & 31 & 92 & 38 \\
 & other & 70 & 73 & 71 & $-$2.9 & \mtint{[$-$7.5,\,$+$0.9]} & 68 & 93 & 79 \\
 & all & 25 & 92 & 27 & $-$66.3 & \mtint{[$-$71.5,\,$-$61.3]} & 67 & 93 & 78 \\
\mtgap
\mtin$\Sdel$ & definition & 2 & 99 & 4 & $-$97.4 & \mtint{[$-$99.4,\,$-$95.0]} & 45 & 100 & 60 \\
 & other & 86 & 68 & 10 & $+$18.4 & \mtint{[$+$14.2,\,$+$23.3]} & 66 & 87 & 80 \\
 & all & 71 & 73 & 9 & $-$1.8 & \mtint{[$-$5.6,\,$+$2.2]} & 65 & 88 & 78 \\
\midrule
\mtgroup{10}{Replication}
\mtin$\Ssrc$ & definition & 7 & 99 & 10 & $-$92.0 & \mtint{[$-$96.3,\,$-$87.2]} & 45 & 100 & 59 \\
 & other & 84 & 90 & 86 & $-$6.0 & \mtint{[$-$8.6,\,$-$3.2]} & 72 & 92 & 76 \\
 & all & 62 & 92 & 64 & $-$30.7 & \mtint{[$-$32.9,\,$-$28.5]} & 70 & 92 & 75 \\
\mtgap
\mtin$\Sdef$ & definition & 7 & 99 & 9 & $-$92.0 & \mtint{[$-$96.0,\,$-$87.7]} & 62 & 100 & 77 \\
 & other & 78 & 86 & 82 & $-$7.5 & \mtint{[$-$14.4,\,$-$1.2]} & 71 & 92 & 81 \\
 & all & 30 & 95 & 32 & $-$64.7 & \mtint{[$-$69.2,\,$-$60.5]} & 71 & 92 & 81 \\
\mtgap
\mtin$\Sdel$ & definition & 3 & 100 & 6 & $-$96.5 & \mtint{[$-$99.0,\,$-$93.4]} & 52 & 100 & 63 \\
 & other & 87 & 67 & 10 & $+$20.4 & \mtint{[$+$14.7,\,$+$25.7]} & 66 & 88 & 80 \\
 & all & 72 & 73 & 9 & $-$0.5 & \mtint{[$-$6.0,\,$+$4.8]} & 65 & 89 & 78 \\
\bottomrule
\end{mttabular}
\mtnotes{Update and Stability levels: point estimates.}
\end{table*}

%% file: generated/tab_switches.tex
\begin{table*}[t]
\caption{Writer switches: how well writer agreement predicts survival of the switch, and what the switch does to retention.}
\label{tab:switches}
\centering\mtsetup
\setlength{\tabcolsep}{4pt}
\begin{mttabular}{@{}l r@{\,}l N{0.00} @{\hspace{12pt}} r@{\,}l @{\hspace{12pt}} r@{\,}l@{}}
\toprule
Writer switch & \multicolumn{2}{c}{Slope} & \multicolumn{1}{c}{AUROC} & \multicolumn{2}{c}{$D$ surv. (pp)} & \multicolumn{2}{c@{}}{Retention $\Delta$ (pp)} \\
\midrule
same tier & 1.37 & \mtint{[0.94,\,2.19]} & 0.76 & $+$0.2 & \mtint{[$-$0.4,\,$+$0.7]} & $-$1.4 & \mtint{[$-$1.9,\,$-$0.9]} \\
upgrade (27B)$^\ast$ & \mtmuted{$-$3.52} & \mtint{[$-$99.29,\,$-$0.96]} & 0.47 & 0.0 & \mtint{[$-$0.4,\,$+$0.4]} & $-$0.4 & \mtint{[$-$0.9,\,0.0]} \\
downgrade & 1.09 & \mtint{[0.95,\,1.39]} & 0.93 & $+$2.7 & \mtint{[$+$1.7,\,$+$3.6]} & $-$0.4 & \mtint{[$-$0.8,\,0.0]} \\
\mtgap
Llama-3.1-8B & 0.69 & \mtint{[0.55,\,0.85]} & 0.64 & $+$7.2 & \mtint{[$+$5.0,\,$+$9.2]} & $-$5.2 & \mtint{[$-$6.4,\,$-$4.1]} \\
gemma-4-12B$^\ast$ & \multicolumn{2}{c}{---} & \mtna & $-$0.1 & \mtint{[$-$0.4,\,0.0]} & $-$0.1 & \mtint{[$-$0.4,\,0.0]} \\
gpt-oss-20b & 0.52 & \mtint{[0.41,\,0.66]} & 0.71 & $-$0.7 & \mtint{[$-$2.0,\,$+$0.5]} & $-$2.4 & \mtint{[$-$3.2,\,$-$1.5]} \\
Ministral-3-14B & 0.62 & \mtint{[0.47,\,0.85]} & 0.62 & $+$7.7 & \mtint{[$+$5.6,\,$+$9.9]} & $-$1.7 & \mtint{[$-$2.5,\,$-$1.0]} \\
Qwen3-8B & 0.46 & \mtint{[0.26,\,0.68]} & 0.64 & $+$1.0 & \mtint{[$-$0.2,\,$+$2.2]} & $-$3.5 & \mtint{[$-$4.7,\,$-$2.3]} \\
\midrule
pooled, 8 switches & 0.85 & \mtint{[0.79,\,0.94]} & 0.83 & & & & \\
\bottomrule
\end{mttabular}
\mtnotes{Slope: calibration slope of survival on writer agreement (90\% interval). AUROC: point estimates. $D$ surv.: survival of $D{=}1$ minus $D{=}0$ items. Retention $\Delta$: the change in retention of \holdmem{}(1,1) (90\% interval). $^\ast$~Almost no item failed, so the slope (gray) is not usable.}
\end{table*}

%% file: generated/tab_families.tex
\begin{table*}[t]
\caption{Other model families as reader of the main writer's stores (5 test worlds), beside Qwen3.5-9B on the same worlds. \Cref{tab:main}b gives them as writer and reader.}
\label{tab:families}
\centering\mtsetup
\setlength{\tabcolsep}{3.5pt}
\begin{mttabular}{@{}l r@{\,}l r@{\,}l r@{\,}l r@{\,}l@{}}
\toprule
As reader only (pp) & \multicolumn{2}{c}{Llama-3.1-8B} & \multicolumn{2}{c}{gemma-4-12B} & \multicolumn{2}{c}{gpt-oss-20b} & \multicolumn{2}{c@{}}{Qwen3.5-9B} \\
\midrule
$B$ effect on Update, $\Ssrc$ & $+$26.4 & \mtint{[$+$18.8,\,$+$33.9]} & $+$13.0 & \mtint{[$+$11.1,\,$+$15.2]} & $+$10.1 & \mtint{[$+$6.7,\,$+$14.5]} & $+$9.1 & \mtint{[$+$5.4,\,$+$13.9]} \\
$B$ effect on Update, $\Sdef$ & $+$17.8 & \mtint{[$+$11.2,\,$+$23.2]} & $+$23.5 & \mtint{[$+$19.3,\,$+$29.3]} & $+$18.1 & \mtint{[$+$9.4,\,$+$24.8]} & $+$17.4 & \mtint{[$+$10.9,\,$+$22.2]} \\
$D$ effect on accuracy on re-derived items, $\Ssrc$ & $+$15.4 & \mtint{[$+$13.3,\,$+$17.7]} & $+$15.2 & \mtint{[$+$13.4,\,$+$16.9]} & $+$13.7 & \mtint{[$+$11.7,\,$+$15.3]} & $+$13.7 & \mtint{[$+$11.9,\,$+$16.2]} \\
$D$ effect on accuracy on re-derived items, $\Sdef$ & $+$34.1 & \mtint{[$+$29.7,\,$+$38.3]} & $+$35.5 & \mtint{[$+$31.5,\,$+$39.4]} & $+$32.1 & \mtint{[$+$28.8,\,$+$35.4]} & $+$33.4 & \mtint{[$+$30.1,\,$+$37.4]} \\
$B{\times}D$ on Update, $\Ssrc$ & $+$9.1 & \mtint{[$+$2.8,\,$+$15.9]} & $+$10.4 & \mtint{[$+$7.0,\,$+$14.1]} & $+$10.1 & \mtint{[$+$3.5,\,$+$15.2]} & $+$6.7 & \mtint{[$+$0.9,\,$+$13.4]} \\
$B{\times}D$ on Update, $\Sdef$ & $+$30.2 & \mtint{[$+$23.8,\,$+$38.3]} & $+$33.6 & \mtint{[$+$29.1,\,$+$38.1]} & $+$28.2 & \mtint{[$+$23.6,\,$+$31.8]} & $+$30.9 & \mtint{[$+$24.5,\,$+$38.0]} \\
\holdmem{}(1,1) accuracy, $\Sdef$ (\%) & 93.3 &  & 96.6 &  & 96.3 &  & 91.6 &  \\
\bottomrule
\end{mttabular}
\mtnotes{$D$'s effect on missed invalidation was 0.0 in every column. \holdmem{}(1,1) accuracy: point estimates. Qwen3.5-9B: the main reader, on the same worlds.}
\end{table*}

%% file: generated/tab_native_null.tex
\begin{table*}[t]
\caption{Native values against their references per system and comparison, with each system's inferred setting (three-month worlds).}
\label{tab:native-diff}\label{tab:native-null}
\centering\mtsetup
\setlength{\tabcolsep}{4pt}
\begin{mttabular}{@{}l l l l r@{\,}l l @{\hspace{12pt}} r@{\,}l@{}}
\toprule
& & & & \multicolumn{3}{c}{Native $-$ cell (pp)} & \multicolumn{2}{c@{}}{} \\
\cmidrule(lr){5-7}
System, cell & \multicolumn{2}{l}{Comparison} & Contained by & \multicolumn{2}{c}{Difference} & \multicolumn{1}{c}{Widened} & \multicolumn{2}{c@{}}{Native $-$ no-store (pp)} \\
\midrule
Mem0 & $\Ssrc$ & Update & no-store, RAG & $+$13.1 & \mtint{[$+$7.4,\,$+$18.9]} & \mtint{[$+$5.3,\,$+$20.9]} & $-$0.5 & \mtint{[$-$5.8,\,$+$4.1]} \\
cell (0,0) &  & Stability & (0,0), (1,0), no-store, RAG & $+$0.8 & \mtint{[$-$7.8,\,$+$8.1]} & \mtint{[$-$8.1,\,$+$8.5]} & $-$5.4 & \mtint{[$-$11.7,\,$+$1.4]} \\
$P$ 54.9\% & $\Sdef$ & Update & (0,0) & $+$3.4 & \mtint{[$-$4.3,\,$+$10.4]} & \mtint{[$-$7.2,\,$+$13.6]} & $-$7.7 & \mtint{[$-$15.7,\,$-$0.4]} \\
\mtint{[49.0,\,60.9]} &  & Stability & (0,0), (1,0), no-store, RAG & $+$0.6 & \mtint{[$-$6.0,\,$+$7.0]} & \mtint{[$-$6.8,\,$+$7.8]} & $-$2.7 & \mtint{[$-$9.4,\,$+$5.6]} \\
inside 5/6 & $\Sdel$ & Update$^\circ$ & (0,0) & $+$13.2 & \mtint{[0.0,\,$+$25.0]} & \mtint{[$-$3.0,\,$+$28.3]} & $-$13.2 & \mtint{[$-$25.6,\,0.0]} \\
 &  & Stability$^\circ$ & (0,0), no-store, RAG & $-$17.0 & \mtint{[$-$23.9,\,$-$8.9]} & \mtint{[$-$25.5,\,$-$7.5]} & $-$6.7 & \mtint{[$-$11.7,\,$-$1.6]} \\
\mtgap
A-Mem & $\Ssrc$ & Update & no-store, RAG & $+$11.7 & \mtint{[$+$6.2,\,$+$16.7]} & \mtint{[$+$6.0,\,$+$16.9]} & $-$1.8 & \mtint{[$-$8.2,\,$+$3.7]} \\
cell (0,0) &  & Stability & none & $+$13.6 & \mtint{[$+$9.3,\,$+$17.9]} & \mtint{[$+$8.3,\,$+$18.9]} & $+$10.5 & \mtint{[$+$4.5,\,$+$16.9]} \\
$P$ 66.8\% & $\Sdef$ & Update & (0,0) & $+$3.4 & \mtint{[$-$2.0,\,$+$8.6]} & \mtint{[$-$2.3,\,$+$8.9]} & $-$7.7 & \mtint{[$-$14.7,\,$-$1.9]} \\
\mtint{[59.9,\,73.3]} &  & Stability & none & $+$12.0 & \mtint{[$+$5.0,\,$+$18.5]} & \mtint{[$+$2.7,\,$+$20.9]} & $+$11.0 & \mtint{[$+$4.1,\,$+$18.3]} \\
inside 1/6 & $\Sdel$ & Update$^\circ$ & none & $+$2.6 & \mtint{[0.0,\,$+$7.7]} & \mtint{[0.0,\,$+$7.7]} & $-$23.7 & \mtint{[$-$33.3,\,$-$13.9]} \\
 &  & Stability$^\circ$ & (1,0) & $-$6.1 & \mtint{[$-$13.3,\,$+$1.1]} & \mtint{[$-$17.1,\,$+$4.9]} & $+$4.3 & \mtint{[$-$1.9,\,$+$10.2]} \\
\mtgap
MemOS & $\Ssrc$ & Update & no-store, RAG & $+$12.8 & \mtint{[$+$7.7,\,$+$17.8]} & \mtint{[$+$7.7,\,$+$17.8]} & $-$0.7 & \mtint{[$-$7.1,\,$+$5.3]} \\
cell (0,0) &  & Stability & (0,0), (1,0), no-store, RAG & $+$2.1 & \mtint{[$-$4.5,\,$+$8.5]} & \mtint{[$-$4.6,\,$+$8.7]} & $-$2.4 & \mtint{[$-$7.9,\,$+$3.0]} \\
$P$ 54.9\% & $\Sdef$ & Update & no-store, RAG & $+$10.2 & \mtint{[$+$3.0,\,$+$16.8]} & \mtint{[$+$2.9,\,$+$16.9]} & $-$0.9 & \mtint{[$-$6.7,\,$+$5.0]} \\
\mtint{[50.0,\,59.9]} &  & Stability & (0,0), (1,0), no-store, RAG & $-$1.5 & \mtint{[$-$8.6,\,$+$5.3]} & \mtint{[$-$8.9,\,$+$5.7]} & $-$2.7 & \mtint{[$-$9.8,\,$+$5.5]} \\
inside 4/6 & $\Sdel$ & Update$^\circ$ & (0,0) & $+$10.5 & \mtint{[0.0,\,$+$22.5]} & \mtint{[$-$4.4,\,$+$26.5]} & $-$15.8 & \mtint{[$-$27.3,\,$-$3.0]} \\
 &  & Stability$^\circ$ & (0,0), no-store, RAG & $-$18.4 & \mtint{[$-$26.2,\,$-$10.6]} & \mtint{[$-$26.4,\,$-$10.3]} & $-$7.7 & \mtint{[$-$12.5,\,$-$2.5]} \\
\mtgap
LightRAG & $\Ssrc$ & Update & none & $-$12.8 & \mtint{[$-$18.2,\,$-$7.6]} & \mtint{[$-$18.9,\,$-$6.8]} & $-$6.2 & \mtint{[$-$12.5,\,$-$0.7]} \\
cell (0,1)c &  & Stability & none & $-$13.5 & \mtint{[$-$18.4,\,$-$8.7]} & \mtint{[$-$20.6,\,$-$6.5]} & $+$16.3 & \mtint{[$+$8.9,\,$+$22.7]} \\
$P$ 10.7\% & $\Sdef$ & Update & (0,0) & $-$14.9 & \mtint{[$-$20.5,\,$-$9.9]} & \mtint{[$-$21.4,\,$-$8.9]} & $-$7.7 & \mtint{[$-$16.2,\,$-$0.8]} \\
\mtint{[7.3,\,14.6]} &  & Stability & none & $-$17.5 & \mtint{[$-$20.8,\,$-$13.9]} & \mtint{[$-$21.8,\,$-$12.9]} & $+$8.3 & \mtint{[$+$1.8,\,$+$15.1]} \\
inside 0/6 & $\Sdel$ & Update$^\circ$ & none & $-$31.6 & \mtint{[$-$48.6,\,$-$17.1]} & \mtint{[$-$48.6,\,$-$17.1]} & $-$26.3 & \mtint{[$-$39.5,\,$-$12.8]} \\
 &  & Stability$^\circ$ & (1,0) & $-$16.6 & \mtint{[$-$20.4,\,$-$12.5]} & \mtint{[$-$27.0,\,$-$6.1]} & $+$6.2 & \mtint{[$+$1.4,\,$+$11.2]} \\
\mtgap
HippoRAG 2 & $\Ssrc$ & Update & (1,0), (0,1)c, no-store, RAG & $-$3.4 & \mtint{[$-$9.5,\,$+$2.9]} & \mtint{[$-$9.8,\,$+$3.1]} & $+$3.2 & \mtint{[$-$2.6,\,$+$8.5]} \\
cell (0,1)c &  & Stability & none & $-$16.0 & \mtint{[$-$19.4,\,$-$12.6]} & \mtint{[$-$19.5,\,$-$12.6]} & $+$11.8 & \mtint{[$+$6.6,\,$+$17.3]} \\
$P$ 9.5\% & $\Sdef$ & Update & no-store, RAG & $-$7.2 & \mtint{[$-$12.0,\,$-$2.7]} & \mtint{[$-$12.0,\,$-$2.7]} & 0.0 & \mtint{[$-$9.5,\,$+$8.9]} \\
\mtint{[5.6,\,13.7]} &  & Stability & none & $-$13.8 & \mtint{[$-$16.8,\,$-$10.9]} & \mtint{[$-$16.8,\,$-$10.8]} & $+$13.6 & \mtint{[$+$5.2,\,$+$23.0]} \\
inside 1/6 & $\Sdel$ & Update$^\circ$ & (0,0) & $-$15.8 & \mtint{[$-$33.3,\,0.0]} & \mtint{[$-$33.3,\,0.0]} & $-$10.5 & \mtint{[$-$22.2,\,0.0]} \\
 &  & Stability$^\circ$ & (1,0) & $-$14.8 & \mtint{[$-$18.2,\,$-$11.7]} & \mtint{[$-$18.2,\,$-$11.7]} & $+$6.7 & \mtint{[$+$1.5,\,$+$12.8]} \\
\mtgap
Cognee & $\Ssrc$ & Update & no-store, RAG & $-$8.9 & \mtint{[$-$14.1,\,$-$4.4]} & \mtint{[$-$14.3,\,$-$4.1]} & $-$2.3 & \mtint{[$-$8.2,\,$+$3.4]} \\
cell (0,1)c &  & Stability & none & $-$3.5 & \mtint{[$-$6.2,\,$-$1.2]} & \mtint{[$-$6.3,\,$-$1.1]} & $+$25.4 & \mtint{[$+$19.5,\,$+$31.0]} \\
$P$ 12.3\% & $\Sdef$ & Update & no-store, RAG & $-$9.4 & \mtint{[$-$13.7,\,$-$5.8]} & \mtint{[$-$14.0,\,$-$5.4]} & $-$2.1 & \mtint{[$-$8.8,\,$+$3.3]} \\
\mtint{[8.6,\,16.2]} &  & Stability & oracle & $-$2.5 & \mtint{[$-$4.9,\,$-$0.4]} & \mtint{[$-$5.1,\,$-$0.3]} & $+$26.1 & \mtint{[$+$16.3,\,$+$37.0]} \\
inside 1/6 & $\Sdel$ & Update$^\circ$ & (0,0) & $-$21.1 & \mtint{[$-$35.1,\,$-$7.9]} & \mtint{[$-$36.2,\,$-$6.7]} & $-$15.8 & \mtint{[$-$27.0,\,$-$3.0]} \\
 &  & Stability$^\circ$ & (0,1)c, (1,1)c & $-$1.8 & \mtint{[$-$4.7,\,$+$1.1]} & \mtint{[$-$5.3,\,$+$1.7]} & $+$20.2 & \mtint{[$+$15.6,\,$+$24.9]} \\
\bottomrule
\end{mttabular}
\mtnotes{Cell: the \holdmem{} cell with the setting inferred from behavior. $P$: deletion persistence, the share of deleted items still answered, with its 95\% interval. Inside $k$/$n$: $k$ comparisons inside the 95\% interval of the cell, each also outside that of the cell with the other $B$. Contained by: the references whose 95\% interval contains the native value ($(D,B)$: \holdmem{} cell, c: citation key, no-store: no-store BM25, RAG: plain RAG, oracle: oracle context). Difference: native minus the \holdmem{} cell with the inferred setting, paired by item, with its 95\% interval. Widened: that interval widened by the variance between repeated ingestions. Native $-$ no-store: paired by item, with its 95\% interval. $^\circ$: after $\Sdel$, from which $B$ is inferred.}
\end{table*}

%% file: appendix/app_errors.tex
\section{Error Analysis}
\label{app:errors}

This section analyzes two failures: the writer's programs after a deletion, and the program for working
hours.

\paragraph{The programs return a value from another live record where the gold has
none.}\label{app:dm-deletion} \Cref{tab:dm-deletion} sorts every asked item-time after a deletion at
$B{=}1$ by what the cell stored and how the reader answered. On the \ExRThreeDelItems{} changed
item-times whose record was deleted, $D_{\mathrm{m}}$ answered \ExRThreeDelDmOkShare{} correctly,
compared with \ExRThreeDelDzeroOkShare{} for $D{=}0$. It lost \ExRThreeDelLossVsDzero{} of these
item-times relative to $D{=}0$. Of this loss, \ExRThreeDelLossShareVsDzeroA{} is a value stored where the
gold has none, nearly all of it a program value that replaced the writer's
(\ExRThreeDelLossShareVsDzeroAReplacement{}). Most of the rest, \ExRThreeDelLossShareVsDzeroD{}, comes
from the reader. When shown the template ``there is no current value'', the reader answers with the
status \texttt{answer} about as often at $D{=}1$ as at $D_{\mathrm{m}}$ ($D_{\mathrm{m}}$ minus $D{=}1$
on this class: \ExRThreeDelDmMinusDoneD{}). None of the loss comes from a failed program.

\paragraph{The programs do not react to the deletion.} Where the program's value replaced the writer's
(\ExRThreeDelValueStored{} item-times, \ExRThreeDelValueStoredShare{}), exact recomputation stored no
value every time. Of these program values, \ExRThreeDelValueOtherRecord{} are the same field of another
live record. For tasks, it is always the due date of the most recently updated task
(\ExRThreeDelValueTaskLastListed{} of \ExRThreeDelValueTask{}). For contacts, it is mostly the field of
the first contact in the list (\ExRThreeDelValueContactFirstListed{} of \ExRThreeDelValueContact{}). On
live records that did not change, the programs are wrong about as often as on deleted ones. Task due
dates were wrong on \ExRThreeDelProgTaskLiveWrong{} of \ExRThreeDelProgTaskLiveItems{} live and
\ExRThreeDelProgTaskDeletedWrong{} of \ExRThreeDelProgTaskDeletedItems{} deleted item-times, and contact
fields on \ExRThreeDelProgContactLiveWrong{} of \ExRThreeDelProgContactLiveItems{} and
\ExRThreeDelProgContactDeletedWrong{} of \ExRThreeDelProgContactDeletedItems{}. The recurring-series
programs return no value for every deleted record (\ExRThreeDelProgSeriesDeletedWrong{} of
\ExRThreeDelProgSeriesDeletedItems{} wrong) and are wrong on \ExRThreeDelProgSeriesLiveWrong{} of
\ExRThreeDelProgSeriesLiveItems{} live item-times. Before any shift, the task programs agree with gold
on \ExRThreeDelTzeroTaskAgree{} of items. While the record that the writer cites exists, the writer's
value passes the span check, and the program's output is only recorded. After a deletion, the writer
states no value, the check fails, and the unchecked program value replaces the correct ``no value''. We
infer which record a program reads from its output.

\paragraph{Failed programs cost Stability instead.} Over all re-derived item-times, the difference in
accuracy of \SynthExCOneOrphan{} combines a gain on definition items (\ExRThreeDelCOneDefinition{}) with
a loss on the other items (\ExRThreeDelCOneOther{}). Failed programs (\ExRThreeDelFailedSlots{} of
\ExRThreeDelSlots{} slots at the last write) account for \ExRThreeDelStableLossShareC{} of what
$D_{\mathrm{m}}$ lost on unchanged item-times (\ExRThreeDelStableAll{}).

\paragraph{The program for working hours never runs.} The writer's program for working hours, a single
program shared by every world, has a syntax error. It ran in none of the
\ExRThreeDelTzeroWorkingHoursItems{} worlds before any shift. The stored item is then unverifiable and
not served, so the stored cell failed all \ExRThreeViewTypeWorkingHoursStabilityItems{} unchanged
working-hours item-times. The query-time view fell back to no-store BM25, which answered them all. This
is the view's only lead over the stored cell (\cref{app:view}).

%% file: appendix/app_datasheet.tex
\section{Datasheet and License}
\label{app:datasheet}

This section gives the datasheet of the release, following \citet{gebru2021datasheets}.
\Cref{app:worlds} specifies the generator.

\paragraph{Motivation.} \memshift{} measures how much of what a memory derived about a person stays correct
after the sources, the person's definitions, the writer model, or the implementation change. Benchmarks
that score answers after new history mix detection and repair, and \memshift{} separates them. It crosses
detection with repair in branches that are identical outside the shift, it includes shifts that add no
text (a redefinition, a writer switch, and an export), and it scores answers against gold computed
before any text exists. The authors created the dataset for this study.

\paragraph{Composition.} An instance is a world: twelve months of one synthetic person across six sources
(e-mail, calendar, tasks, contacts, notes, and chat). A world has \WorldEventsMin{}--\WorldEventsMax{}
events, about \WorldTokensM{}M tokens, \ExAttrWorldSlots{} items, and its branches. Probes are questions
about one item at a given time, with a gold value and a gold status (\texttt{answer}, \texttt{abstain}, or
\texttt{blocked}). The
gold labels also record the dependency objects of every item at every time. Splits use disjoint seeds:
development worlds 0--9, calibration worlds 900--909, test worlds 1000--1019 (\TestWorlds{} worlds), and
replication worlds 1020--1039 (\RepWorlds{} worlds). Three-month worlds on seeds 1000--1014 serve the
native systems. The dataset is a sample from the generator, and further seeds give further worlds. It
holds no personal data, no record of a real person, and no human subjects.

\paragraph{Collection process.} The data are fully synthetic: a seeded generator writes every world,
and no person was recorded, surveyed, or paid. A language model writes world text only in the rewritten
worlds, where it rewrites message bodies and keeps every gold fact verbatim (\cref{app:rewritten}).

\paragraph{Labeling.} Deterministic evaluators compute gold from the source objects before any text
is rendered. No annotator labels an instance, and no language model scores an answer. The source
objects are stored as generated.

\paragraph{Uses.} The intended use is to measure the carryover of memories and memory systems under
controlled shifts through the seven-method interface. Further uses are to run the four diagnostic tests
on a memory and to reproduce the analyses of this paper. The worlds are synthetic evaluation data, so
results on them describe synthetic worlds and the models tested, not how a system treats the data of
real people. Seeds 1000--1039 are evaluation seeds, and results of a system tuned on them are not
comparable.

\paragraph{Distribution and license.} Code, generator, worlds, gold labels, and run manifests are
released at \url{\repourl} as one package with the probes and the analyses. The worlds come with the seeds
and the generator version that reproduce them byte for byte. The code is released under the MIT license,
and the data (worlds, gold labels, and model outputs) under CC BY 4.0. The open-weight models are not
redistributed and keep their own licenses. For Claude Sonnet 5.5 and Opus 5.5, which we accessed through
a commercial API, we release their responses.

\paragraph{Maintenance.} The authors maintain the package. A change to the generator, a scorer, or a test
threshold is a new tagged version. A world is identified by its seed and generator version. Every
run manifest records the code version it ran on, so each published result stays tied to the version that
produced it. Further seeds extend the dataset without changing existing worlds.